\documentclass[11pt, logo, twocolumn, copyright]{deepmind}

\usepackage[authoryear, sort&compress, round]{natbib}
\usepackage{makecell} %
\usepackage{afterpage}
\usepackage{tcolorbox}
\renewcommand{\today}{March 17, 2022}

\title{The Frost Hollow Experiments: Pavlovian Signalling as a Path to Coordination and Communication Between Agents}

\correspondingauthor{ppilarski@deepmind.com}

\author[1,3,4]{Patrick M. Pilarski}
\author[1]{Andrew Butcher}
\author[1]{Elnaz Davoodi}
\author[1]{Michael Bradley Johanson}
\author[1]{Dylan J. A. Brenneis}
\author[3,4]{Adam S. R. Parker}
\author[1]{Leslie Acker}
\author[2]{Matthew M. Botvinick}
\author[1]{Joseph Modayil}
\author[1,3,4]{Adam White}

\affil[1]{DeepMind, Edmonton, Canada}
\affil[2]{DeepMind, London, UK}
\affil[3]{University of Alberta, Edmonton, Canada}
\affil[4]{Alberta Machine Intelligence Institute (Amii), Edmonton, Canada}

\begin{abstract}
{\small  Learned communication between agents is a powerful tool when approaching decision-making problems that are hard to overcome by any single agent in isolation. However, continual coordination and communication learning between machine agents or human-machine partnerships remains a challenging open problem. As a stepping stone toward solving the continual communication learning problem, in this paper we contribute a multi-faceted study into what we term {\em Pavlovian signalling}---a process by which learned, temporally extended predictions made by one agent inform decision-making by another agent with different perceptual access to their shared environment. We seek to establish how different temporal processes and representational choices impact Pavlovian signalling between learning agents. To do so, we introduce a partially observable decision-making domain we call the Frost Hollow. Extending from classical animal learning experiments, in this domain a prediction learning agent and a reinforcement learning agent are coupled into a two-part decision-making system that seeks to acquire sparse reward while avoiding time-conditional hazards. We evaluate two domain variations: machine prediction and control learning agents interacting in a simulated linear walk, and, as a key case of interest, a prediction learning machine interacting with a human participant via Pavlovian signalling in a virtual reality environment. Our results showcase the speed of learning for Pavlovian signalling, the impact that different temporal representations do (and do not) have on agent-agent coordination, and how temporal aliasing impacts agent-agent and human-agent interactions differently. As a main contribution, we establish Pavlovian signalling as a natural bridge between fixed signalling paradigms and fully adaptive communication learning between two agents. We further show how to computationally build this adaptive signalling process out of a fixed signalling process, characterized by fast continual prediction learning and minimal constraints on the nature of the agent receiving signals. Our results therefore point to an actionable, constructivist path towards continual communication learning between reinforcement learning agents, with potential impact in a range of real-world settings.
 
}
\end{abstract}

\begin{document}

\maketitle

\section{Introduction}

Communication learning by machines promises substantial benefits when compared to hand-engineered communication systems for machine-machine or human-machine interaction \citep{lazaridou2020,crandall2018}. Further, when different interacting agents have different perceptual access to their shared environment, and different affordances within that environment, the potential for benefit from coordination increases dramatically, as is well noted in work on human-human and human-machine coordination and joint action \citep{pezzulo2013,pezzulo2011,sebanz2006,sebanz2009,vesper2010,pesquita2018,candidi2015,grynszpan2019,knoblich2011}. However, despite the promise of increased flexibility, decreased human design effort, and opportunity for ongoing adaptation, emergent communication and coordination between learning by machines remains challenging \citep{lazaridou2020}, as does emergent signalling, coordination, and joint action between humans and machines \citep{pezzulo2011,grynszpan2019}. 

 In this paper, we build on prior research to explore the space between hand-engineered signalling approaches and fully learned agent-agent communication. Early work by Pavlov showed how signals, and signals of signals, elicit reflexive action in animals \citep{pavlov1927,windholz1990}. Further, signalling via multiple modalities has been established as more than just information transmission but in fact as a natural and powerful bridge between agents that enables alignment, coordination, and joint action between cooperating agents \citep{scottphillips2009,scottphillips2014,pezzulo2013,pezzulo2008coordinating,grynszpan2019,knoblich2011}. We here examine reflexes in response to predictions made by an agent can play the role of useful signals (feature outputs) intended to inform decision-making by another agent or a component of a single agent. Specifically, we consider the case of reflexive signals that are produced in response to predictions learned by one agent during its interactions with another agent and their shared environment. As will be demonstrated in the principal experiments of this work, our resulting approach captures some of the flexibility and adaptability of machine-learned communication while also providing rapid learning and requiring minimal assumptions to be made about the nature of the interacting agents. 

\subsection{Nexting, Predictions, and Reflexes}

Animal signalling, and animal decision-making in general, is intimately linked to the predictions learned by the animal. Animals continually make predictions about the world, in particular they make a great many predictions about what will happen next in their sensorimotor experience~\citep{clark2013whatever,gilbert2006stumbling,pezzulo2008coordinating}.  When taking a walk, we are continually making predictions about the pressure on the soles of our feet, the sound from each footfall, and the next note of a bird's song.  Such predictions seem to be computed in the background, and we become aware of their presence only when the predictions are violated: when a misstep causes our foot to slip, the sound from the gravel changes, and the bird song stops.  
Such predictions are naturally tightly coupled, where information from one sensor modality can be used to make predictions about another sensor modality.
This background prediction process has been termed {\em nexting}: the process of continually learning and making many short-timescale predictions about a range of different personal sensory signals ~\citep{gilbert2006stumbling,modayil2014multi}.  

There are several reasons why making predictions that relate the many signals in a mind could be useful to an agent and its communication with other agents.  One role is to facilitate conventional reward-based learning. \citet{rescorla1988pavlovian} argues that the predictions made by latent classical conditioning underpin most instrumental learning.  Another role for predictions is to coordinate fine motor skills, commonly associated with the cerebellum in animals~\citep{miall1996forward}.  A third role is to support Pavlovian response mechanisms, where a fixed reflexive response is produced by an animal that has learned to anticipate the imminent arrival of a stimulus  (as in Pavlov's experiments where a dog learns to salivate after hearing the bell, in anticipation of the food).  The related learning process of eyeblink conditioning has been directly traced to neural circuits in the cerebellum~\citep{jirenhed2007acquisition}.

The computational approach to learning and making predictions may take different forms, which may also have multiple biological instantiations~\citep{balleine2010human,dayan2014model}. One common division lies between model-based and model-free mechanisms.  Another common division lies between reward-driven or reward-free mechanisms.  A third division is occurs more in computational models, which often come in frequentist or Bayesian formulations, depending on what domain knowledge and computational substrate is available for the prediction algorithms, and whether the focus is on learning the predictions or the use of the predictions.  The common computational approach to nexting has been a reward-free, model-free, frequentist approach~\citep{modayil2014multi}.

The final pieces to this puzzle are the internal representations that support nexting processes and prediction learning.  One common approach in reinforcement learning is to restrict the internal representation of the environment to a function of the current sensory experience, or a short history of sensory experience~\citep{sutton1988learning,mnih2015human}.  More generally, an agent requires internal representations that support flexible behavior in the absence of an external stimulus, for problems ranging from the construction of cognitive maps~\citep{tolman1948cognitive} to bridging the short intervals between stimuli in trace conditioning experiments~\citep{ludvig2012evaluating}.   The particular forms of those internal representations of time have direct impacts on how predictions are adapted by agent experience.

In animal group or social decision making, predictions about future events or outcomes are also common, numerous, and grounded in agents' sensorimotor stream of experience. At first glance, coordination between agents may appear straightforward: two humans look at each other, nod, and pick up opposite ends of a large table so as to move it out into its desired location the middle of a yard \citep{sebanz2006}. However, underpinning even the most simple coordinated acts is also an elaborate process of nexting, prediction and action that unfolds at multiple timescales \citep{pesquita2018,knoblich2011}. In pursuit of shared objectives and desired change in their environment, humans and other animals are well known to coordinate their signalling and actions according to predictions about each other and about spatial and conceptual patterns in their environment and the way those patterns change over time \citep{pezzulo2013,knoblich2011,pesquita2018}. 

 In the present work we study the way that nexting in the form of simple sensorimotor predictions can be learned, adapted, and coupled to the generation of grounded signals used by other agents or components of the same agent---a process that we term {Pavlovian signalling}. Primarily, {\em we contribute evidence as to the degree to which this signalling approach can be rapidly learned and be useful during online or continual learning} between machine-machine and human-machine partnerships across different domain variations and representational choices. Our results aim to shed specific light on how different representations of time can impact learned temporally extended predictions, the resulting signalling between agents, and policy learning based on these learned predictions and signals. While we frame this investigation in terms of agent-agent interaction, we note that this study is equally relevant in the context of two discrete parts of a single decision-making system (similar to the signalling between the cerebellum and other parts of the central nervous system in a single animal).

This manuscript proceeds as follows. In Sec. \ref{sec:pavlov-sig} we provide greater context and a definition for Pavlovian signalling, followed in Sec. \ref{sec:time} by background on the use and representation of time by animals and machines. In Sec. \ref{sec:env-description} we describe the core problem domain for this work: the Frost Hollow. This is followed in Sec. \ref{sec:methods} by a description of our methodological choices for prediction learning, namely, general value functions (GVFs) and the specific representations and Pavlovian signalling choices that relate to them. In the empirical body of this work, Sec. \ref{sec:nexting-experiments} presents a first study into the way that different GVF predictions interact with different temporal representations and signal generation choices, followed in Sec. \ref{sec:control-experiments} by a study into the ways these prediction can be used to support control learning in a linear walk implementation of the frost hollow domain. We finish our empirical contributions in Sec. \ref{sec:vr-experiments} with an extension to human-agent interaction in a virtual reality version of the Frost Hollow. We then conclude in Secs. \ref{sec:discussion} and \ref{sec:conclusions} with observations that connect the individual experiments and outline potential areas for future extension.  %

\section{Pavlovian Signalling}%
\label{sec:pavlov-sig}

What we here term as Pavlovian signalling is a dominant theme connecting the main empirical contributions of this manuscript. In this section we define Pavlovian signalling. To do so, we first introduce ideas from the literature on the subject of Pavlovian control, followed by a survey of signalling as used in agent-agent communication; we then bring these two ideas together to illustrate the specific case of Pavlovian signalling we explore in detail and expand upon in this work, along with contact points in related literature.

\subsection{Pavlovian Control}

The primary means for studying prediction learning in animals has been the observed changes in the animal's behavior, namely how a learned prediction impacts control.  This was extensively studied in classical conditioning experiments. Unlike the later instrumental learning experiments of Skinner~\citep{rescorla1988pavlovian} which requires the animal to make a decision that is associated to differences in later reward, the learning mechanism of classical conditioning involves neither rewards nor decisions.  In the absence of such an explicit reward maximizing mechanism for behavior change, it is natural to inquire how learning a prediction by classical conditioning can change the behavior.

One compelling answer is that the learned prediction of a stimulus is associated to an unconditioned, fixed response.  For the example of Pavlov's dog, the unconditioned stimulus (US) of food in the dog's mouth is associated with the later unconditioned response (UR) of salivation.  Note that there are consistent delays in the animal's response to this stimulus, which delays its ability to successfully swallow the food.  If the animal can learn to predict when food will arrive, then it could start salivation earlier, and be faster to acquire a beneficial resource.  A second sensory signal, such as a bell that predicts the onset of the food, is referred to as the conditioned stimulus (CS).  This simple form of learning in classical conditioning creates temporal associations from an arbitrary external stimulus (CS) to a biologically relevant signal (US), with no direct influence on the nature of the response, but only affects its timing, ideally initiating the UR just as the US arrives.  Interestingly, this process operates in the absence of an external reward in animals~\citep{jirenhed2007acquisition}, and it is also independent of the decisions made by the animal, unlike instrumental learning which selectively rewards a conditioned response (CR).  The absence of these two factors makes Pavlovian conditioning a robust and efficient learning process.

This approach to Pavlovian control has a simple computational realization~\citep{modayil2014prediction, dalrymple2020pavlovian}.   Namely, it is possible to create a fixed policy that emits an action $a_1$ when a stimulus $s_1$ is predicted above some threshold $\tau$, and an action $a_2$ otherwise. Here, the stimulus (US) is $s_1$ and the response (UR) is to emits $a_1$.  Now we extend this with a prediction of the onset of the stimulus in the near future, starting from the current time $t$.  Then, the Pavlovian control will emit the response to either the direct presence of the stimulus or when the prediction crosses a threshold $\tau$ that indicates the stimulus will arrive imminently. The ideal setting of $\tau$ depends on the timescale of the prediction, the lead time needed for the response action to be effective, and the accuracy of the prediction. Thus, Pavlovian control emits a fixed reflex-like response to the prediction of an event.

This simple form of coupling the prediction to behavior is only one of many possibilities.  The predictions could modulate the vigor of the response (the amount of salivation) in addition to the timing.  Multiple predictions could also be combined to specify a more complex behavioral response (salivate for meat but not when running).  The predictions could also be coupled with rewards as in conventional instrumental learning. 

\subsection{Signalling}
\label{sec:pavlov-signalling}
 
 Signals carry information. In the simplest form, we can think of signals as means of transmitting information. However, the {\em informational content} and {\em quantity of information} in a signal are the two main aspects of signal which makes it an interesting to study for philosophers, information theorists, linguists, cognitive scientists and computer scientists~\citep{dretske1981}. Extending further, signals are known to be used in contexts beyond the raw encoding and decoding of information, and form a fundamental part of more elaborate systems of coordination and alignment between interacting agents \citep{pezzulo2013,scottphillips2009,scottphillips2014}. Taking into account both simple and complex forms of signaling, a signaling system facilitates efficient use of signals to convey content or to change the internal state of different parts of a system. Signalling systems are not limited to humans, but all levels of biological organization. For instance, monkeys~\citep{cheney2018monkeys}, birds~\citep{charrier2005call}, bees~\citep{riley2005flight, seeley2006group} and even bacteria~\citep{taga2003chemical} have signaling systems. Signaling systems can be conventional, learned and evolved gradually or even naturally exist \citep{scottphillips2014}. For humans (and arguably some other species) for example, cultural evolution and social learning evolves their signaling systems \citep{scottphillips2014}. 

In their simplest form signals transmit information from a sender (i.e source) to a receiver (i.e., destination). We can say there are two types of information in a signal~\citep{lewis2008convention}: the information that sender intends to transmit, and the information that receiver perceives to act on. It is worth noting that a maximal signaling system is the one where the sender transmits the most informative information regarding the sender’s {\em intent}, and the receiver takes the most information out of the signal (i.e., understand sender's intent and acts accordingly). Signalling has further been described by \cite{pezzulo2013} as ``an intentional strategy that supports social interactions,'' that  ``acts in concert with automatic mechanisms of resonance, prediction, and imitation, especially when the context makes actions and intentions ambiguous and difficult to read.''

An important aspect of the informational content of a signal is associated with the symbol grounding problem. The symbol grounding problem~\citep{vogt2007language} concerns the association between symbols and their meanings (i.e., semantics). In the context of Pavlovian conditioning (classical conditioning), a neutral stimuli becomes associated with a significant event. During this learning process (i.e., associating between neutral stimuli and occurrence of an event), signals in the brain are transmitted which contains information regarding prediction about an event. As in social signalling, the signal may be concretely linked to a sensorimotor perception or internal state for one agent or part of the system (grounded) while unassociated in any a priori way (ungrounded) for the agent or sub-system receiving the signal. This is the case we primarily consider in the present work.  %

\subsection{Converting Predictions into Tokens}

With an understanding of how learned predictions might be mapped to hard-wired responses or actions, and how signals may be grounded on the part of the sender or the receiver, and we now provide a definition for the combination of these two ideas: Pavlovian signalling as we explore it in this manuscript.  

\begin{tcolorbox}[colback=yellow!7!red!5!white,colframe=yellow!35!red!25!white]
\textbf{Pavlovian signalling} is a process wherein learned, temporally extended predictions are mapped in a defined way to signals intended for receipt by a decision-making agent, and where these signals are grounded for the sender in the definition of the predictive question and mapping approach that generated them.
\end{tcolorbox}

In the context of this study, a signal conveys information about the occurrence of the next event and we consider a conventional signaling system to ground the signal. In order to represent the informational content of a signal, we consider a signal as taking the form of a vector of tokens (i.e., symbols)  where each token conveys a piece of information about the occurrence of the next event. Following the Gricean maxims of communication~\citep{grice1975logic, grice1989studies}, and to provide a straightforward lens for investigation, we specifically focus on using one binary token to represent prediction of the next event in light of it being sufficient to convey information regarding the occurrence of the next event.

\begin{figure}[!th]
\centering
\includegraphics[width=2.9in]{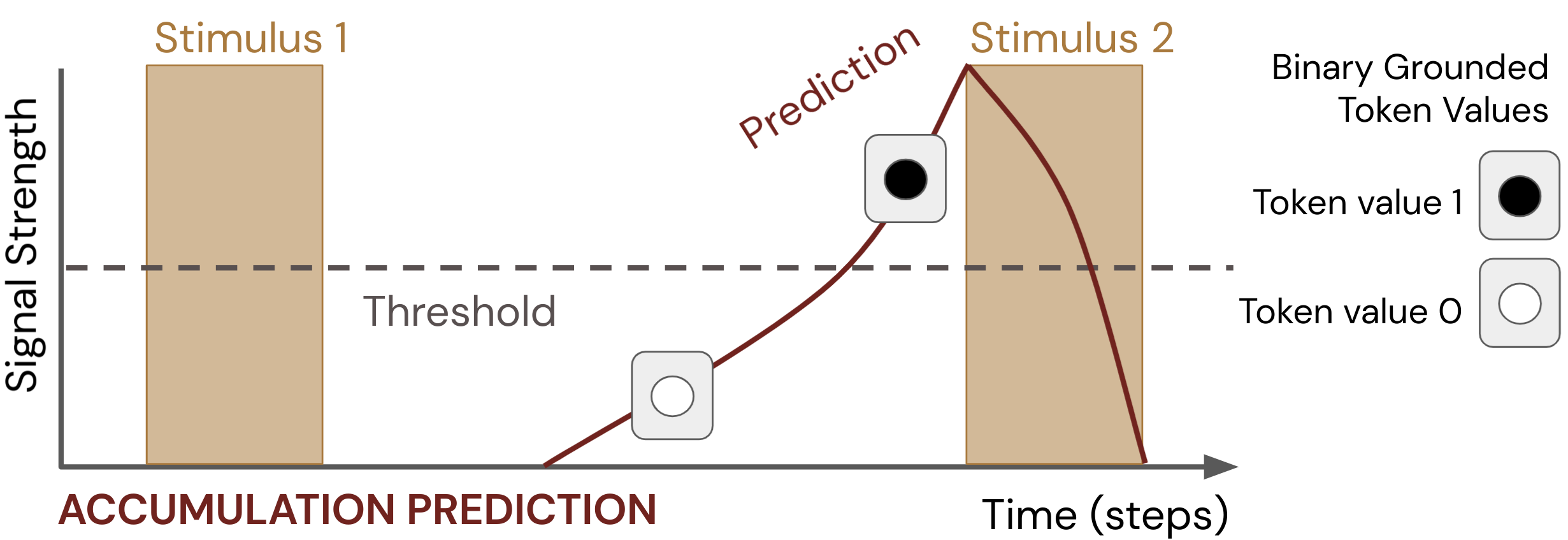}\\
(a)\\
\includegraphics[width=2.9in]{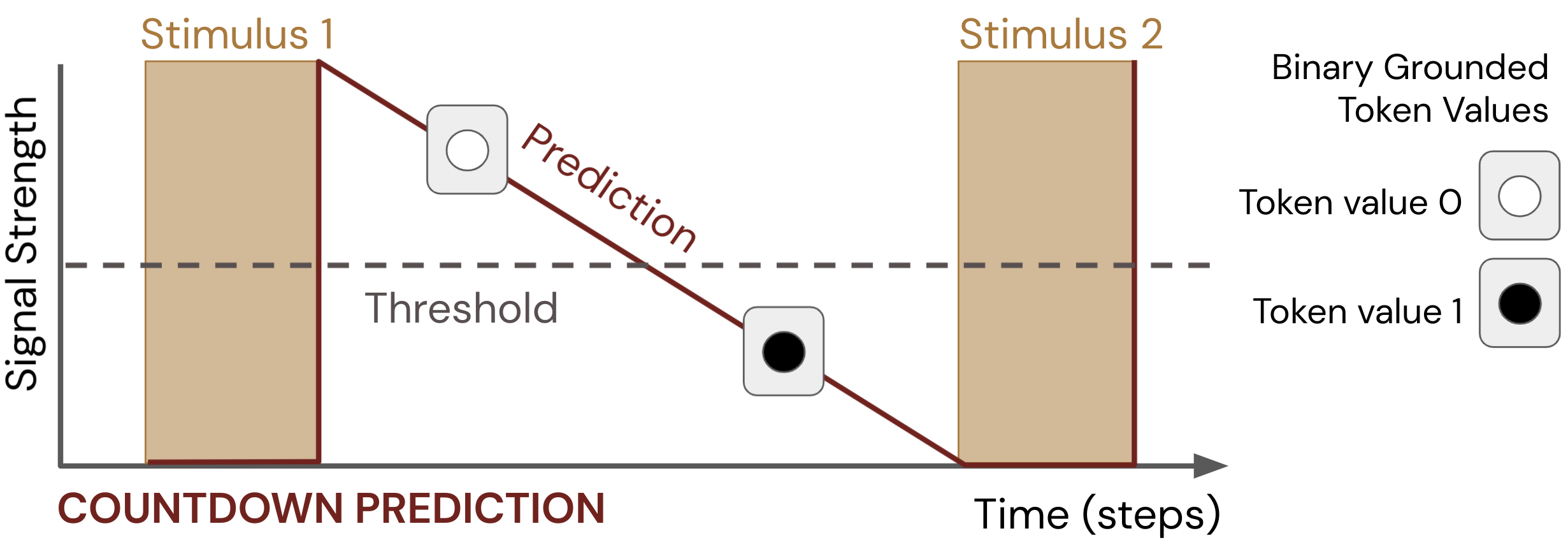}\\
(b)
\caption{{\bf Pavlovian signalling in schematic form}. Shown here are two different idealized predictions (red traces) that either (a) rise in advance of an impending stimulus or (b) decrease in a consistent way to forecast the time until the impending stimulus. Both predictions are shown with a threshold-based tokenization scheme for Pavlovian signalling: when the learned prediction magnitude relating to an impend stimulus crosses a threshold, the grounded Boolean token (i.e., the emitted signal) related to that prediction changes in sign.}
\label{fig:pred-scematic}
\end{figure}

As can be seen in Fig.~\ref{fig:pred-scematic}, we used a conventional signaling system based on signal strength and a threshold to assign meaning to the signal (i.e., to ground the signal). Depending on the informational content of the signal (i.e., token value), the signal has the potential to carry information regarding prediction of the occurrence of an event. For example in an accumulation-style prediction (Fig. \ref{fig:pred-scematic}a), if the value of a prediction is larger than a threshold, it contains information which predicts the occurrence of an event in the future. Conversely, in a countdown-style prediction \ref{fig:pred-scematic}b), if the value of a prediction is smaller than a threshold, the tokens in the signal contain information about occurrence of a future event. As can be seen in the signalling system used in the two cases in Fig.~\ref{fig:pred-scematic}, the grounding rule used to ground the tokens are different, however in both cases token value of 1 means occurrence of the next event is close, and token value of 0 means no event is expected to occur in the near future. 

As examples from the literature that provide case demonstrations of this approach, \cite{edwards2016} mapped machine-learned predictions to vibrotactile signalling tokens (scalar tokens in their case) intended for receipt for a human user of a robotic arm, \cite{pilarski2019} mapped machine-learned predictions about viable foraging locations to audio feedback to a human participant, and \cite{parker2019} mapped learned predictions of robot sensor activation to vibrotactile signalling tokens via a fixed threshold that was grounded in the values of real-world motor sensors. In all of these examples, a hard-coded mapping was created from learned predictions to emitted signalling tokens, where the tokens were defined in terms of the learning parameters of the predictive questions being learned by a machine and the intrinsic, sensory values being used to create the mapping process. The present paper provides a foundation for understanding the impact of such prediction and mapping choices on the efficacy of resulting agent-agent interactions. 

\section{Representing Time}
\label{sec:time}

We seek to establish in this work how different temporal processes and representational choices impact Pavlovian signalling between learning agents. Signalling between agents is tightly connected to the ability of agents to perceive and generate patterns that unfold in time; further, experimental approaches relevant to our study of agent-agent decision-making involve agents maintaining a sense of elapsed or anticipated time until past or future events. This section therefore summarizes current thinking from the literature on models and mechanisms for timekeeping in the animal brain, including specific neural models for interval and pattern timing with and without contextual input from other parts of the sensorium. Our survey of the neuroscience literature is followed by a brief overview of computational mechanisms for interval and pattern timing in the machine learning literature, with a focus on models that can be applied to setting of continual learning from a stream of changing sensorimotor experience.  These foundations provide a basis and inspiration for the methods and empirical comparisons that follow the remainder of our manuscript.

\subsection{Representations of Time in Animals}
\label{sec:time-animal}

As a foundation for the computational approaches we develop in the remainder of the manuscript, this subsection presents a survey of current perspectives on time as it is represented and processed in animal brains. Time is a multifaceted concept that continues to draw focused attention from philosophers, biological scientists, physicists, and many other academic communities (for an excellent presentation of current thinking on time from all the communities, please see \cite{buonomano2017}). It is no surprise that humans have thought a lot about time, how to measure it and how to mark its passing, and how to use it in making choices. At all levels, temporal patterns govern human and other animal's abilities to survive and thrive in their respective environments. This has led to a proliferation of biological mechanisms for addressing time, and more recently a proliferation of technological innovations to offload timekeeping to the environment around an organism to enhance both the resolution and scope of timekeeping---from measuring the movement of the sun and planets to capturing the motion through space of droplets in a water clock, from evaluating nuclear decay to measuring the minute oscillations of crystals or signal transmissions in connected networks, time keeping has proved a diverse and consuming pursuit for humans across the ages \citep{buonomano2017,andrewes2006,harrison1767,copernicus1543}.
   
In this work, we draw ideas heavily from the biological (specifically neuroscientific) implementations of and uses of time and timekeeping. As well described by \cite{buonomano2017}, animals and humans in particular have a set of common and distinct uses for time: humans process time to remember the past in order to predict the future, and use time to recognize and generate temporal patterns. Further, humans create and maintain a subjective perception of time, making the flow of time part of a their stream of experience; this perception plays a role in remembering past experience and simulating future experience. Humans have the capacity to represent time since events or stimuli, estimate time until a future events, scale and adapt their perception to changing event intervals, align their behaviour to rhythmic events, and importantly, use and change the environment to implement a measure of time external to themselves \citep{buonomano2017}. In terms of an organism's subjective or internal perception of time, experts have largely grouped capabilities into  retrospective and prospective timekeeping (estimating the time since an event or stimuls, and tracking or measuring the time until a future anticipated stimulus, respectively) and into capabilities that relate to different temporal spans (e.g., milliseconds, seconds, minutes, hours, days, and years) \citep{buonomano2017,eichenbaum2014,eichenbaum2017,paton2018}. 

Evidence for these capabilities in humans and other animals has been drawn from a rich history of animal learning and behavioural experimentation. In addition to recent work on timekeeping at the cellular level (c.f., \cite{buonomano2017}), there is a growing body of specific evidence as to how time and timekeeping is instantiated in the tissue of animal central nervous systems including in the cerebrum, in deep brain structures, and in hind brain structures such as the cerebellum. Early work by \cite{tolman1948cognitive} on cognitive maps, followed by new understanding on place cells and grid cells in animal brains and computational models of their brains by \cite{oKeefe1971}, \cite{hafting2005}, \cite{cueva2018}, \cite{banino2018}, and \cite{stachenfeld2014} showed that the brain is readily able to maintain and adapt a representation of an animals position, orientation, or relationship to the space around it. In a similar fashion, time cells have been identified in the animal brain that allow an organism to represent and adapt its relationship or association to events and stimulus that unfold in the flow of time \citep{macdonald2011}; researchers have identified temporal processing multiple cell populations (e.g., ramping cells, \cite{tsao2018}), in humans and well as other animals \citep{umbach2020}, and in multiple brain regions including the cerebellum,  hippocampus, and  striatum \citep{lusk2016}. 

Time and timekeeping as found in the brain has further been described in terms the key mechanisms or models suspected to be implemented in the tissue of the nervous system, with different mechanisms speculated to exist to deal with different scales or spans of time \citep{paton2018,eichenbaum2014,eichenbaum2017}\footnote{In particular, \cite{paton2018} note that mechanisms for spans of time between tens of millisesconds and tens of seconds remain less clear, and that this temporal processing span is very important as it, in their words, ``allows for the recognition and generation of complex temporal patterns that cannot be characterized by the duration of any one element.''}. As described by \cite{tsao2018} and \cite{paton2018}, mechanisms have been grouped roughly into those that are {\em explicit} or {\em dedicated} (that generate timestamps within neural tissue in a clock-like way) v.s. those that are {\em inherent} or {\em intrinsic} (that extract the representation of time from changes in the stream of an animal's sensorimotor experience), with current evidence largely supporting the inherent or intrinsic models of temporal processing. Further categorizations that have been applied to temporal processing in the brain include {\em interval timing} v.s. {\em  pattern timing}, for {\em sensory timing} vs {\em motor timing}, and {\em subsecond} v.s. {\em suprasecond timing} \citep{paton2018}. 

One dominant model to emerge from prior work in this area is the {\em pacemaker-accumulator model}, as reviewed by \cite{paton2018}. In this model, time-based oscillations or the presence of an oscillator is coupled with a downstream accumulator or integrator that collects and tallies the momentary inputs of the oscillator \citep{paton2018} (c.f., the {\em temporal context model} wherein oscillations/tors in other brain areas are integrated in the striatum to represent the passage of time \citep{eichenbaum2014}). Within a pacemaker-accumulator model, \cite{paton2018} note that it is also possible to provide the contribution of an oscillator instead using a background stochastic input that is then accumulated (accumulation of noise or other unrelated signals). Pacemaker-accumulator models retain their connection to seminal work by \cite{gibbon1977} on {\em scalar-expectancy theory}, wherein an animal was suggested to accumulate an estimate of elapsed time following some cue or event and compare it to the expected time until a future event as a basis for decision making.

Further distinctions have been made between {\em ramping models}, {\em population clocks}, and {\em oscillators} \citep{tsao2018,paton2018,eichenbaum2014}. Population clocks are considered a chain or collection of cells that sequentially fire, either in response to outside inputs or in an oscillatory fashion \citep{paton2018}. The {\em sparse population clock model}, for example, takes the form of a simple feed-forward or synfire chain wherein each neuron would only activate once in a repeatable sequence \citep{paton2018}; the {\em chaining model}, similarly, posits that an internal sequences of neurons in the hippocampus fire in a fixed pattern, possibly modulated or scaled by contextual events from other brain regions, in what has been called the {\em combined model} \citep{eichenbaum2014}. In contrast, ramping models postulate that time is represented in changes to the tonic firing rate of different cells or cell populations \cite{paton2018}. In addition to single oscillators, other alternatives have been suggested such as the {\em multiple oscillator model}, {\em striatal beat frequency model}, and others, where neurons are then dedicated to detecting the coincidence of the multiple oscillators at different frequencies \citep{paton2018}. 

Animal brains have been shown to be able to perform temporal perceptual learning, perform interval identification (detecting different length of gaps between events), re-scale timing in motor production, use temporal events as ``singposts'' to judge the duration until other future events, and utilize context to modulate and sculpt the internal representations of the flow of time \citep{paton2018,eichenbaum2014,tsao2018}. The similarity to animal spatial processing is noteworthy. As put by \cite{eichenbaum2017}, it may in fact be that that space and time are not different and special in the brain, and that the mechanisms involved simply represent two modalities with associative order and relationships, or, in other words that a ``common thread [in both space and time] is ‘association by proximity’'' \citep{eichenbaum2014}.

\subsection{Representations of Time in\\ Learning Machines}
 
In machine learning, though the majority of work ignores time via the independent and identically distributed (i.i.d.) random variables setting, there have been significant efforts in predicting and controlling partially observable domains where the world changes with time. Classically, in supervised learning and regression, all training and testing data are sampled randomly and we assume there are no temporal correlations between any two data points. However, there has also been substantial work on reinforcement learning, recurrent learning, and time-series forecasting, all of which make use of time in the specification of the solution's update mechanism and in the case of reinforcement learning in the specification of the target of learning.

In reinforcement learning, time plays a central role in both problem formulation and solution mechanisms. A reinforcement-learning agent maximizes the total reward observed over temporally correlated trajectories of experience; it adjusts its way of behaving to get more reward. To do so, most algorithms learn value functions which relate the long-term utility of a state to the states that follow it in time. As we will see later, these value functions are well thought of as multi-step predictive questions and that reinforcement learning methods can be used to represent predictive knowledge with temporal extent. 

The concept of state is a key assumption in the formulation value functions and Markov Decision Processes used to mathematically define reinforcement learning. The idea is that the state summarizes all previous interactions with the system: the next state and reward are dependent only on the previous state and action---that is the state is Markov. However, in many problems the agent does not have access to all the relevant information in order to obtain a Markov state. Perhaps some key state variables evolve according to some seasonal process not directly observable to the agent. The agent would do better by remembering part or all of its previous interactions with the world. Remembering all previous interactions is intractable and wasteful, whereas determining what to remember and how long is a challenging learning problem. This is the task of state construction.

The history of state construction in machine learning is as old as reinforcement learning itself. Early efforts studied computing summaries of the history of interactions with simple functions like tap delay lines, exponential averages and gamma function \citep{mozer1993neural}. These approaches, though general, rely on people to design the history function, whereas recurrent learning systems attempt to learn to summarize the history via gradient descent. The hidden layers Recurrent Neural Network (RNN) use recurrent connections to previous layers allowing the network to represent complex non-linear functions that have a temporal extent. For example, an RNN can represent represent a tap delay line memory of a long sequence of inputs. Updating the recurrent weights of the RNN in principle requires storing the activations of the hidden until from the beginning of time (BPTT), and thus the update is truncated after $T$ steps in the truncated BPTT time algorithm. It will well known that such truncation is not appropriate for modelling long temporal dependencies. Several algorithms have been proposed to improve this aspect via better state initialization and improved optimizations \citep{nath2019training}. Synthetic gradients and BP($\lambda$) \citep{jaderberg2017decoupled} estimate the gradient by bootstrapping gradient estimates from other layers. 

We can avoid storing all previous activations using a recursive update rule called Real Time Recurrent Learning (RTRL). This approach requires quadratic computation and thus much research has focused on approximations of RTRL such as Unbiased Online Recurrent Optimization (UORO) \citep{tallec2017unbiased}. There are many variants of RNN that explore different architectures including learned gating mechansims on the hidden state, such as LSTMs \citep{hochreiter1997long}, phased LSTMs \citep{neil2016phased}, and clockwork RRNs \citep{koutnik2014}. These architectural variants still require computing or approximating the gradient back through time in some way.            

Finally, sparse attention learning mechanisms avoid BPTT and have generated considerable excitement in Natural Language Processing and sequence modelling \citep{vaswani2017attention}. Current methods require very large models that store a history of the hidden state activation's for many previous time-steps. Furthermore, learned attention models have not been explored in the fully online setting we explore in this paper \citep{parisotto2020stabilizing}.

As noted non-exhaustively in this section, there are many ways time manifests in different machine learning architectures. We sample from and use these approaches, both computational and neuroscientific, as foundations for the different parameters of the experimental domain that follows and the representational comparisons we make in the experimental sections that follows.

\section{The Frost Hollow Domain}
\label{sec:env-description}

In the present section we describe the formulation and mechanics of the principle domain we use for the different empirical studies presented in this manuscript. We extend this domain from standard temporal decision-making benchmarks in the animal learning literature.

For many years, researchers in machine learning have used simulation problems inspired by animal learning to better understand the capabilities of artificial agents. Drawing on this rich history of animal and machine learning experimentation, and in particular recent work by \cite{rafiee2021} on trace conditioning and a suite of problems inspired by experiments in animal learning, we now introduce a domain for our study called {\em the Frost Hollow}. 

The Frost Hollow environment is a partially observable domain designed to evaluate our agents' ability to predict and generate signals relating to events that unfold over time---in this case, collecting heat from sunlight and avoiding heat loss due to the hazardous winter wind. The environment places a player in a forest glade in winter. A warm sunbeam shines into the center
of the clearing, and if an player stands in the sunbeam they accumulate heat points.
If the player collects a sufficient amount of heat while standing in the sunbeam,
they exchange it for a point of reward. However, an intermittent cold wind blows through
the clearing on a somewhat regular schedule, and if the player is caught exposed,
they lose any accumulated heat. The player can hide from the wind by taking shelter closer
to the surrounding trees, thus retaining any accumulated heat until it is safe to
return to the sunbeam. The trees are several steps away from the sunbeam, and so
in order to reach shelter in time, the player must start moving before the wind starts.
In each episode, the player's goal is to gain as much reward as possible by
collecting sufficient heat several times over. 

Within each episode the wind blows 
somewhat regularly according to a set of parameters, such that it is possible for the agent to learn from
past observations of the wind to try to predict when it will happen next. Thus, the core challenge of the Frost Hollow environment is predicting events over time:
using timing features about the past (e.g., how long since the previous wind gust) to make
useful predictions about the future (e.g., how long until the next gust) to guide behaviour. Predictions
must be both accurate and determined far enough in advance to be actionable by the player.

\begin{figure*}[!th]
\centering
\includegraphics[height=1.8in]{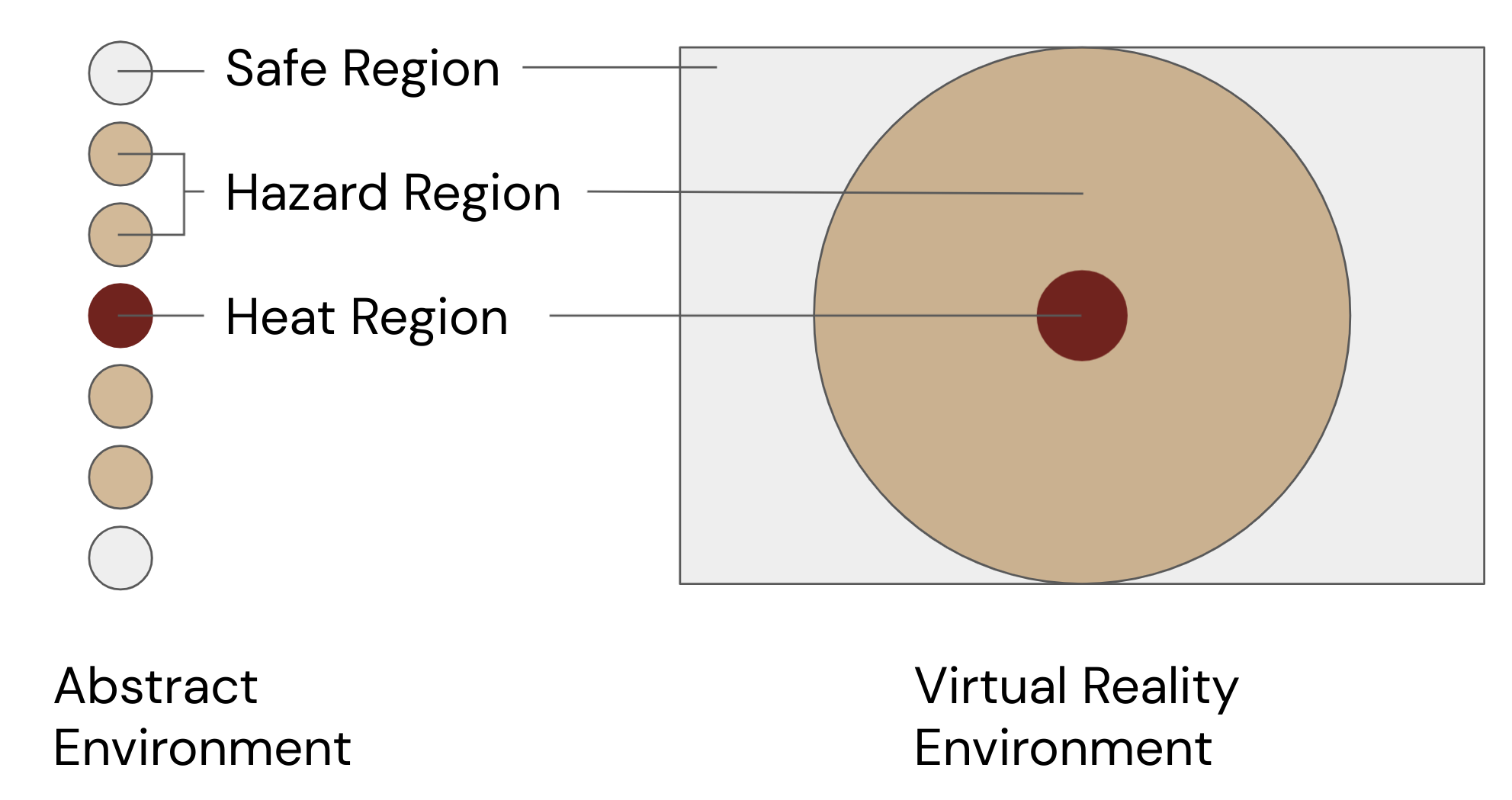}
\includegraphics[height=1.8in]{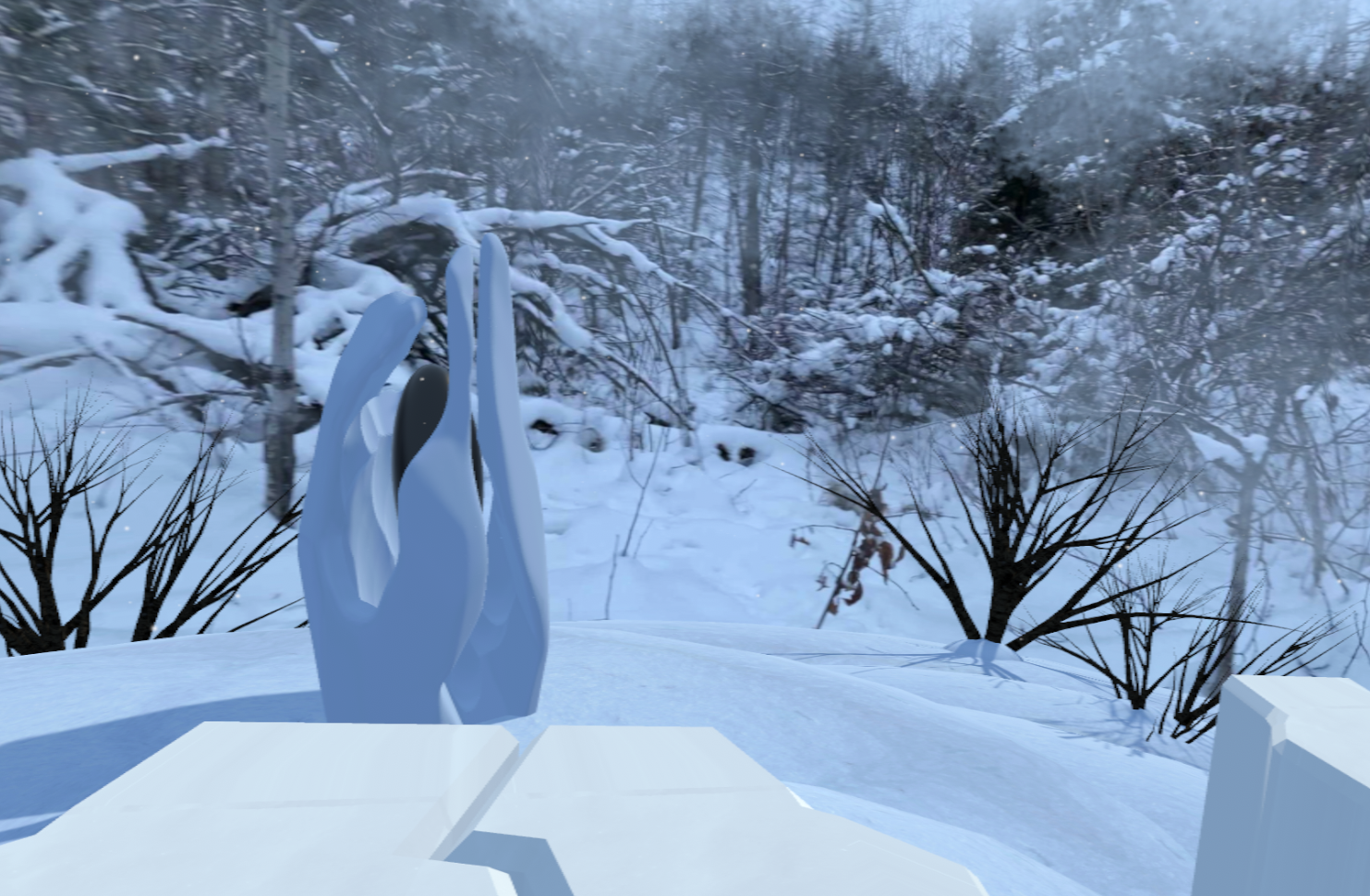}\\
{ (a) \hfil (b)}\\
\caption{{\bf Schematic showing the relationship between abstract and virtual reality variants of the Frost Hollow domain}, along with the relative location of heat regions, hazard regions, and safe regions in the Frost Hollow domain,
(a) shown in terms of the spatial layout of the abstract (linear walk) and virtual reality implementations of the domain, and
(b) alongside an example of the first-person view of the virtual reality environment perceivable by a human participant.
}
\label{fig:frosthollow-environments}
\end{figure*}

For our experiments, we have used this general description to create two environmental variants.
The first variant is an abstract setting suitable for simple machine reinforcement learning agents, in 
which there is a linear chain of locations representing  a cross section through the center
of the glade, with shelter at either end and a sunbeam in the middle. On each timestep the
agent can move left, right, or stand still. 
The second variant is a first-person virtual reality environment for human participants.
In this setting the forest glade is rendered in detail, and human subjects
physically walk around a 3m x 2m space to move between shelter and the sunbeam
while avoiding the wind.
Figure \ref{fig:frosthollow-environments} provides an illustration of these environmental variants.

\begin{figure*}[!th]
\centering
\includegraphics[width=5in]{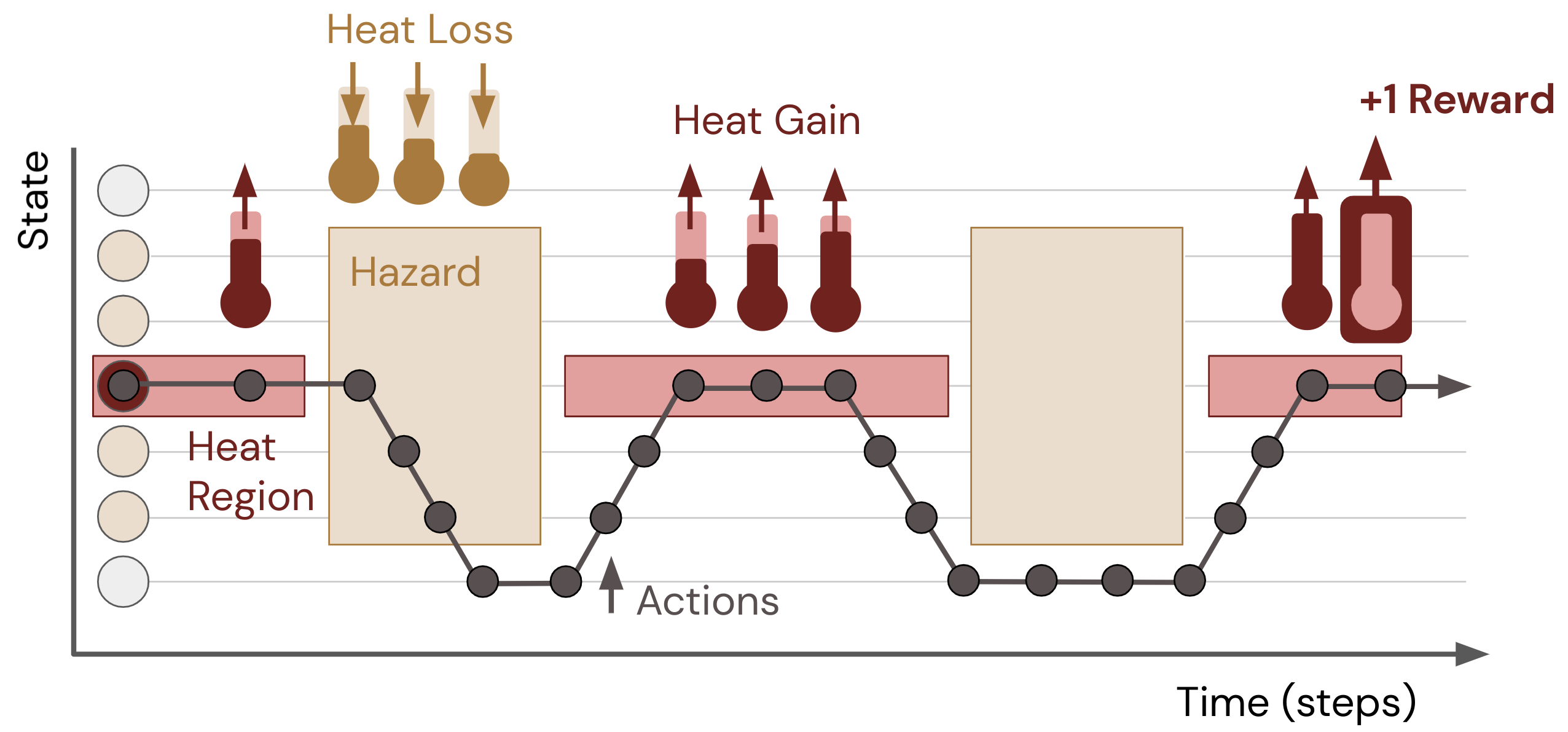}
\caption{{\bf Schematic representation of the abstract Frost Hollow domain} and one example trajectory showing how an agent might interact with it over time.}
\label{fig:abstract-frosthollow}
\end{figure*}

Figure~\ref{fig:abstract-frosthollow} illustrates how the player (i.e., the 
agent or human participant) must move through time and space to earn reward.
When the agent stands in the central sunbeam, they accumulate heat points.
Intermittently, a wind hazard occurs, which removes any heat points if the
player is standing anywhere in the hazard or heat regions. By predicting the
onset of the wind and moving into the safe regions on the edge of the environment,
the agent or human can retain their heat, then return to the center to collect
enough to cash it in for one point of reward.

\subsection{Task Description, Definitions,\\and Parameters}
\label{sec:env-constants}

We will now give a more formal description of our abstract and
virtual reality environments, and describe which parameters are
held constant or varied throughout our experiments. This section
will describe the spatial layout, reward mechanisms, and
observation and action streams of the environment, while the next
section will describe how we vary the timing mechanism of hazards
across our experiments. 

The player's goal is to maximize the amount of \textit{reward} that they
earn in each episode. One point of reward is gained when the player
accumulates a number of \textit{heat points} equal to a configurable
\textit{heat capacity} (e.g., a capacity of 6 heat points), 
after which their heat points are reset to zero. The heat capacity 
is held constant within the player's interaction with the environment, 
but we can configure it for each experiment to adjust the difficulty 
of the task.

Each environment's map is divided into three regions. In the center
is the \textit{heat region}, where the player accumulates heat at
a constant rate: $0.5$ per timestep in the abstract environment,
and $0.1875$ per second in the virtual reality environment. 
The \textit{hazard region} is a larger area that contains the heat region.
According to a possibly stochastic schedule that we can configure,
a cold gust of wind, which we call a \textit{hazard}, 
will blow through the hazard region for some interval of time.
If the player is standing in the hazard region (including the heat region,
which it contains), they lose all of their heat points, but do not
suffer any other penalty such as a negative reward. Finally, the remaining
locations in the environment are the \textit{safe region}, where
the player is protected from the hazard.

In all of our experiments, the dimensions and layouts of the environments
are constant as shown in Figure~\ref{fig:frosthollow-environments}. The
abstract environment has 7 locations, with the regions labelled
in the figure. The virtual reality environment has a 3m by 2m space
in which the player can move, divided into a 0.165 meter radius 
heat region at the center, surrounded by a 1 meter radius hazard region, and
a small safe area outside of that at either edge of the space.

One key difference between our abstract and virtual reality environments
is the passage of time. In the abstract environment, time advances in large
and discrete increments, and the environment pauses until the player submits
their action on each timestep. In the virtual reality environment, time advances
at 120 frames/second in real time. Using a basic assumption that a human
can safely move about 1 meter/second in virtual reality, we designed our
7 location abstract domain to represent about 0.5 seconds of real time
per timestep, and chose the region dimensions, heat capacity and rate of gain,
and hazard intervals to require a similar interaction by the players in
each environment. In each environment, the player receives the following observations
and can take the following actions, described here in overview form for reference and again in more detail in the respective empirical sections relating to the abstract and virtual reality environments.

{\bf Abstract environment:} Observations for the player include the current {\em position}, a one-hot vector with one entry per location, {\em hazard} presence or absence via a boolean indicating if the wind is currently blowing, and {\em heat}, a scalar in [0, heat capacity] indicating how much heat the player has collected. Actions for the player take the form of \textit{movement}: an integer in [-1, 0, 1], that moves the agent deterministically to an adjacent location (-1, 1) or allows the agent to stay in the same location (0).

{\bf Virtual reality environment:} Observations for the player include {\em vision} via A 1440x1600 pixel display running at 120 frames/second, presented to the human participant using the Valve Index virtual reality headset. The visual scene conveys the position, hazard, and heat information to the human participant. The player also receives input via {\em controller vibration}; by turning the controller vibration on or off, the player can perceive hazards and other input sources (e.g., signals from another agent). Actions for the player include {\em movement}, here effected by the human participant walking and shifting using their body, where the coordinates of their headset is used to determine their position in the virtual reality environment. Using their body movement, the player can also {\em convert heat} while standing in the heat region by raising their hand to trigger the exchange of a full heat capacity into a point of reward. Additional detail on the virtual reality environment are provided in Sec. \ref{sec:vr-experiments}.

\subsection{Experimental Conditions: Normal,\\Random,and Drift Intervals}
\label{sec:env-drift-conditions}

\begin{figure}[!t]
\centering
\includegraphics[width=2.8in]{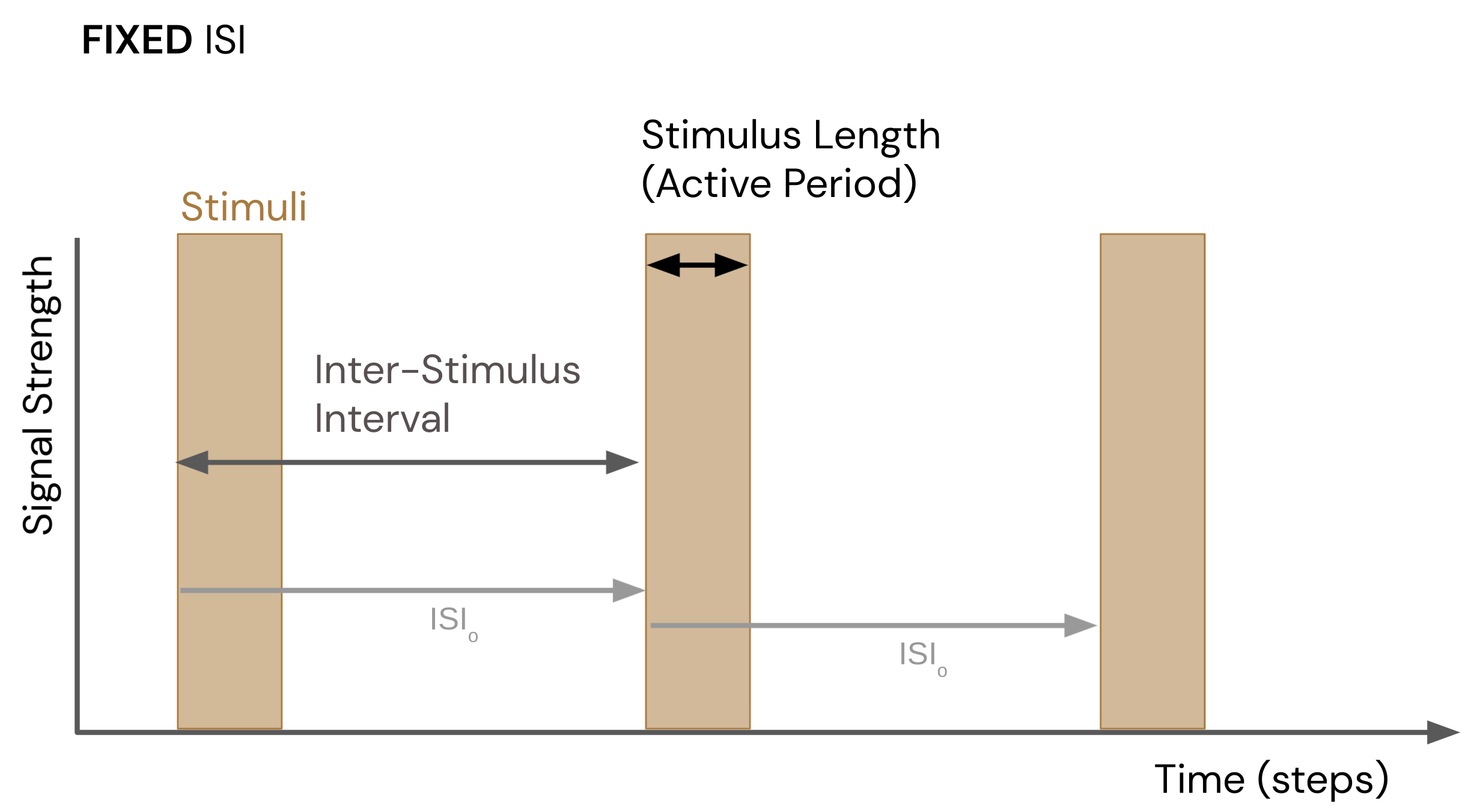}\\
(a)\\
\includegraphics[width=2.8in]{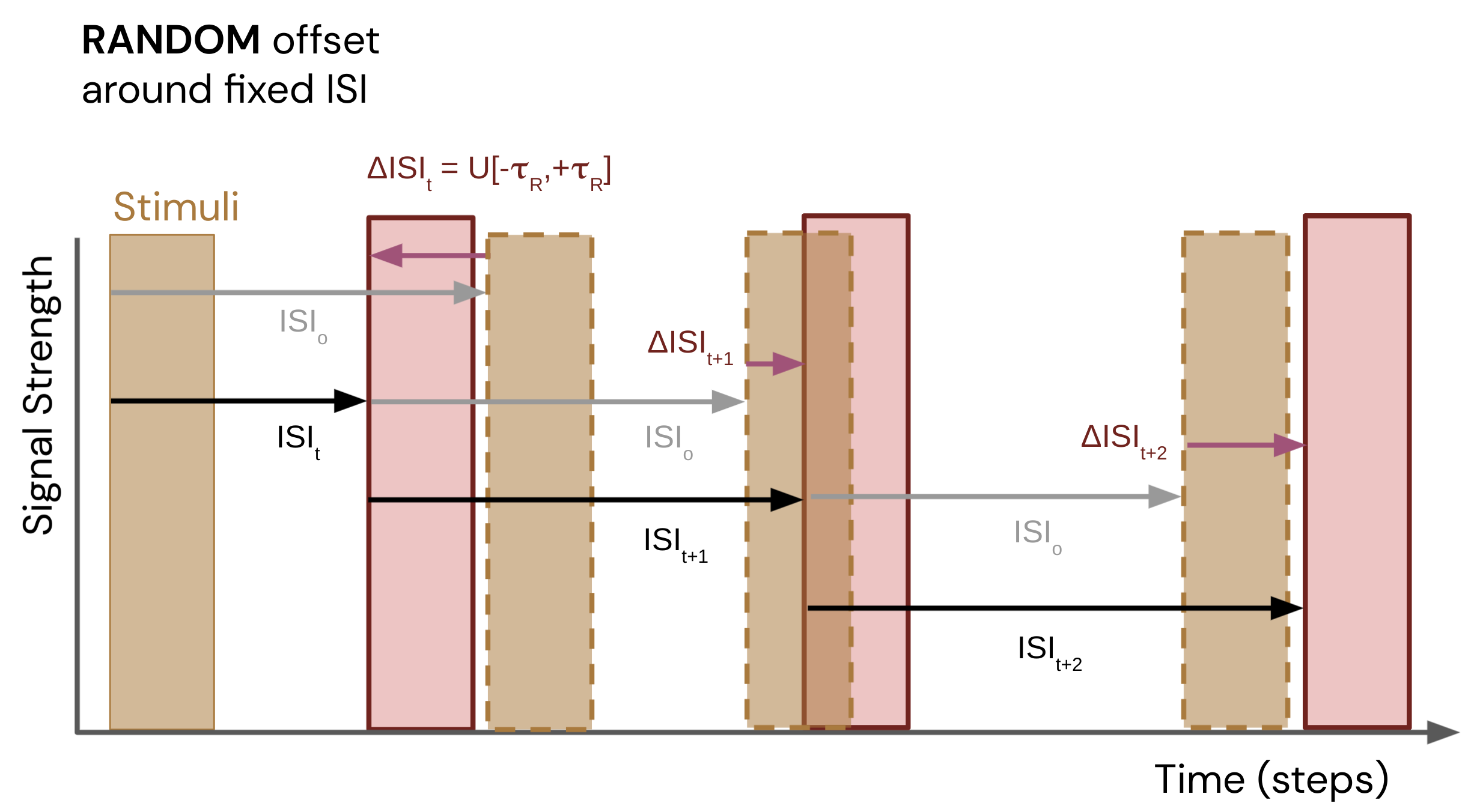}\\
(b)\\
\includegraphics[width=2.8in]{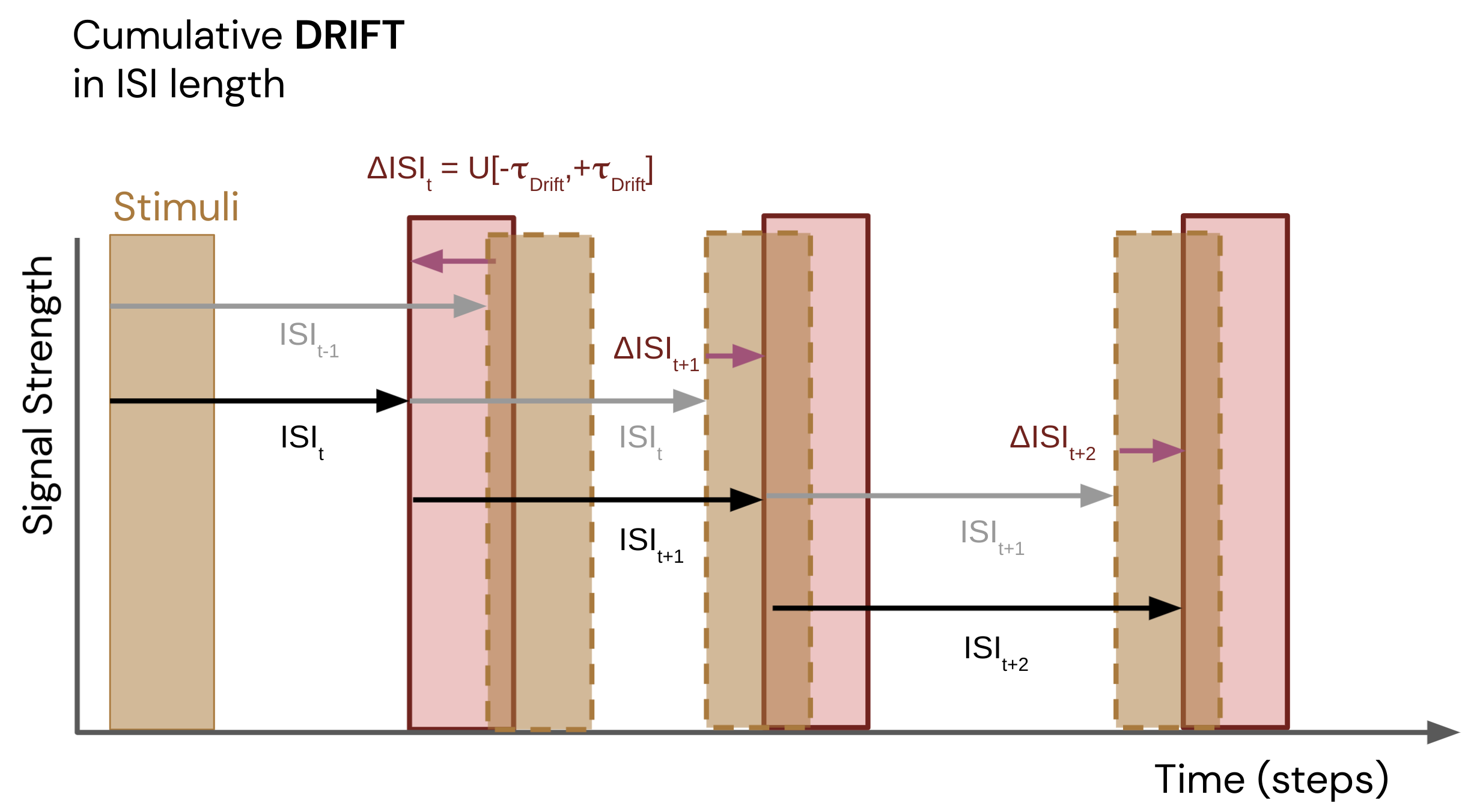}\\
(c)
\caption{{\bf Depiction of the three experimental conditions used in this work}: the {\em fixed condition}, wherein the inter-stimulus interval (ISI) remains constant, the {\em random condition} wherein the ISI changes randomly around a fixed interval each time it occurs, and the {\em drift condition} wherein the ISI undergoes cumulative changes in duration each time it occurs.}
\label{fig:fixed-random-drift-schematic}
\end{figure}

Throughout the experimental results which we will present, a 
key element that we varied was the schedule and stochastic 
nature of the wind hazard. The hazard is described by two key
variables: the \textit{inter-stimulus interval} which is the
time between hazards, as measured from the start of one hazard to the start of the following hazard, and the \textit{stimulus length} which is
the duration of the hazard; see Figure~\ref{fig:fixed-random-drift-schematic}a for an example.

With this framework, we considered three types of hazard, illustrated
in Figures \ref{fig:fixed-random-drift-schematic}a-c. In each
experiment, the type and parameters for the hazard were held constant during the player's interaction with the environment, and only varied across independent experiments. These conditions are as follows:

{\bf Fixed:} The hazard is parameterized with fixed values for
    the inter-stimulus interval and stimulus length (e.g., 8 and 2
    steps respectively in the abstract environment).
    
{\bf Random:} The hazard has a fixed stimulus length (e.g., 2 steps
    in the abstract environment), but the inter-stimulus interval
    is chosen at uniform random from a range [x-m, x+n] (e.g., a total ISI range of
    $[8, 13]$ steps in the abstract environment).
    
{\bf Drift:} The hazard has a fixed stimulus length and
    initial inter-stimulus interval (e.g., 2 and 8 steps in the abstract
    environment, as in the Fixed condition), but the inter-stimulus
    interval varies permanently by $\pm n$, within a minimum and
    maximum range, before each hazard (e.g., by $\pm 1$ steps within the bounds
    [8, 13] steps in the abstract environment).

\section{Methods: Predictions,\\Representations, and Tokens}
\label{sec:methods}

Having established a basis in the neuroscience and computing literature, and introduced a domain expressly constructed for studying temporal decision making, we now discuss the specific computational approaches we will use in our empirical comparisons. In this section we describe nexting with generalized value functions, present a selection of linear representations to use in nexting, and present a concrete specification for turning predictions into tokens for Pavlovian signalling between agents.  

\subsection{Nexting with General Value Functions}

We now turn to the specific computational framing we use for predictions and signals in this work. We herein model an agent's predictions about the world using General Value Functions (GVFs) which are value functions applied to non-reward based targets \citep{sutton2011}. A GVF formally specifies a predictive question, which can be understood informally as: {\em what will be the total accumulation of some signal of interest, if I follow some policy until termination?} A GVF is a value function where the target is the discounted sum of some cumulant $C_{t+1} \in \mathbb{R}$, that would be observed if the agent followed policy $\pi(A_t|S_t) \doteq Pr(A_t = a| S_t = s)$---a {\em what would happen if} form of question. The elements of the sum are weighted by earlier discounts $\gamma_t \in \mathbb{R}$. The discount becomes zero if a termination event occurs, and is typically less than one otherwise corresponding to the horizon of the predictive question. Taken together, we can  specify a predictive question by defined $C, \gamma$, and $\pi$. We first define the future return,
\begin{equation}G_t \equiv \sum_{k=0}^\infty \Bigg(\prod_{j=1}^k\gamma_{t+j}\Bigg)C_{t+k+1},\label{eq:gvf}\end{equation}
where the question is then defined as
\begin{equation}v(s) \equiv \mathbb{E}_\pi \{G_t | S_t = s\}.\label{eq:gvf_question}\end{equation}

The agent must learn answers to the GVF question to obtain knowledge of the world; that is, approximate $v$ from data. Given a batch of data, we could compute the right-hand side of Eq. \ref{eq:gvf} directly. In practice, the agent will observe a stream of states, actions, cumulants, and terminations as it interacts with the world. In this online setting, we can approximate $v$ in each state with a parametric function updated via temporal difference learning. Let $x \in \mathbb{R}^d$ be features summarizing the current state $x_t \equiv x(S_t)$, perhaps a state aggregation or a collection of radial basis function outputs. We define the prediction to be $V_t \equiv w_t ^\intercal x_t$, where $w_t \in \mathbb{R}^d$ and $V_t \approx v(S_t)$. Although more complex methods are possible, we follow prior work~\citep{sutton2011,modayil2014multi} and use the TD($\lambda$) algorithm to update $w_t$ on each timestep:  
\begin{align*}
e_t &\gets e_{t-1} + x_t\\
\delta_t &\gets C_{t+1} + \gamma(x_{t+1}) w_t ^\intercal x_{t+1} -  w_t ^\intercal x_t\\
w_{t+1} &\gets w_t + \alpha \delta_t e_t\\ 
e_{t} &\gets \gamma(x_{t+1}) \lambda e_{t},
\end{align*}

where $\alpha$ is a scalar learning rate parameter and $e \in \mathbb{R}^d$ is exponentially decaying memory of previous feature activations. This approach has been shown to learn accurate approximations of GVF questions in a variety of settings and is computationally frugal---requiring computation and memory linear in the number of features $d$---and is thus ideal for our problem setting of interest.  

\subsection{Temporal Representations}

\begin{figure}[!th]
\centering
\includegraphics[width=2.8in]{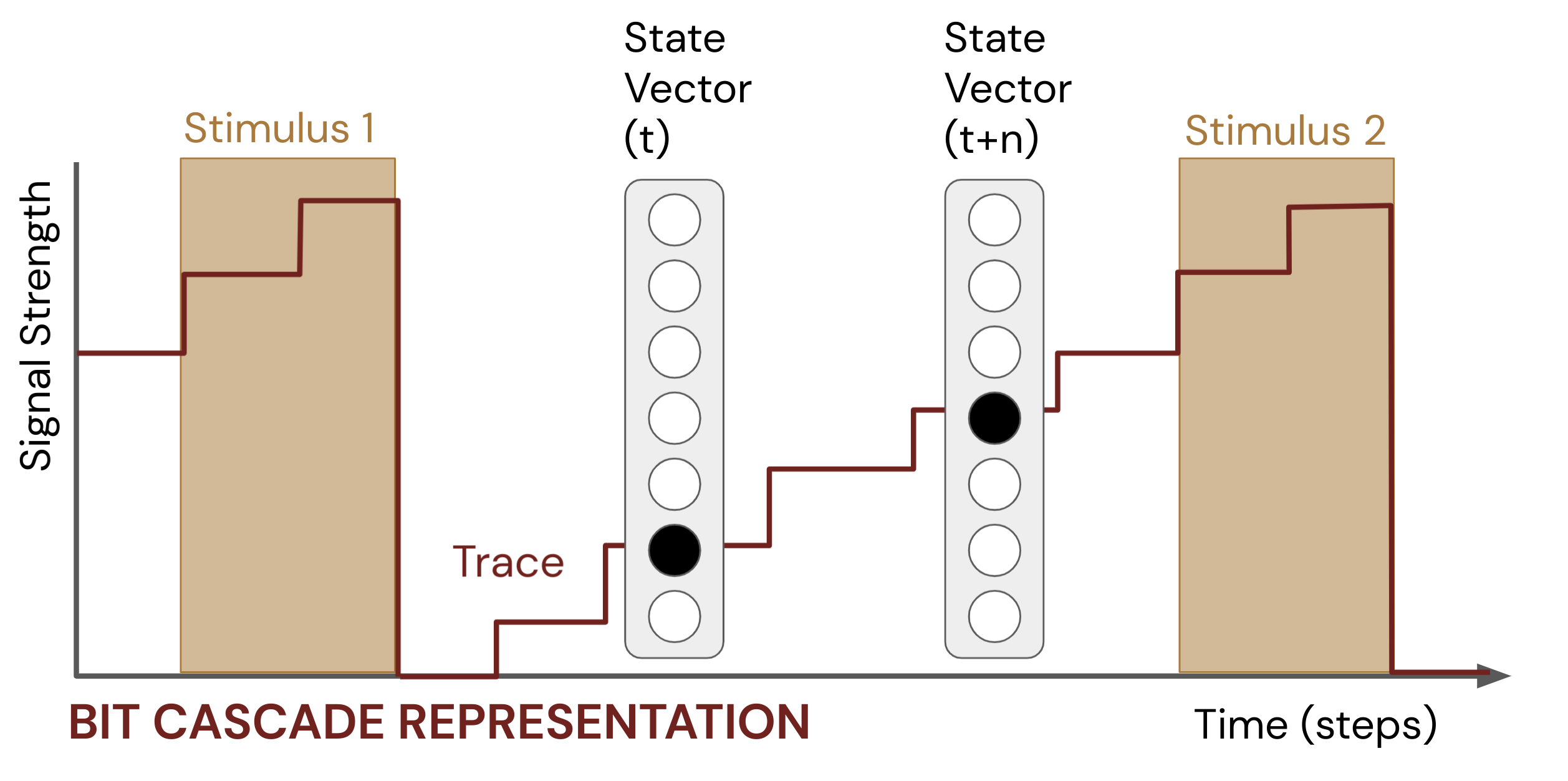}\\
(a)\\
\includegraphics[width=2.8in]{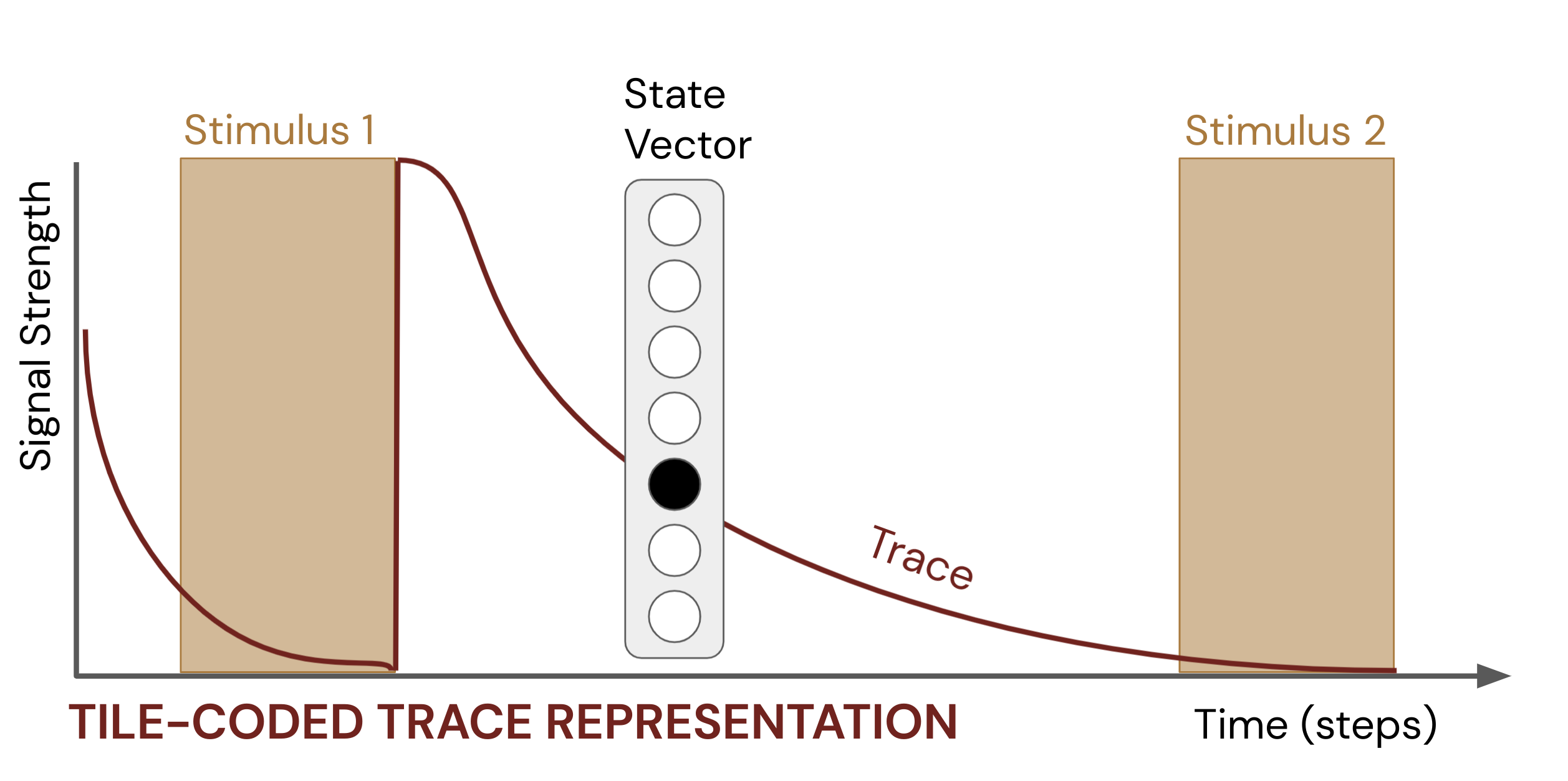}\\
(b)\\
\includegraphics[width=2.8in]{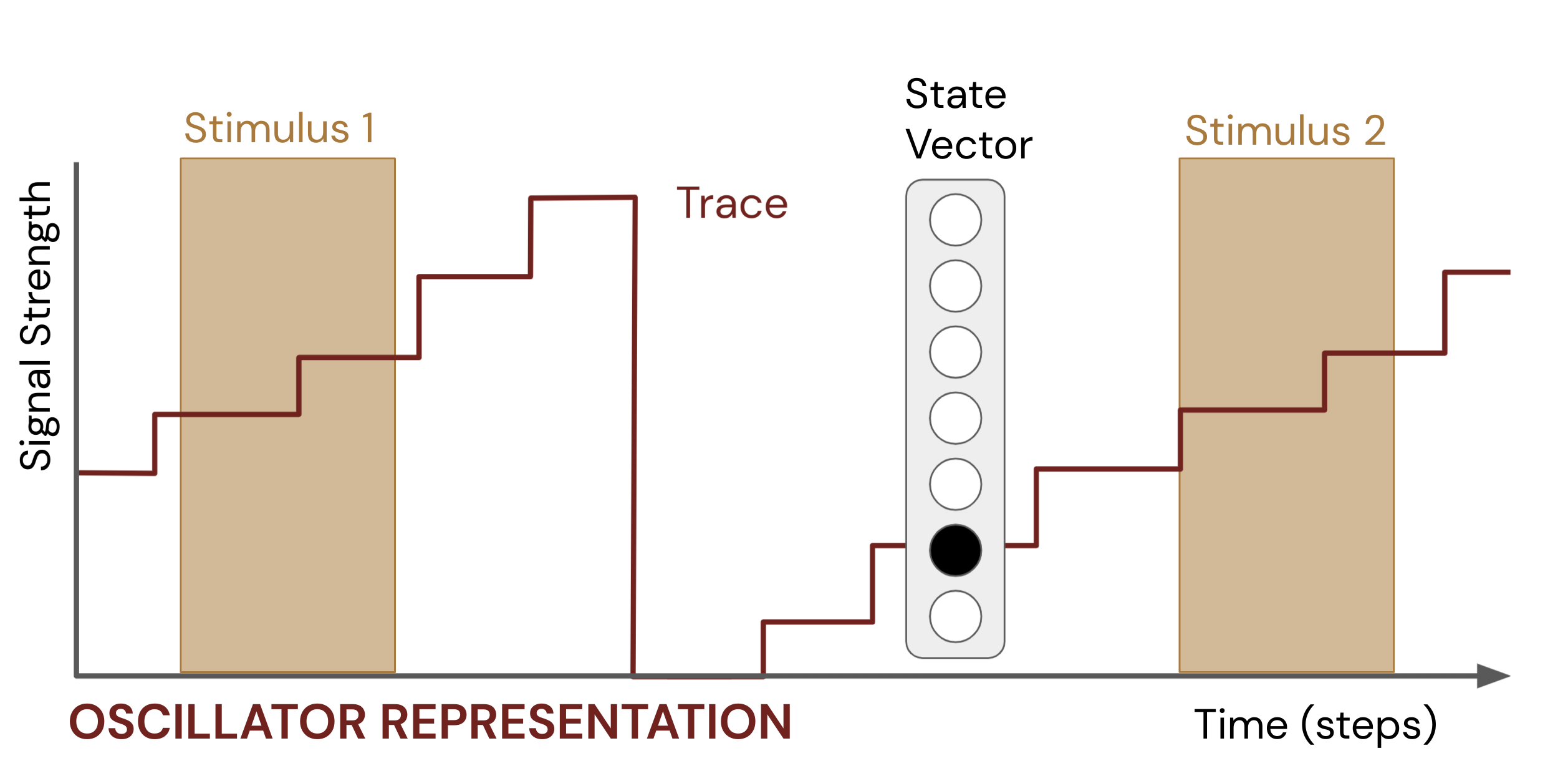}\\
(c)\\
\includegraphics[width=2.8in]{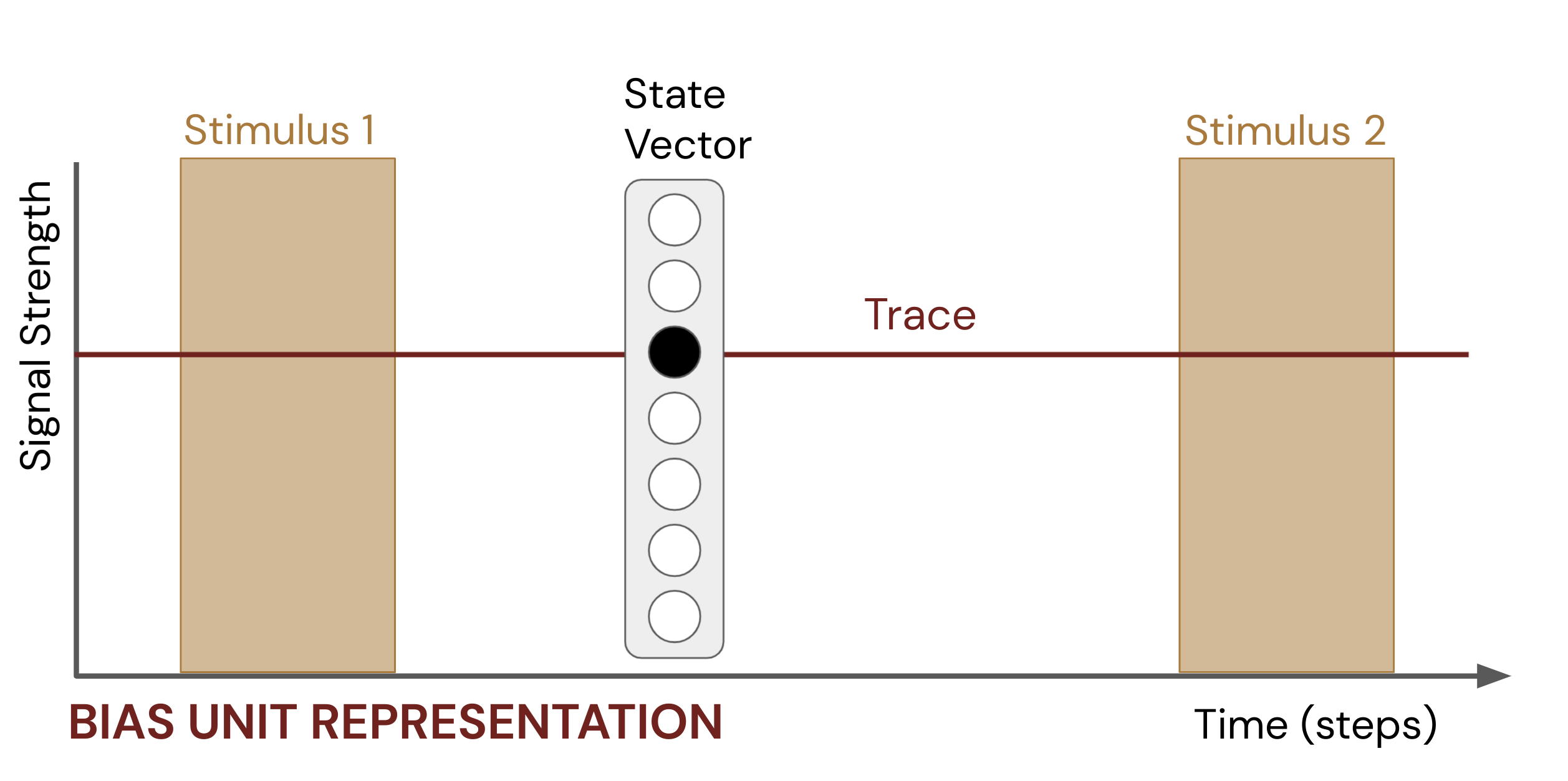}\\
(d)
\caption{{\bf Different temporal representations explicitly considered in this work} and derived from known activity models in the central nervous system. These include (a) a {\em bit cascade} that resets on the presentation of a given stimulus or pattern, and steps or progresses in a roughly linear way from state to state, b) a {\em tile-coded trace} that resets on presentation of stimulus and decays from an initial starting value, c) an {\em oscillator} mechanism that acts as the bit cascade but resets upon a given periodicity and not upon the presentation of stimulus, and d) a {\em bias unit} that retains a constant value across time. All representations in this work were considered to be coded or aliased into discrete state units.}
\label{fig:rep-schematic}
\end{figure}

Agent-agent temporal decision-making tasks are the key focus of the present work. Making predictions about events that unfold over time requires a sufficiently informative representation, $x_t$. We therefore implement a subset of biologically and computationally motivated temporal representations as a basis for GVF learning (sampled from present literature as surveyed in Sec. \ref{sec:time}), such that we can understand the sensitivity (or lack thereof) of prediction learning speed and agent performance to temporal representation.

We now describe the way our selected representations were concretely realised, drawing heavily on the suggested mechanics of population clocks, ramping models, and context-sensitive pacemaker-accumulator models surveyed in Sec. \ref{sec:time-animal} above. Temporal representations included in this study all specifically take the form of a parameterized one-hot vector---a Boolean vector of fixed length where all elements are of value zero except a single element that is one. The length of the vector, the change in position within the vector of the active element, and the boundary conditions vary with each treatment, but each is a special case of this unified representation. All of the base temporal one-hot vectors in this work also had a further bit appended to them, the presence representation (c.f., \cite{rafiee2021})---a Boolean element that represents the presence or absence of a given stimulus, in this case the Frost Hollow wind hazard.

The simplest temporal representation included in the study, the {\em bias unit representation}, was implemented as a vector with one constant feature concatenated with the presence representation $\mathbb{B}^1\frown\mathbb{B}^1$. This corresponds to a length-1 one-hot vector, with the active index advancing through the vector at a rate of one element each time step, wrapping at the boundary (i.e., when the index of the active element exceeds the vector length, the index of the active feature is reset to the beginning of the vector). With the added presence representation, the bias unit representation thus was a vector of length two, with between one and two active bits.

The implementation of the {\em oscillator representation} was an extension of the bias representation, here with the length of the vector is extended to match the number of time steps $n$ between the falling edge of hazards in each fixed-condition environment $\mathbb{B}^n\frown\mathbb{B}^1$. Following on the general form of population clocks from the animal brain, as noted in Sec. \ref{sec:time-animal}, our implementation of the {\em bit cascade representation} was similar to that of the oscillator representation, but with one change: in the bit cascade representation, the active element of the vector was reset to the beginning of the vector upon the transition of the presence representation from one to zero (the step the hazard ends). The maximum length of the vector was chosen to exceed the maximum period of the hazard for each condition. In other words, the oscillator and the bit cascade differ in that {\em the oscillator representation advances independent of any stimulus in the environment, while changes in the bit cascade representation are tied to presence representation.}%

In each of the above temporal representations, the active element advances through the vector at a rate of one element per time step. Most closely aligning to the changes in tonic firing rate of ramping models from temporal representations in the brain (c.f., Sec. \ref{sec:time-animal}), the implementation of the {\em tile-coded trace representation} is an extension of the bit cascade with advancement rates of the exponential form \(e^{-at}\), with \(a < 1.0\) representing the exponential decay constant, and where $t$ is the number of time steps since the last reset of the active element to the beginning of the vector. At t=0, \(e^{-at} = 1.0\), and the function decays towards 0 as t approaches infinity. Setting \(x = (1.0 - e^{-at})\), and clipping x to the range [0.0, 1.0), the feature index was found by multiplying x by the length of the representation vector.

\afterpage{\clearpage}

\subsection{Predictions and Tokens of Interest:\\Counting Down and Ramping Up}
\label{sec:gvf-defs}

Building on the idea of prospective and retrospective timing presented above, for this work we consider two specific types of predictive nexting questions: predictions about the onset of an impending signal or stimulus in terms of the expected future accumulation of that signal (a {\em rising} prediction about a future event), and a prediction of the expected time remaining until a future signal or stimulus pattern (a {\em falling} prediction or learned countdown timer until an event will occur). Both of these predictive questions of interest can then be specified in terms of the three GVF question parameters---the signal of interest is specified as the cumulant $C$, the timescale of interest as specified by $\gamma$, and the policy $\pi$ of interest---and use as their foundation the difference temporal representations described above and shown in Fig. \ref{fig:rep-schematic}.

First, for the case of a continual accumulation of an observed stimulus, we can identify GVF question parameters as follows, in what we will refer to as a fixed-timescale question or prediction, or more informally, as an {\em accumulation question}. For accumulation GVF questions used in this work, the cumulant takes the value of a specific stimulus, and the gamma-discounted sum of the cumulants gives lower emphasis on stimuli farther in the future:

\begin{align}
C_t &= 
\begin{cases}
    1.0,& \text{if } \text{stimulus\  present}\\
    0.0, & \text{otherwise}
\end{cases}\\
\gamma_t &= 0.9\\
\pi &= \text{on policy}
\end{align}

Second, we consider a learned expectation of the steps until as an accumulation of the steps until an observed stimulus, in what we hereafter refer to as state-conditional question or more informally based on our use in this work, a {\em countdown question}. In our countdown GVF formulation, a cumulant of 1.0 is sampled on every step, with the question termination value $\gamma$ now depending on stimulus (e.g., hazard) state as follows:

\begin{align}
C_t &= 1.0\\
\gamma_t &= 
\begin{cases}
    0.0,& \text{if } \text{stimulus\  present}\\
    1.0, & \text{otherwise}
\end{cases}\\
\pi &= \text{on policy}
\end{align}

A schematic of these two prediction types can been seen in Fig. \ref{fig:pred-scematic}.  As we will now describe, the answers to these accumulation and state-conditional questions (predictions), as learned through processes of temporal-difference learning on a stream of experience, can be turned into tokens or further signals to support agent decision making. In other words, following from the idea of Pavlovian control, we can create a fixed mapping from learned predictions based on temporal representations to signals that might be used by another agent or part of an agent.   

With these questions as a basis, we can directly implement the process for Pavlovian signalling depicted in Fig. \ref{fig:pred-scematic} wherein predictions are mapped to Boolean tokens according to fixed thresholds. For example, in accumulation predictions, if the value of the prediction rises to exceed a fixed, scalar threshold, a token of 1 is emitted. If the prediction is less than or equal to the threshold, a token of 0 is emitted. Similarly, in the countdown prediction, if the value of the prediction decreases to be equal to or less than a threshold, a token of 1 is emitted (with the token being emitted as 0 otherwise).

\section{Exploration of Nexting and Pavlovian Signalling in Frost Hollow}
\label{sec:nexting-experiments}

We now turn to the first empirical section of this paper. As a first goal of this section, we explore the practical implications and qualitative differences of different representational choices on nexting-style GVF predictions made in the Frost Hollow domain. As a second goal, we then demonstrate the way these representational choices impact the way that tokens might be generated from these predictions---i.e., the way that representation may impact Pavlovian signalling. {\em This section's main contribution is therefore to build intuition for the reader} as to the way different representations of time operate, the way different GVF predictions unfold when forecasting temporal sequences of events, the way that Pavlovian signalling manifests and changes due to variation in both predictions, representations, and domain conditions. This intuition is important, as the specific predictions, representations, and Pavlovian signalling approaches described in this section will be used as the foundation for the remaining empirical studies presented in this paper.

One important aspect of the present section is that it is primarily concerned with the way predictions depend on time and the sequence of temporal events; we do not here address the impact of location or space in the generation of predictions, and specifically do not consider the case where control or policy learning by an agent depends on predictions. Control learning based on learned predictions will be the focus of the following section, so in this introductory set of experiments we study in detail those predictions, their form and variation, and the way representation impacts predictive ability.

\subsection{How do Different Representational\\Choices Impact Learned Predictions?}

\begin{figure*}[!th]
\centering
{\bf REPRESENTATION \hfil FIXED \hfil RANDOM \hfil DRIFT}\\
\ \\
\includegraphics[width=1.5in]{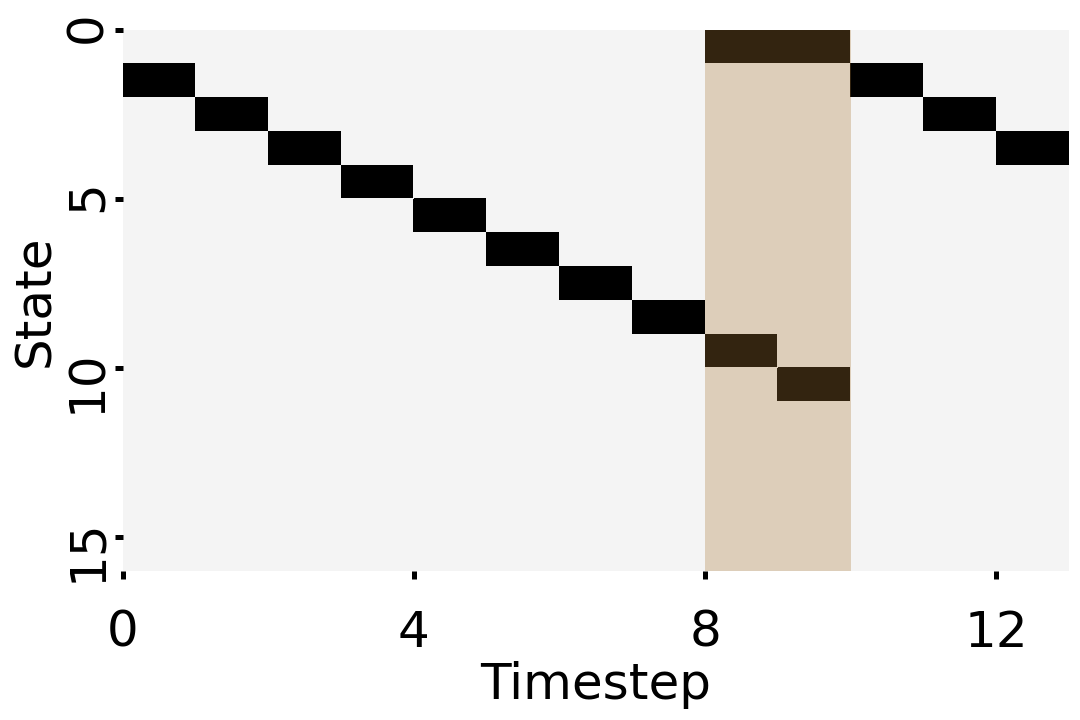}
\includegraphics[width=1.5in]{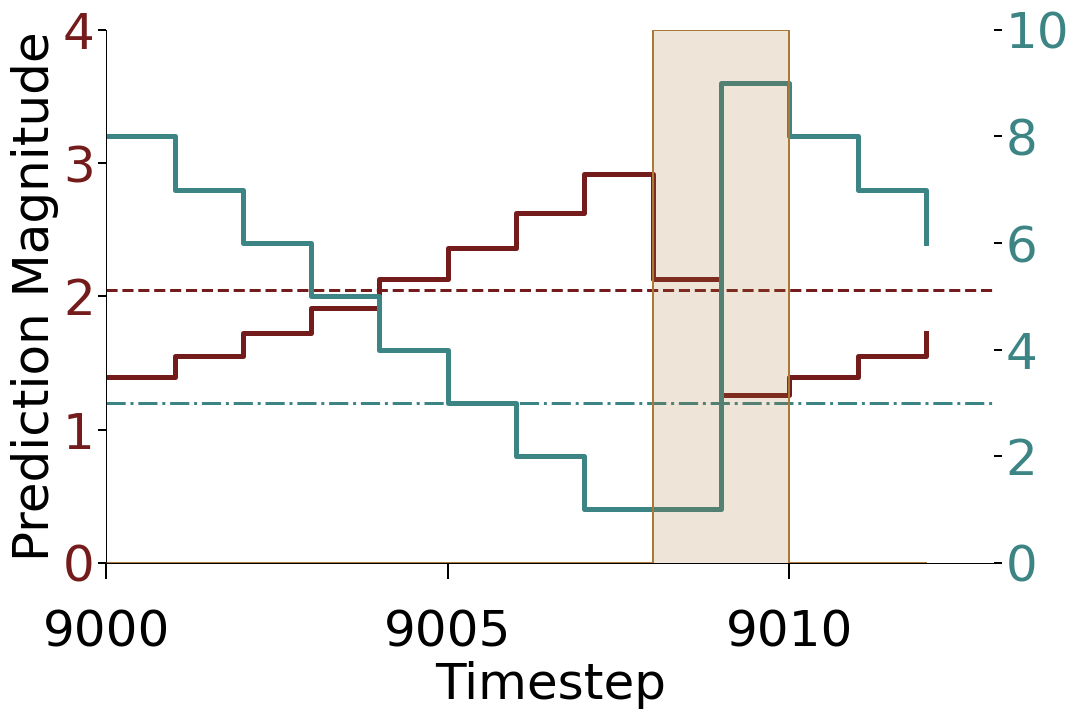}
\includegraphics[width=1.5in]{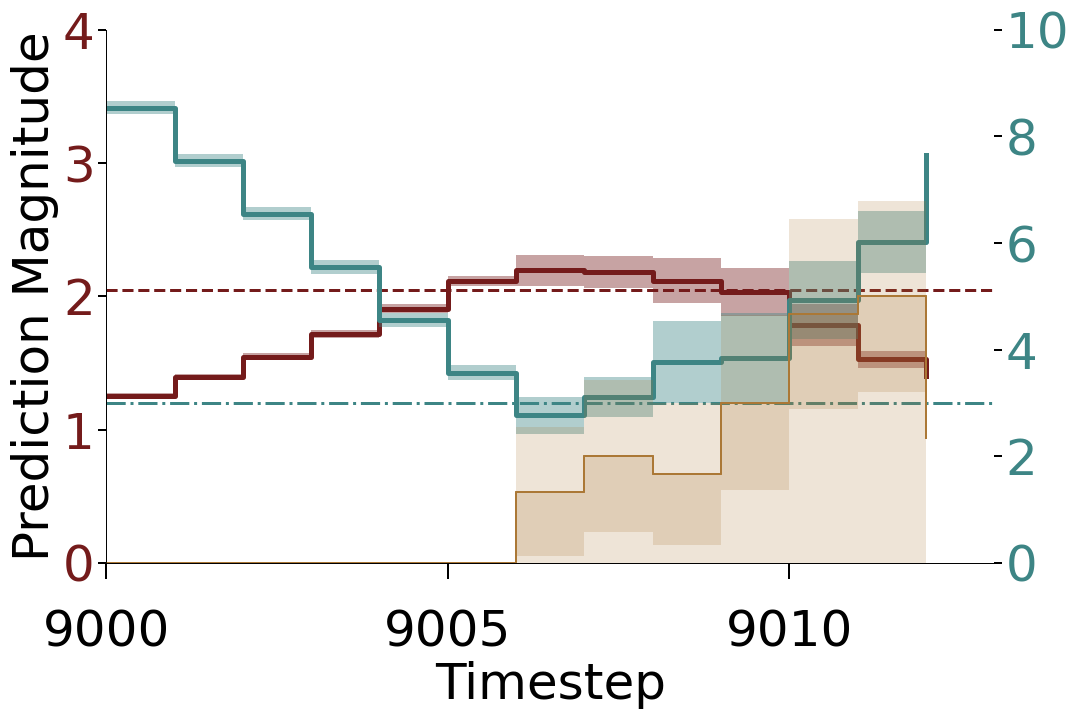}
\includegraphics[width=1.5in]{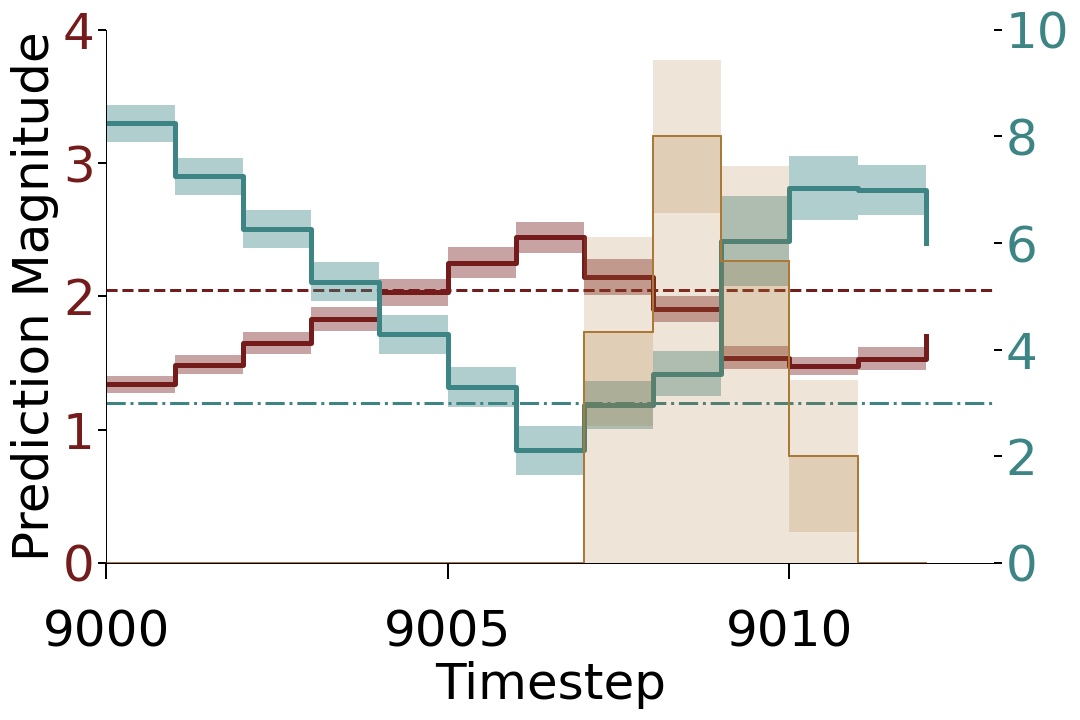}\\
(a) { Bit Cascade + PR}\\
\ \\
\includegraphics[width=1.5in]{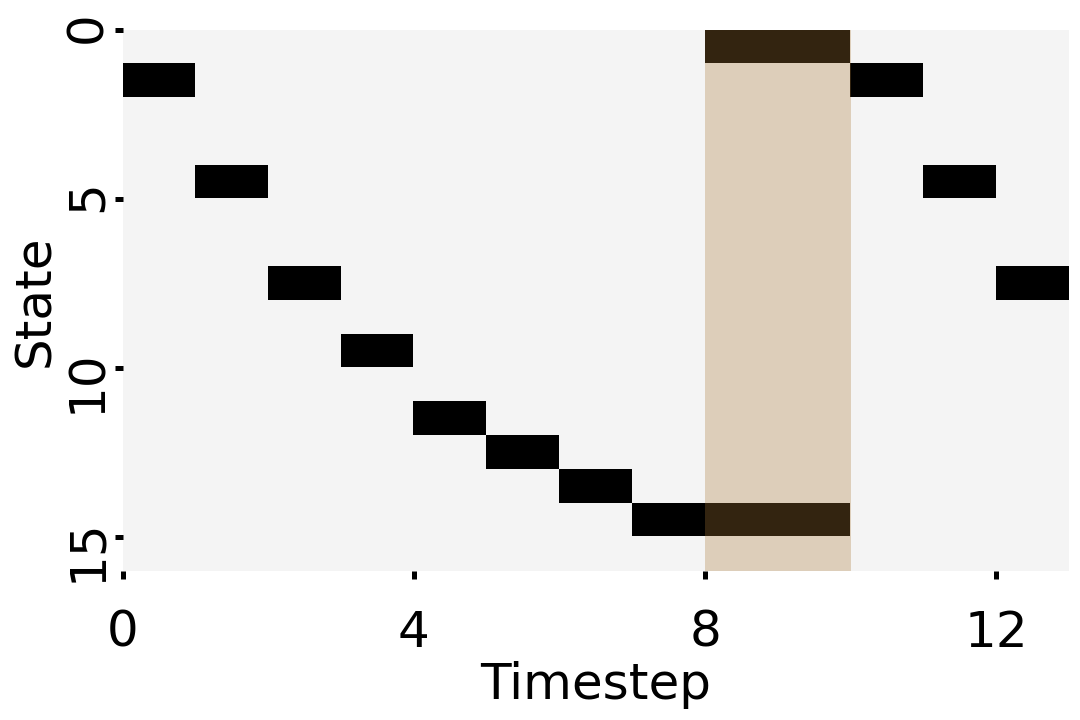}
\includegraphics[width=1.5in]{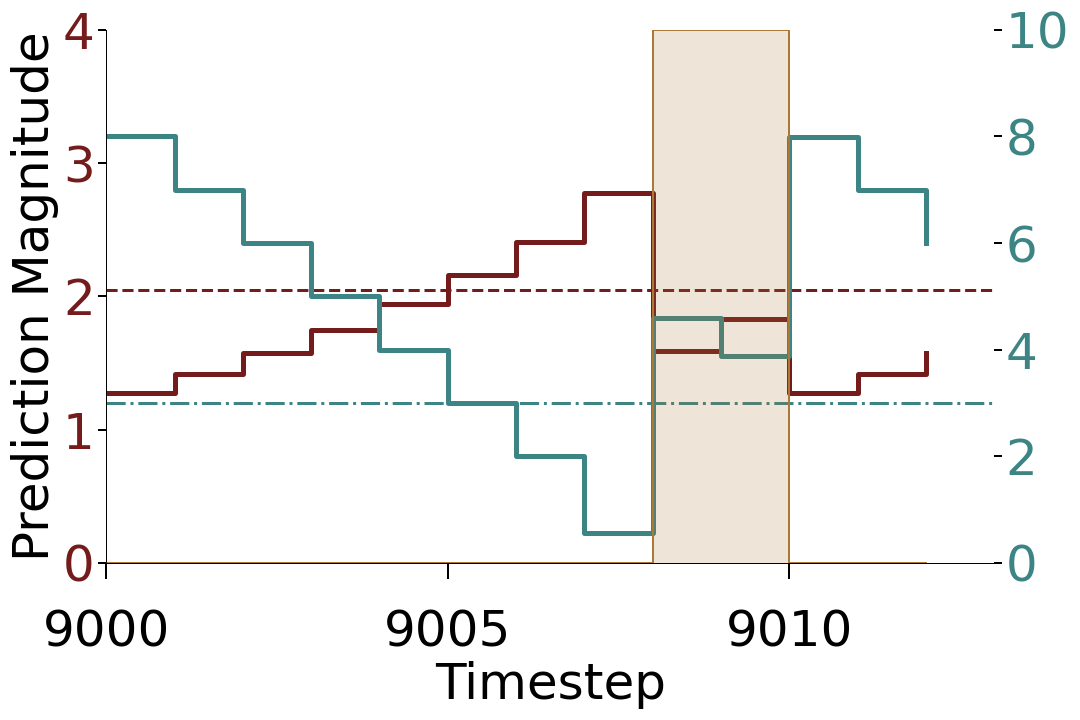}
\includegraphics[width=1.5in]{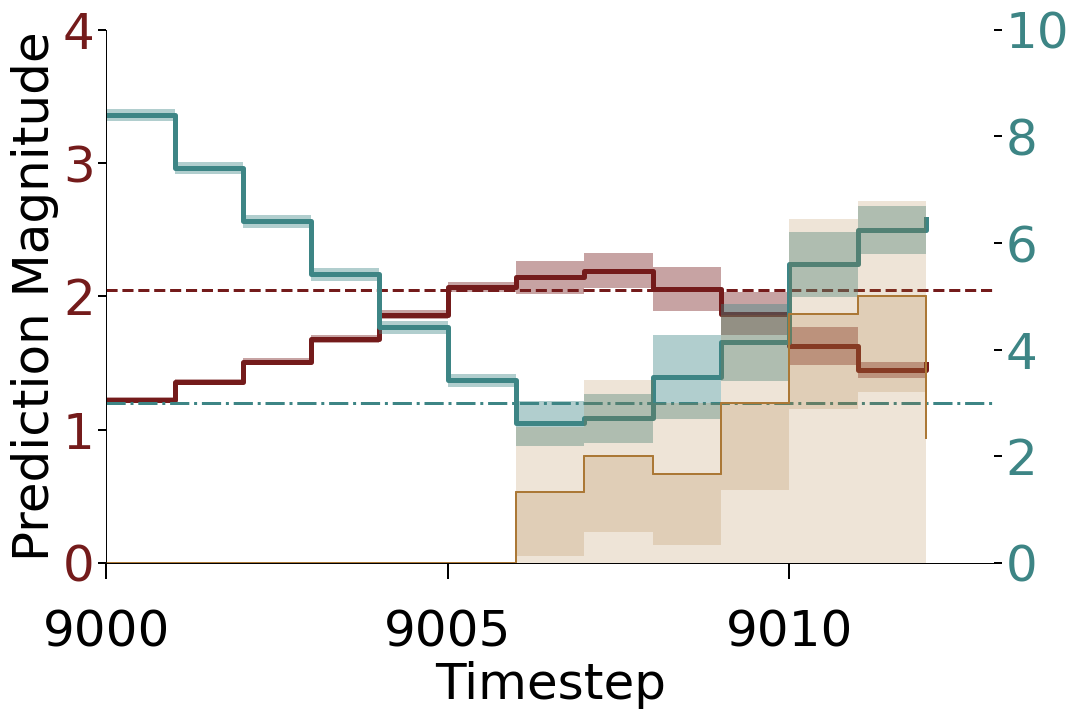}
\includegraphics[width=1.5in]{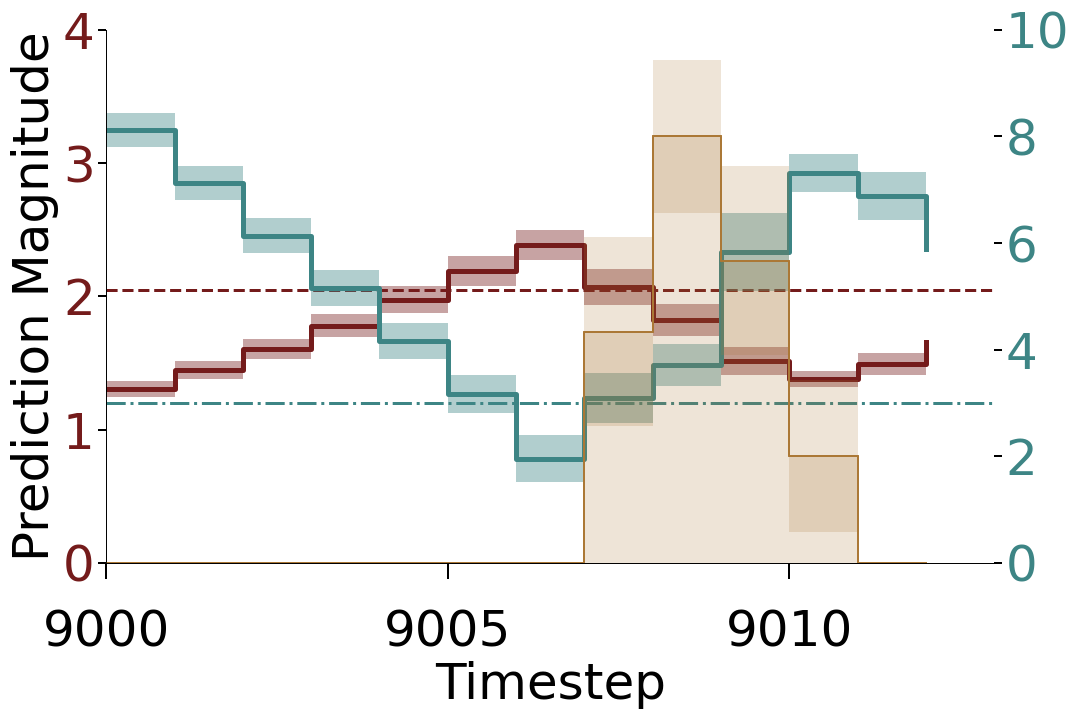}\\
(b) { TCT + PR} (decay = $e^{-0.3t}$)\\
\ \\
\includegraphics[width=1.5in]{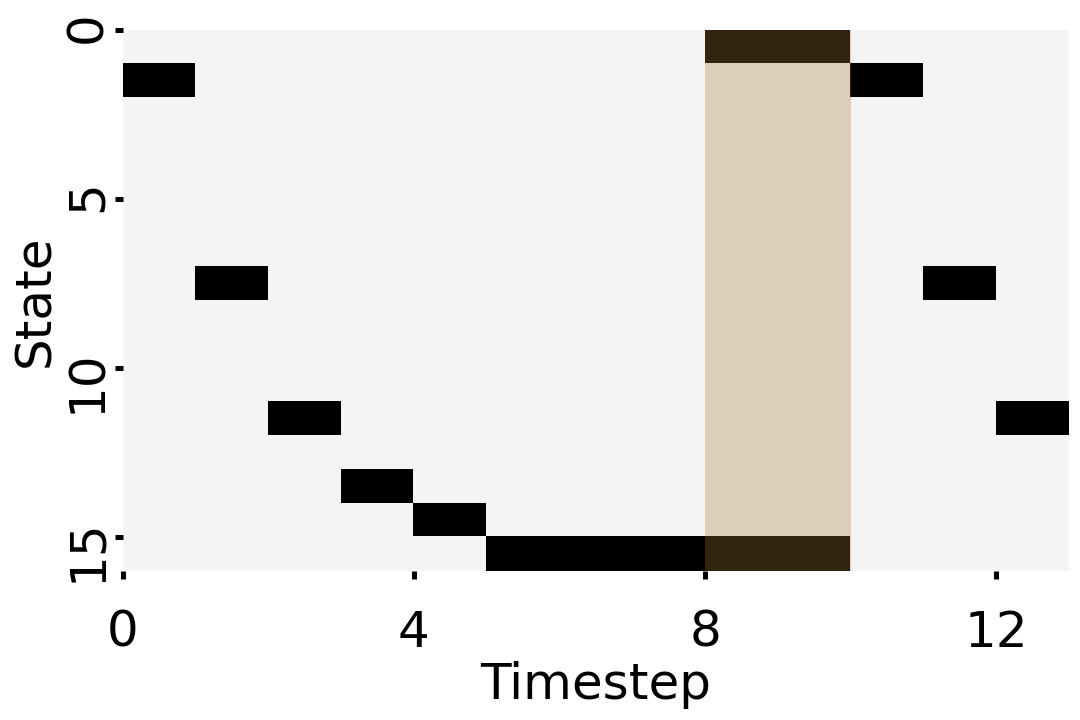}
\includegraphics[width=1.5in]{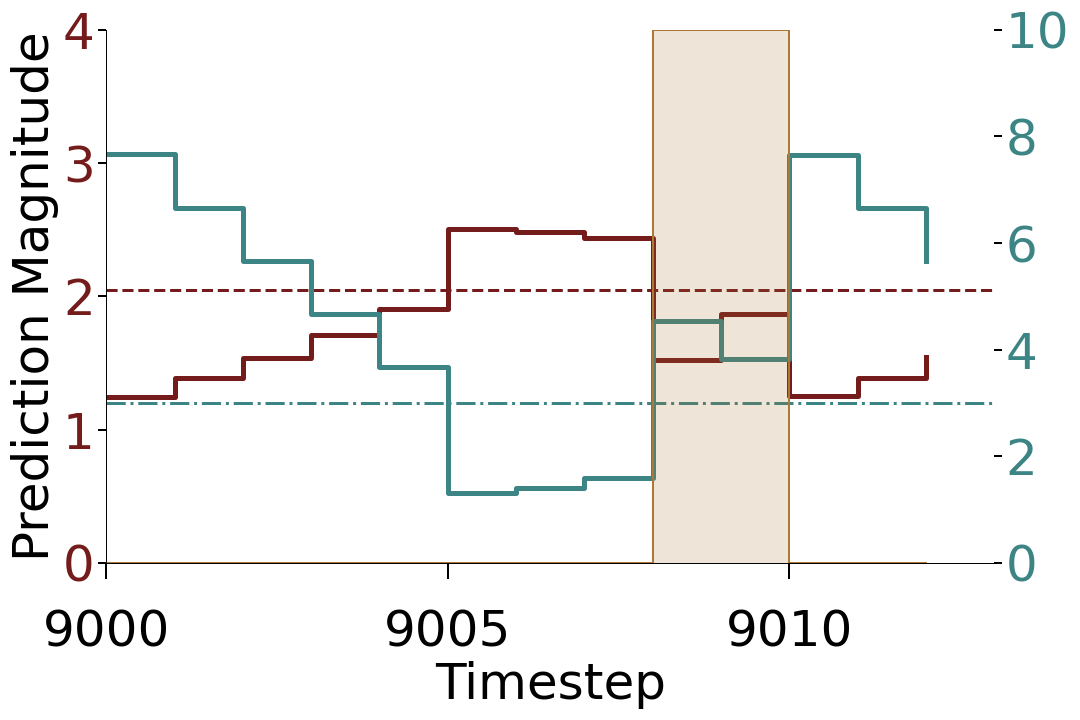}
\includegraphics[width=1.5in]{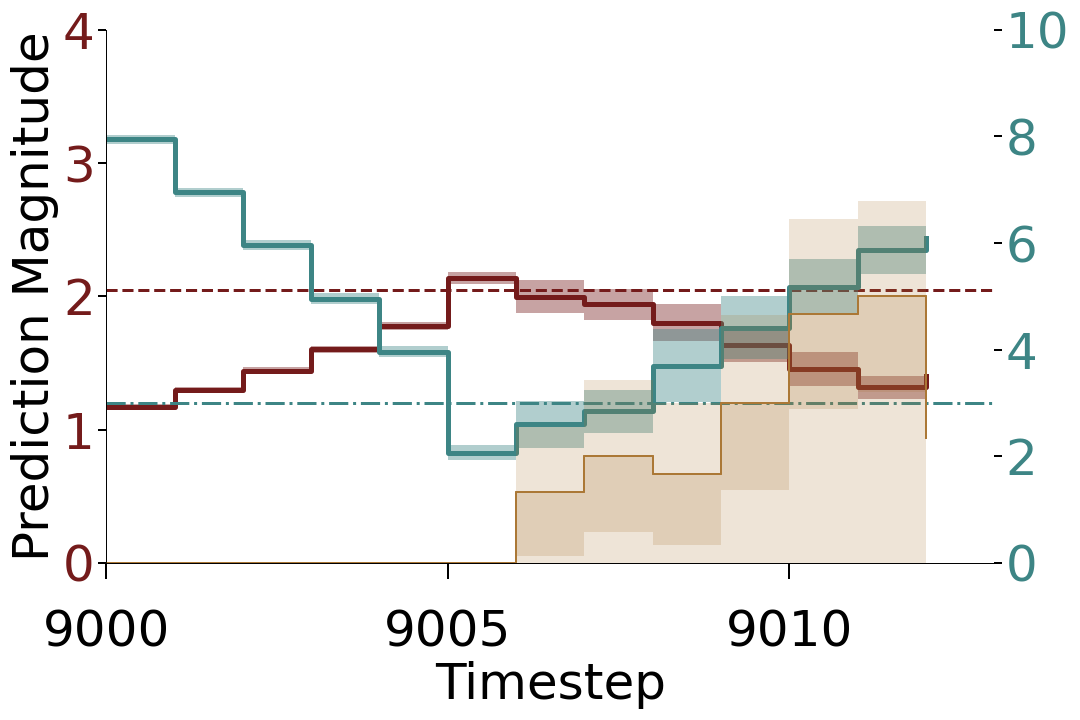}
\includegraphics[width=1.5in]{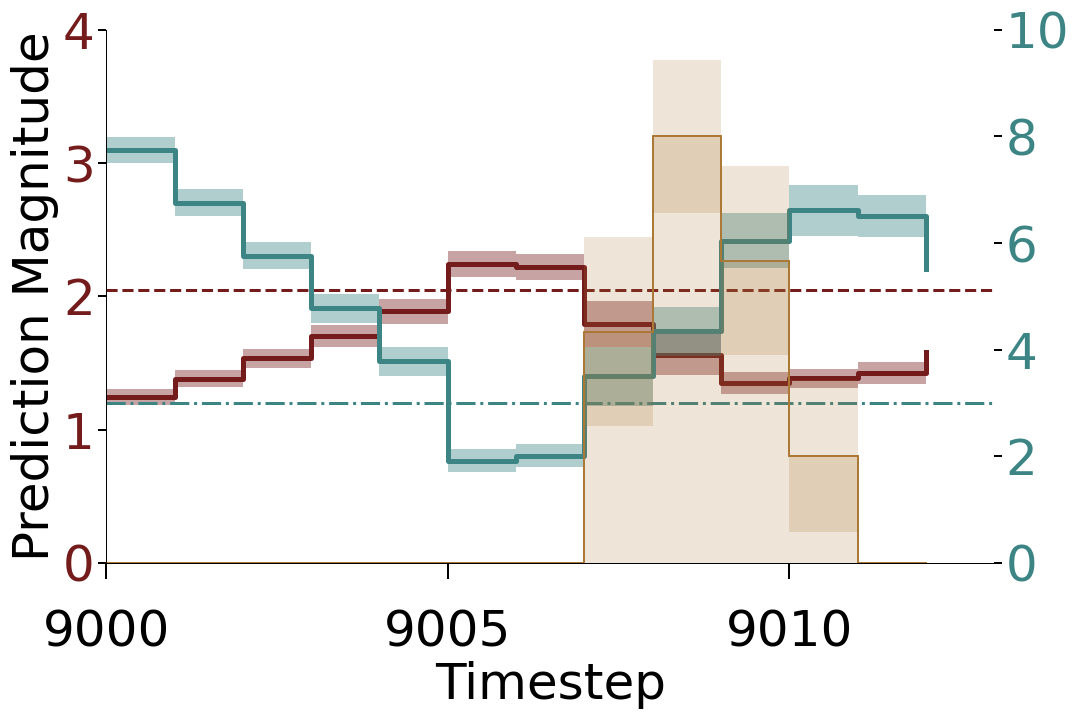}\\
(c) { TCT + PR} (decay = $e^{-0.6t}$)\\
\ \\
\includegraphics[width=1.5in]{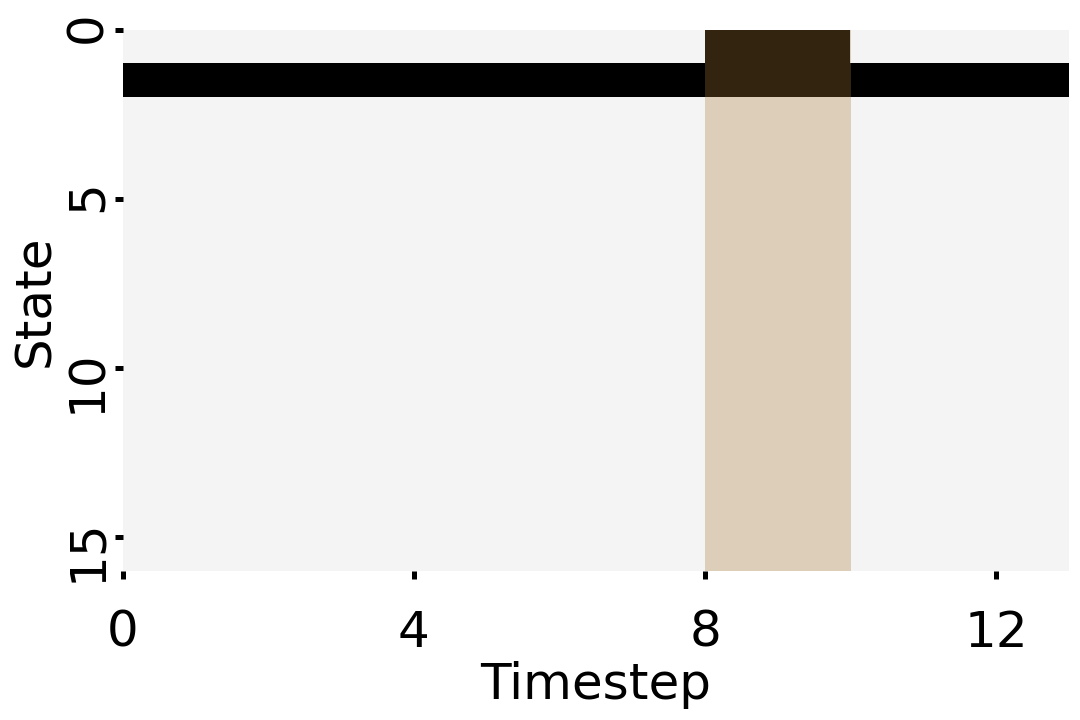}
\includegraphics[width=1.5in]{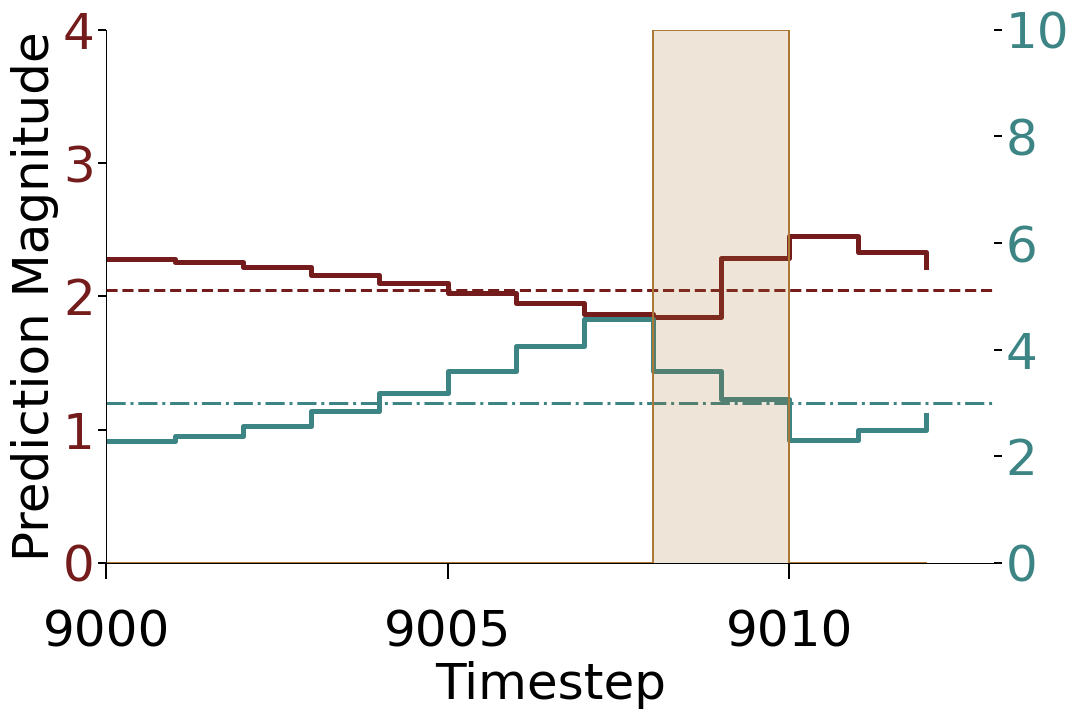}
\includegraphics[width=1.5in]{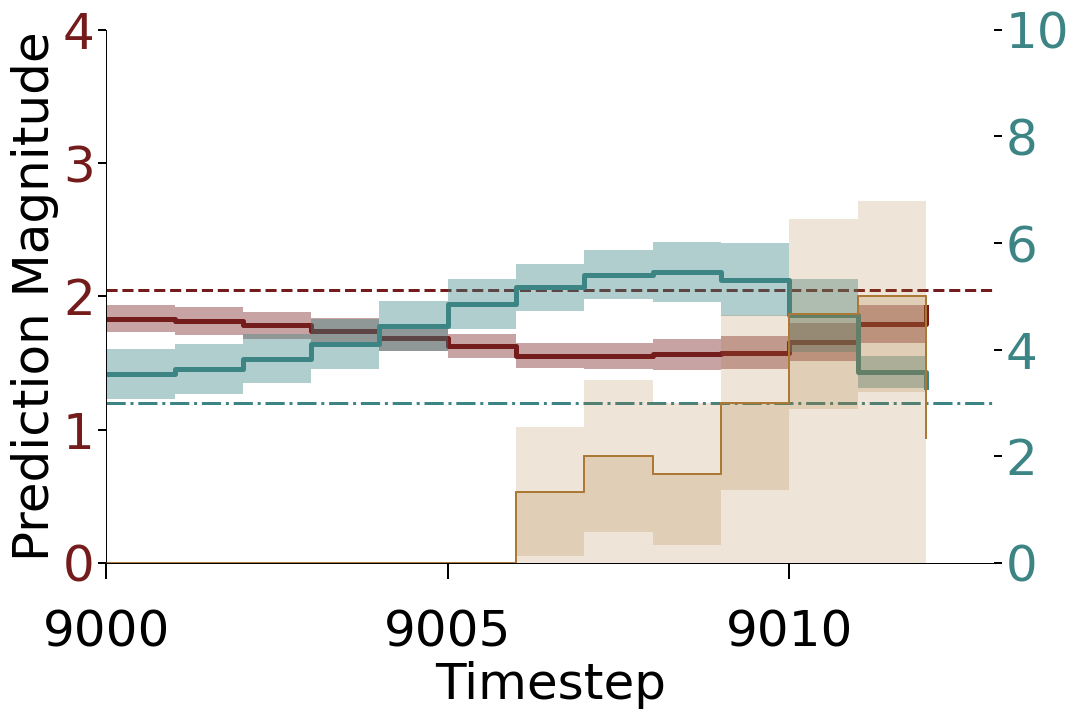}
\includegraphics[width=1.5in]{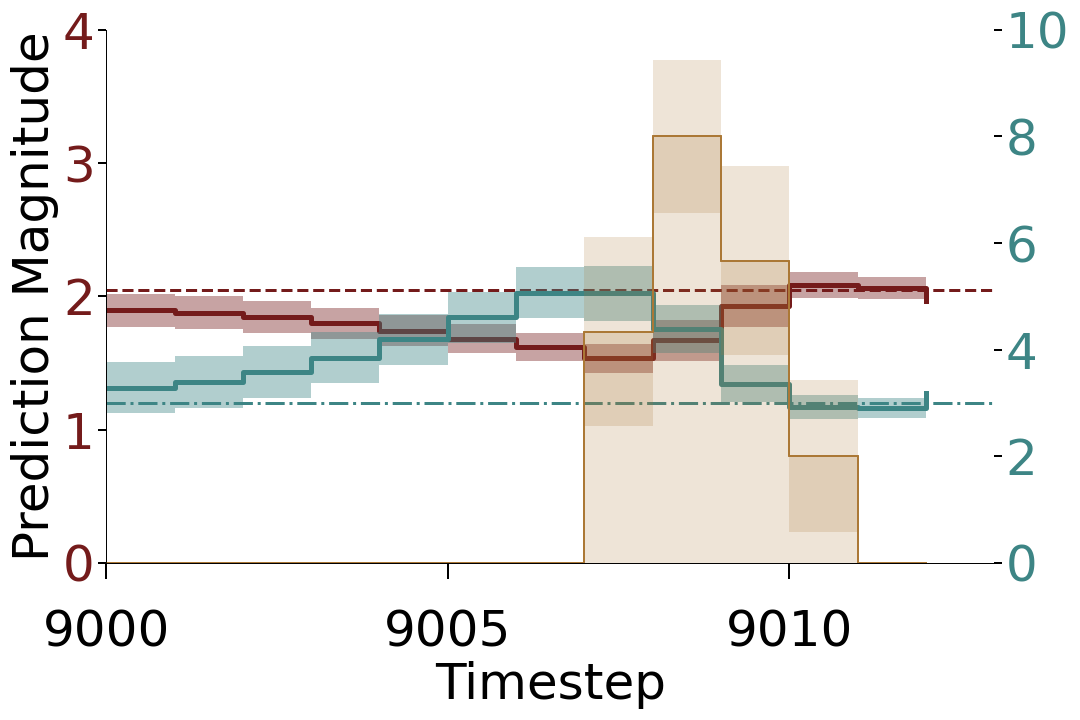}\\
(d) { Bias Unit + PR}\\
\ \\
\includegraphics[width=1.5in]{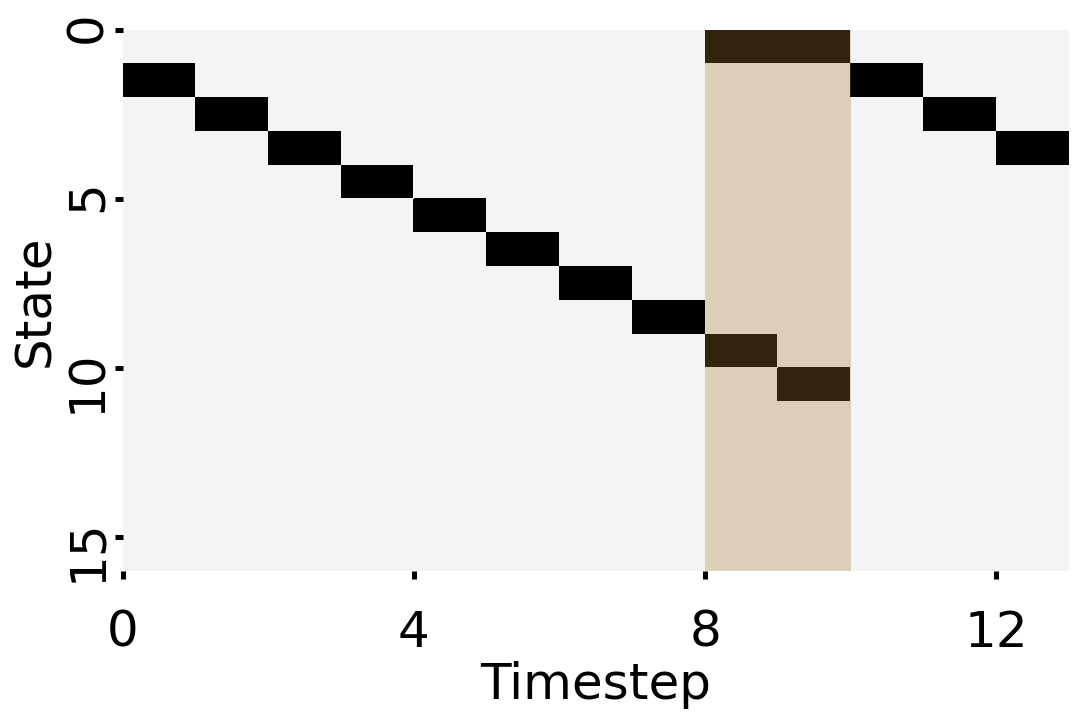}
\includegraphics[width=1.5in]{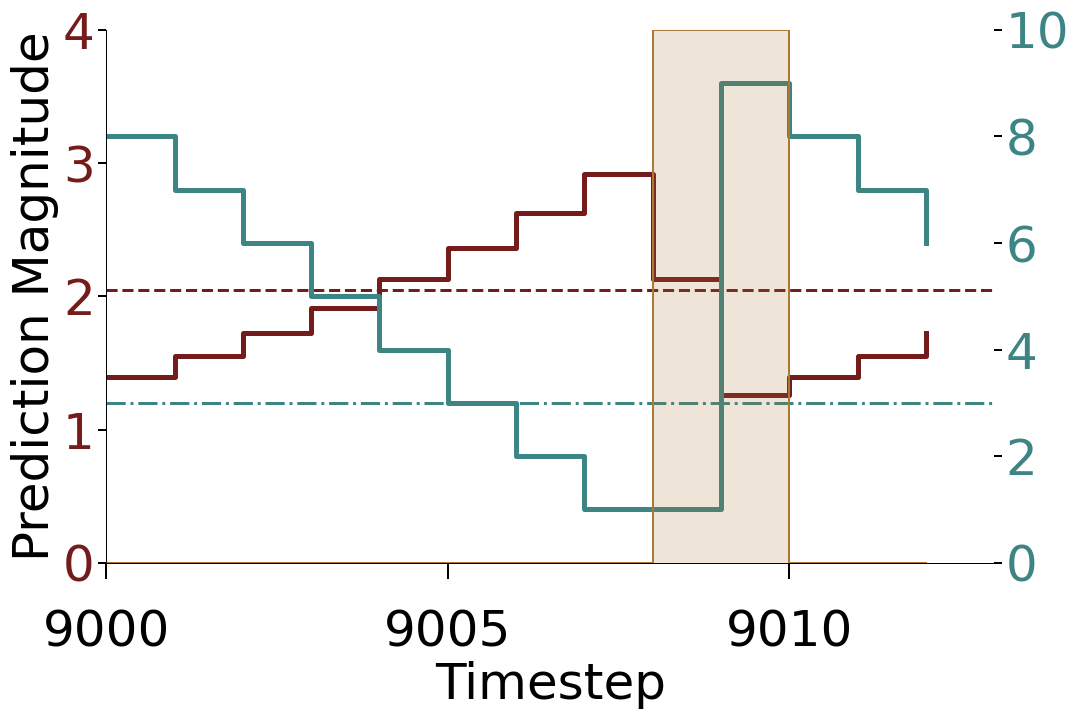}
\includegraphics[width=1.5in]{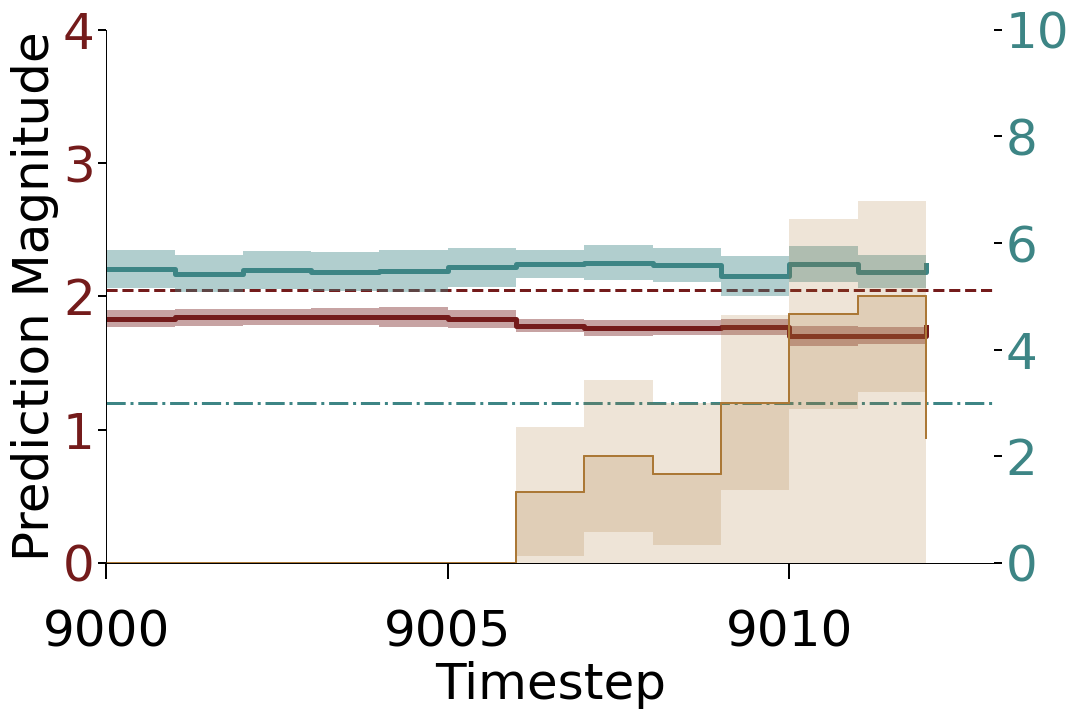}
\includegraphics[width=1.5in]{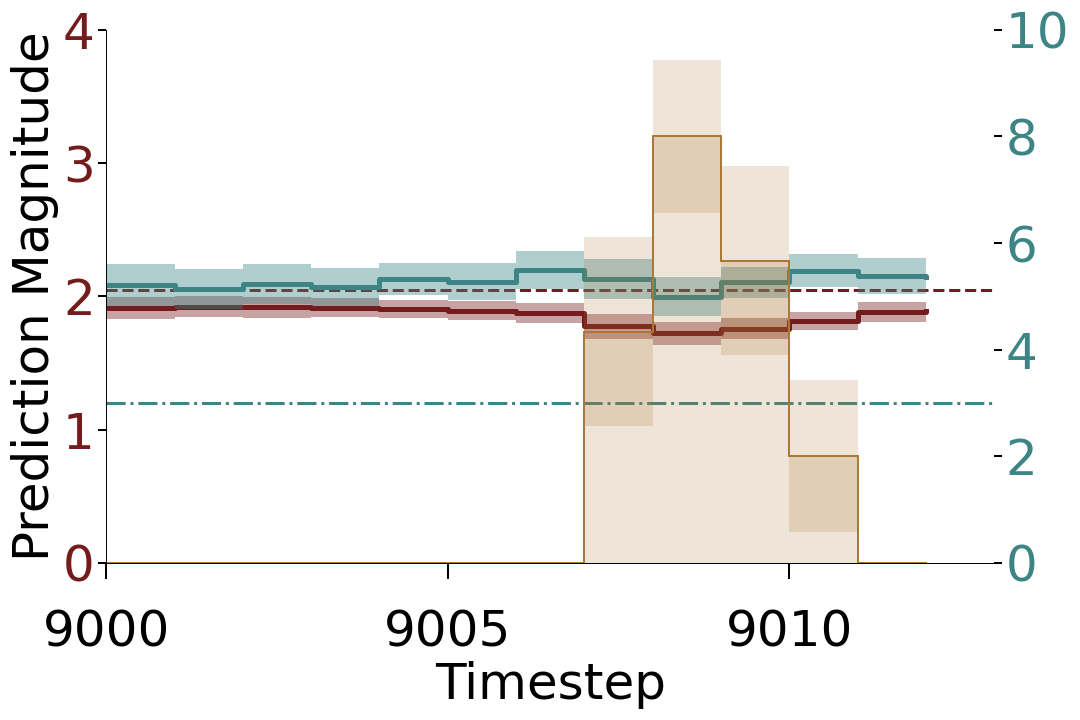}\\
(e) { Oscillator + PR}\\
\caption{{\bf Examples of GVF learning in the Frost Hollow domain for short inter-stimulus intervals} (ten steps, with an active period of two steps). Shown here (left column) are the recorded bit progressions for each representation over spans of single ISIs of equal length, including the presence representation (black square in the top row of each plot) and the temporal representation (black squares in rows 1+). {\bf Right columns show an example of prediction learning for both the accumulation (red trace) and countdown (blue trace) GVF questions} with five different representations across the fixed, random, and drift conditions (three rightmost columns, hazards shown in yellow). Shaded trace envelopes indicate the standard error of the mean. Threshold levels for Pavlovian signalling for each prediction type are shown as horizontal dashed lines. For random and drift conditions, hazard stimulus is also averaged to illustrate variability in the ISI. All results are averaged over 30 independent trials, with results shown for the tenth episode (block of 1000 steps) of learning so as to align the ISI starting conditions for all prediction learners.}
\label{fig:nexting-short-isi-comp}
\end{figure*}

\begin{figure*}[!th]
\centering
{\bf REPRESENTATION \hfil FIXED \hfil RANDOM \hfil DRIFT}\\
\ \\
\includegraphics[width=1.5in]{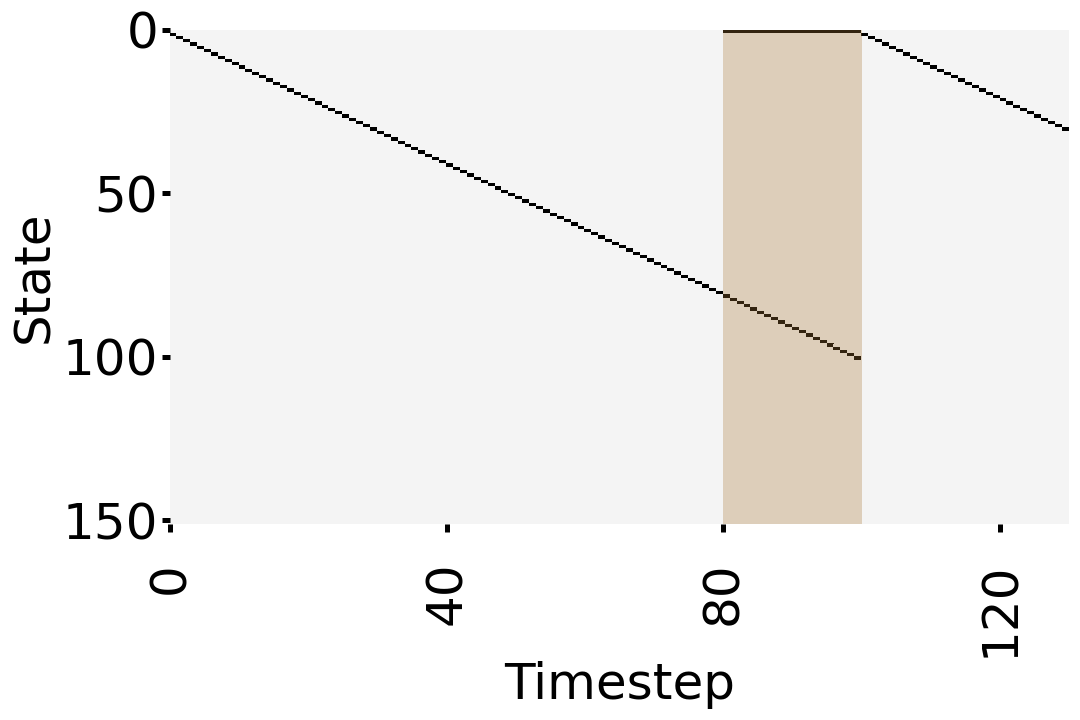}
\includegraphics[width=1.5in]{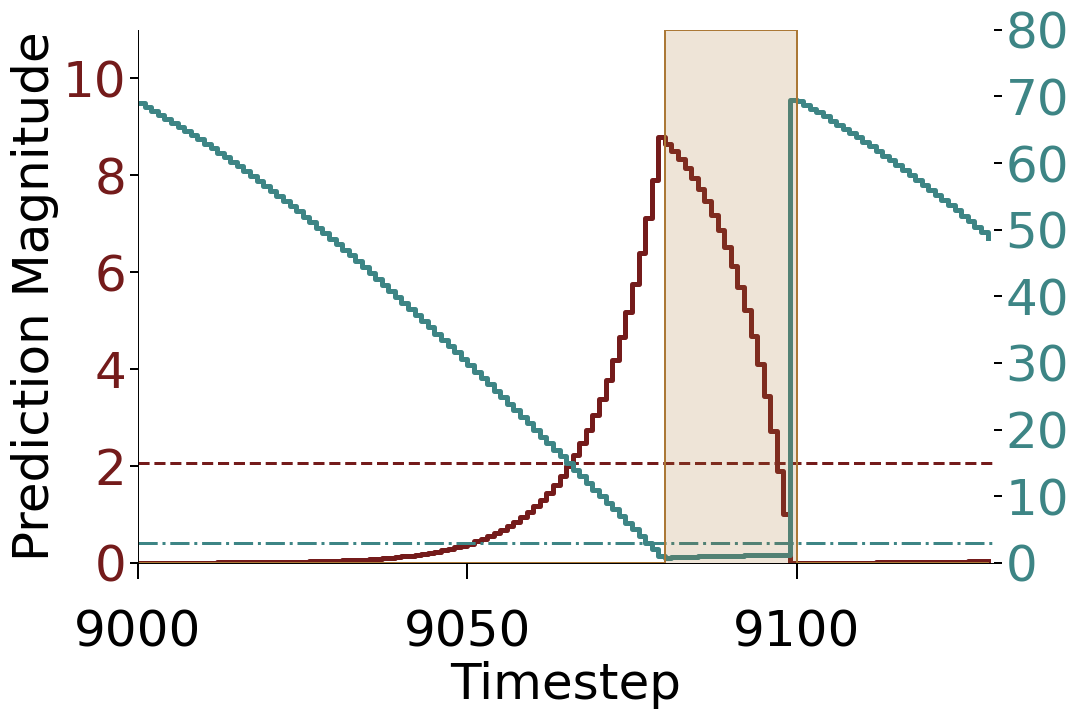}
\includegraphics[width=1.5in]{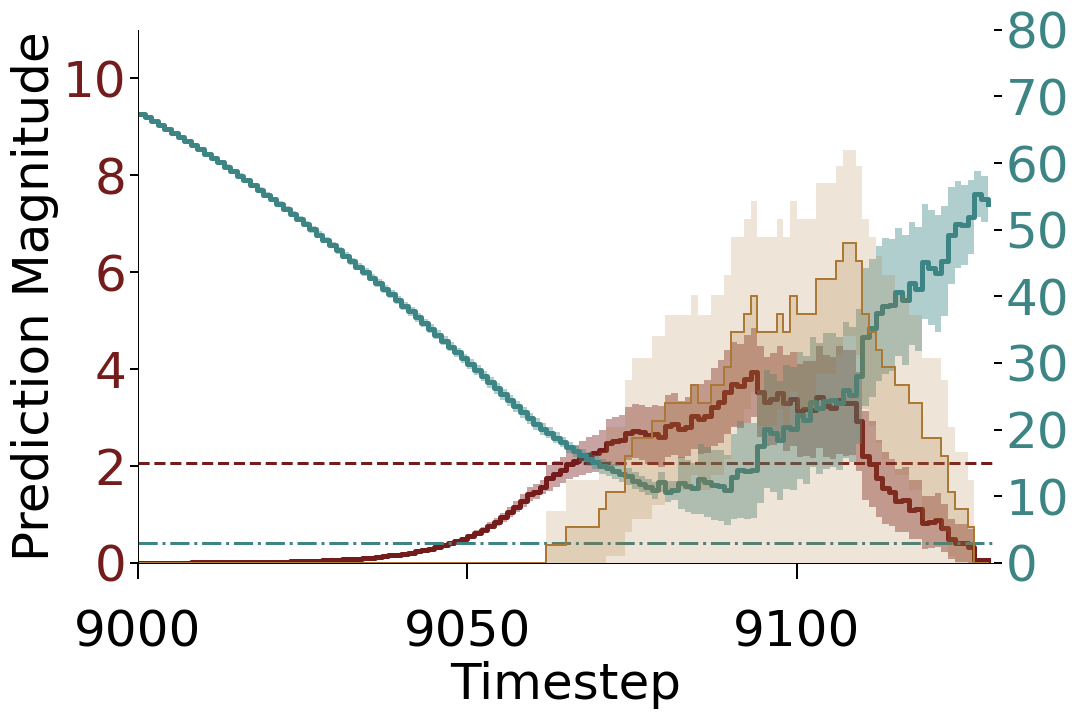}
\includegraphics[width=1.5in]{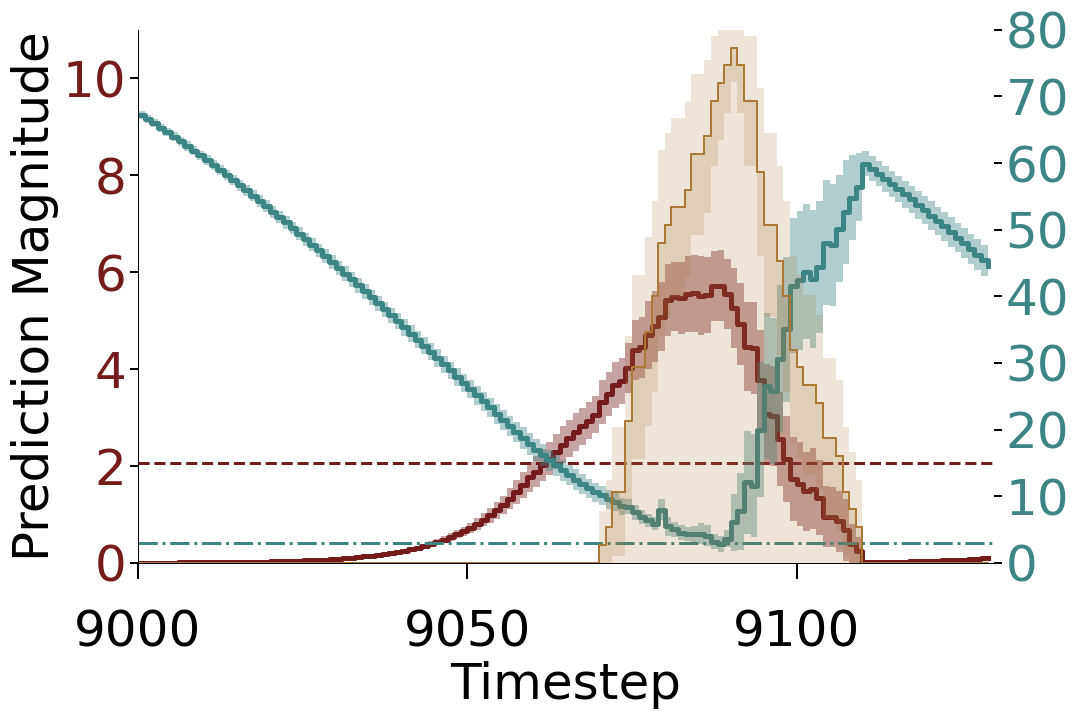}\\
(a) Bit Cascade + PR\\
\ \\
\includegraphics[width=1.5in]{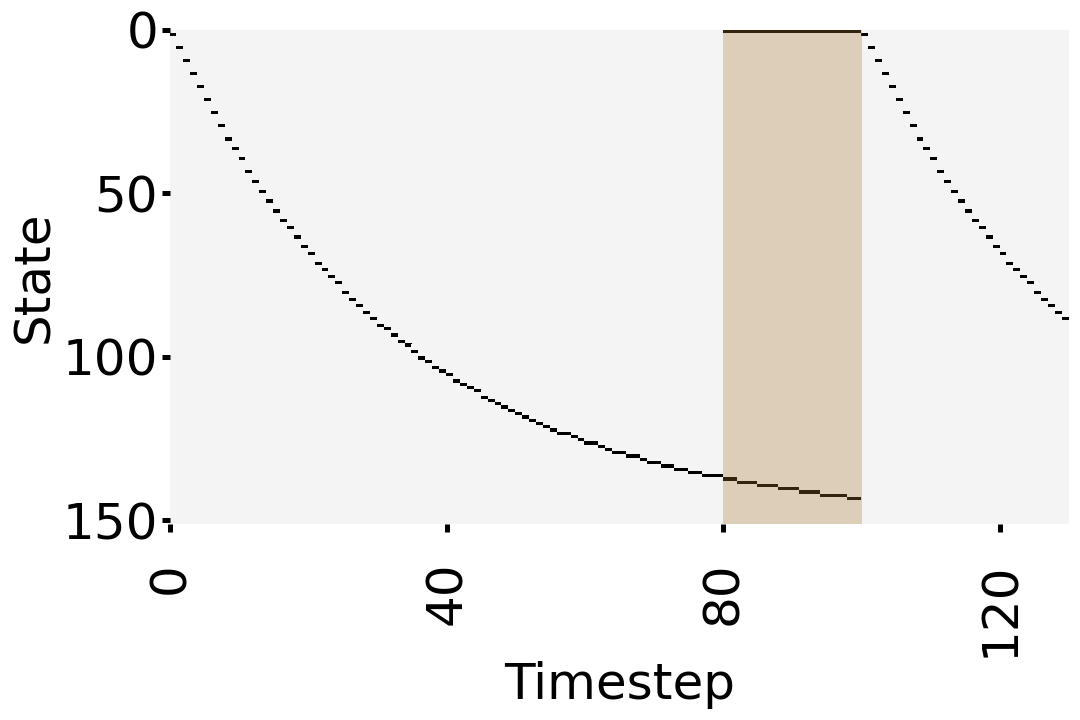}
\includegraphics[width=1.5in]{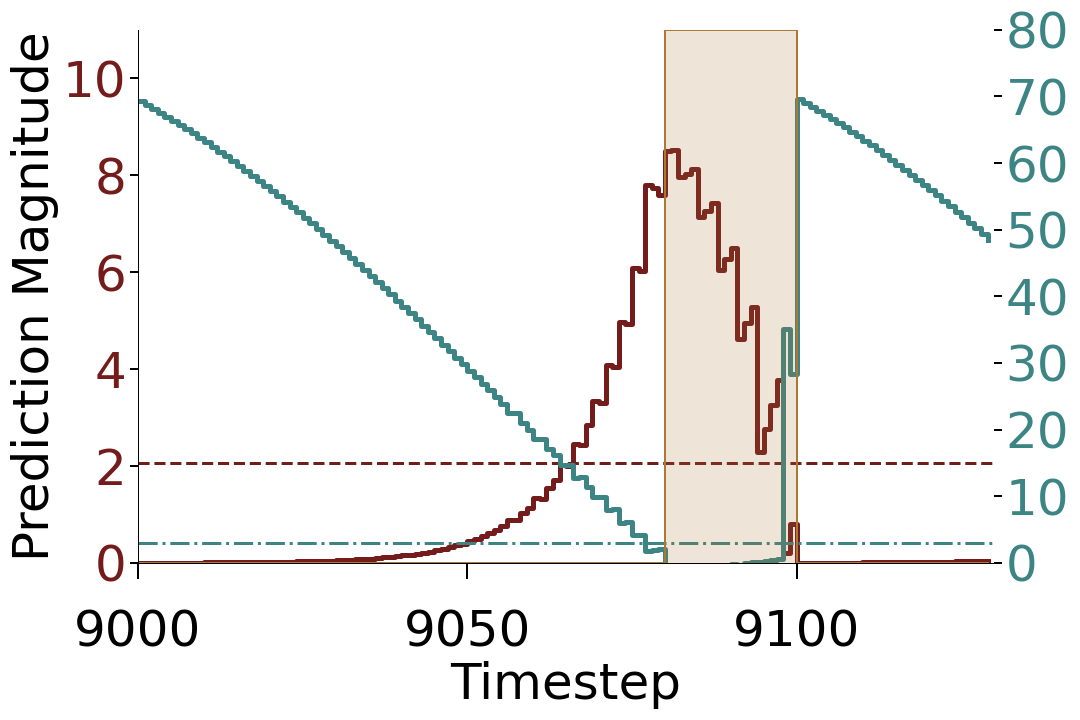}
\includegraphics[width=1.5in]{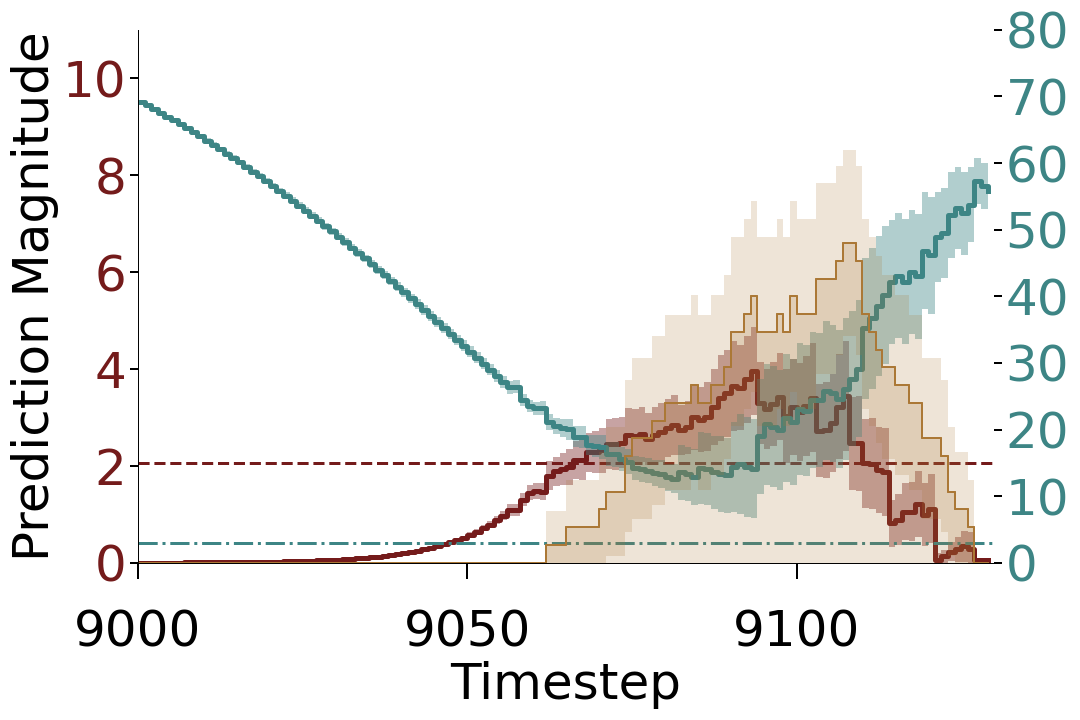}
\includegraphics[width=1.5in]{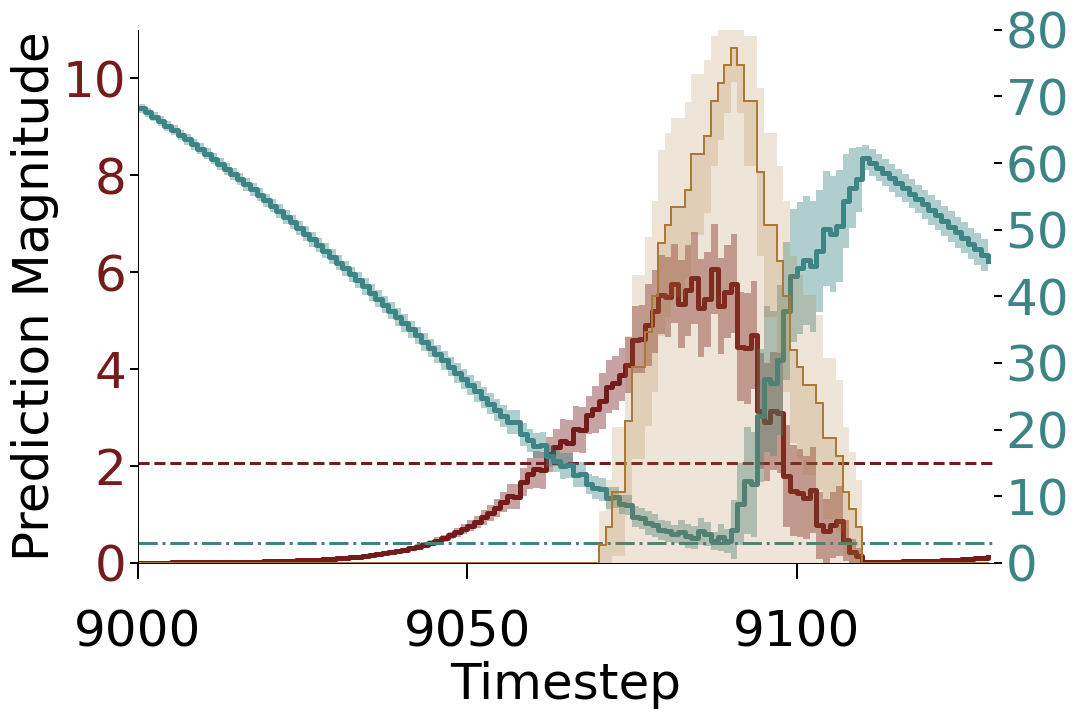}\\
(b) TCT + PR (decay = $e^{-0.03t}$)\\
\ \\
\includegraphics[width=1.5in]{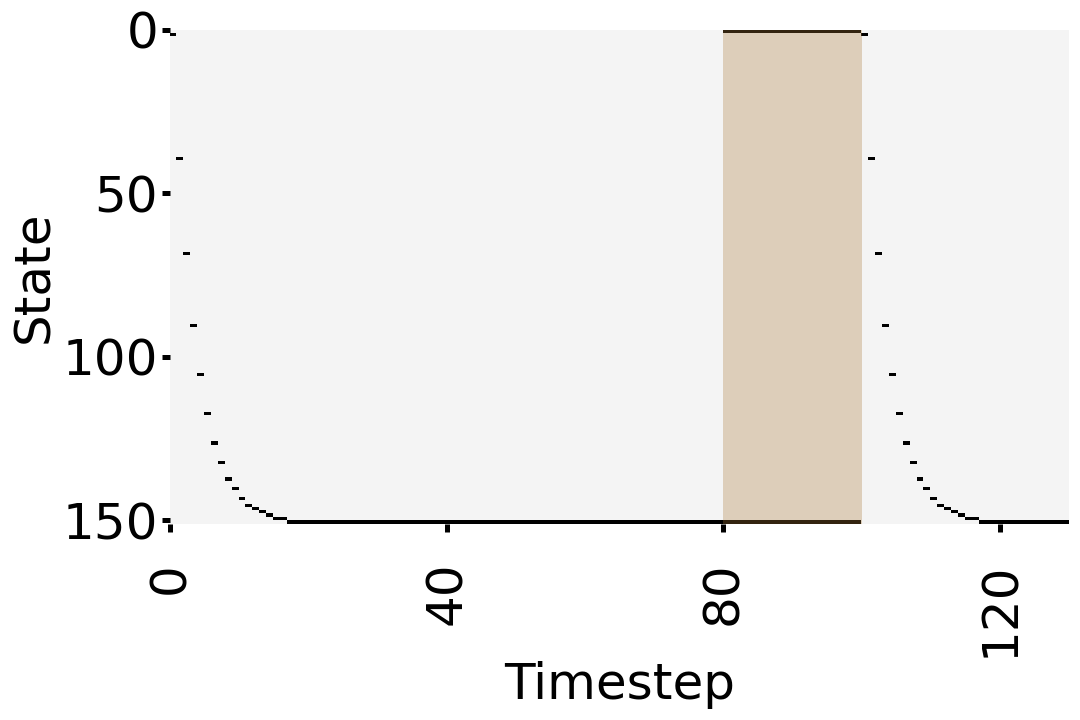}
\includegraphics[width=1.5in]{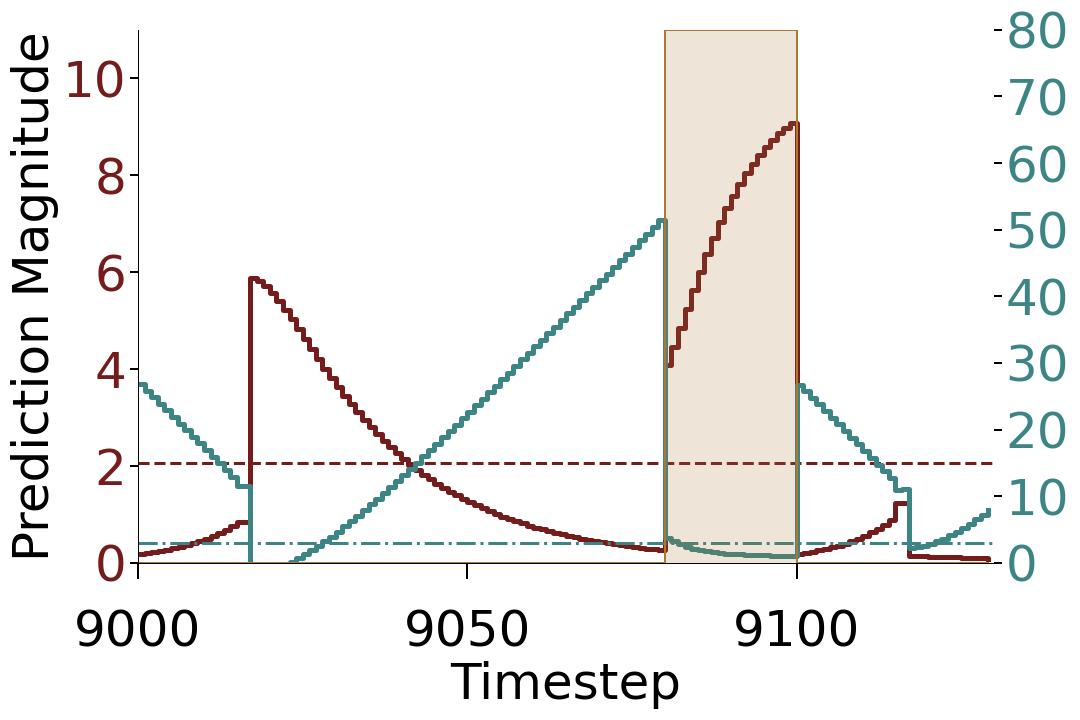}
\includegraphics[width=1.5in]{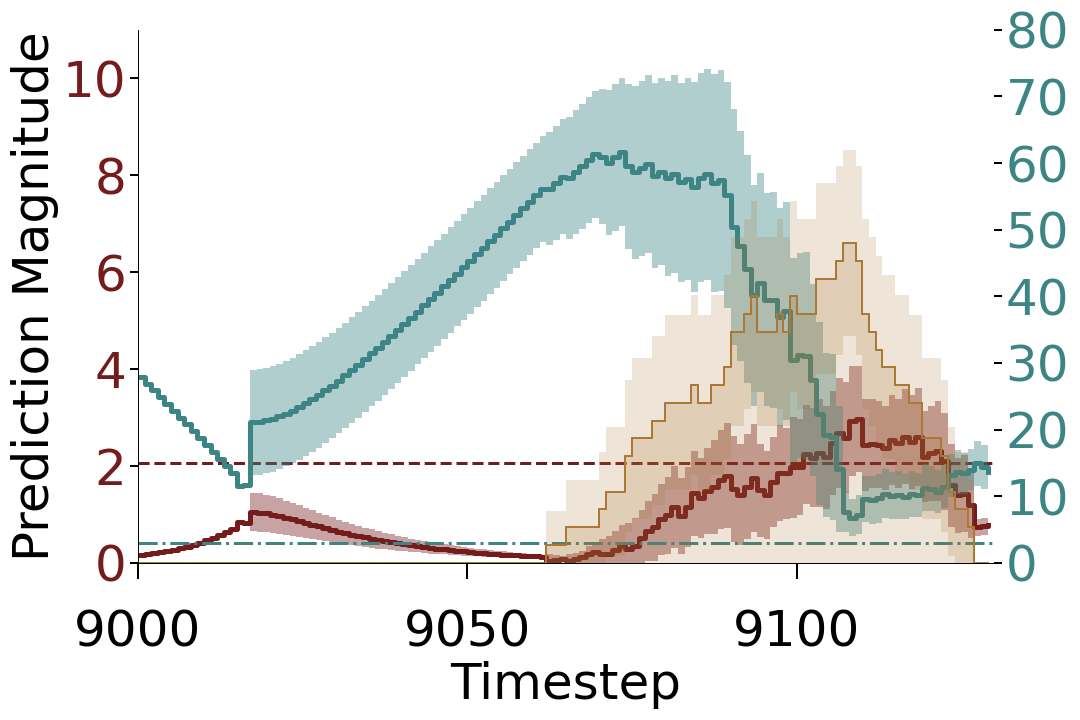}
\includegraphics[width=1.5in]{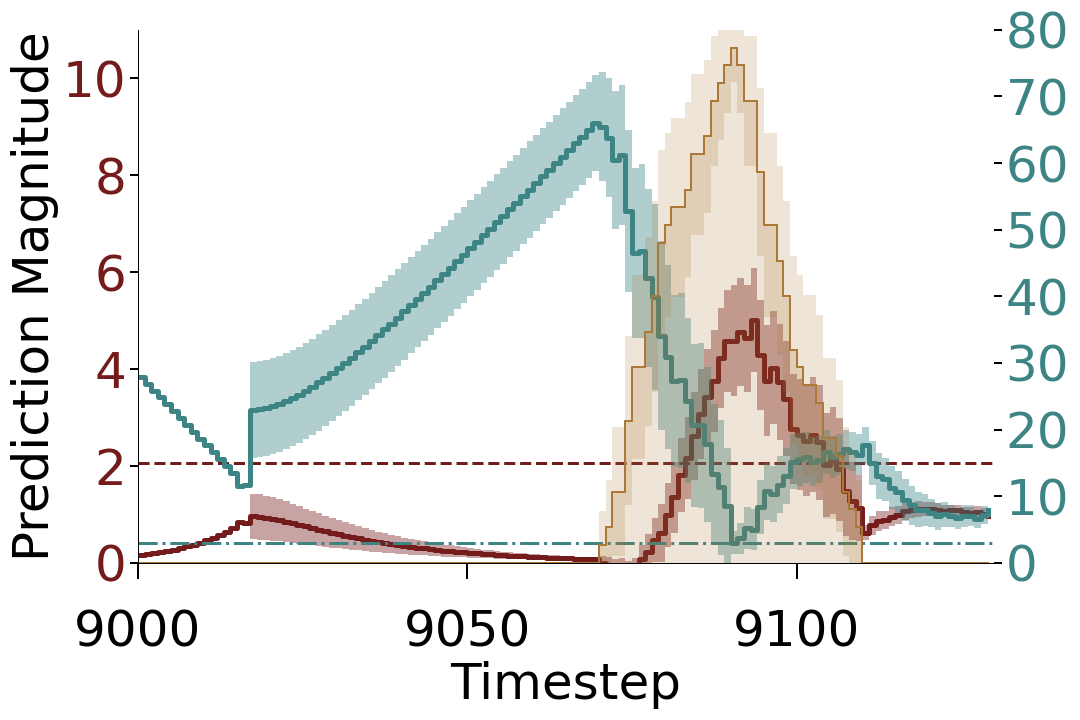}\\
(c) TCT + PR (decay = $e^{-0.6t}$)\\
\ \\
\includegraphics[width=1.5in]{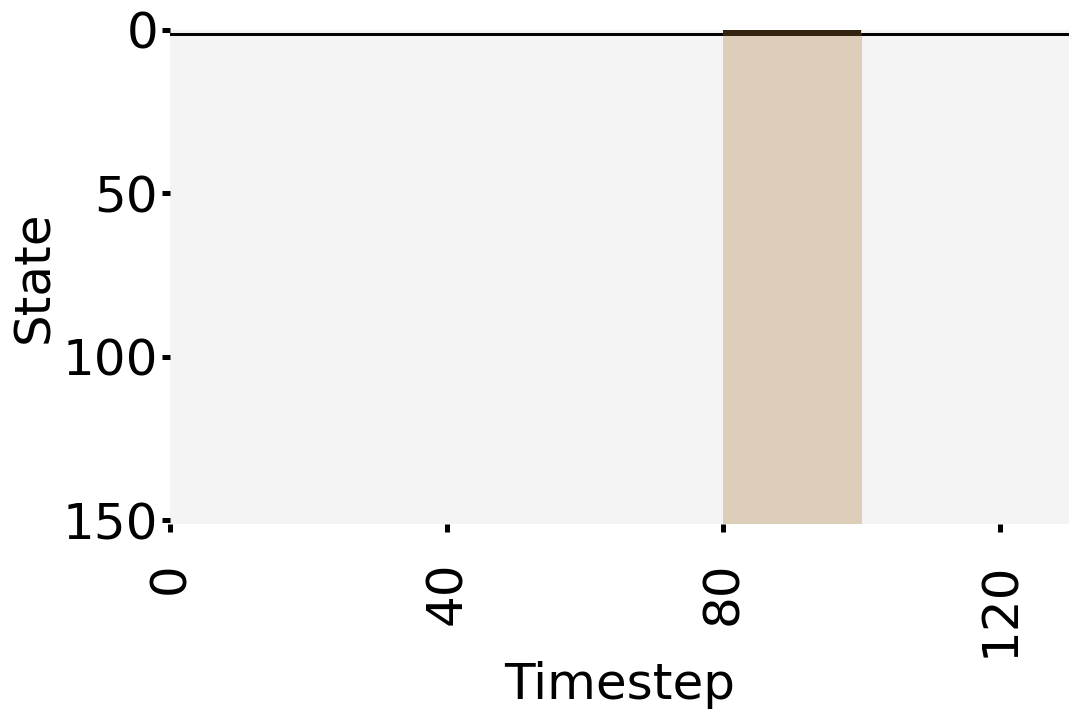}
\includegraphics[width=1.5in]{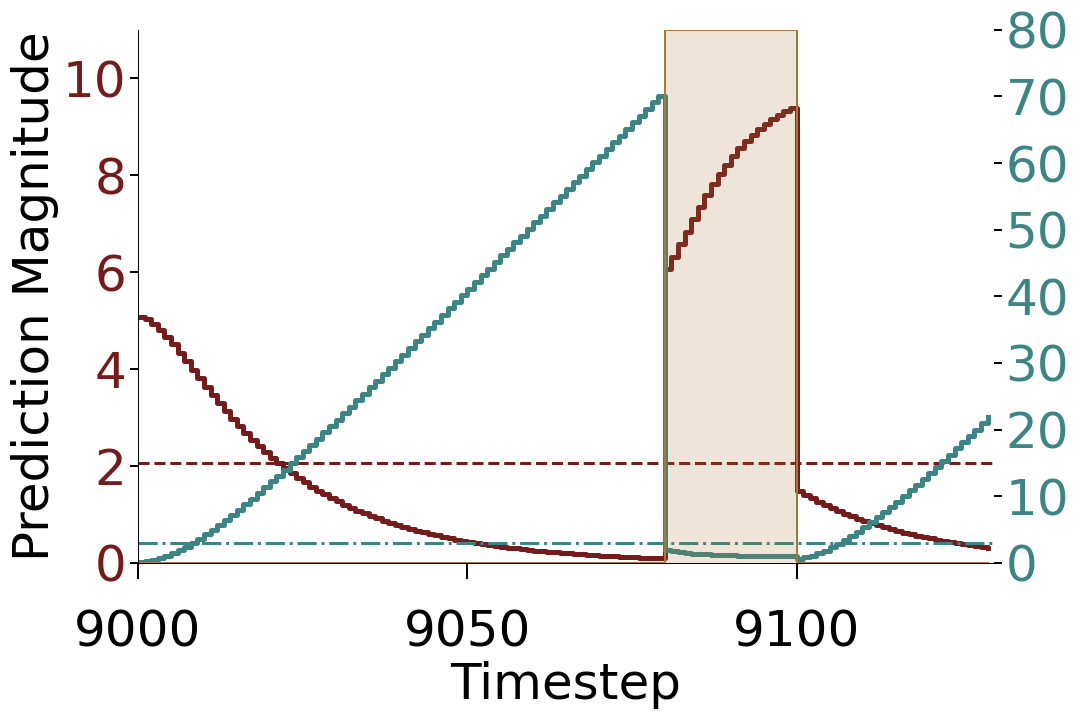}
\includegraphics[width=1.5in]{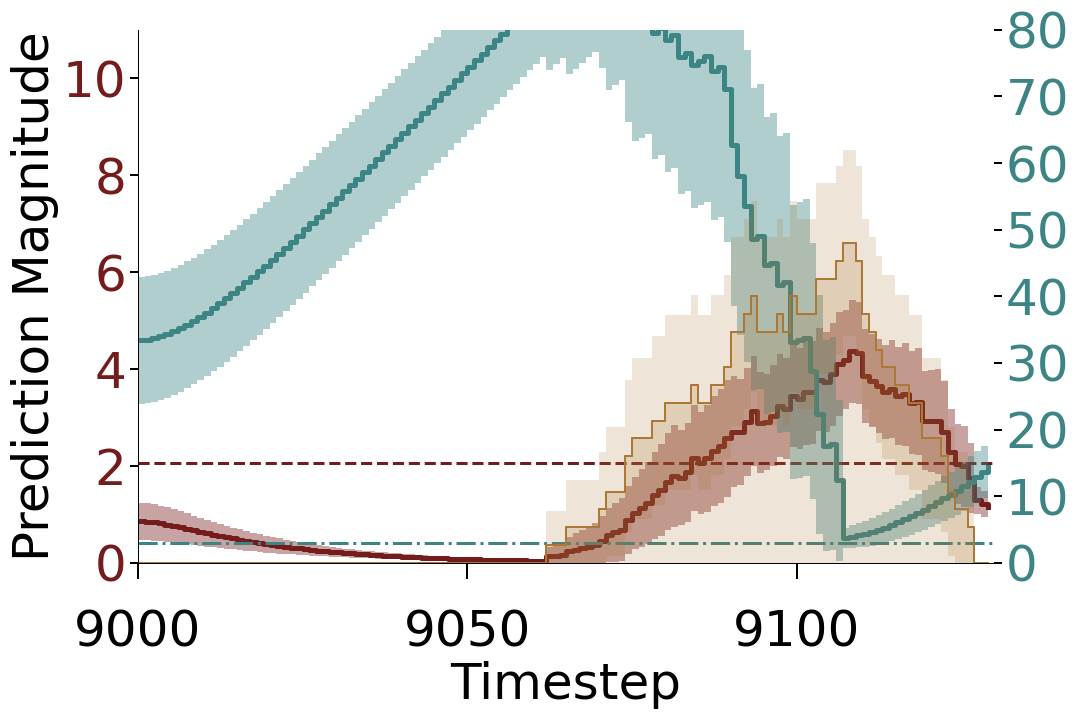}
\includegraphics[width=1.5in]{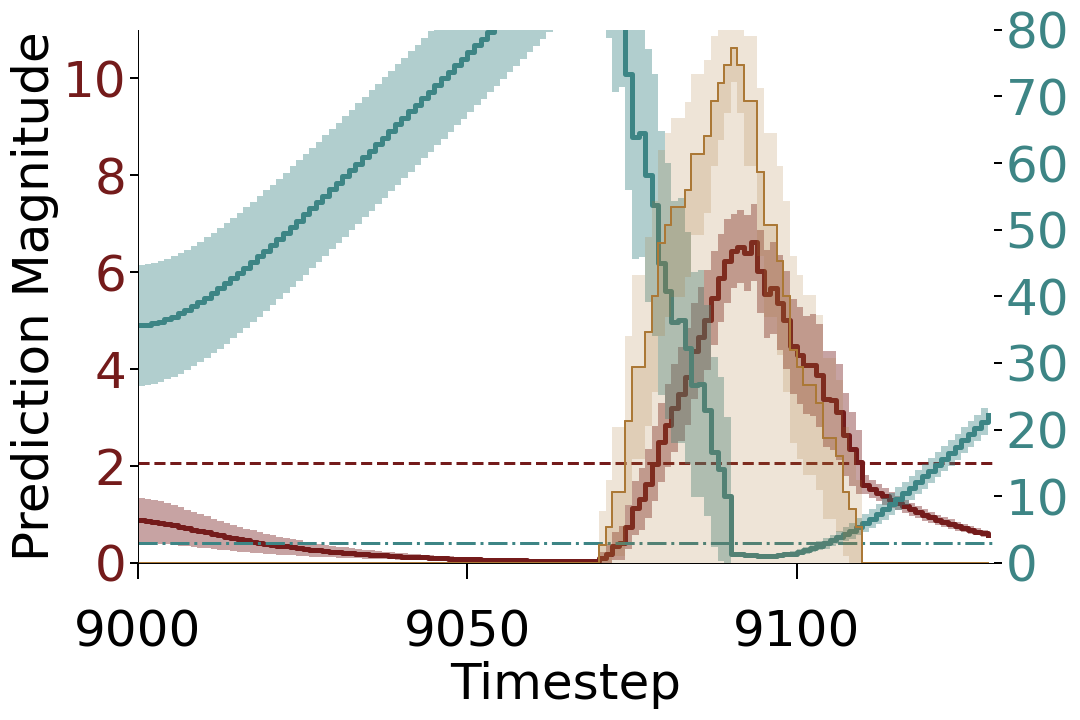}\\
(d) Bias Unit + PR\\
\ \\
\includegraphics[width=1.5in]{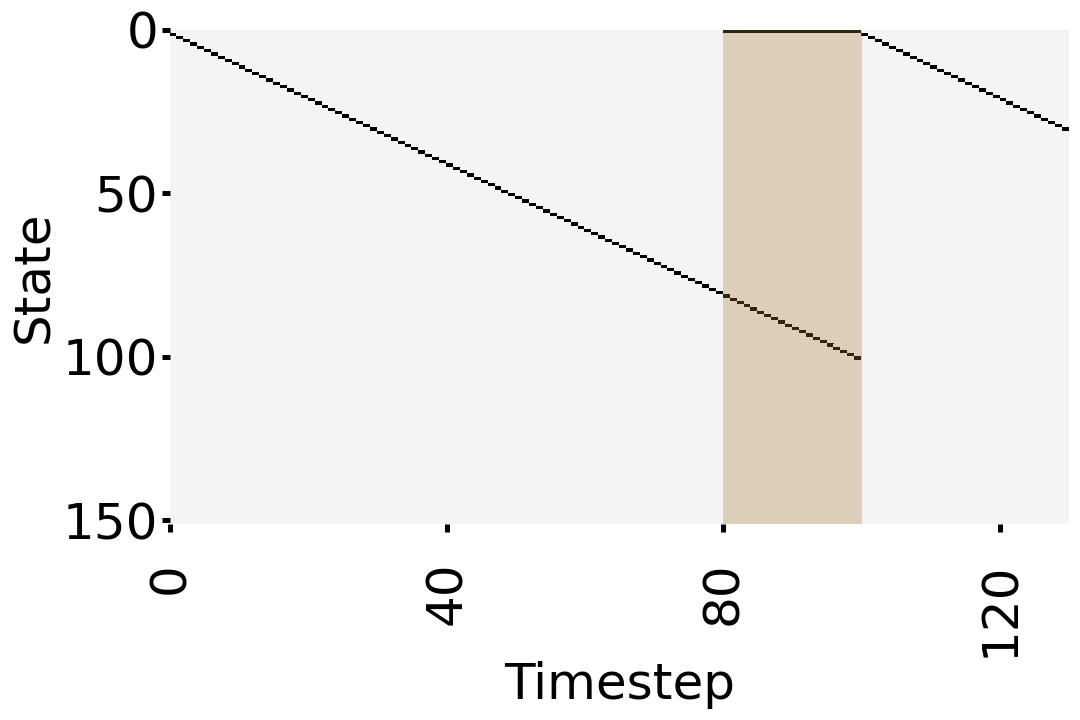}
\includegraphics[width=1.5in]{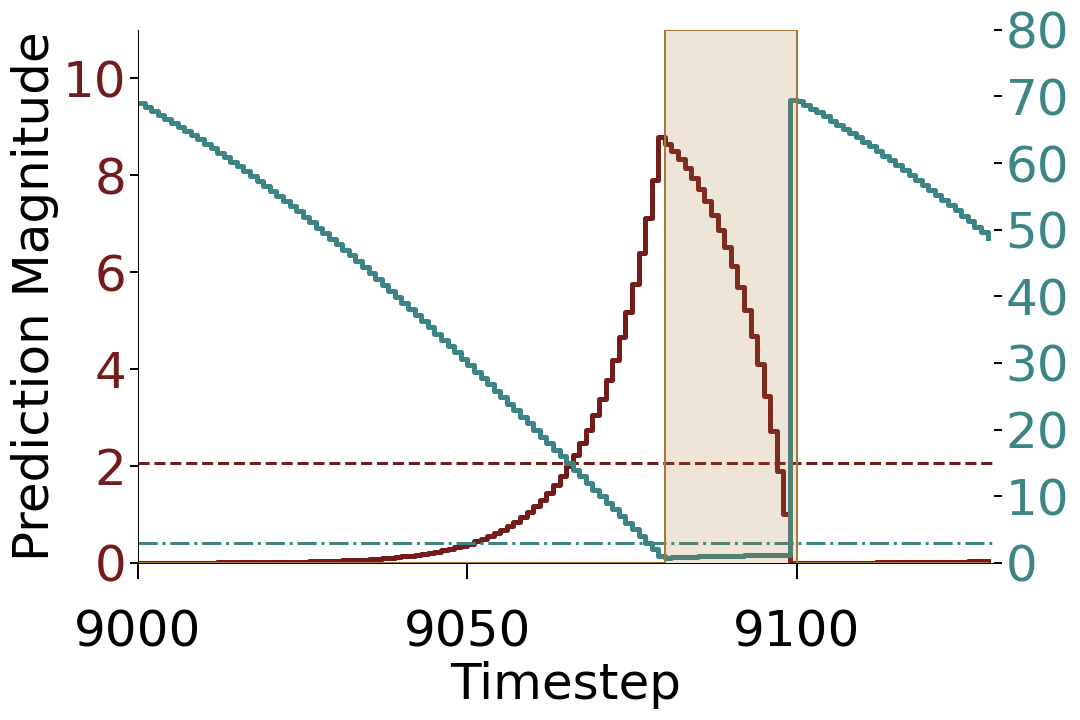}
\includegraphics[width=1.5in]{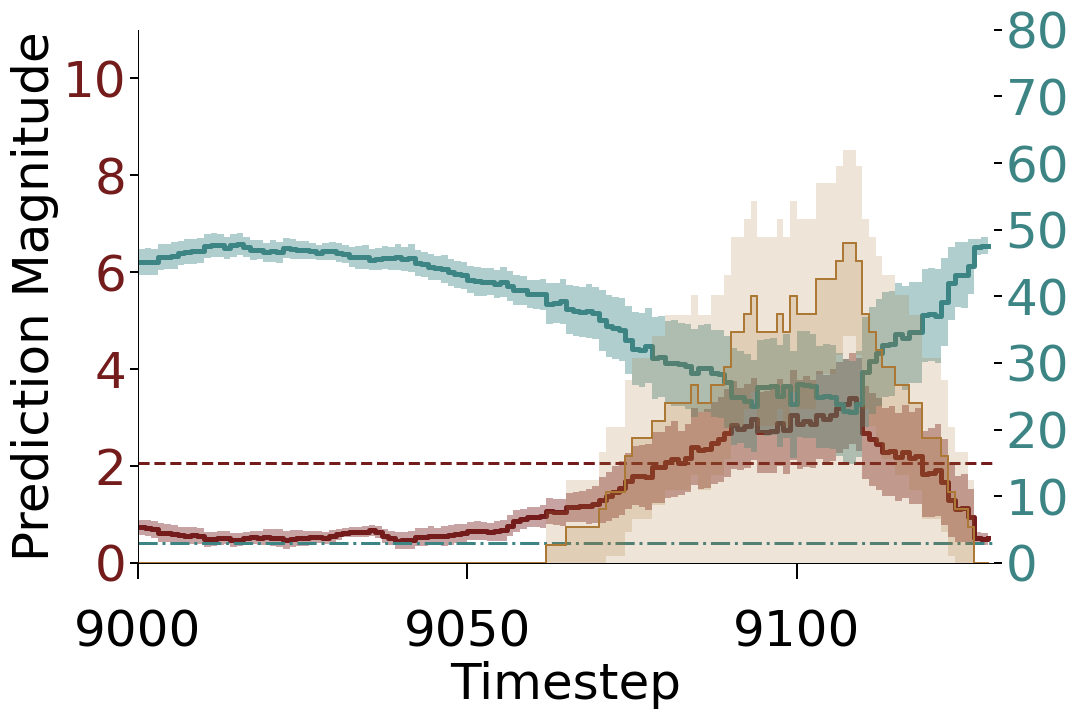}
\includegraphics[width=1.5in]{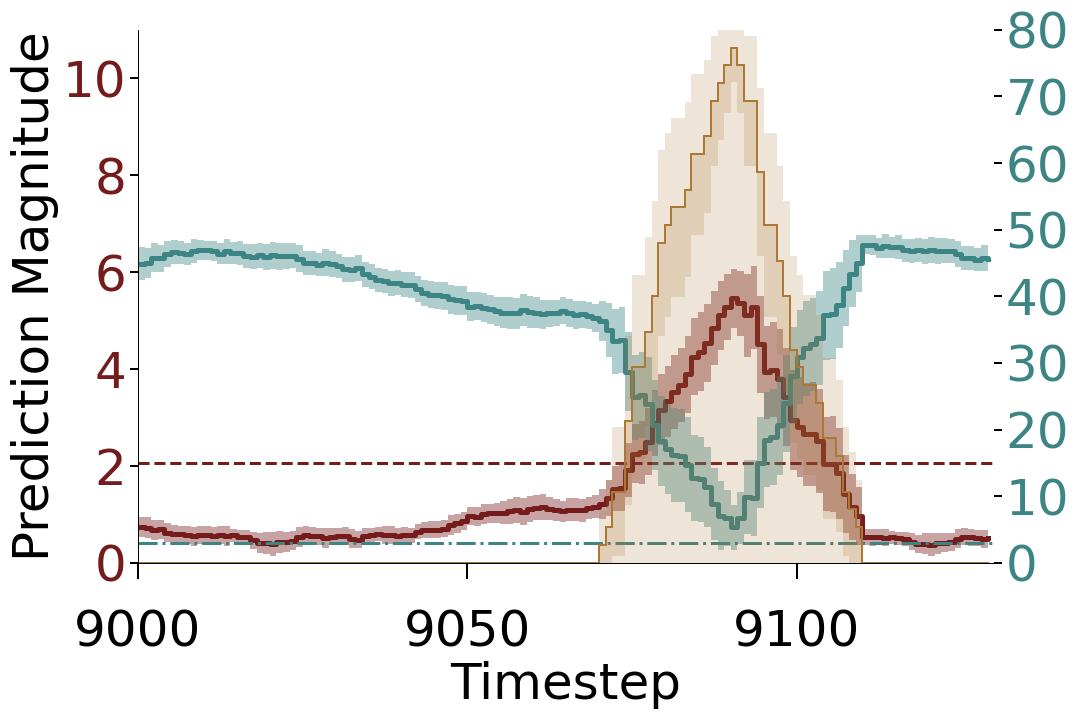}\\
(e) Oscillator + PR\\
\caption{{\bf GVF learning in the Frost Hollow domain for long inter-stimulus intervals} (100 steps, with an active period of 20 steps). In keeping with the approach of Fig. \ref{fig:nexting-short-isi-comp}, shown here is an example of prediction learning for both the accumulation (red trace) and countdown (blue trace) GVF questions with five different representations across the fixed, random, and drift conditions (three rightmost columns, hazards shown in yellow). Note the different decay constants for the tile-coded traces, selected to accommodate and highlight the effect more steps falling within the longer ISI.}
\label{fig:nexting-long-isi-comp}
\end{figure*}

As a first set of experiments, we present the way five different representations of time provide a basis for GVF-based prediction learning with regard to a continuing sequence of hazard pulses within the Frost Hollow environment, with the presence representation as the prediction target (as above, best thought of as a continuing sequence of CS/US pairing or inter-stimulus intervals from the animal learning literature).

For these of experiments, we consider the four different representations depicted in Fig.\ \ref{fig:rep-schematic} above---the bit cascade, the tile-coded trace, the oscillator, and the bias unit---with two different decay rates for the tile coded trace (one fast and one slow). For each of these representations, we study the two different types of prediction identified in Fig. \ref{fig:pred-scematic}: an accumulation prediction that rises in advance of an impending hazard, and a countdown prediction that tracks the number of steps remaining until hazard onset. Finally, we examine combinations of these conditions for both a short ISI (10 steps) and a long ISI (100 steps) between hazard onsets, in the fixed, random, drift hazard conditions introduced above (Fig. \ref{fig:fixed-random-drift-schematic}).

For this section we explicitly examine the first 10k learning steps for each run (broken down into 1k-step episodes to provide periodic periods of alignment that were helpful for analysis); we performed 30 independent runs for each different combination of conditions and hyper-parameters. Further, for this section we specifically examined the step size (learning rate) $\alpha_{gvf}=0.1$, with $\lambda_{gvf}=0.9$ and all starting GVF weight vector values in $w_0$ initialized to zero.\footnote{More detailed hyper-parameter comparisons were also conducted, and will be examined in the following section within the context of control learning experiments.} For the decay rates in the TCT representation, in all short ISI experiments (Fig. \ref{fig:nexting-short-isi-comp}), the decay constants were chosen as \( a \in \{ 0.3, 0.6 \} \). In all long ISI experiments (Fig. \ref{fig:nexting-long-isi-comp}), the interval between hazards and the length of each hazard was extended. The extended period was accompanied by modified decay constants, \( a \in \{ 0.03, 0.3 \} \).

Figures \ref{fig:nexting-short-isi-comp} and \ref{fig:nexting-long-isi-comp} show the results of prediction learning on the five different representations for the short ISI case and long ISI case, respectively, at start of tenth episode (i.e., after 9000 steps of learning), with all data averaged over the 30 independent random seeds. In these plots, the active hazard is shown via grey shading. Shown are the active bits in the representation (left column, black pixels indicate active bits). The presence representation is displayed as the 0th bit of State and the active representation bit is therefore displayed in bits 1 through 15 below the bit for the presence representation. Learned predictions are plotted in the three rightmost columns for the fixed, random, and drift conditions; prediction magnitude for countdown questions is shown in blue, and for cumulant questions in red. Shaded regions around the predictions and hazard show the standard error of the mean over all 30 runs. For reference, the dashed horizontal lines show the tokenization thresholds for Pavlovian signalling corresponding to each question. 

From Figs. \ref{fig:nexting-short-isi-comp} and \ref{fig:nexting-long-isi-comp} we see that temporal representation affects the quality and variability of predictions made by co-agents differently, and the effect is different depending on the GVF question being asked. The Bit Cascade representation enables robust predictions across all environment treatments and GVF questions, approaching for the fixed condition idealized predictions as can be computed post-hoc for these predictive questions by applying the forward view return computation in Eq. \ref{eq:gvf} at every time step (this can also been seen in the TD-error trace Figs. \ref{fig:token-fixed-gamma-comp}a and \ref{fig:token-countdown-comp}a, green trace, decreasing to zero) \footnote{We note that the idealized predictions can be computed and plotted post-hoc for all points in a recorded data set as per the return calculation in Eq. \ref{eq:gvf}; bit cascade plots for predictions in the fixed condition well approximate this true return.}. Due to the reduced rate of change in the active element later in an ISI, the tile-coded traces demonstrates aliasing that affects the prediction magnitudes close to the hazard onset; aliasing can be well seen in the visible impact of the step size on predictions that share a similar representation, presenting as a visible increments/decrements (Fig. \ref{fig:nexting-short-isi-comp}) or ramps (Fig. \ref{fig:nexting-long-isi-comp}) of fixed slope in the value of the prediction---a hallmark of tracking and aliasing. 

\afterpage{\clearpage}

The bias unit representation exhibited what we expected for a representation that is largely tracking and not forecasting a signal of interest: unlike the form of an ideal prediction for this GVF question, the accumulation prediction with a bias unit representation {\em falls} as it approaches the oncoming stimulus, while the countdown prediction {\em rises} in anticipation of the stimulus---{\em the bias unit representation generates a mirror image in time of the ideal predictions} shown in Figs. \ref{fig:nexting-short-isi-comp}a and \ref{fig:nexting-long-isi-comp}a. Both show simply that the cumulant is being added to prior expectations, leading to a historical trace of the cumulant and not in fact predictive information. This is akin to the latter part of the ISI for TCT representations, wherein the last active bit of the representation functions as a bias unit and shows a similar converse profile to expected ideal or computed predictions.

The oscillator representation behaved identically to the bit cascade on the fixed environment, but lost its representational power when the environment is stochastic and was found to have no appreciable predictive utility in the random and drift conditions, remaining far from the ideal predictive values (consistently high TD-error) and not demonstrating anticipatory behaviour. In other words, the oscillator representation is as expected able to act as a perfect time keeper when wrapping occurs at exactly the period of the temporal signal being tracked. Wrapping that is not coincident with period of the temporal signal of interest results in aliasing.

\subsection{How do Different Representational\\Choices Impact Token Generation in\\Pavlovian Signalling?}
 
\label{PavSigExps}

As a second set of experiments, we illustrate the way that these different predictions are mapped via a fixed threshold to the generation of tokens for use in Pavlovian signalling, creating intuition as to how the fixed, random, and drifting conditions impact token generation over the time-course of multiple learning episodes. These experiments follow the same experimental approach as the prior section.

Co-agents produced a single binary token used by the agent to maximize reward. As described in Sec. \ref{sec:pavlov-signalling}, predictions were transformed into tokens by use of a threshold function. Different threshold functions were used for the accumulation and countdown GVF questions. The value of predictions about accumulation questions rose in anticipation of stimulus onset, while the value of predictions of cumulant questions decreased in response the approaching the stimulus. For the accumulation question, the token was produced by evaluating the prediction p, against the threshold h, and was returned as logical value of the expression (p > h). Similarly for the countdown question, the token was produced as (p <= h).\footnote{ More complex tokenization strategies (e.g., vector tokenization) are interesting, but are outside the scope of the present study; will be briefly describe anecdotal results for multi-token signalling in the discussion in Sec. \ref{sec:discussion-multitoken}.} In our experiments in this section and those that follow, we selected threshold values based on the advance notice needed for an agent in the abstract domain to avoid hazards and thus accumulate heat for reward, as can be arrived at empirically or from the idealized return of Eq. \ref{eq:gvf}: $\tau=2.05$ for the accumulation GVF and $\tau=3.0$ for the countdown GVF---e.g., for a countdown GVF specified by $C$ and $\gamma$, when the expected steps until the hazard $V_t=3.0$ an agent would have the needed 3 steps to move from the heat region to safety.

\begin{figure*}[!th]
\centering
\includegraphics[width=2in]{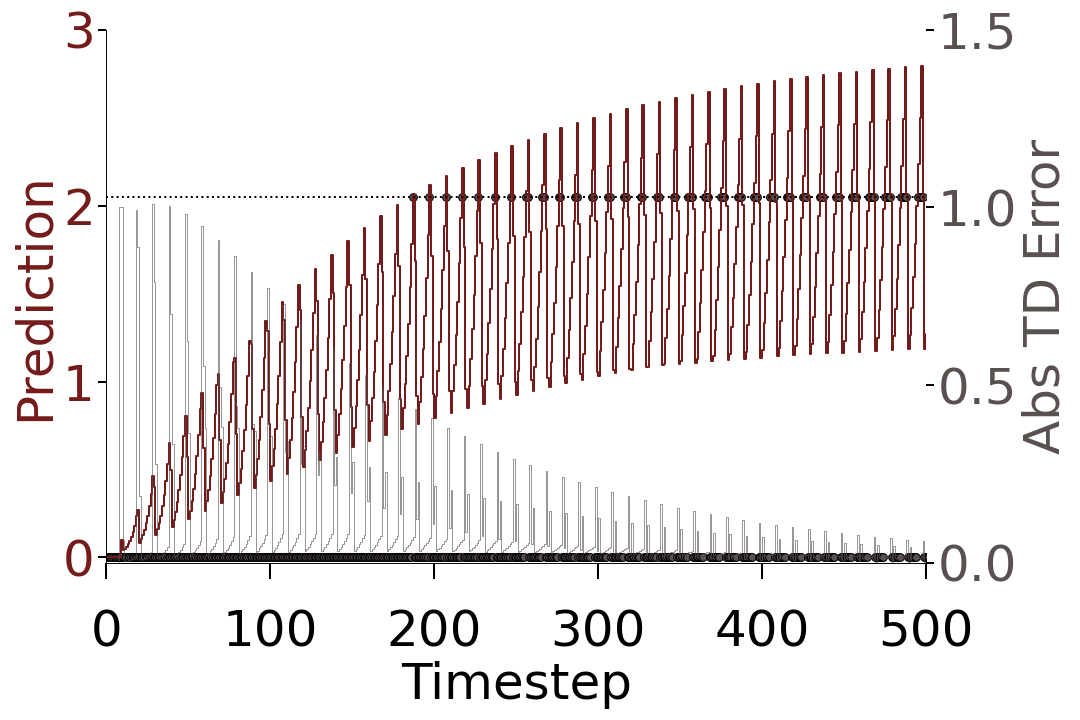}
\includegraphics[width=2in]{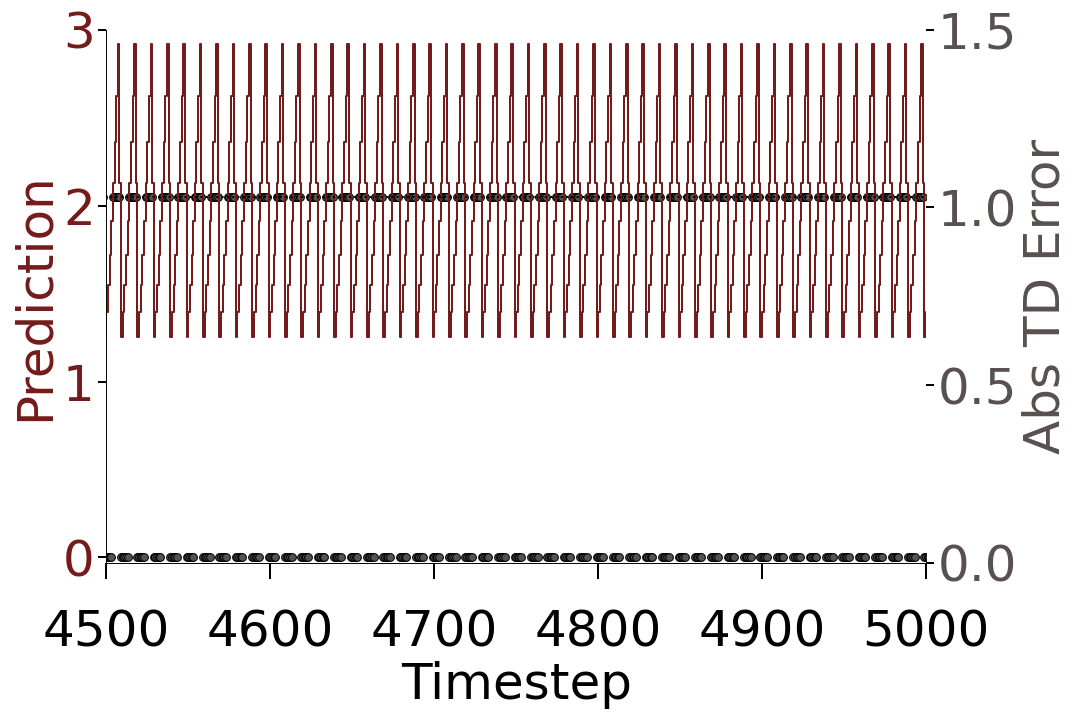}
\includegraphics[width=2in]{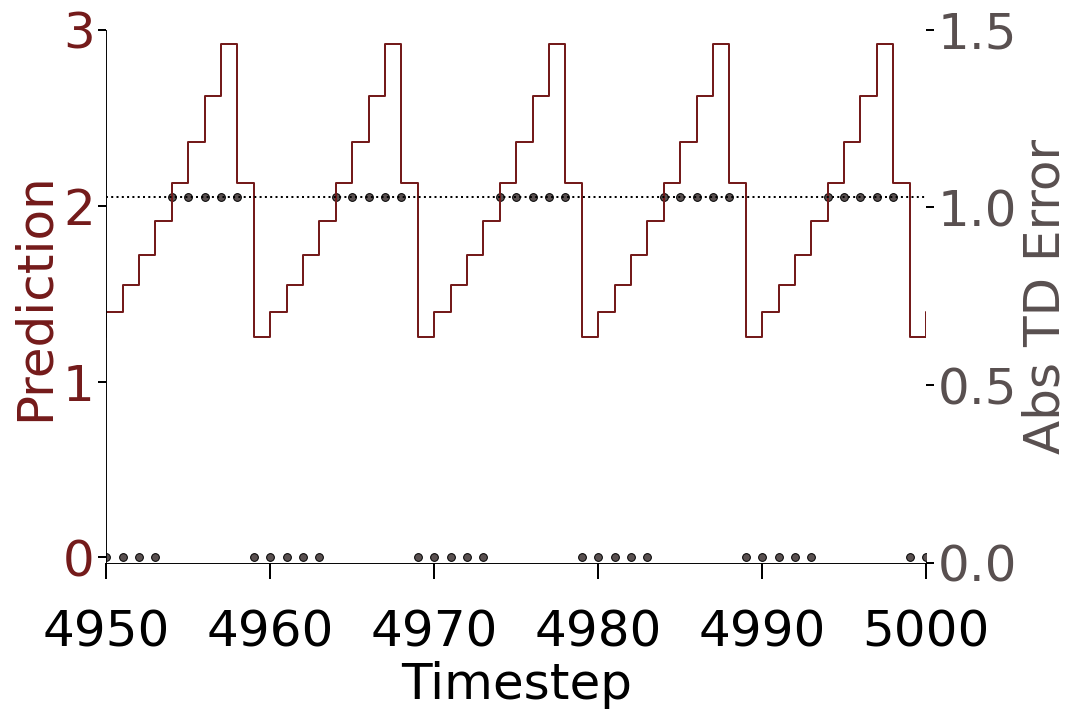}
\\
(a) FIXED\\
\ \\
\includegraphics[width=2in]{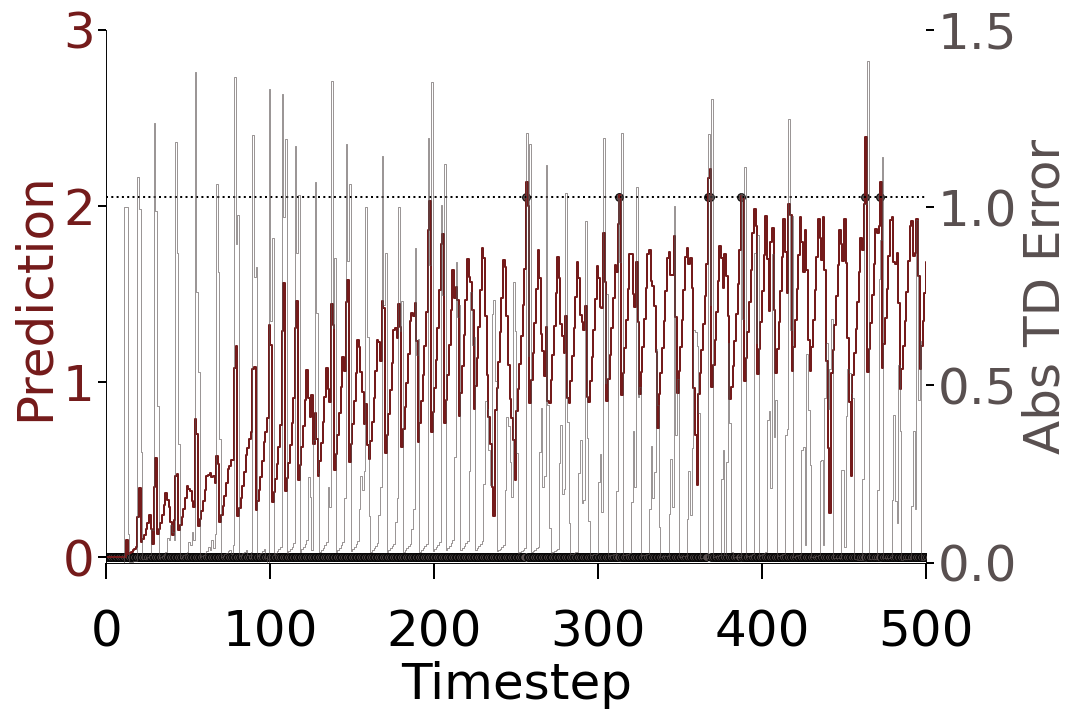}
\includegraphics[width=2in]{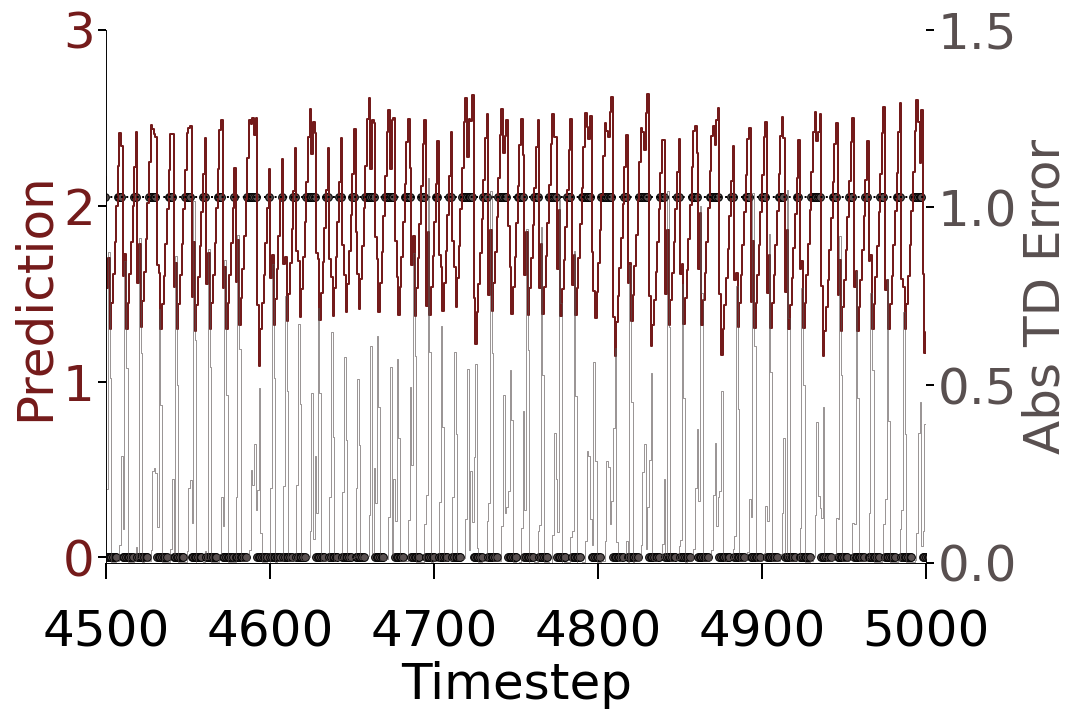}
\includegraphics[width=2in]{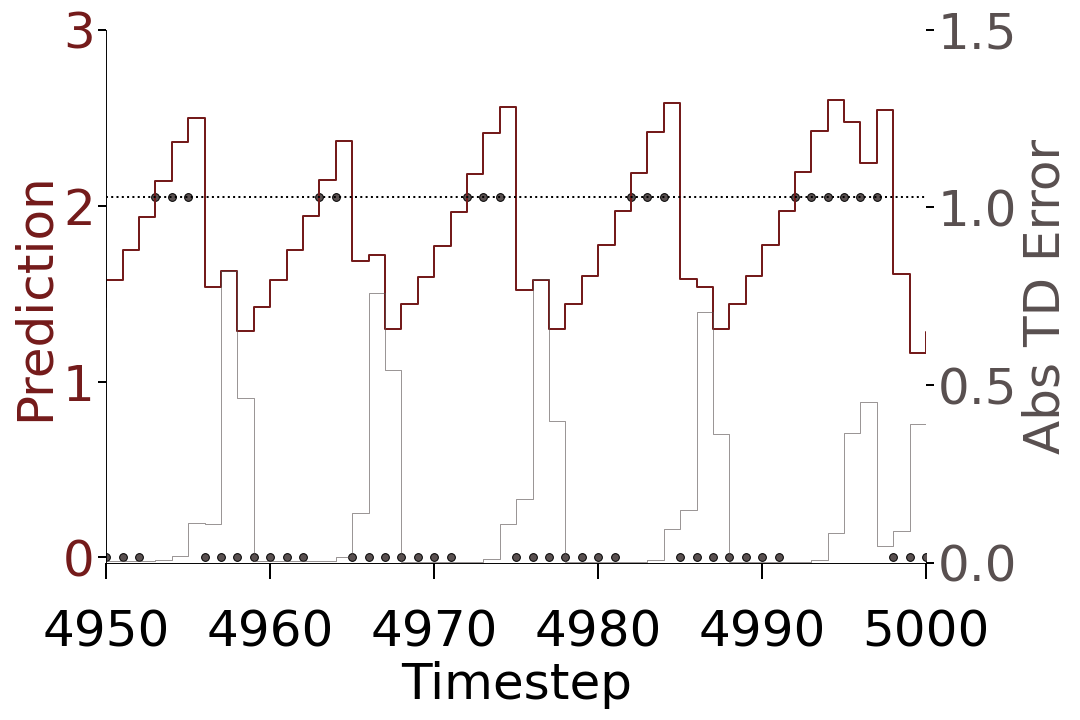}\\
(b) RANDOM\\
\ \\
\includegraphics[width=2in]{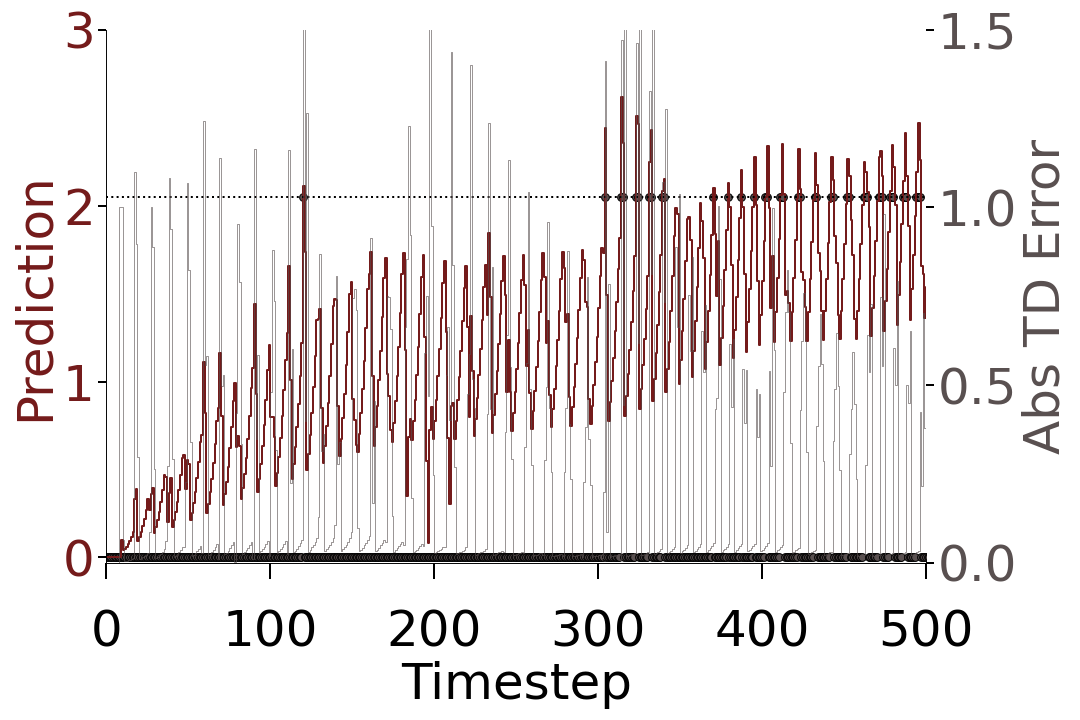}
\includegraphics[width=2in]{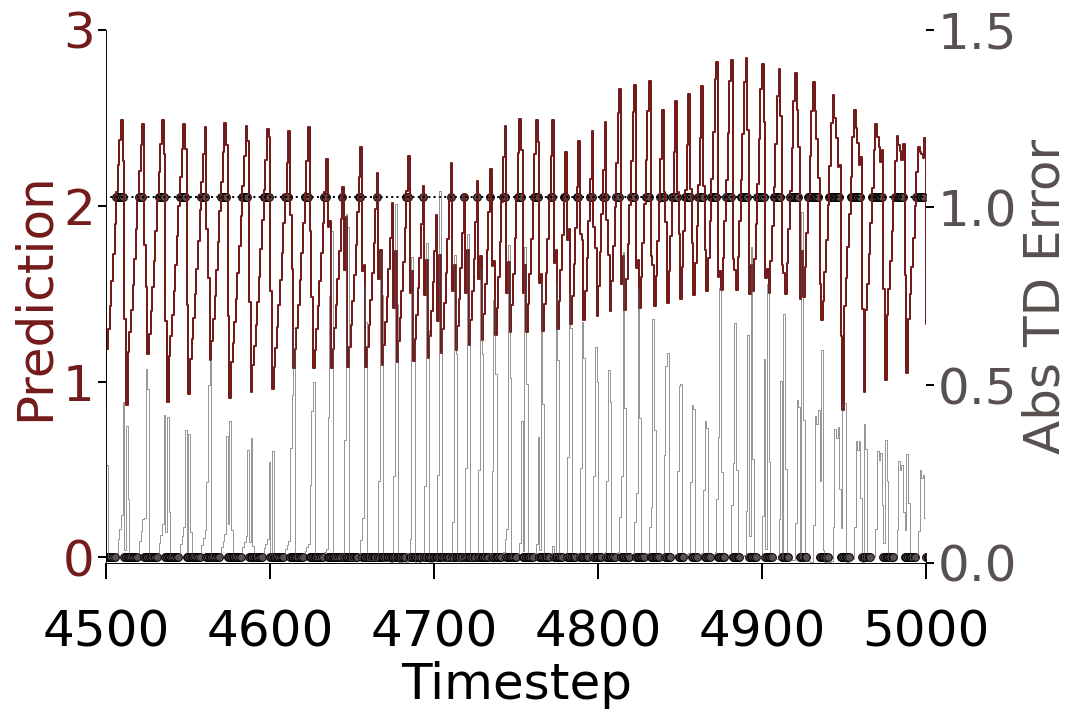}
\includegraphics[width=2in]{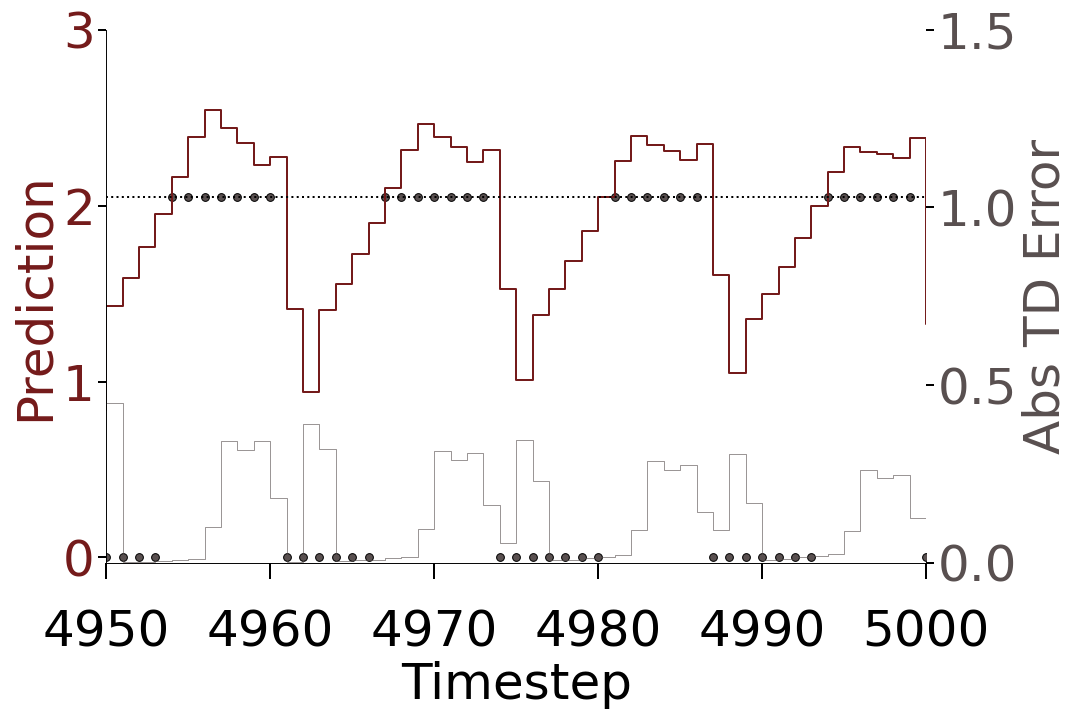}\\
(c) DRIFT 
\caption{{\bf Comparison of accumulation predictions to tokens} (red traces and black circles, respectively) being generated by a threshold (2.05) in the (a) fixed, (b) random, and (c) drift conditions over the first 5000 steps of learning for one representative trial. Predictions are shown for early learning (left column), and then steady state learning at two different zoom levels (right and middle columns). Predictions are cross-plotted with absolute temporal-difference error (grey).}
\label{fig:token-fixed-gamma-comp}
\end{figure*}

\begin{figure*}[!th]
\centering
\includegraphics[width=2in]{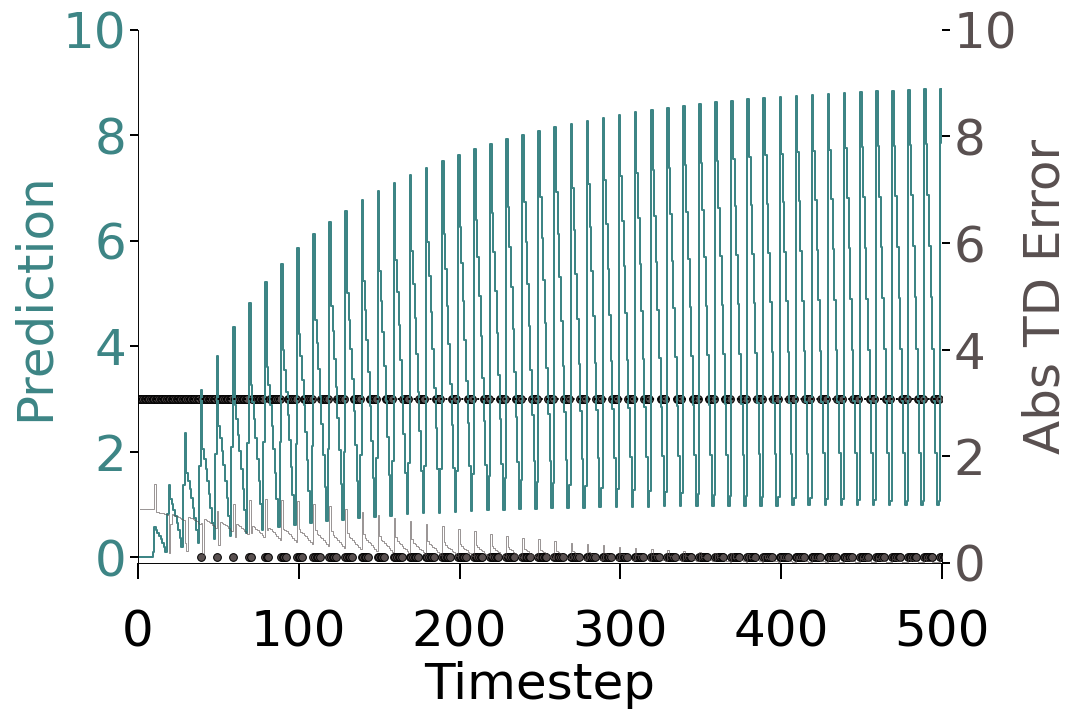}
\includegraphics[width=2in]{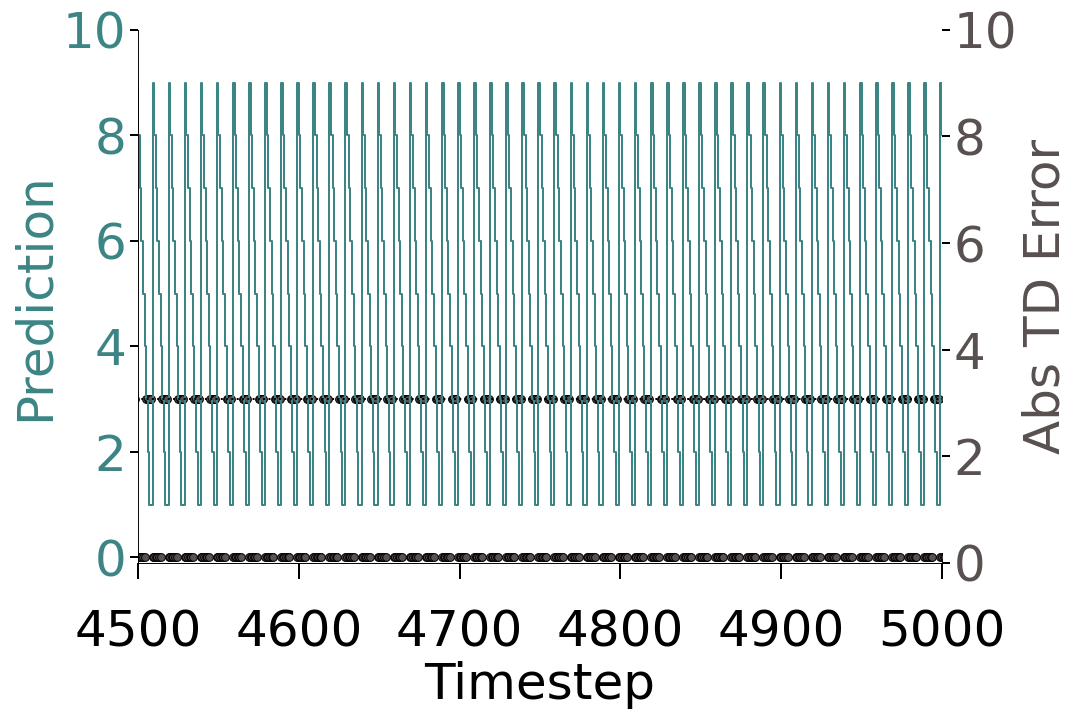}
\includegraphics[width=2in]{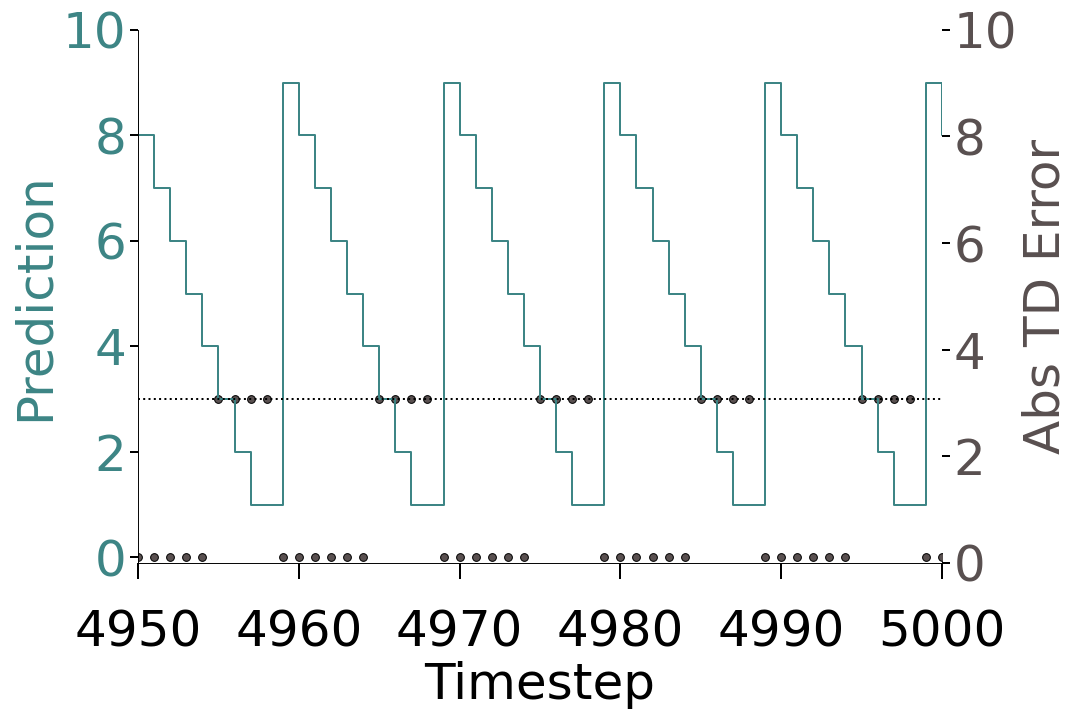}
\\
(a) FIXED\\
\ \\
\includegraphics[width=2in]{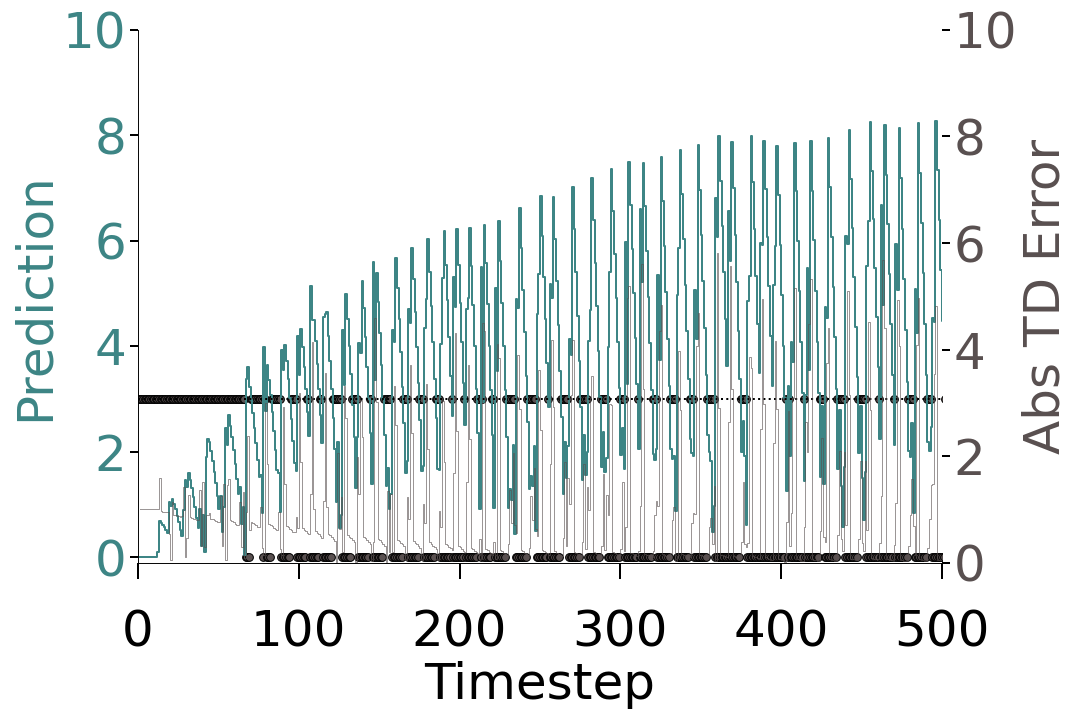}
\includegraphics[width=2in]{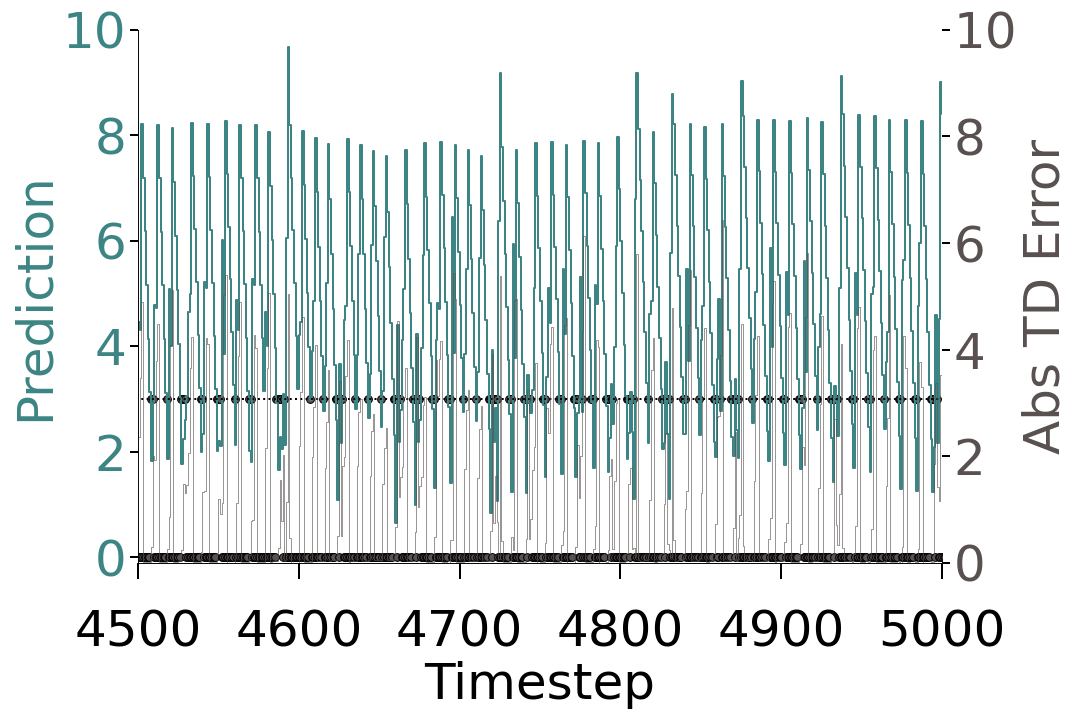}
\includegraphics[width=2in]{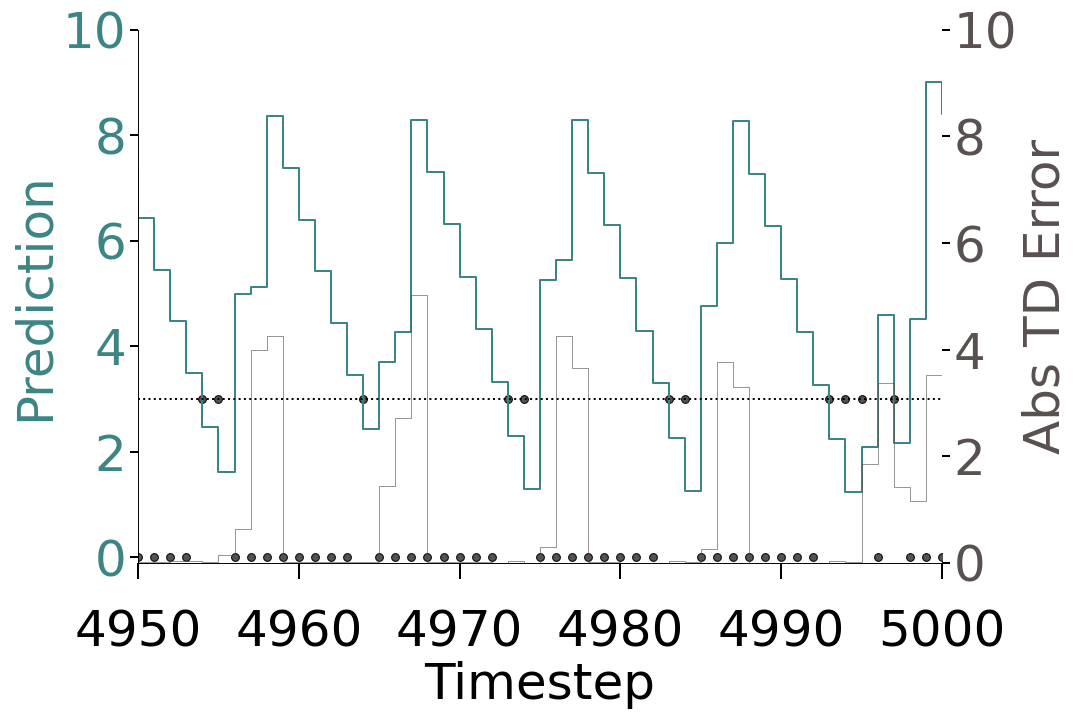}\\
(b) RANDOM\\
\ \\
\includegraphics[width=2in]{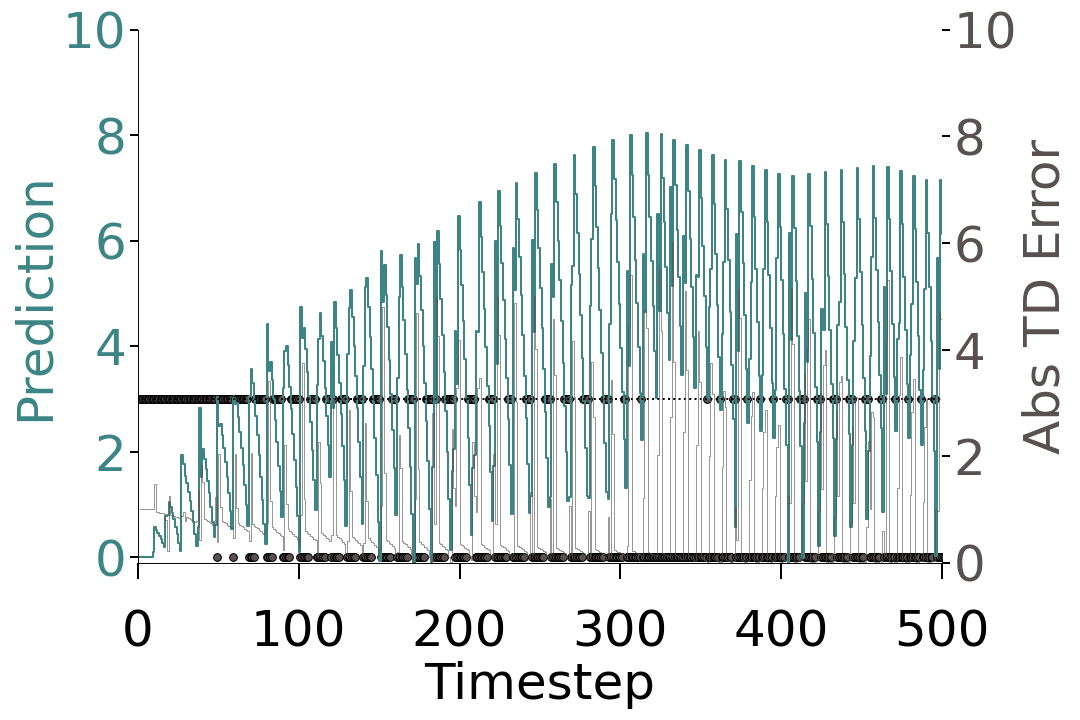}
\includegraphics[width=2in]{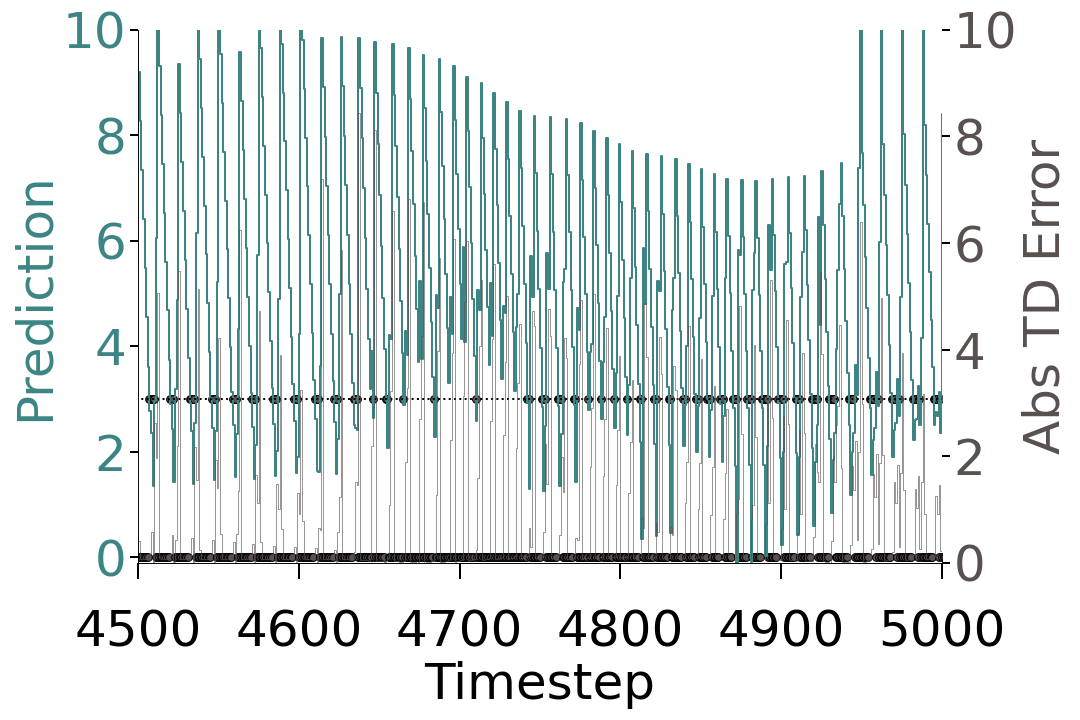}
\includegraphics[width=2in]{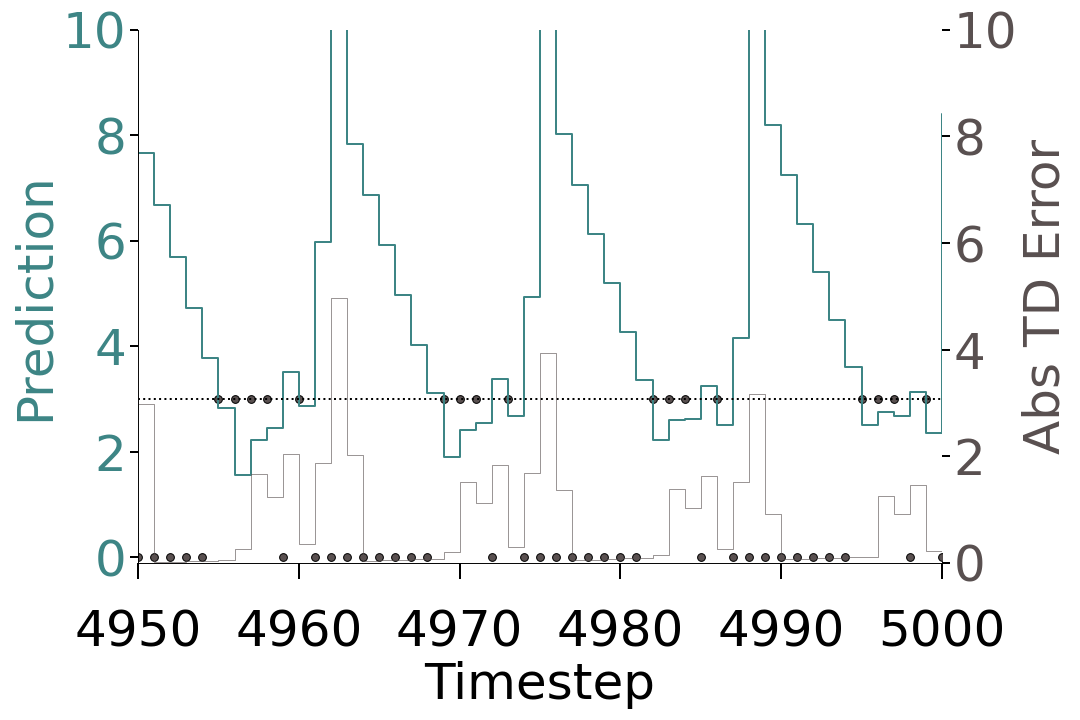}\\
(c) DRIFT 
\caption{{\bf Comparison of countdown predictions to tokens} (blue traces and black circles, respectively) being generated by a threshold (3.0) in the (a) fixed, (b) random, and (c) drift conditions over the first 5000 steps of learning for one representative trial. Predictions are shown for early learning (left column), and then steady state learning at two different zoom levels (right and middle columns). Predictions are cross-plotted with absolute temporal-difference error (grey).}
\label{fig:token-countdown-comp}
\end{figure*}

As shown in Figs. \ref{fig:token-fixed-gamma-comp} and \ref{fig:token-countdown-comp}, for the fixed condition the learned predictions (red traces) stabilized within the first $\sim$500 steps of learning, with the TD-error (green traces) decreasing to almost zero over this time and remaining near zero for the remainder of the 5000 steps considered in this experiment. For the thresholds studied here, this resulted in consistent token generation (black dots) after $\sim$200 steps for the accumulation question (Fig. \ref{fig:token-fixed-gamma-comp}); for the countdown question, the initialization of the value function to zero resulted in what might be considered overly cautious token generation: tokens relating to impending hazards were generated immediately, due to predictions of step-until-hazard being lower than the threshold, and began including tokens relating to a lack of hazard only after roughly 100 steps of learning. By $\sim$500 steps of learning, the countdown GVF-based tokenization was also stable and consistently presented signals of the impending hazard. The precise relationship between predictions and tokens can be seen in greater detail for late learning in Figs. \ref{fig:token-fixed-gamma-comp} and \ref{fig:token-countdown-comp}.

In contrast to the stability of predictions and tokens shown for the fixed conditions, predictions and tokens in the random and drift conditions of Figs. \ref{fig:token-fixed-gamma-comp} and \ref{fig:token-countdown-comp} were not surprisingly much more varied, with consistently high TD-errors throughout learning due to the GVF learning algorithm tracking the target concept through a process of continual learning (notable in late learning in the zoomed-in panels of Figs. \ref{fig:token-fixed-gamma-comp}c and \ref{fig:token-countdown-comp}c). Of note, for the GVF learning rate of $\alpha_{GVF}=0.1$ shown in these figures, we see that, largely in the countdown case as opposed to the accumulation question case, there are cases even later in learning where tracking the cumulant leads to tokens not being generated when the updated predictions are compared with a fixed threshold that no longer accurately captures the relationships between a prediction and the time until an impending hazard. We also see more protracted shifts in prediction magnitude in the drift condition, as would be expected if the GVF was appropriately tracking the question cumulant return of interest. These deflections were, as expected, less for lower GVF learning rates.

Taken as whole, we see token generation is consistent with a GVF learner that is engaging in a process of continual learning and the adaptation of predictions. Especially for the accumulation case, these results suggest that using learned GVF predictions for generating tokens in Pavlovian signalling will provide viable state information for Frost Hollow control learning agents; this expectation will be explored in the section that follows.

\section{Control Learning in Frost Hollow with Pavlovian Signalling}
\label{sec:control-experiments}

\begin{figure*}[!th]
\centering
\includegraphics[height=1.4in]{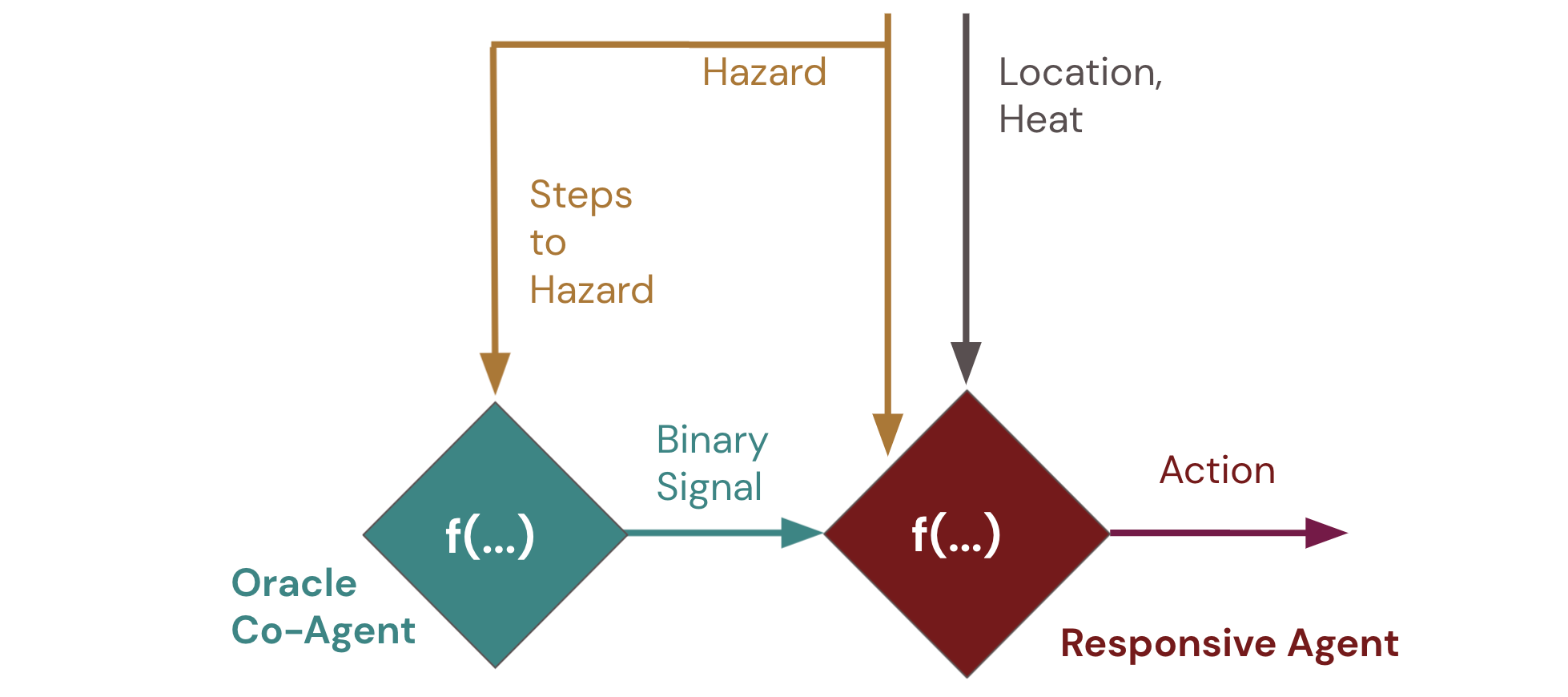}
\includegraphics[height=1.4in]{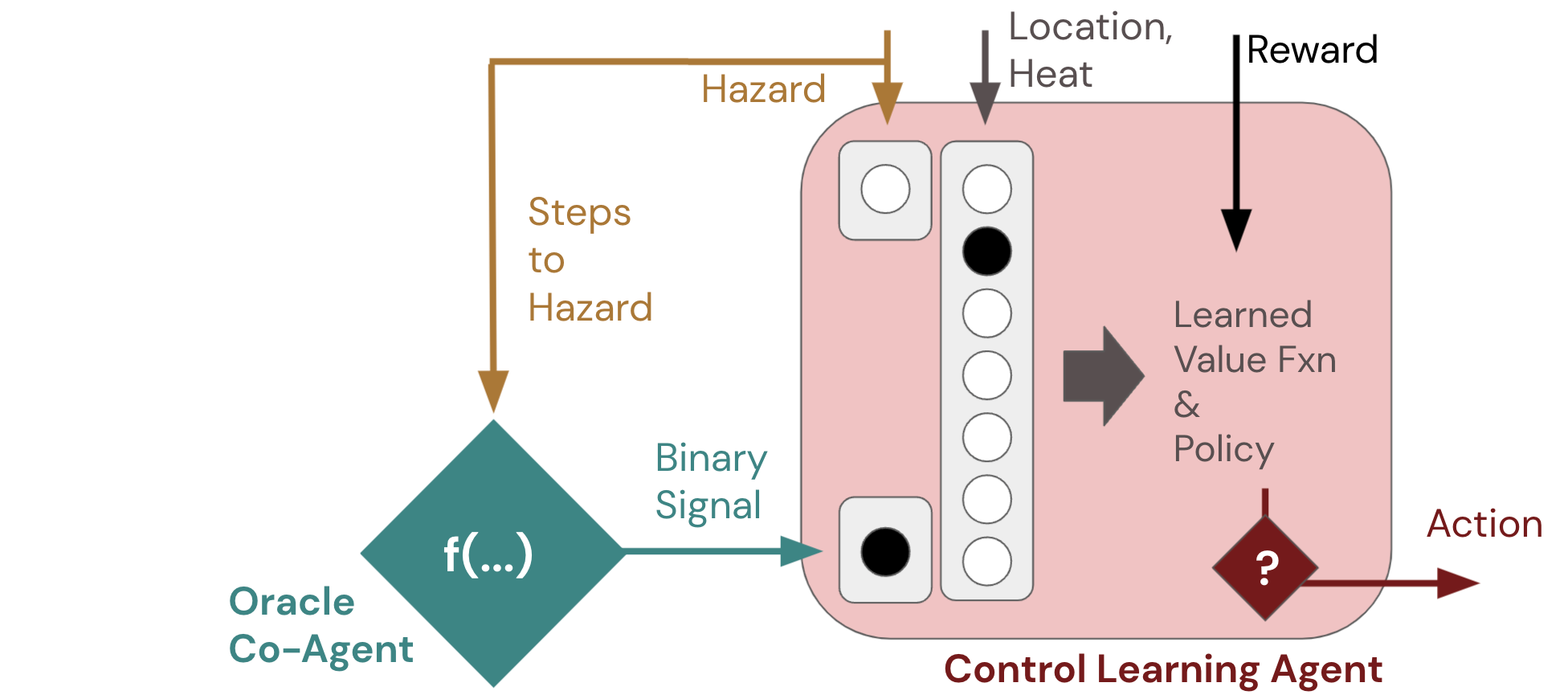}\\
(a) \hfil (b)\\
\ \\
\includegraphics[height=1.4in]{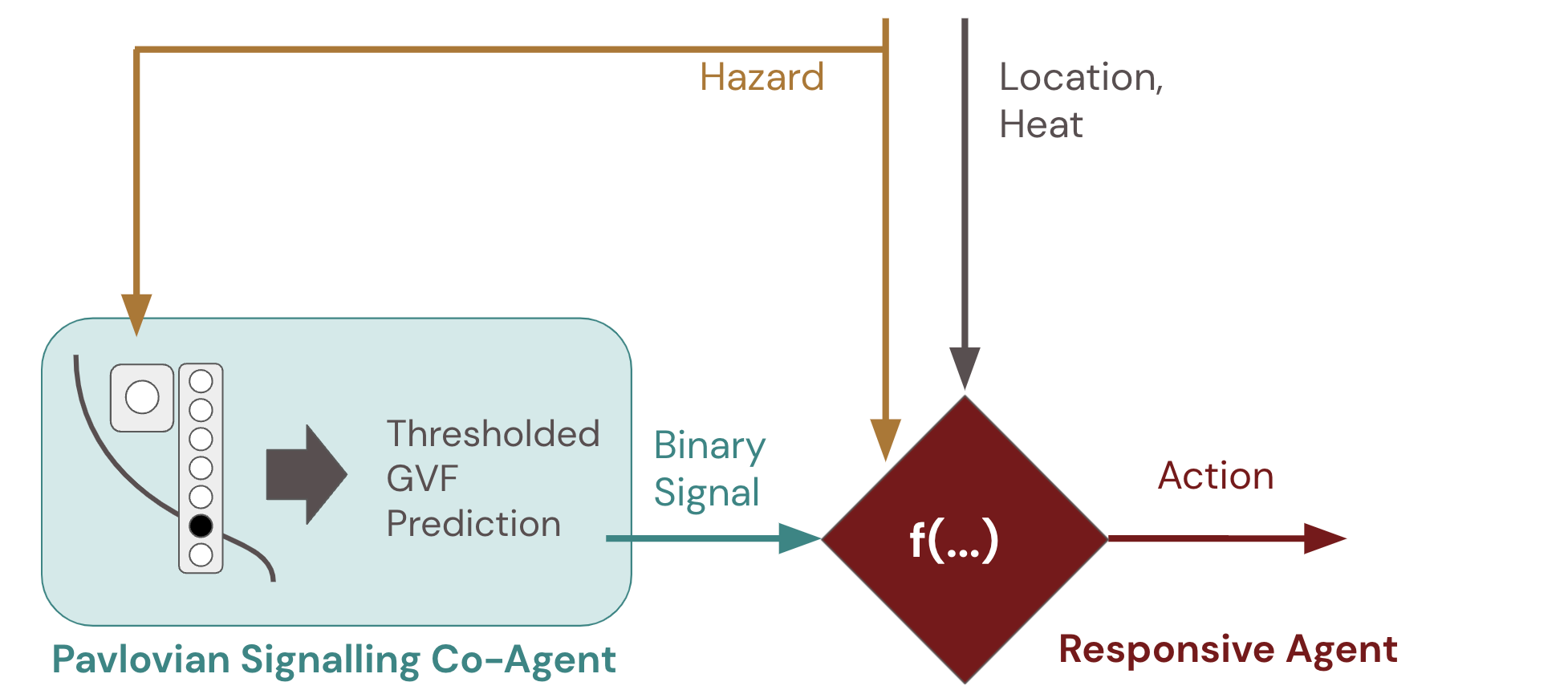}
\includegraphics[height=1.4in]{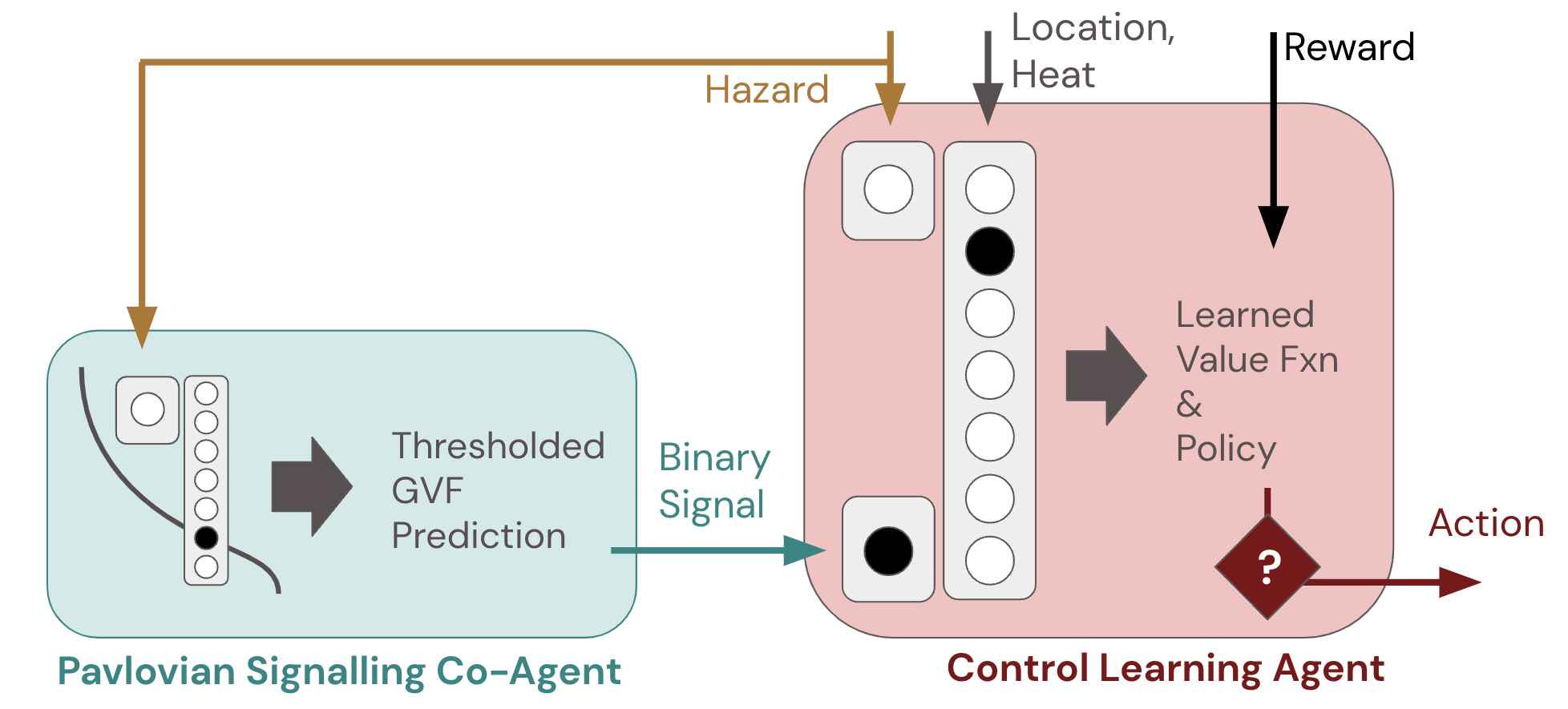}\\
(c) \hfil (d)
\caption{{\bf Combinations of agents and co-agents we consider in this study}. Shown are the integration of (a) an oracle co-agent with a hand-designed responsive agent, (b) an oracle co-agent with a control learning agent, (c) a learning Pavlovian signalling co-agent coupled with a responsive agent, and (d) the full learning case: a Pavlovian signalling co-agent that learns to predict a hazard and passes tokens to a control learning agent.}
\label{fig:agent-agent-schematic}
\end{figure*}

We now turn to the second empirical section of this paper. In the previous empirical section, we identified key differences in how temporal representations were able to support the prediction learning with respect to phenomena unfolding in time, and the way these predictions might be turned into tokens that could be used by a second agent or a discrete decision-making unit within a single agent. The intuition built in this previous section regarding predictions, representations, and Pavlovian signalling will now be used in the study of agent-agent interaction, and specifically to {\em understand how tokens generated through a process of Pavlovian signalling can be used to support control learning} in the abstract Frost Hollow domain when two interacting agents have different perceptual and control interactions within that domain, and to {\em understand how different temporal representation choices impact this process of agent-agent learning and interaction}.

Building on the previous section, the objective of this section is therefore to now address the interactions between space and time, where control or policy learning by an agent depends in one way or another on an agent's location in Frost Hollow and also on learned predictions about an impending hazard. We first examine the five different representations presented in Sec. \ref{sec:nexting-experiments} in terms of their use in generating tokens for a control learning agent, including comparisons with oracle prediction and control learners. We then examine the impact different control learning choices have or do not have on the learning process, and finish with an examination of how the learned polices and performance change with respect to changes in the difficulty of the Frost Hollow environment.

\subsection{Agent-Agent Combinations}
 
For this section we specifically consider the case where we have two interacting learning machines that are situated together within the Frost Hollow domain and that must interact so as to collect reward. While it is natural to think of these two learning machines as two tightly coupled parts of a single, larger learning machine (e.g., the analogy we drew in earlier sections to cerebral and cerebellar interactions) here we depict them as concrete, independent learning machines. We do this so as to be able to clearly and concretely describe the comparisons we make between different conditions, perceptual spaces, and environmental affordances exposed to each agent without creating assumptions about other parts of a single-agent composite architecture, and also to closely parallel the human-machine interactions virtual reality experiments presented in Sec. \ref{sec:vr-experiments}, our final empirical study.
 
We here denote the two learning machines under consideration as the {\em agent} and the {\em co-agent}. The agent is responsible for policy learning so as to solve the Frost Hollow control challenge, while the co-agent is responsible for prediction learning so as to provide accurate and relevant forecasts about the Frost Hollow hazard stimulus. As shown in Fig. \ref{fig:agent-agent-schematic}, and mirroring the representations examined in the Sec. \ref{sec:nexting-experiments}, the only state input to the co-agent in our experiments is the presence representation, a single bit that indicates the presence or absence of the hazard on a given timestep. All other state information for the co-agent is derived from its internal temporal representations, if any. The output of the co-agent is a single bit: a token that takes either the value of 1 or 0 depending on the magnitude of the predictions generated by the co-agent. In contrast, the agent is a control learner that takes as input this token from the co-agent, the presence representation for the hazard, the reward from the environment, and also its location (one of seven states) in the Frost Hollow abstract domain (also shown in Fig. \ref{fig:agent-agent-schematic}). The agent produces, as output, an action that can either be to move up, down, or stay stationary in the world. In all but one of the cases considered, we note that the token that passes between the agent and co-agent is grounded by the co-agent in the GVF question parameters and the tokenization threshold, but is {\em ungrounded} by the agent in its receipt of the Pavlovian signal (c.f., Sec. \ref{sec:pavlov-signalling}). We now define the exact form of the two co-agent classes and two agent classes we compare in our experiments. 
 
{\bf Oracle Co-agent}: the first, most rudimentary co-agent we consider is the oracle case: a comparison baseline that receives perfect information about the number of time steps remaining until the next hazard (it has the ability to see with perfect accuracy into the future hazard states of the environment); across fixed, random, and drift conditions, this results in an integer that is mapped to an output token according to the threshold described for the countdown GVF in Sec. \ref{sec:nexting-experiments} and Fig. \ref{fig:token-countdown-comp} (in this case, generating a token of 1 if the number of steps to the hazard is three steps or less).

{\bf Pavlovian Signalling Co-agent}: as the primary co-agent of interest, we also consider the Pavlovian signalling case: a co-agent that learns to predict the hazard presence representation input bit based on the presence representation and the state of one of the five representations examined in Sec. \ref{sec:nexting-experiments}; the form of the GVF prediction learned by this co-agent is, as above, either that of the accumulation question, or the countdown question. Following the tokenization depicted in Figs. \ref{fig:token-fixed-gamma-comp} and \ref{fig:token-countdown-comp}, this learned prediction is mapped according to a fixed threshold to the single output bit that is provided to the control learning agent. 

{\bf Responsive Agent}: Similar to the oracle co-agent, we also consider a control learning equivalent with a known, stable policy that takes into account the specific semantics of the token being provided by the co-agent (i.e., an agent that has by design also grounded its interpretation of the 0 or 1 token value in terms of an impending hazard or the lack of an impending hazard). This responsive agent has a hard-coded understanding of the location of the goal region, the safe region, and the hazard region; it reacts in a procedural way to the bit coming from the co-agent, such that its policy responds to impending danger by moving towards and staying in the safe region, while in the absence of a token signalling danger it moves toward and stays on the goal region. We note that, for some configurations of the Frost Hollow environment, this hand-designed policy is not optimal, but it serves as a consistent control baseline for the harder difficulty levels of the domain we primarily examine.

{\bf Control Learning Agent}: our primary learning agent considered in this work is a straightforward, on-policy reinforcement learner that has a tabular state representation (i.e., no function approximation, with a unique state for each possible configuration of the agent's position in the environment, the hazard's presence, collected heat, and the token provided by the co-agent). %

The primary control learning agent for this work follows the standard Expected Sarsa($\lambda$) learning algorithm 
\citep{sutton2018}, chosen to minimize complexity on the agent side and focus on guiding our understanding of paired agent/co-agent learning and the role of temporal abstractions in learning dynamics. For reference here, Expected Sarsa($\lambda$) calculates its temporal-difference error $\delta_t$ via a summation over all actions weighted by their probability under the current policy $\pi$ for the future state $x(S_{t+1},A_{t+1})$, as combined with the reward $R_{t+1}$ and the action values for the current state $x(S_{t},A_{t})$, and uses this error and eligibility traces $e_t$ to updates the weights $w_t$ associated with its action values as follows:
\begin{align*}
e_t &\gets e_{t-1} + x(S_{t},A_{t})\\
\delta_t &\gets R_{t+1} + \gamma \sum_a [\pi(a|S_{t+1}) w_t ^\intercal x(S_{t+1},A_{t+1})]\\
w_{t+1} &\gets w_t + \alpha \delta_t e_t\\ 
e_{t} &\gets \gamma \lambda e_{t}
\end{align*}
Here $\alpha$ is the step size for learning weights, $\lambda$ is the eligibility trace decay rate, and $\gamma$ is the discounting rate applied to future action values; action values used in action selection for a given state $S_t$ and action $A_t$ are approximated via the linear combination $Q(S_t,A_t) = w_t ^\intercal x(S_t,A_t)$, where action values were optimistically initialized and were selected on each step according to epsilon-greedy action selection. We examined the performance of co-agents with countdown and accumulation GVF questions with bias, bit cascade, and tile-coded trace representations with GVF learning rates $\alpha_{gvf} \in \{0.01, 0.1\}$ to study fast and slow tracking of non-stationarity in the environment. Control learning algorithm parameters were determined via empirical sweeps, with results below shown for the best-case values of $\alpha=0.01$, with exploration via an $\epsilon$-greedy exploration policy, $\epsilon \in \{0.01, 0.1\}$, and optimistic initialization with weights initialized to 1.0. As described in Sec. \ref{PavSigExps} above, we used signalling thresholds of $\tau=2.05$ for co-agents with accumulation GVFs and $\tau=3.0$ for co-agents with countdown GVFs; token generation with respect to prediction magnitudes and these thresholds can be seen in Figs. \ref{fig:nexting-short-isi-comp}--\ref{fig:token-countdown-comp}.

 \subsection{How Does a Co-agent's Temporal Representation Impact an Agent's Ability to Learn the Frost Hollow Domain?}

\begin{figure*}[!th]
\centering
{\bf First 800 Episodes} \hfil {\bf All 5000 Episodes}\\
$\alpha_{gvf}=0.01$ \hfil $\alpha_{gvf}=0.1$  \hfil $\alpha_{gvf}=0.01$ \hfil $\alpha_{gvf}=0.1$\\
\ \\
\includegraphics[width=1.6in]{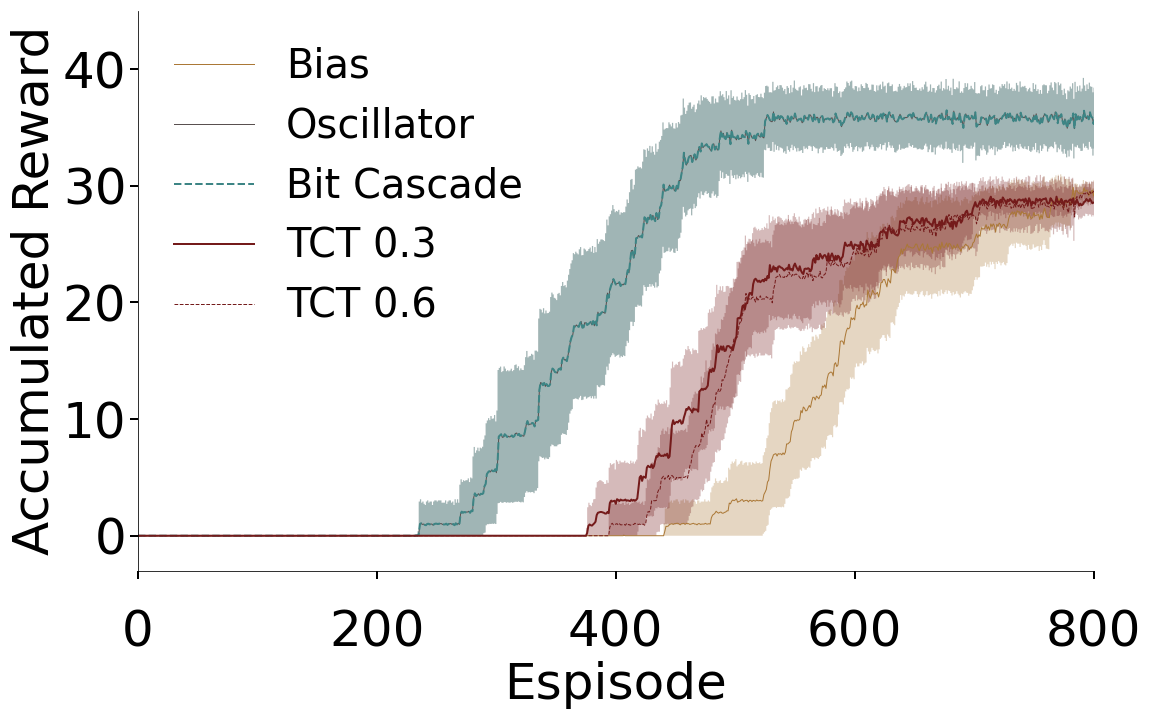}
\includegraphics[width=1.6in]{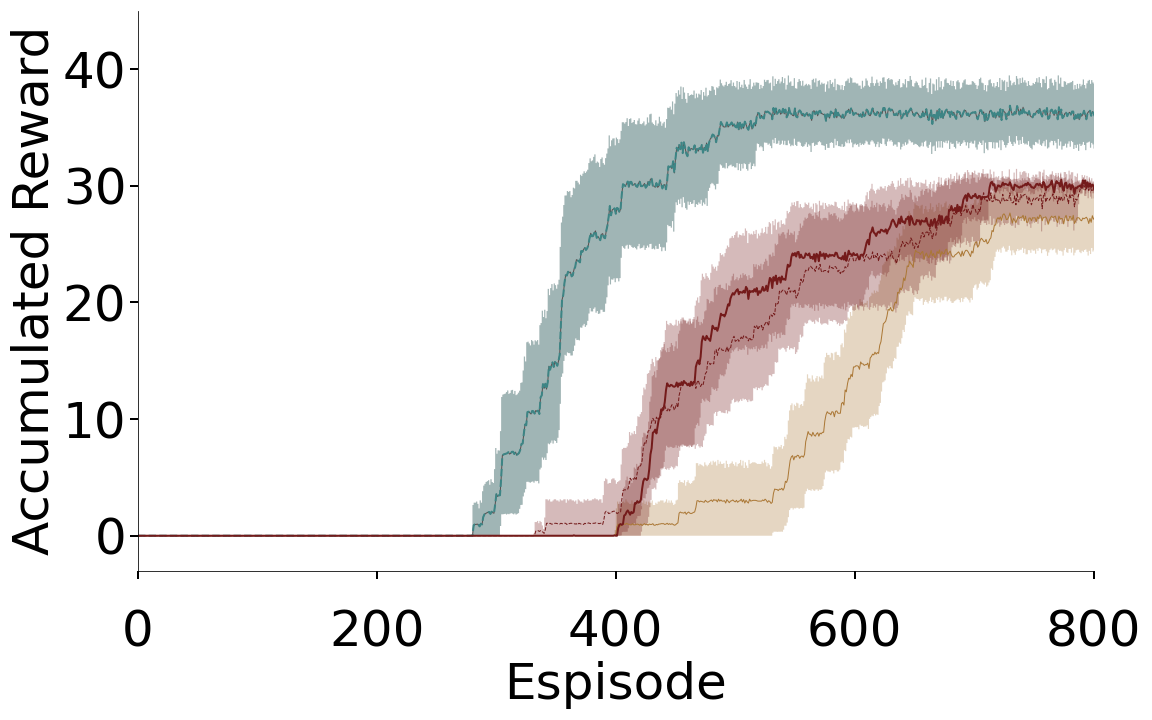}
\includegraphics[width=1.6in]{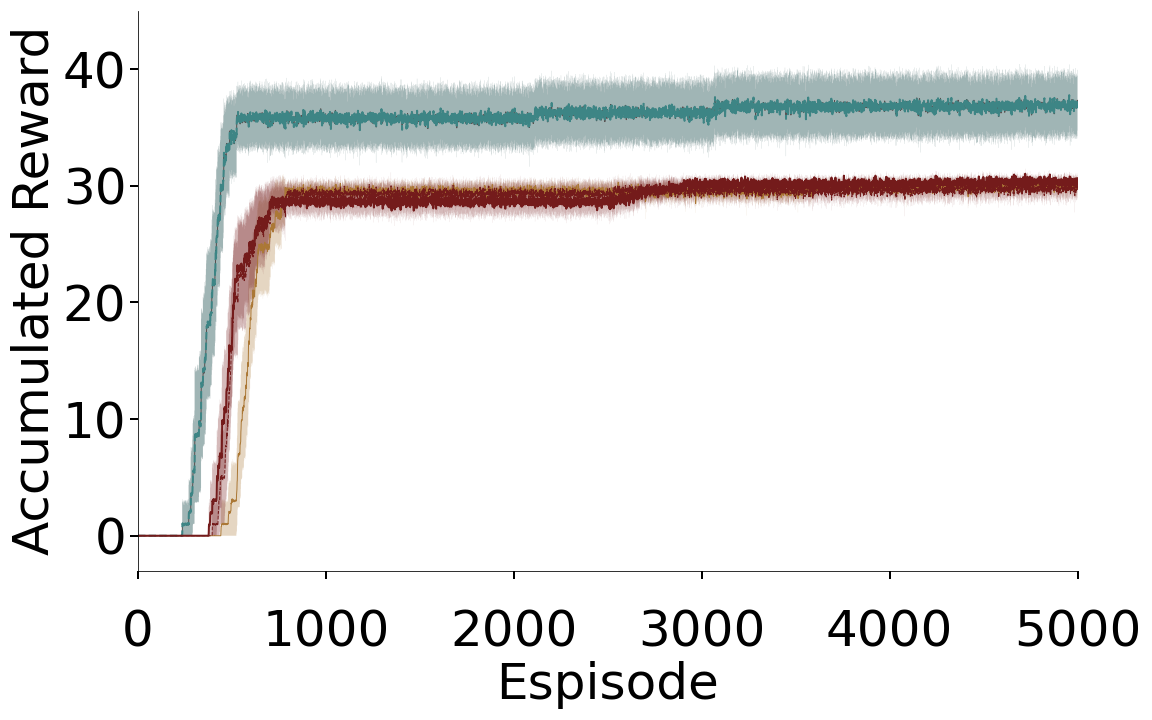}
\includegraphics[width=1.6in]{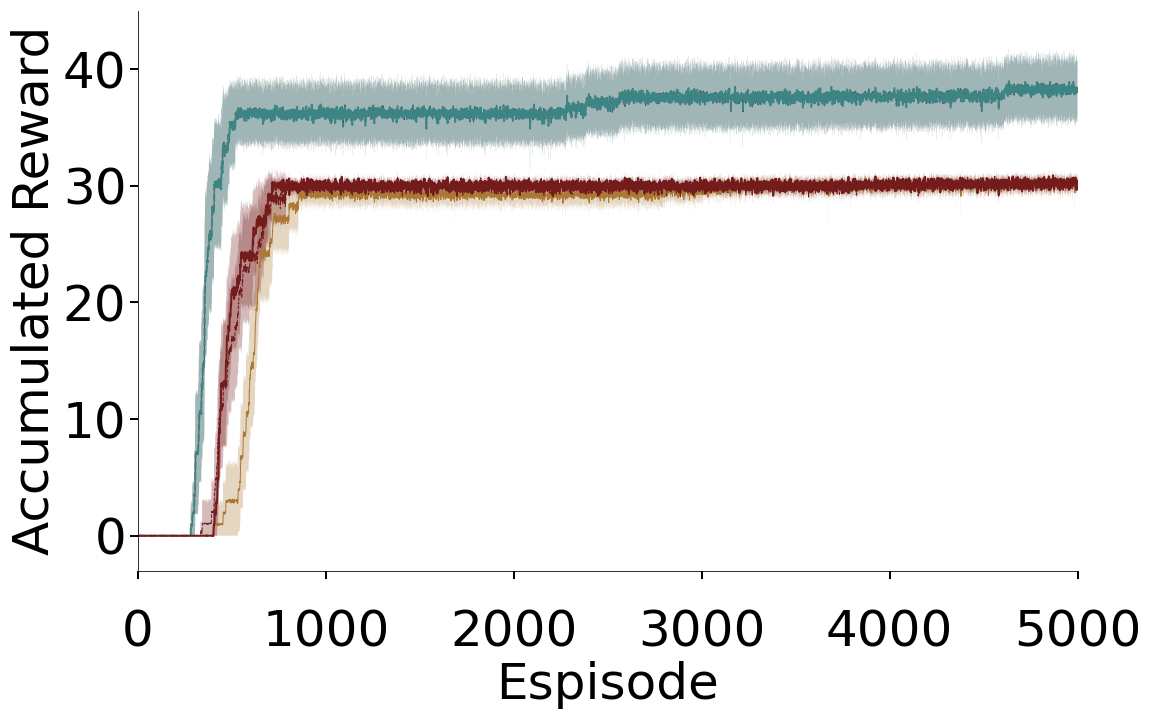}\\
(a) { FIXED}\\
\ \\
\includegraphics[width=1.6in]{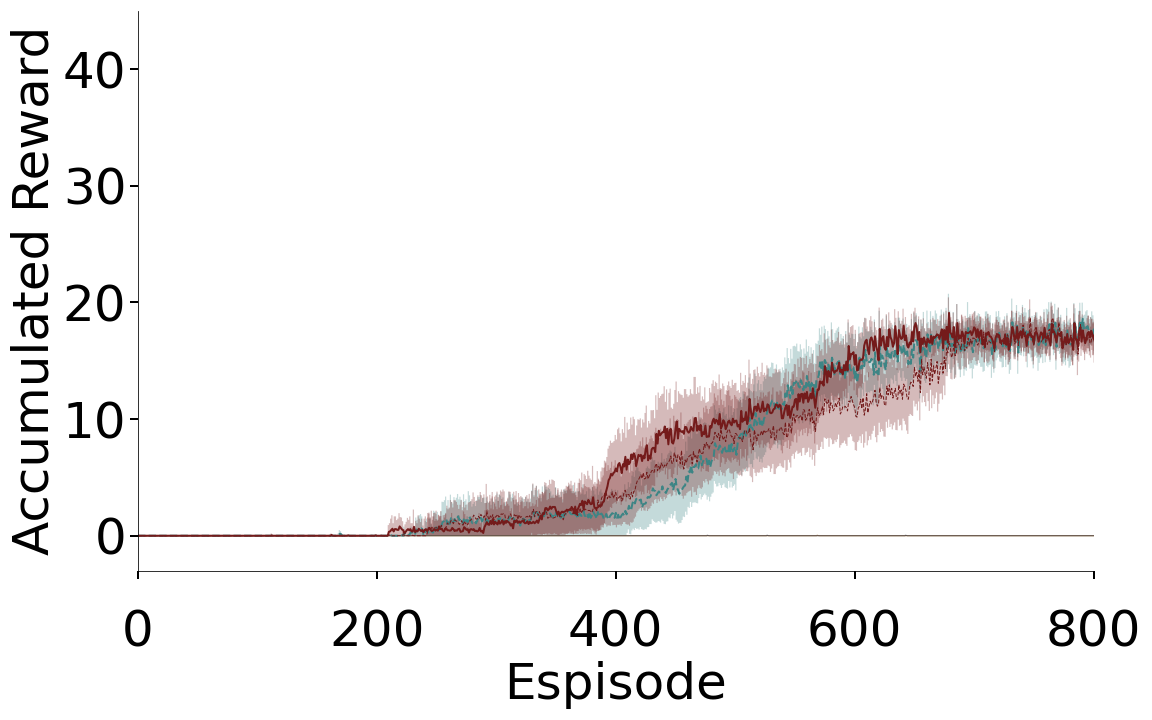}
\includegraphics[width=1.6in]{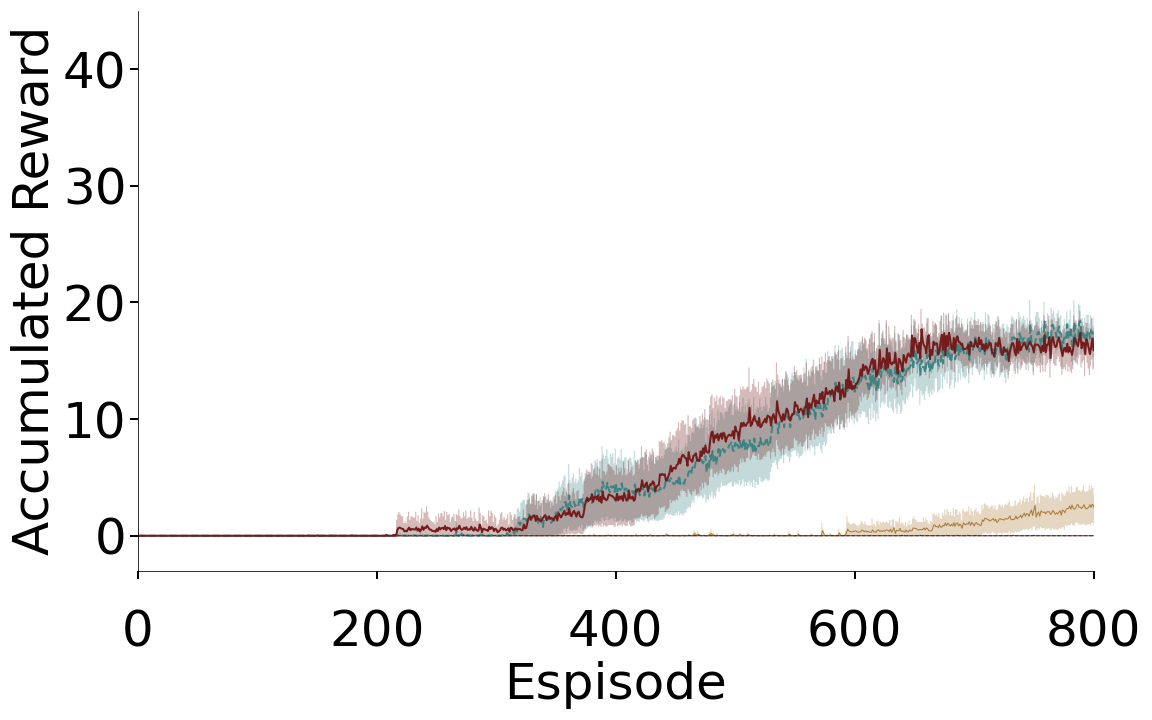}
\includegraphics[width=1.6in]{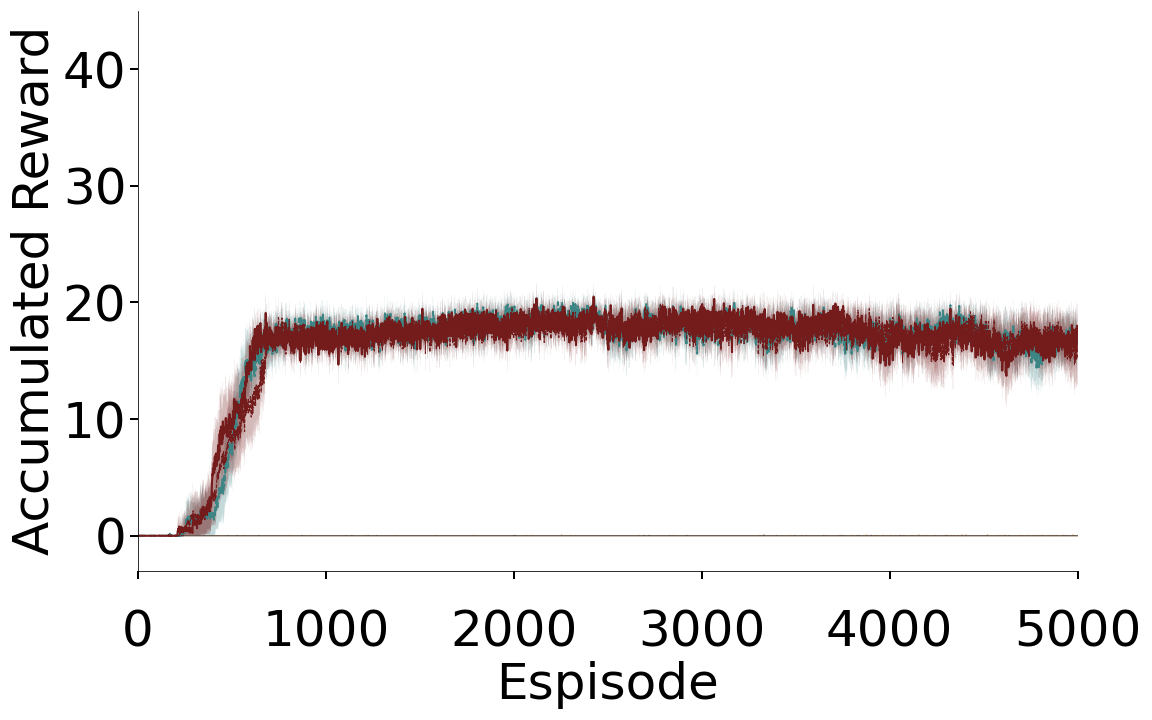}
\includegraphics[width=1.6in]{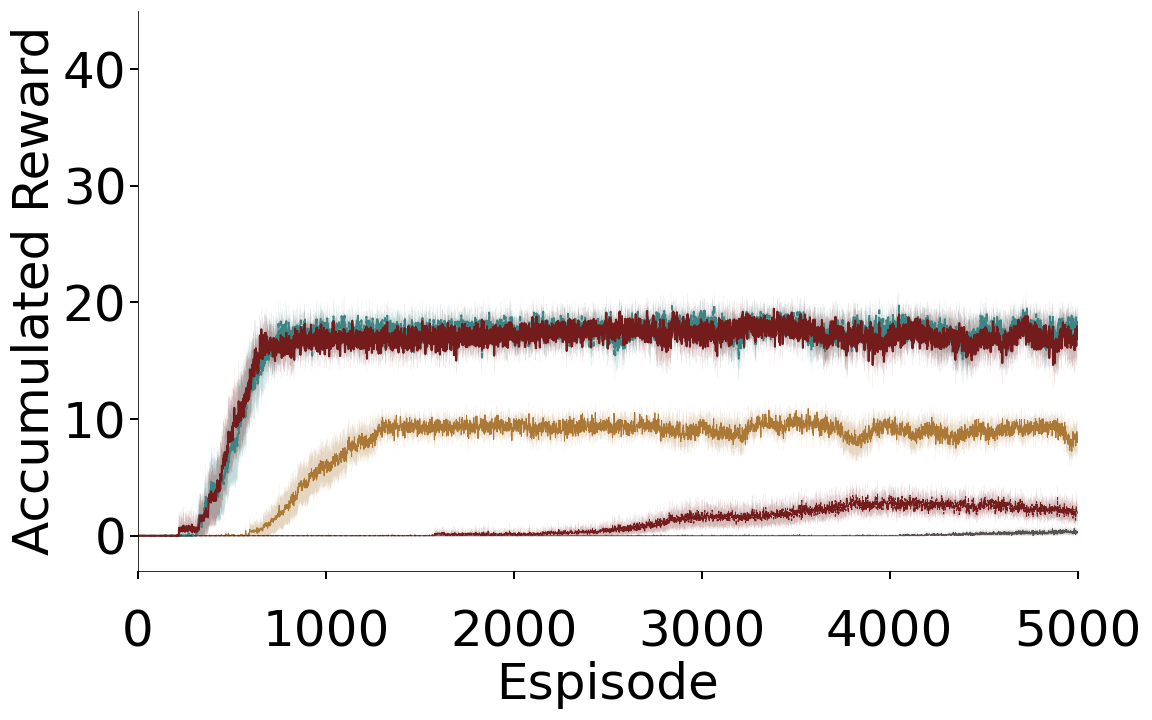}\\
(b) { RANDOM}\\
\ \\
\includegraphics[width=1.6in]{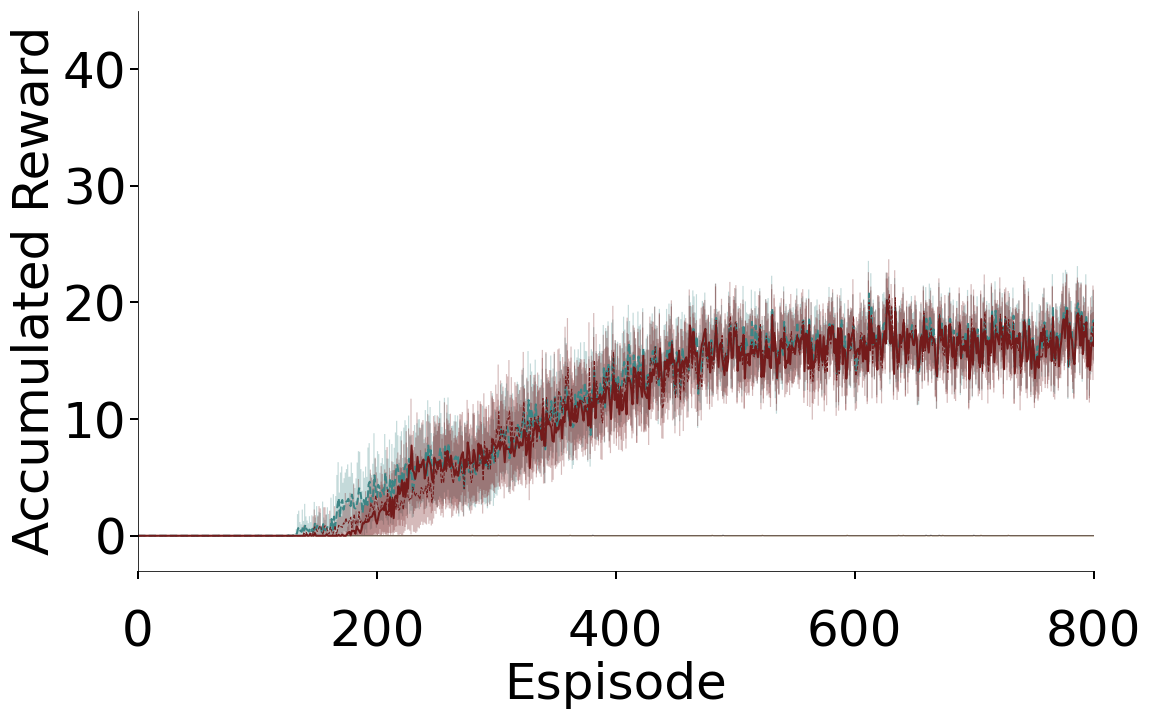}
\includegraphics[width=1.6in]{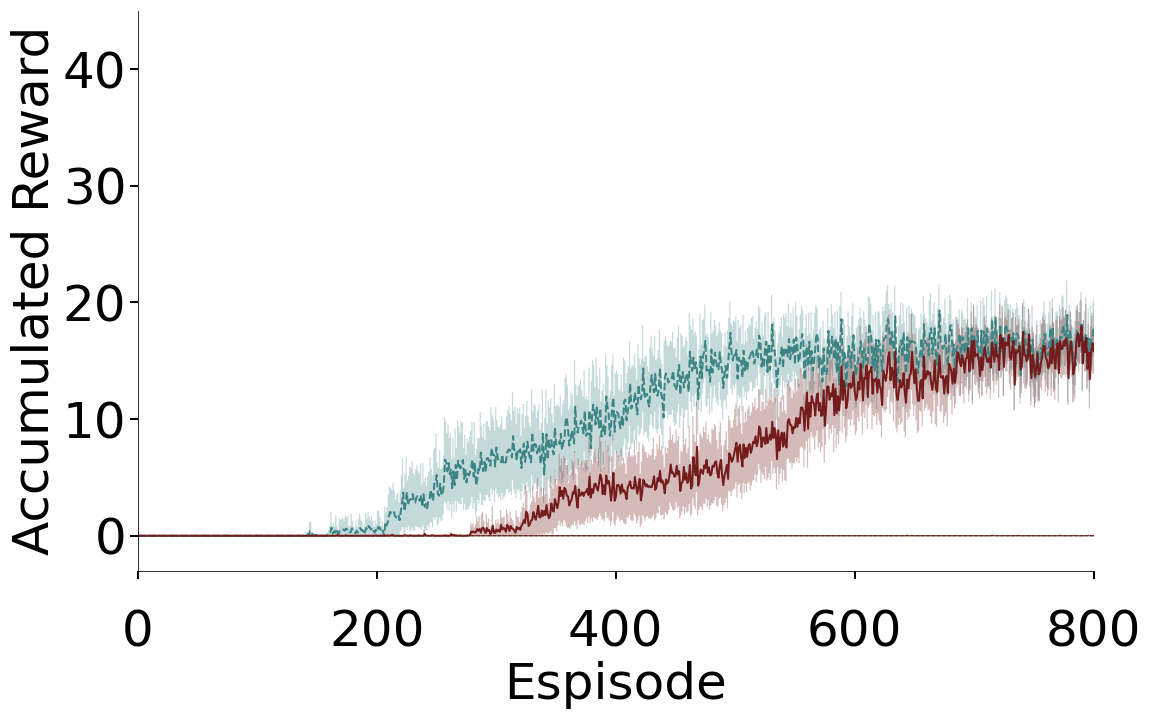}
\includegraphics[width=1.6in]{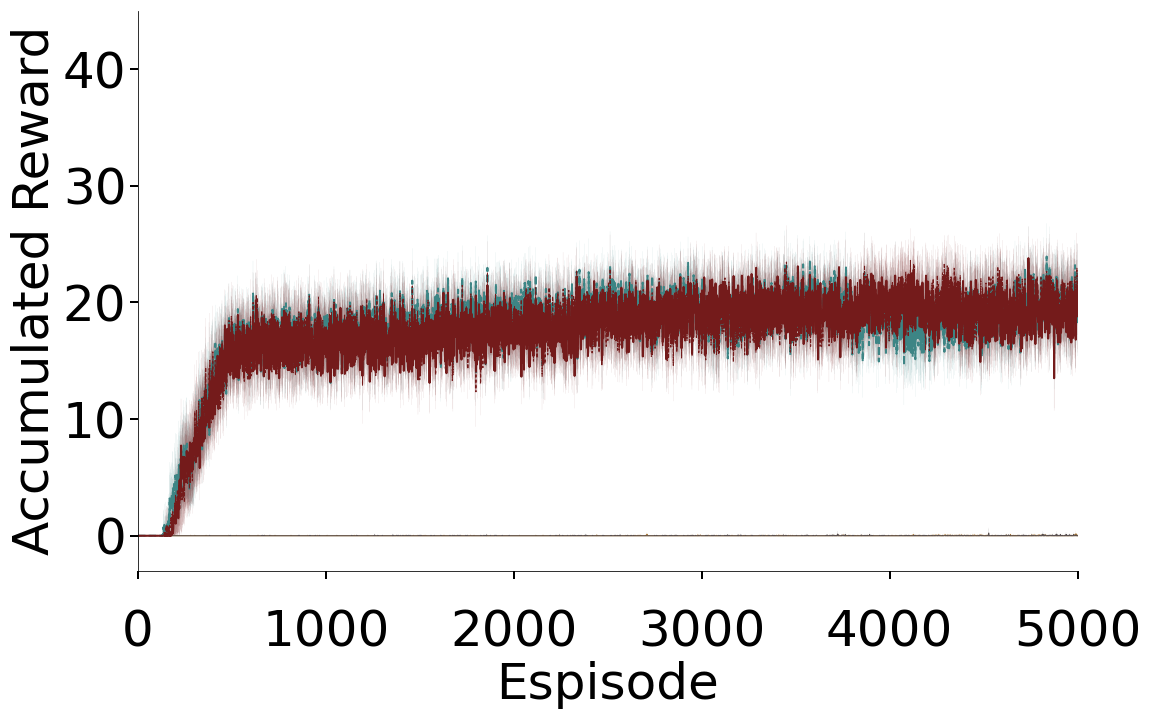}
\includegraphics[width=1.6in]{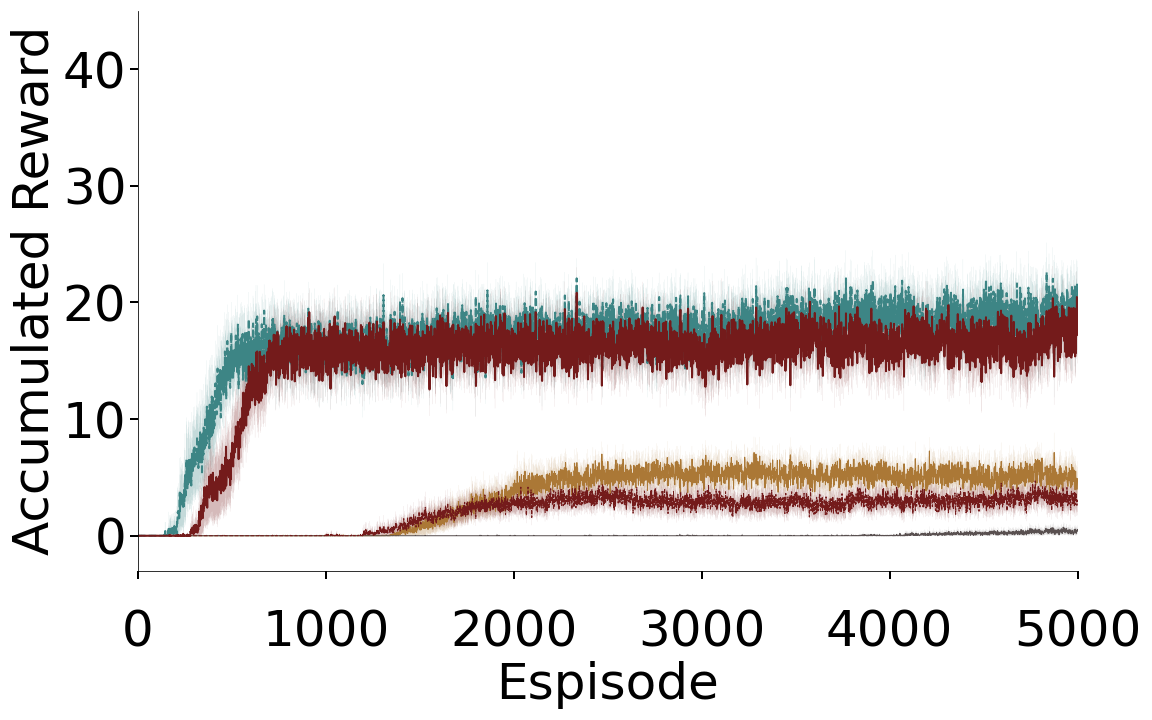}\\
(c) { DRIFT}\\
\caption{{\bf Temporal representation and accumulation GVF learning rate impacted task performance}. Comparison of GVF co-agent representation with a accumulation GVF prediction as coupled to an Expected Sarsa learning agent over the first 800 episodes (8000 steps) and for the entire 5000 episodes (5M steps), averaged over 30 runs with a maximum heat capacity of 6, in the (a) fixed, (b) random, and (c) drift conditions. Shown are the Bias Unit (solid yellow line), Oscillator (solid grey line), Bit Cascade (dashed blue line), TCT with a decay of $e^{-0.3t}$ (solid red line) and TCT with a decay of $e^{-0.6t}$ (dashed red line).}
\label{fig:rep-fixed-gamma-comparison}
\end{figure*}

\begin{figure*}[!th]
\centering
{\bf First 800 Episodes} \hfil {\bf All 5000 Episodes}\\
$\alpha_{gvf}=0.01$ \hfil $\alpha_{gvf}=0.1$  \hfil $\alpha_{gvf}=0.01$ \hfil $\alpha_{gvf}=0.1$\\
\ \\
\includegraphics[width=1.6in]{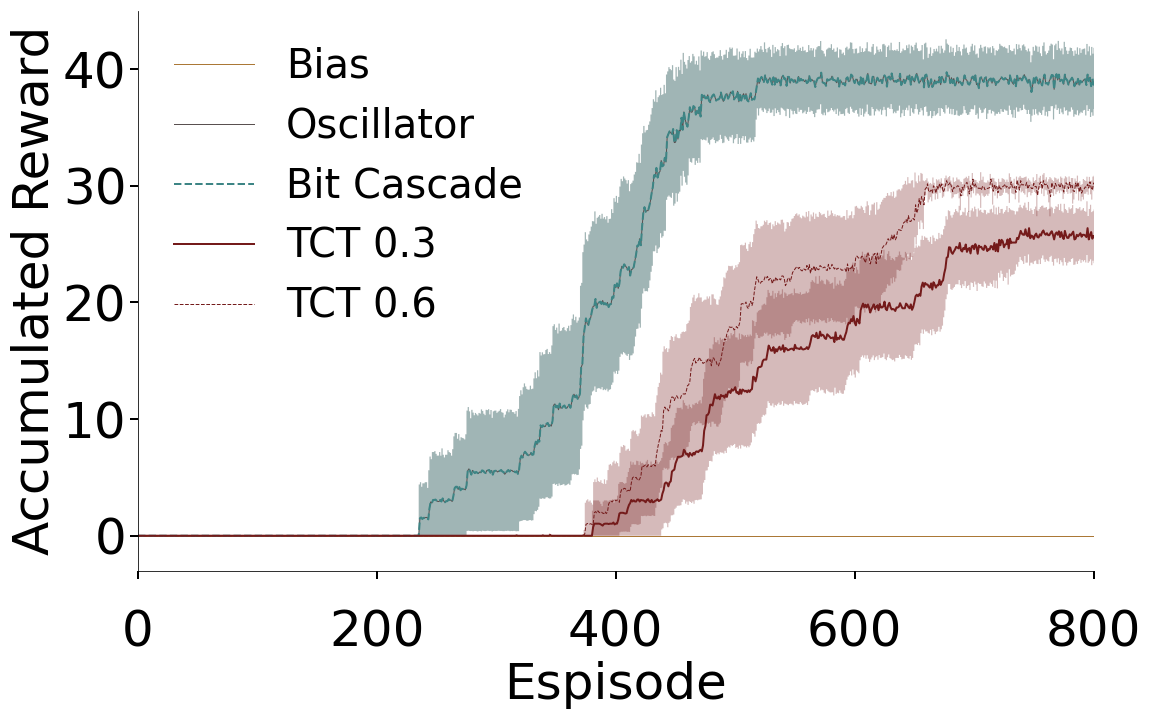}
\includegraphics[width=1.6in]{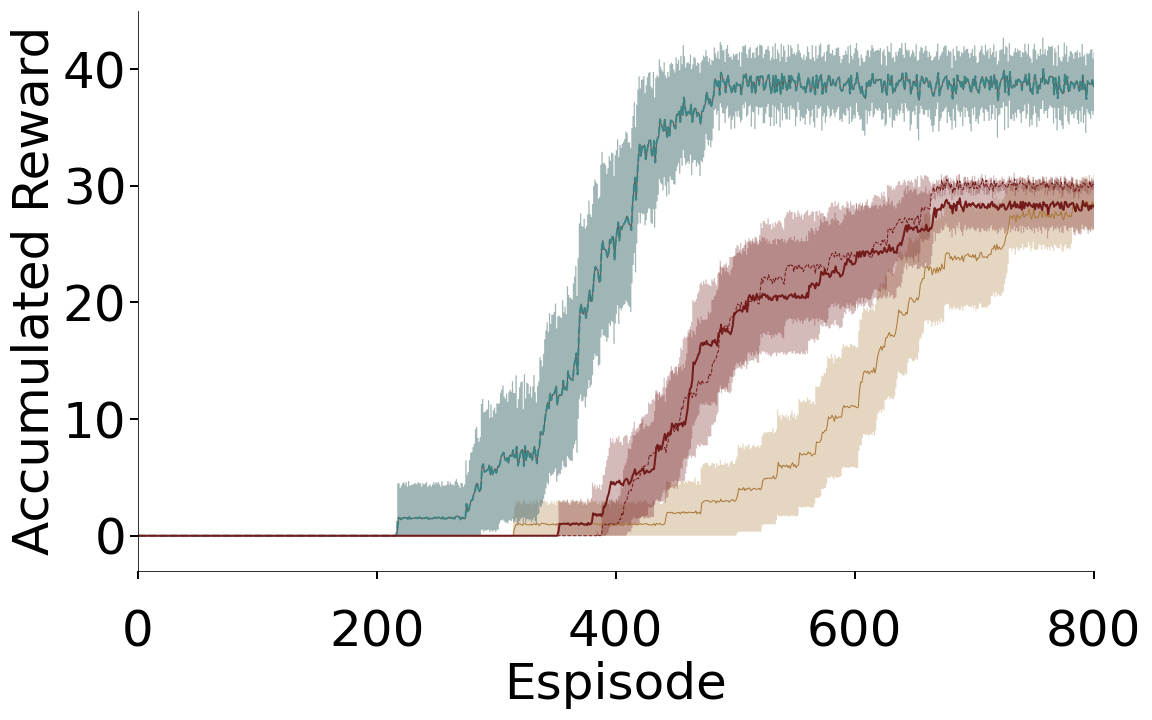}
\includegraphics[width=1.6in]{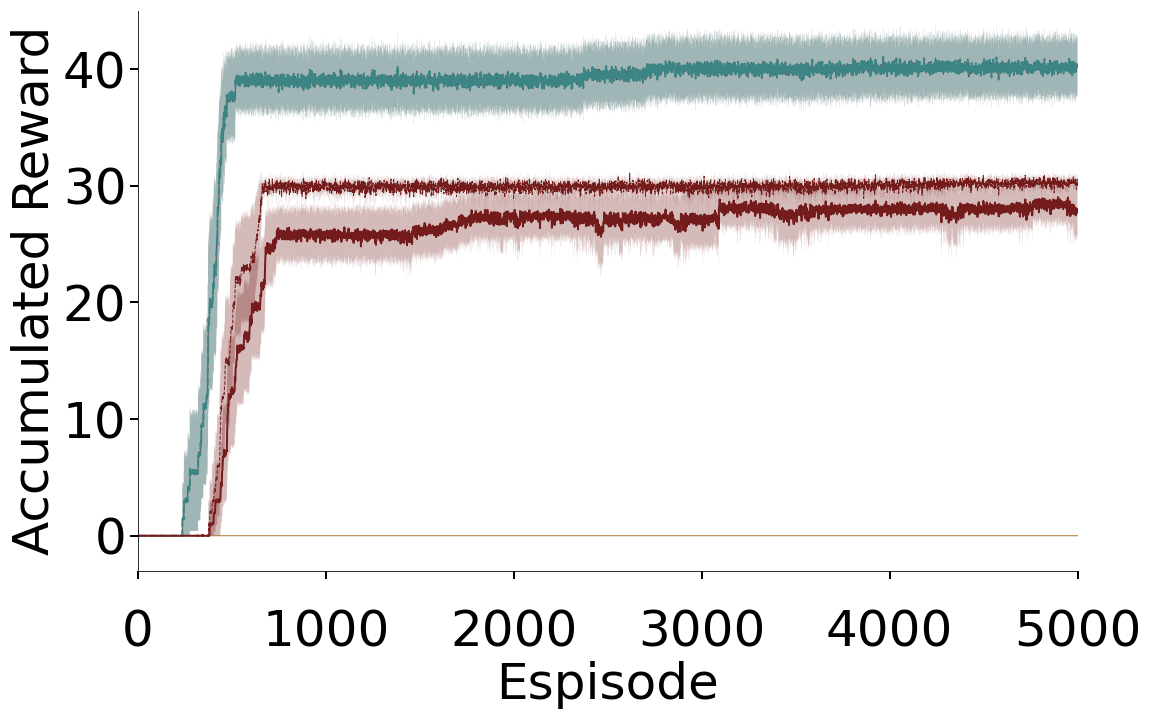}
\includegraphics[width=1.6in]{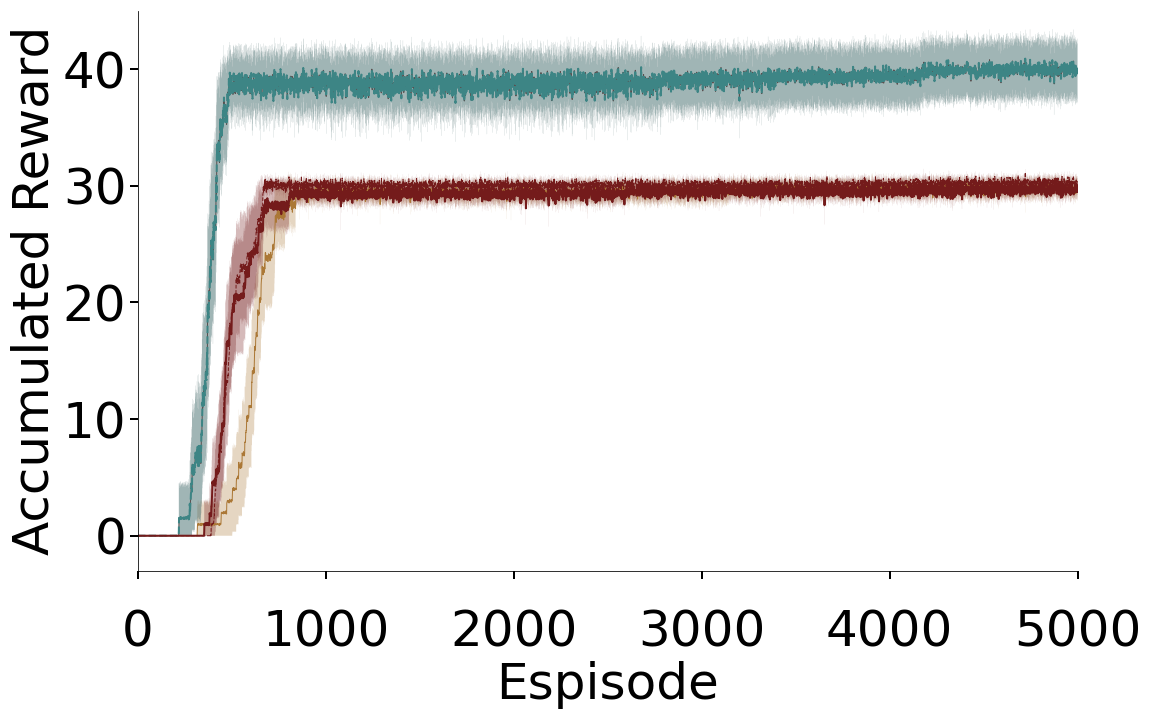}\\
(a) { FIXED}\\
\ \\
\includegraphics[width=1.6in]{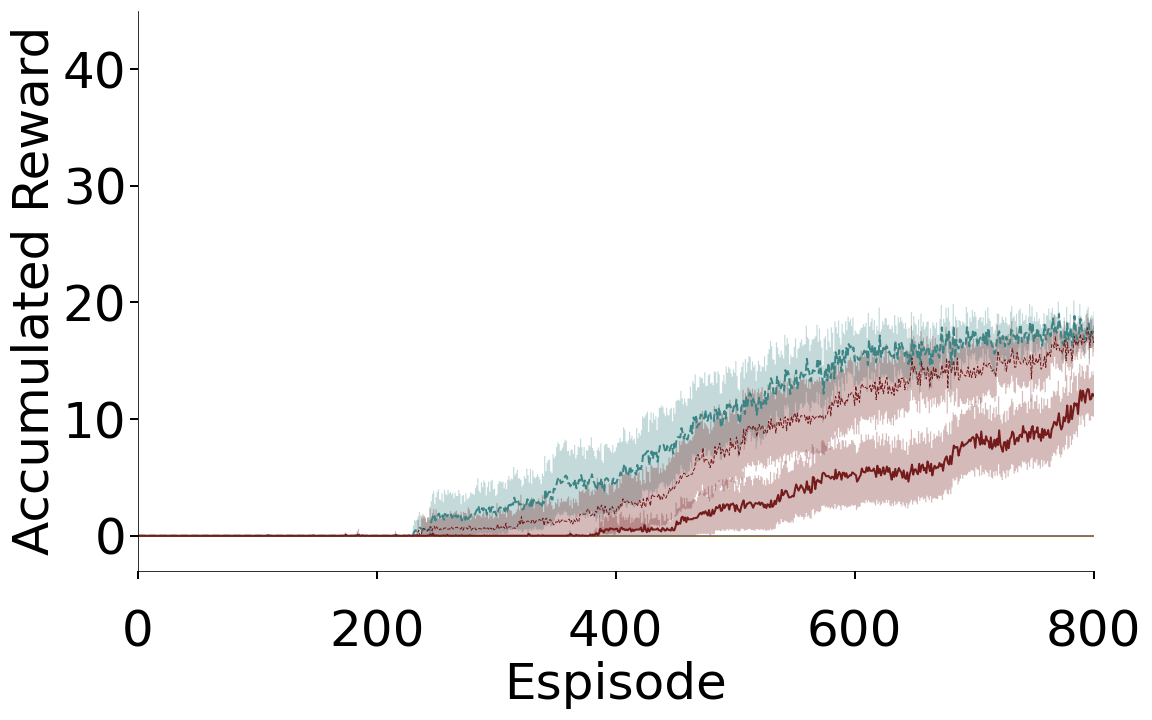}
\includegraphics[width=1.6in]{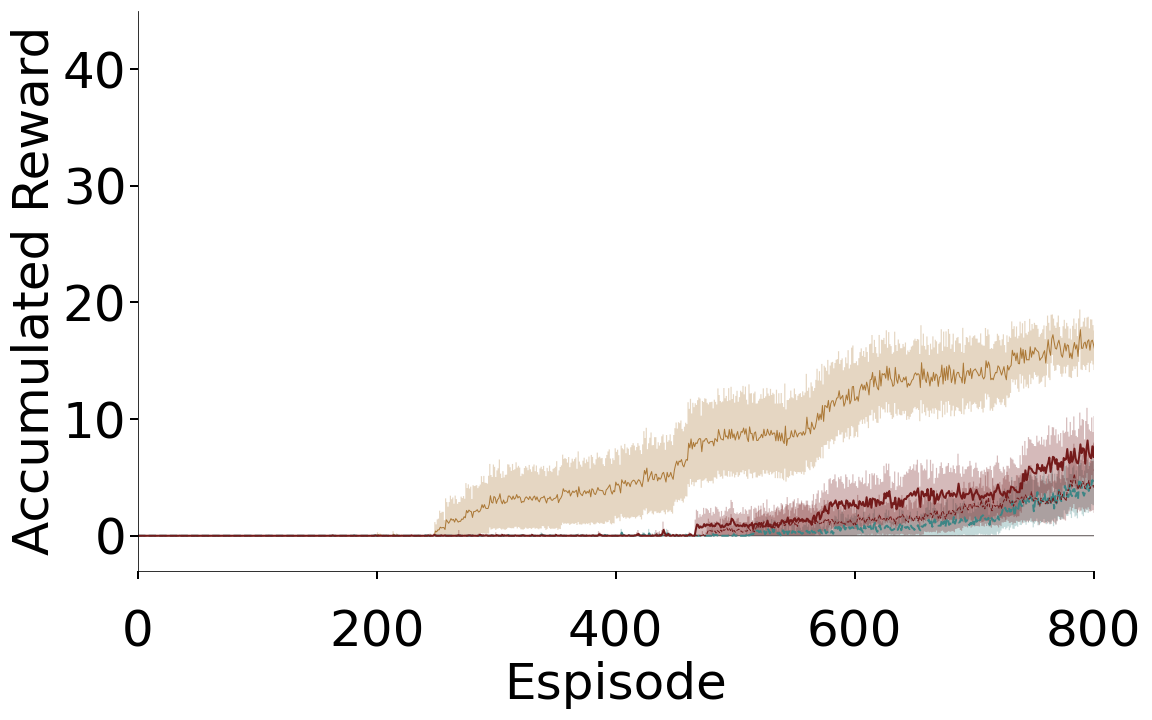}
\includegraphics[width=1.6in]{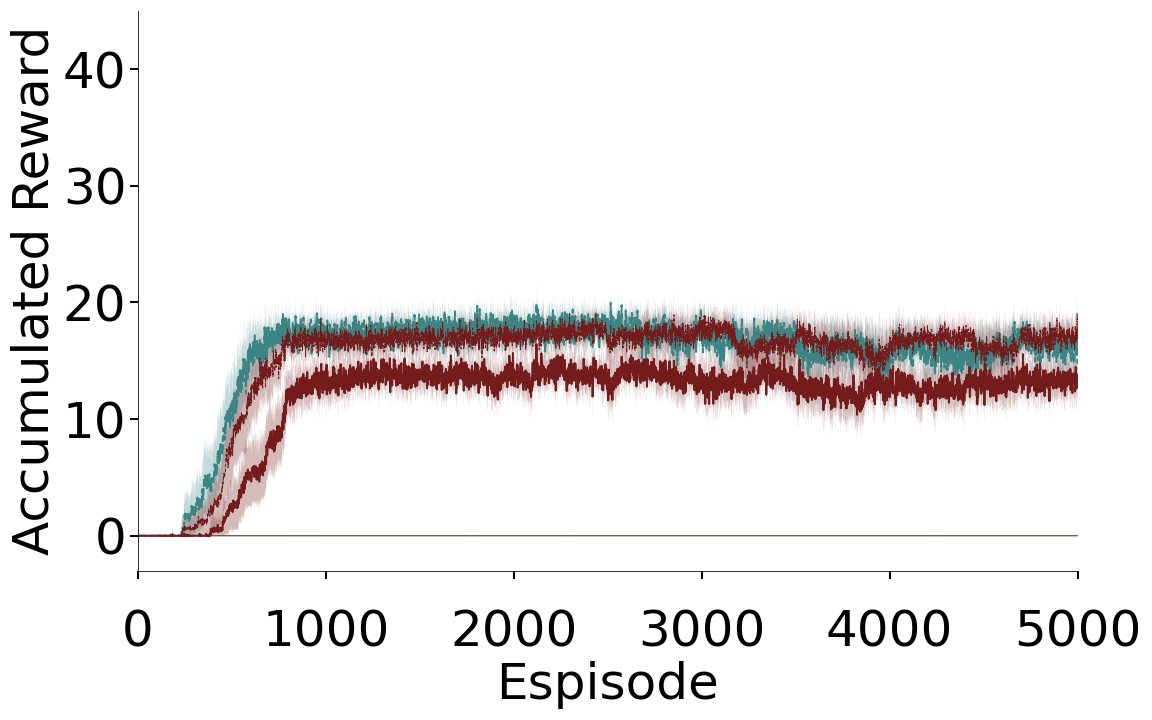}
\includegraphics[width=1.6in]{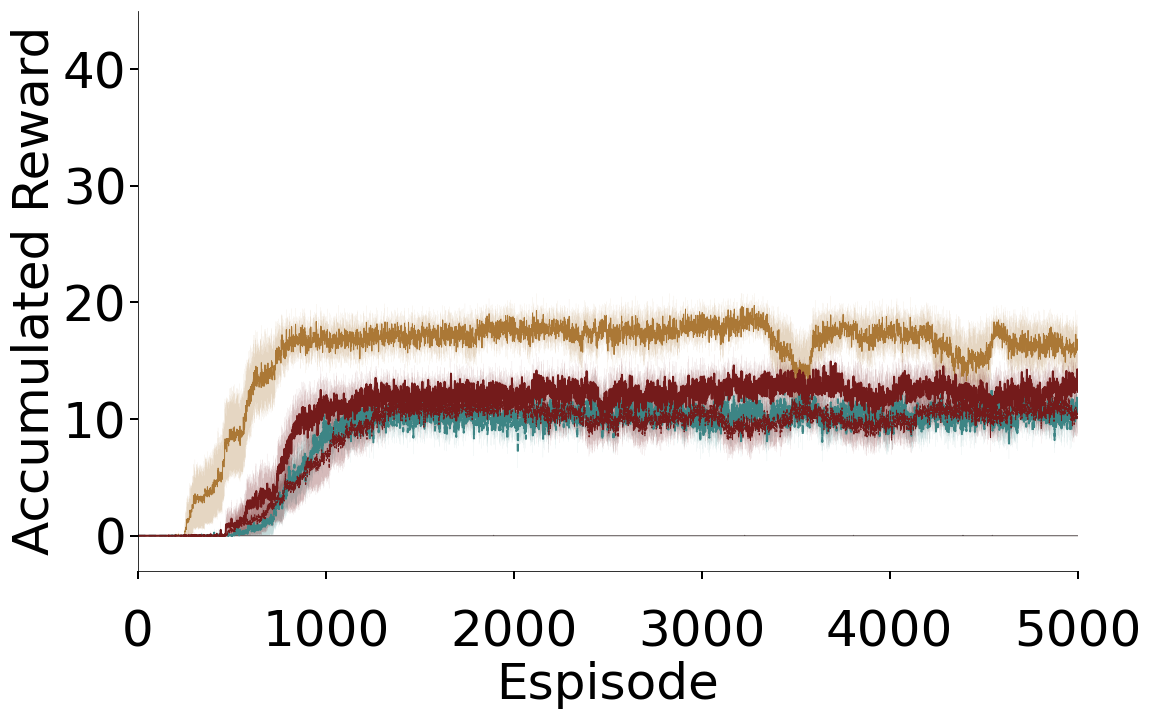}\\
(b) { RANDOM}\\
\ \\

\includegraphics[width=1.6in]{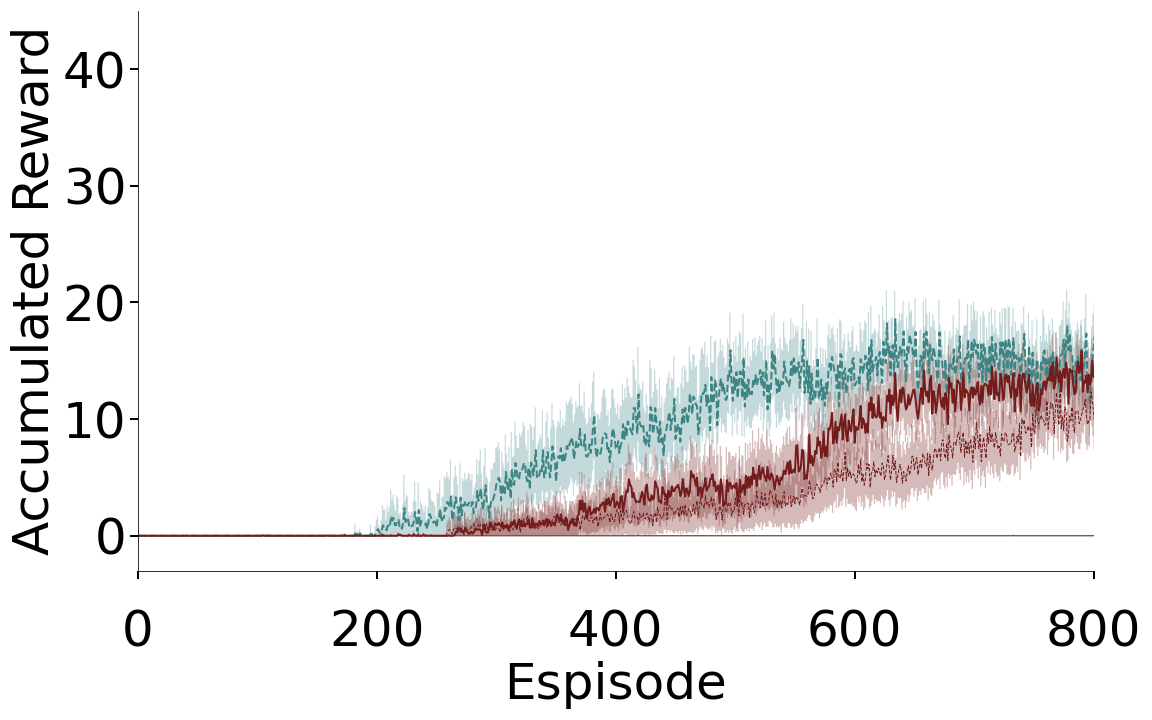}
\includegraphics[width=1.6in]{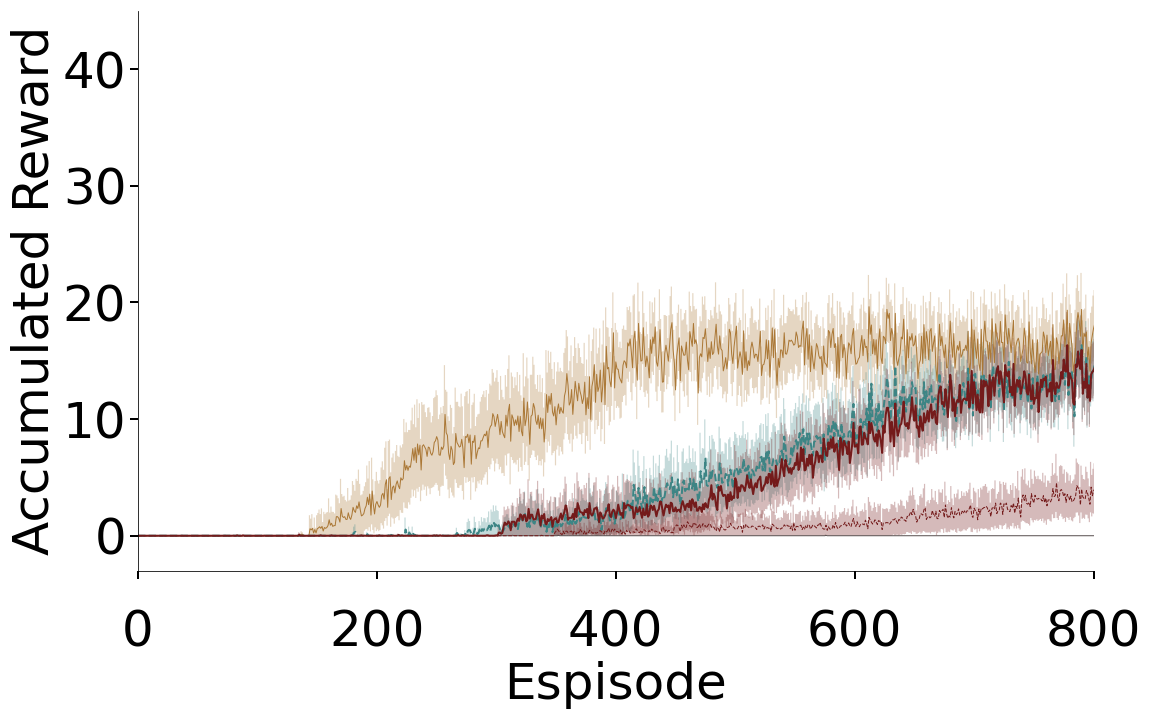}
\includegraphics[width=1.6in]{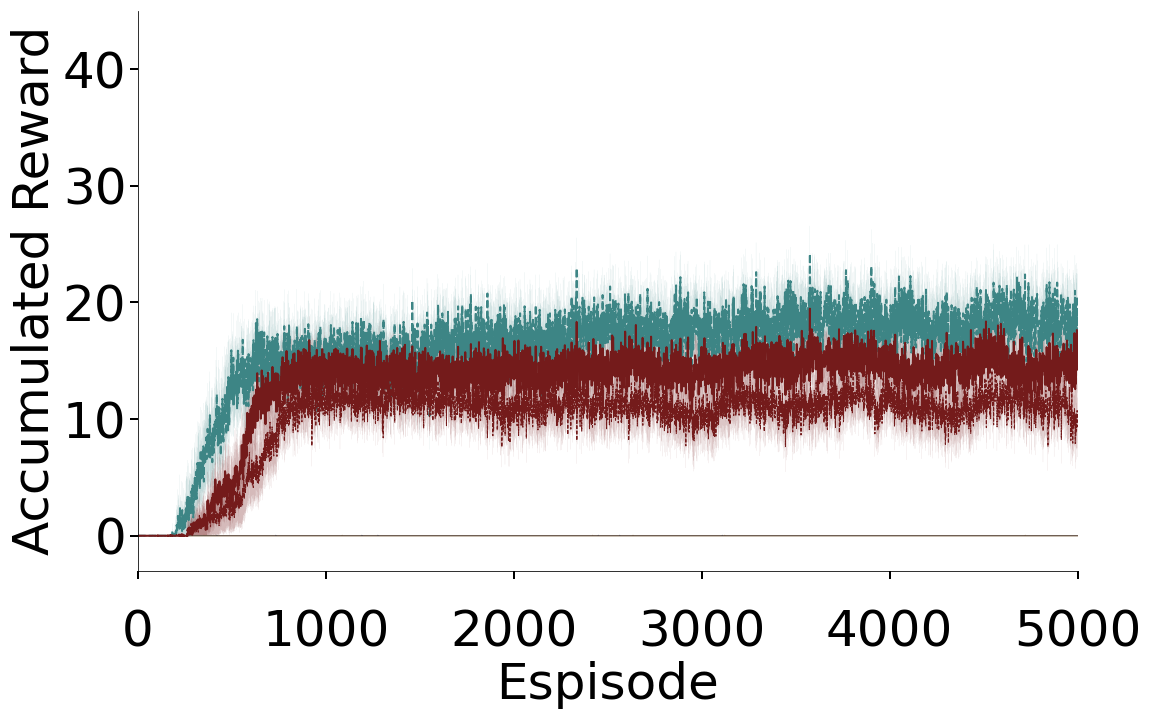}
\includegraphics[width=1.6in]{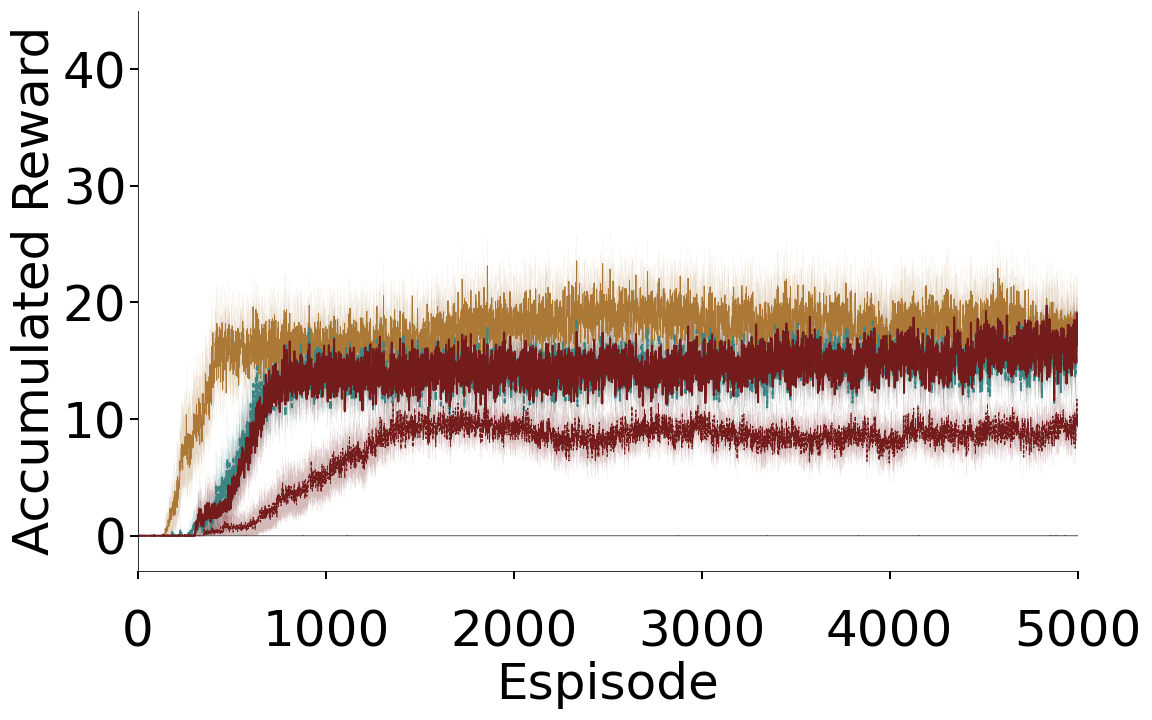}\\
(c) { DRIFT}\\
\caption{{\bf Temporal representation and state-conditional (countdown) GVF learning rate impacted task performance}. Comparison of GVF co-agent representation with a countdown GVF as coupled to an Expected Sarsa learning agent over the first 800 episodes (8000 steps) and for the entire 5000 episodes (5M steps), averaged over 30 runs with a maximum heat capacity of 6, in the (a) fixed, (b) random, and (c) drift conditions. Shown are the Bias Unit (solid yellow line), Oscillator (solid grey line), Bit Cascade (dashed blue line), TCT with a decay of $e^{-0.3t}$ (solid red line) and TCT with a decay of $e^{-0.6t}$ (dashed red line).}
\label{fig:rep-countdown-comparison}
\end{figure*}

\begin{figure*}[!th]
\centering
{\bf Accumulation GVF} \hfil {\bf Countdown GVF}\\
\includegraphics[width=0.49\textwidth]{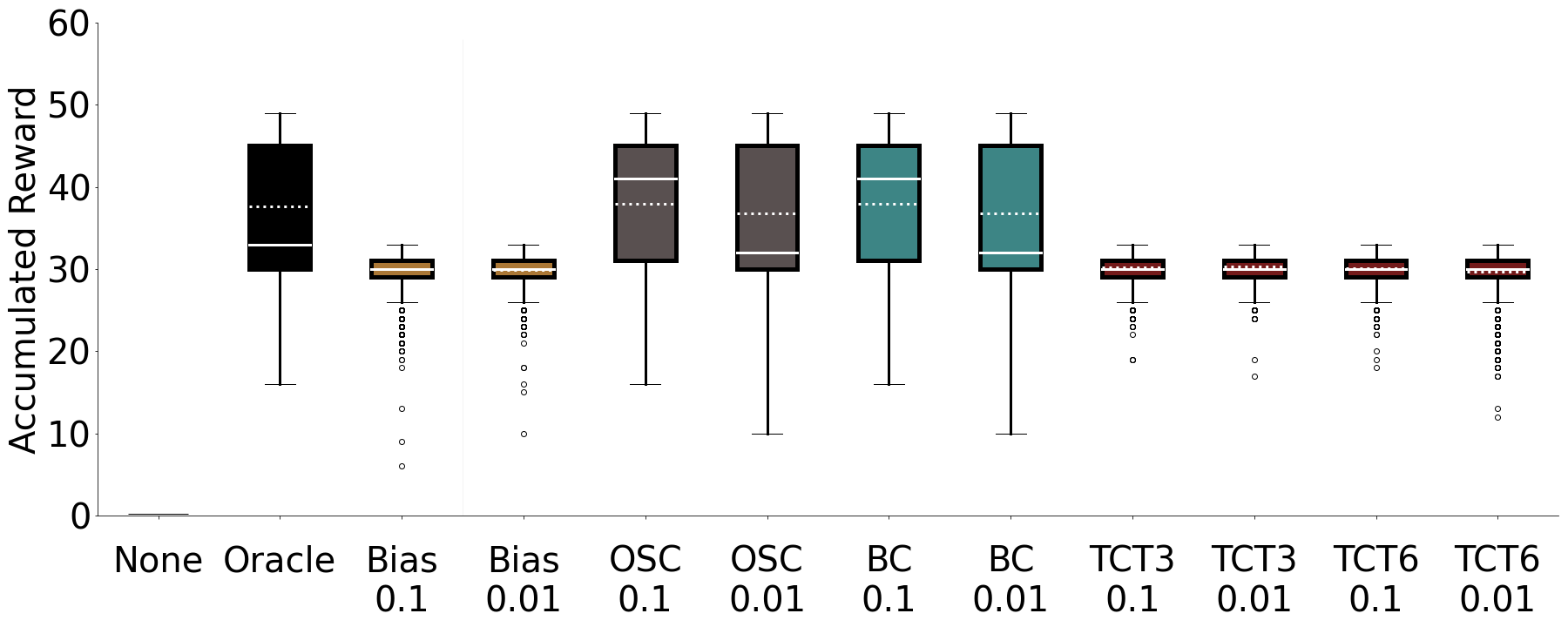} \includegraphics[width=0.49\textwidth]{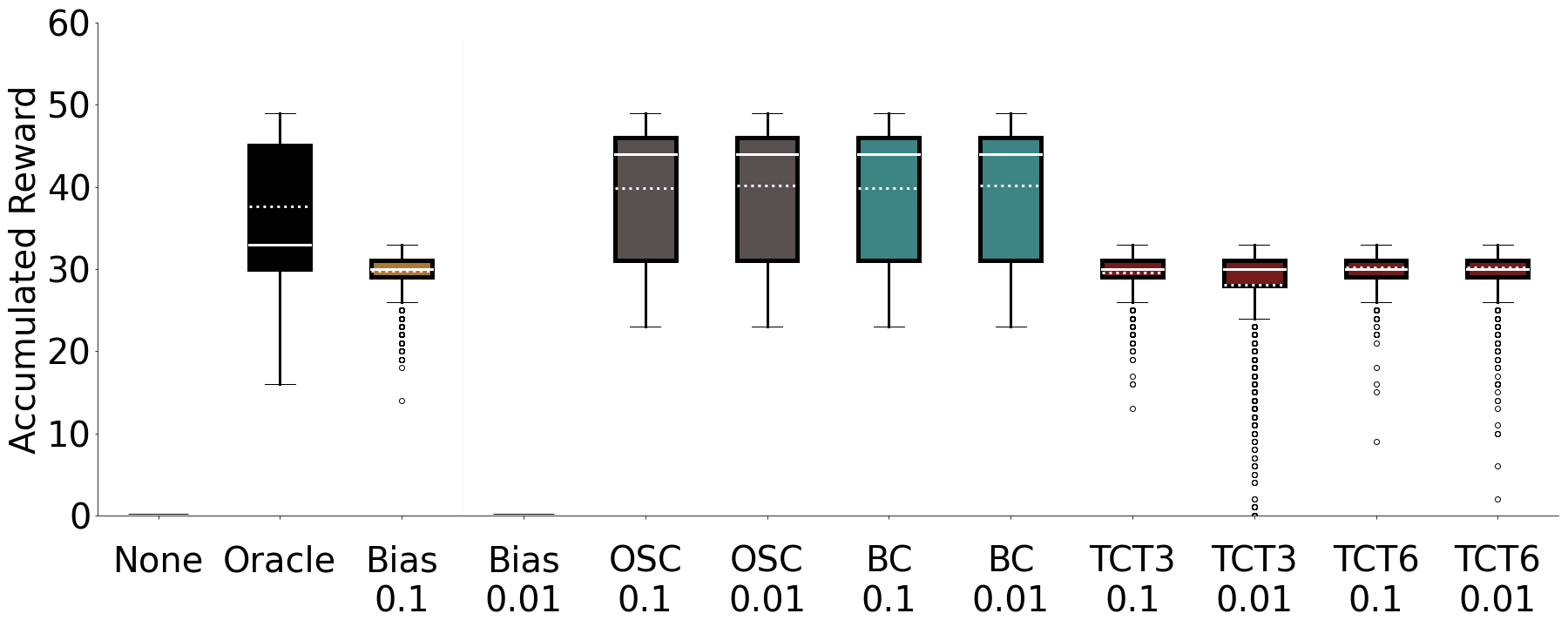}\\
(a) { FIXED}\\
\ \\
\includegraphics[width=0.49\textwidth]{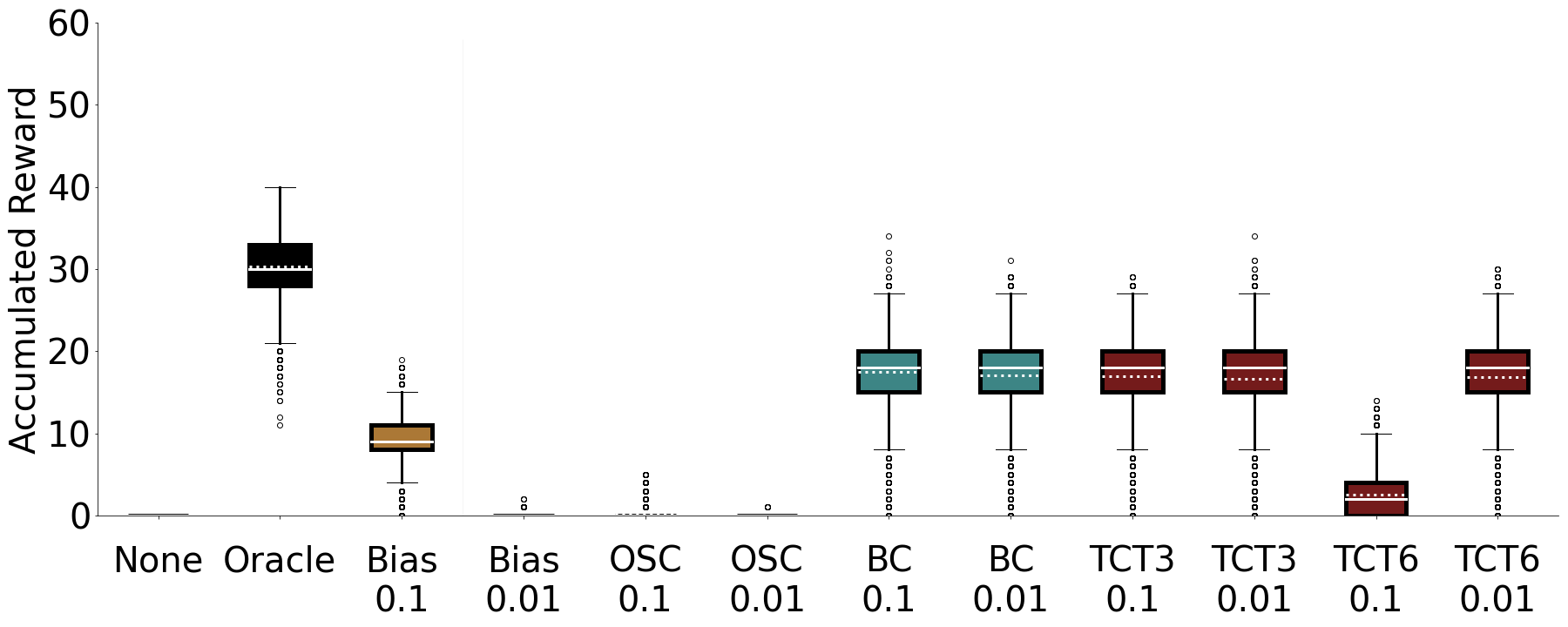}
\includegraphics[width=0.49\textwidth]{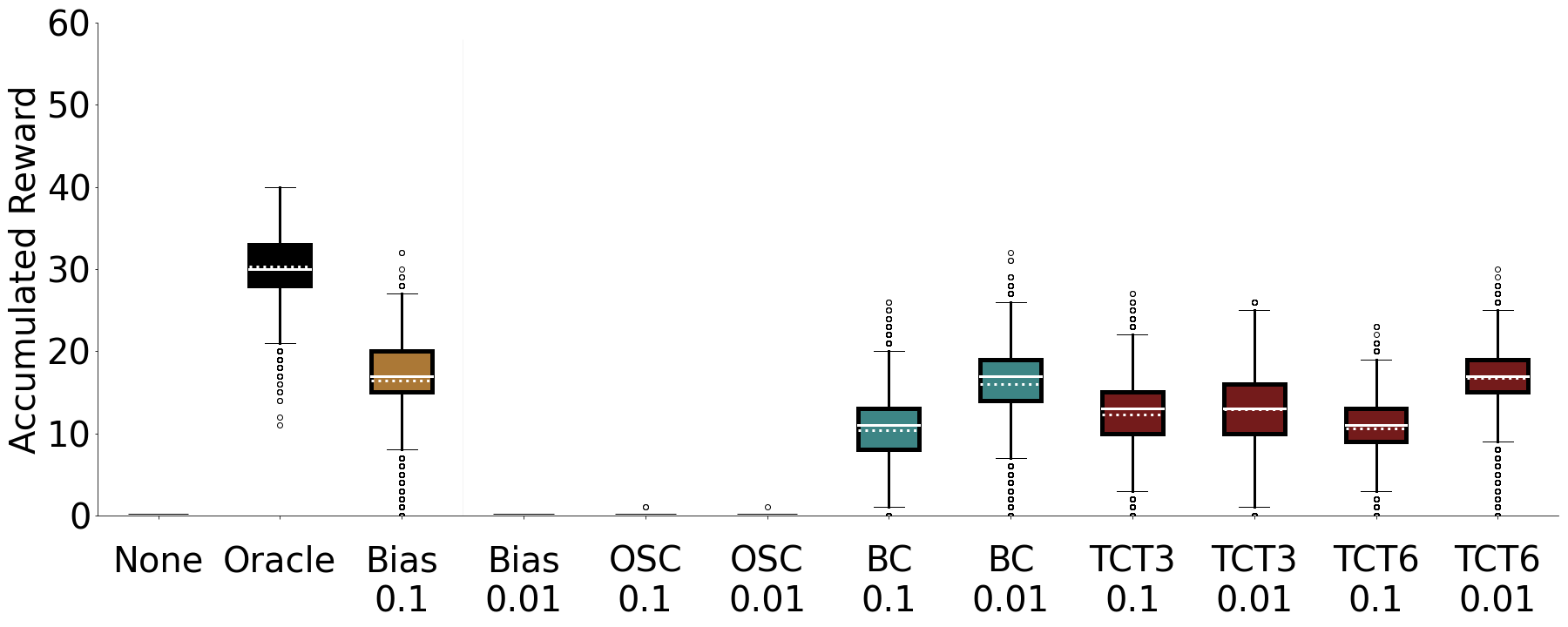}\\
(b) { RANDOM}\\
\ \\
\includegraphics[width=0.49\textwidth]{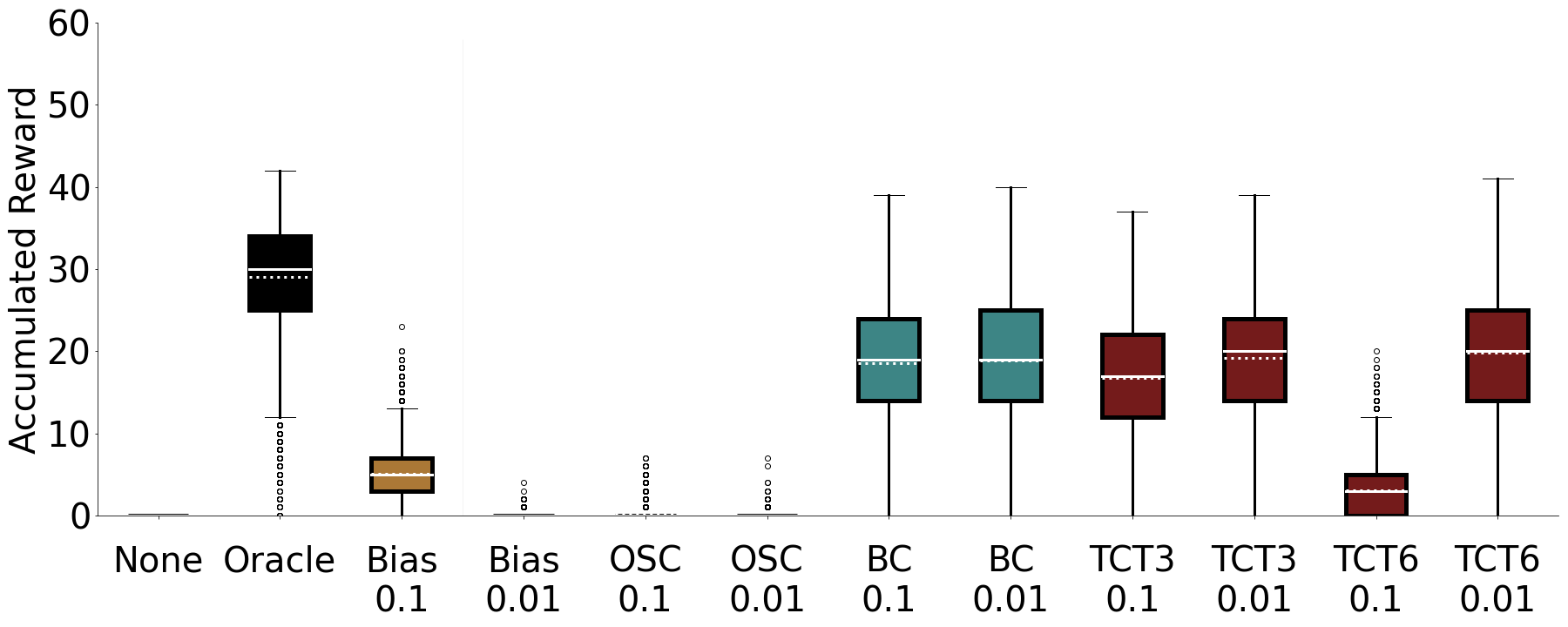}
\includegraphics[width=0.49\textwidth]{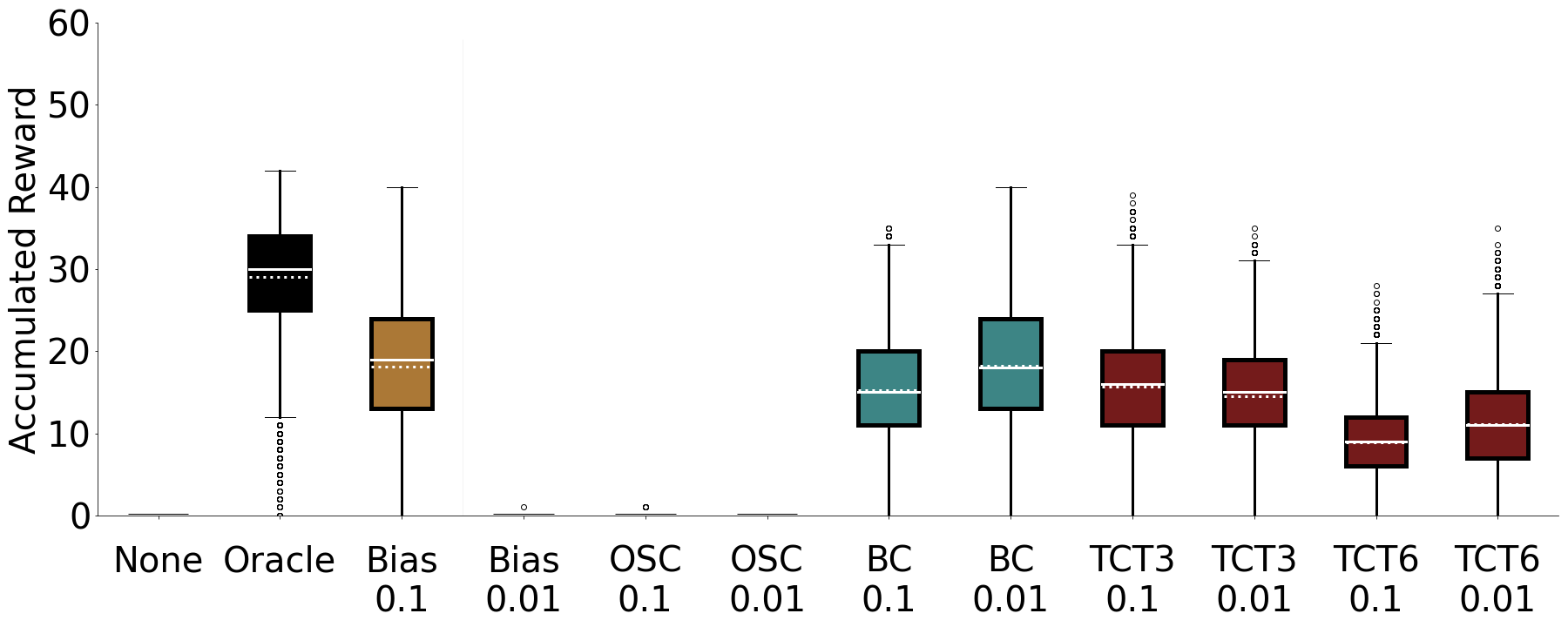}\\
(c) { DRIFT}
\caption{{\bf Differences in asymptotic accumulated reward across representations and GVF questions}, as shown by median plots for accumulated reward for last 1000 episodes (1000k steps) at heat gain 6, across 30 random seeds, across temporal representations, with oracle and GVF-based co-agents and Expected Sarsa as the learning agent for (a) fixed, (b) random, and (c) drift conditions. Median shown as a solid white line, and average shown as a dotted white line on each box. GVF co-agent step size indicated as 0.1 or 0.01 on the second line; tile-coded trace decay rates of 0.3 and 0.6 are indicated as TCT3 and TCT6 respectively.}
\label{fig:rep-comparison-boxplot}
\end{figure*}

\begin{figure*}[!th]
\centering
{\bf Accumulation GVF} \hfil  {\bf Countdown GVF}\\
\includegraphics[width=0.49\textwidth]{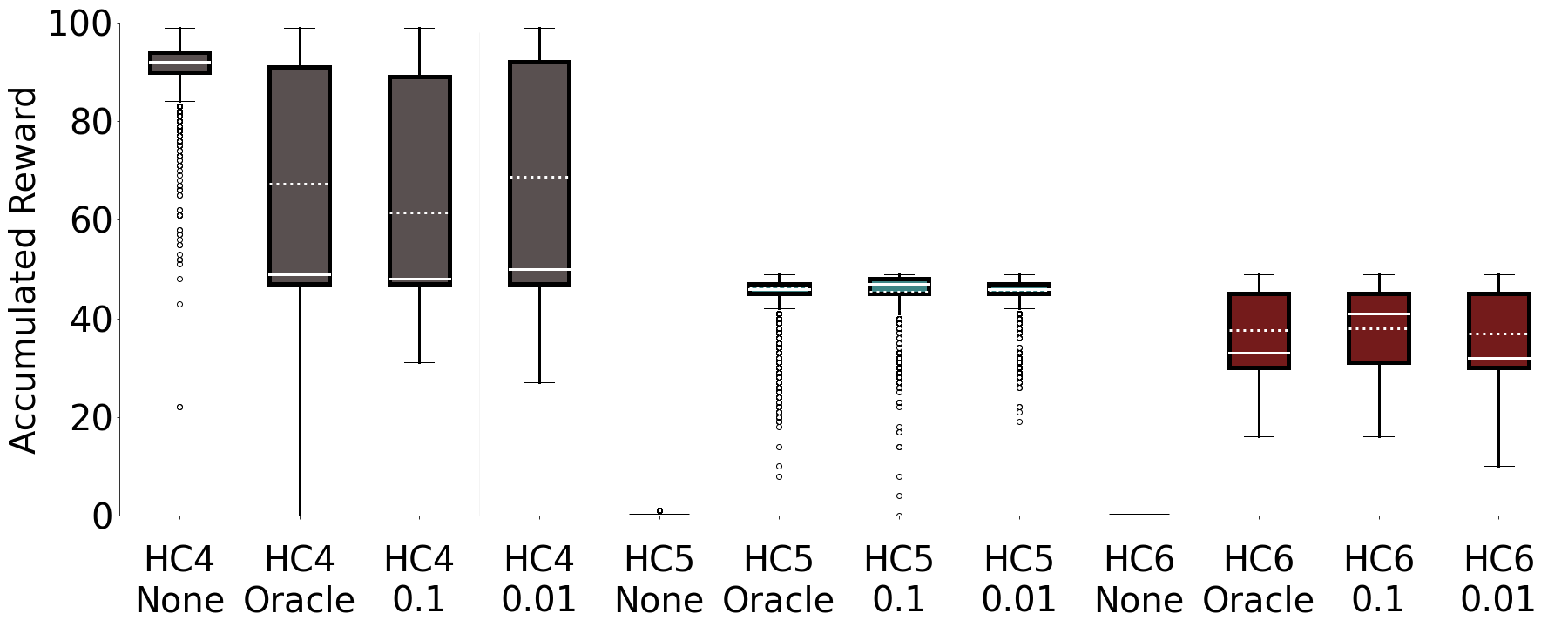}\includegraphics[width=0.49\textwidth]{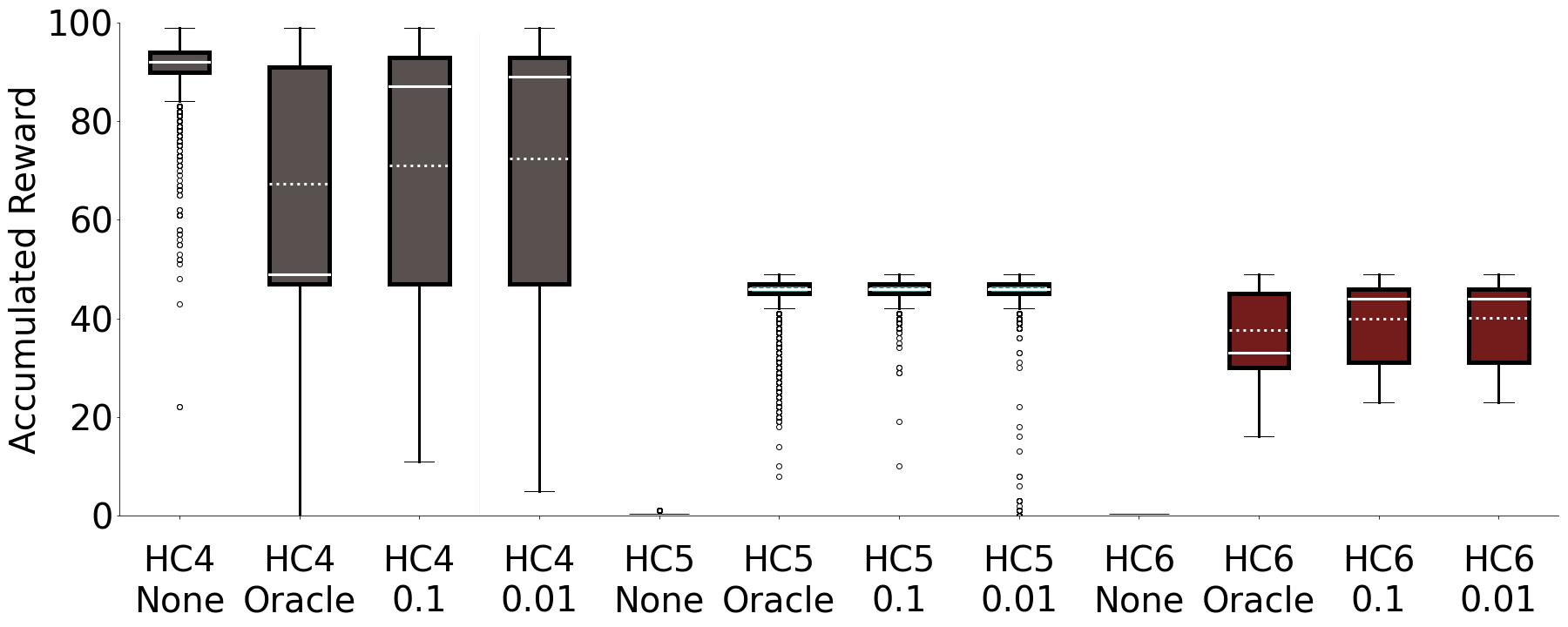}\\
(a) { FIXED}\\
\ \\
\includegraphics[width=0.49\textwidth]{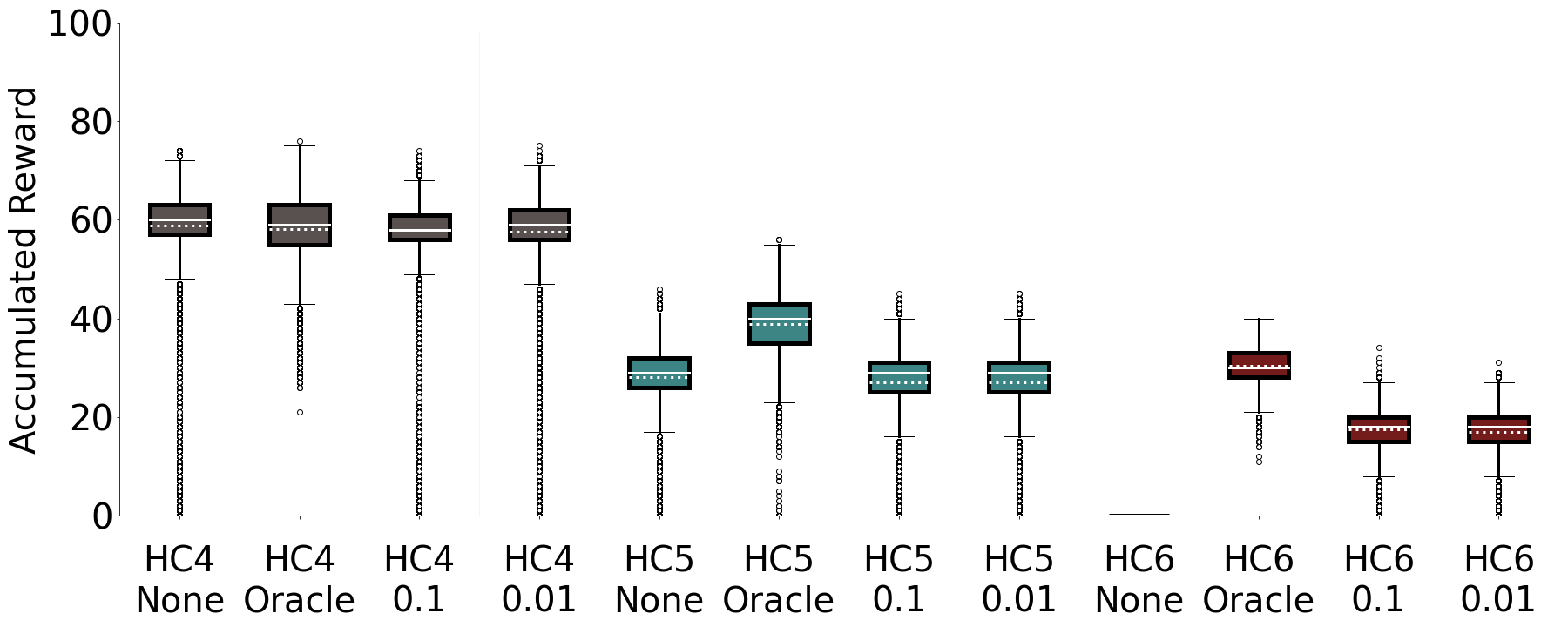}\includegraphics[width=0.49\textwidth]{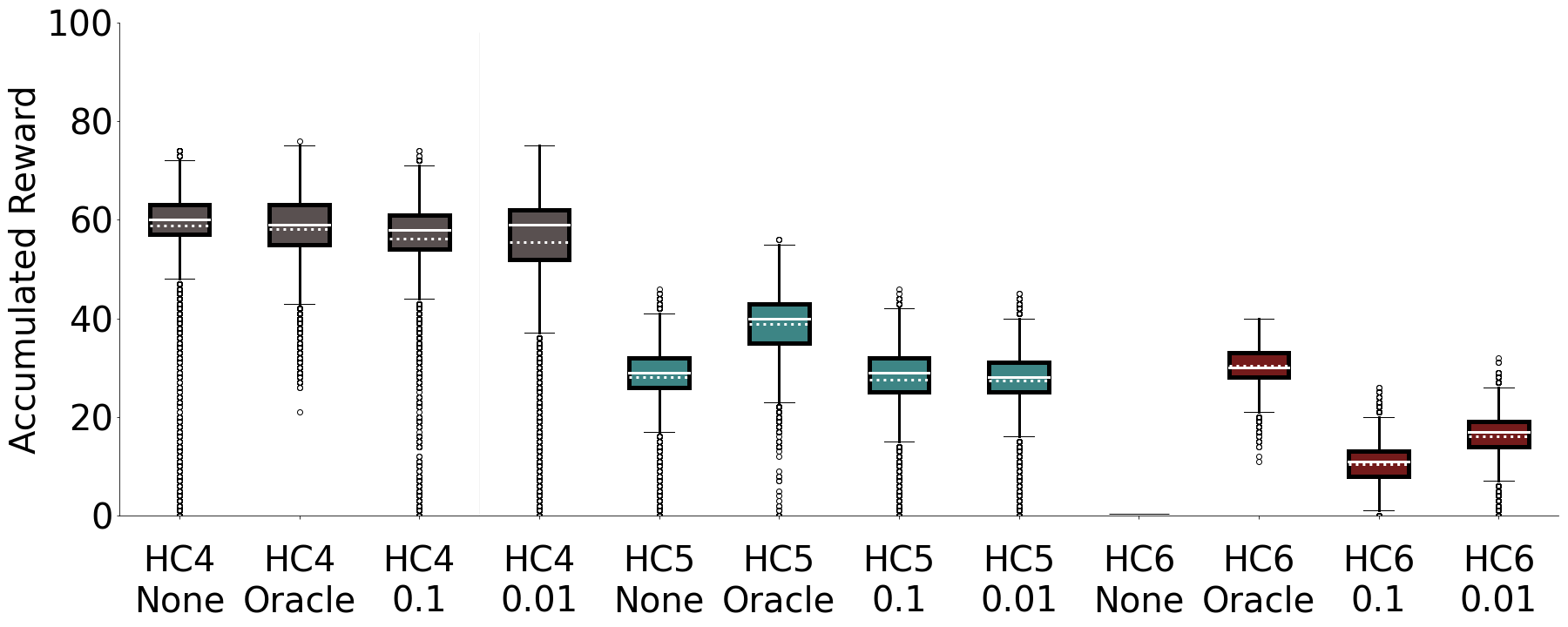}\\
(b) { RANDOM}\\
 \ \\
\includegraphics[width=0.49\textwidth]{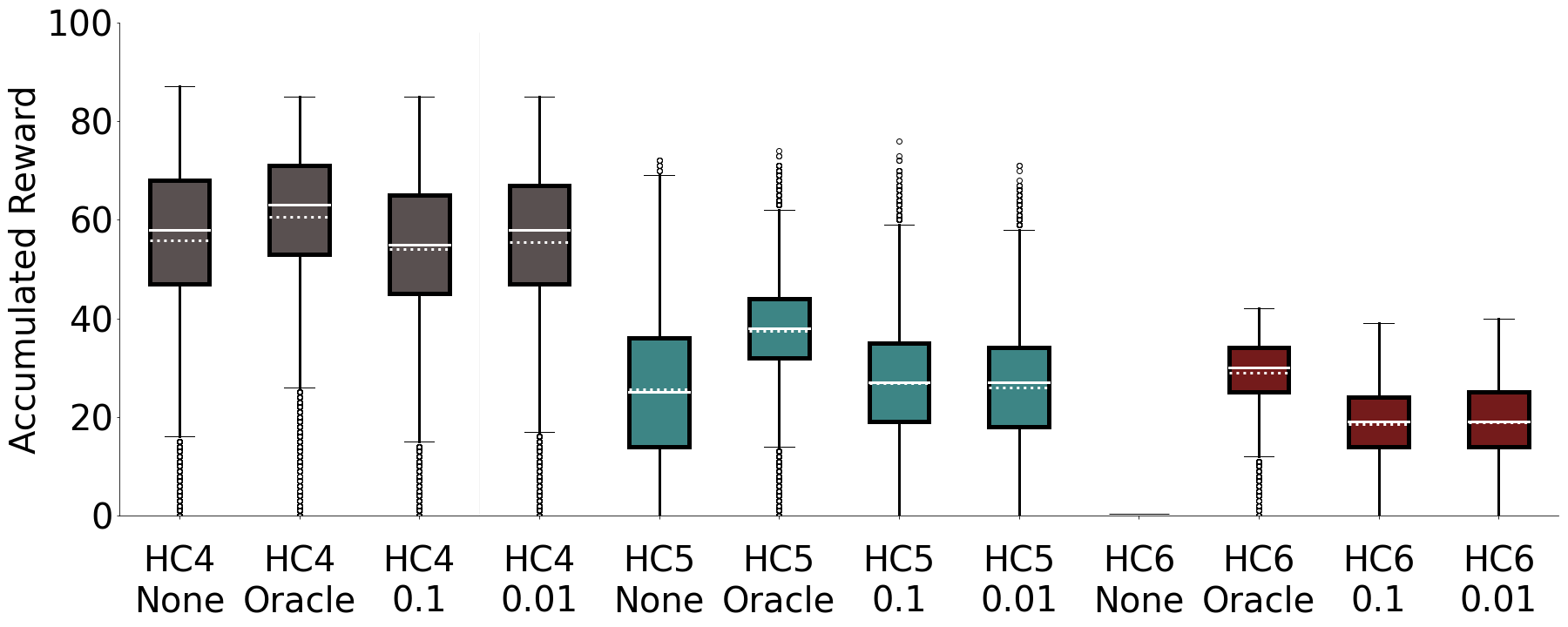}\includegraphics[width=0.49\textwidth]{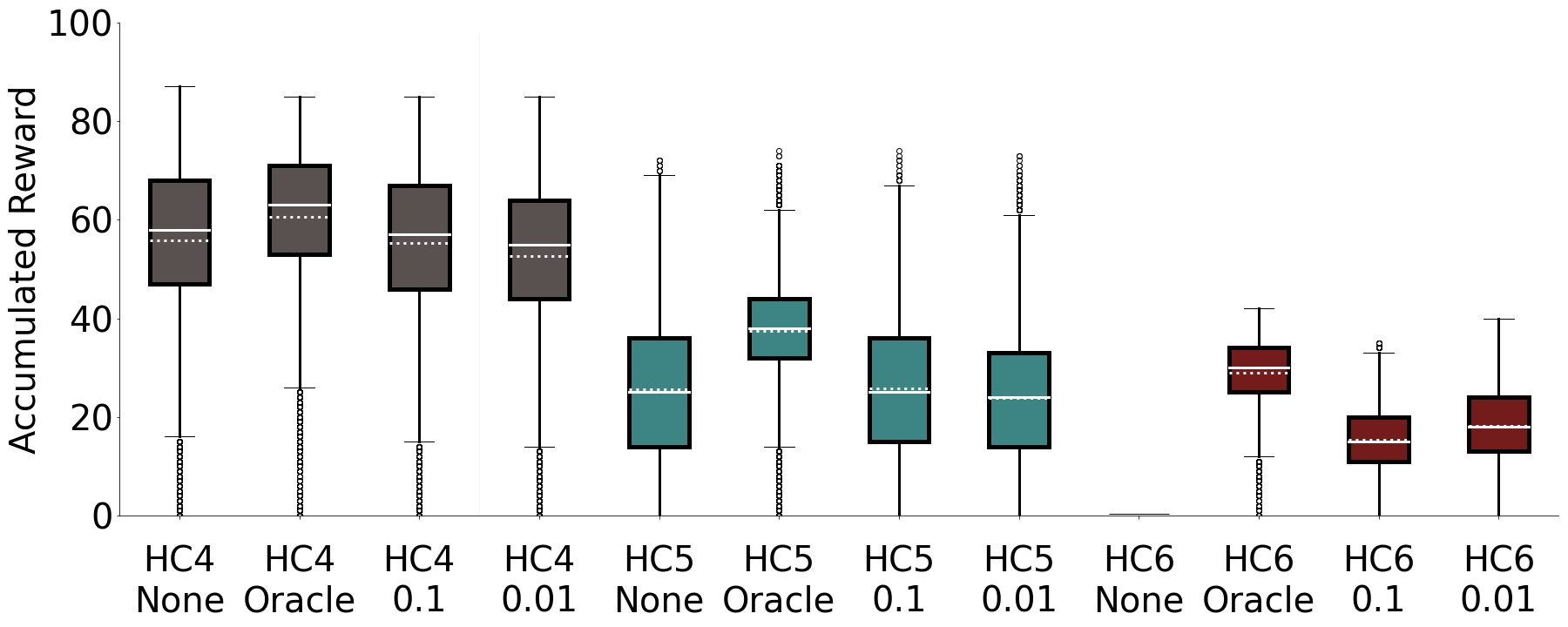}\\
(c) { DRIFT}
\caption{{\bf Differences in asymptotic accumulated reward across heat capacity settings for the bit cascade representation}, as shown by median plots for accumulated reward for last 1000 episodes (1000k steps) across 30 random seeds, with oracle and Bit Cascade GVF-based co-agents and Expected Sarsa as the learning agent for (a) fixed, (b) random, (c) drift conditions. Median shown as a solid white line, and average shown as a dotted white line on each box. GVF co-agent step size indicated as 0.1 or 0.01 on the second line.}
\label{fig:heat-capacity-sweep}
\end{figure*}

Figure \ref{fig:nexting-short-isi-comp} shows that the choice of temporal representation affects predictions made by the co-agent, including their timing and degree of aliasing at different points in an ISI's span with respect to token generation thresholds. We here focus on performance differences induced by these choices. We ran sweeps of 5000 learning episodes each of 1000 steps in length, across agent/co-agent pairs, in the fixed (ISI 8 with 2 hazard steps), random (ISI is in the range by 5 to 10 steps, selected independently after each hazard), and drift conditions (ISI changes by -1 to 1 steps after each hazard). Early learning differences in agent performance for Pavlovian signalling co-agent / control learning agent combinations (average accumulated episodic reward) are shown in Figs. \ref{fig:rep-fixed-gamma-comparison} and \ref{fig:rep-countdown-comparison}, while asymptotic learning performance across representations and with the oracle co-agent are shown in Fig. \ref{fig:rep-comparison-boxplot}. Of note, these results can be well considered with respect to the different forms of aliasing in our representations for accumulation and countdown predictions, as seen in Fig. \ref{fig:nexting-short-isi-comp}. 

First, we found that, unsurprisingly, Pavlovian signalling on the part of the co-agent was not just a benefit for solving the Frost Hollow tasks, but essential---agents without a co-agent were unable to obtain reward (Fig. \ref{fig:rep-comparison-boxplot}, leftmost data field). In the Fixed condition, it is possible to obtain a maximum accumulated reward of 50 per episode by earning one point every second hazard cycle. The Oracle co-agent is unrealistically strong in that it directly observes the time to the upcoming hazard instead of learning it, and learning agents connected to it are often the top performers. The remaining columns in the figure present a learning co-agent with a learning agent, and we find that, with the exception of one Bias unit co-agent, all are able to consistently obtain reward in all three environmental conditions. The Bias-0.1 co-agent is interesting in that it is competitive with the other learning co-agents in all environmental conditions. Although the feature vector output by the time representation is the same in all states, the GVF does contain a weight parameter that is updated on each timestep, and with an appropriately tuned learning rate it can oscillate across the fixed threshold to output useful Pavlovian signals for the learning agent to act on (c.f., Fig. \ref{fig:nexting-short-isi-comp}d). 
Paired with a countdown GVF question, the Bias representation with a learning rate of 0.01 was unable to track the stimulus, and obtained virtually no reward in the environment, except for a single reward in one of 30 drift trials. Paired with an accumulating GVF, it obtained approximately 29.937 reward per episode in the fixed condition, but only about 0.0019 and 0.0092 reward per episode in the random and drift conditions. Other representations performed comparably, indicating that the representation was less of a factor than learning rate and environmental condition on overall agent performance. Finally, we note that the oscillator, as expected, performed identically to the bit cascade representation when the number of steps between hazards was equal to its period in the fixed condition, but was unable to support a GVF in providing utility in the random and drift conditions (Fig. \ref{fig:rep-comparison-boxplot}).

In Fig.~\ref{fig:heat-capacity-sweep} we see the effect of varying heat capacity across conditions. In each condition the agent received 0.5 heat as it stood on the goal location. In the heat capacity 4 condition, the agent was rewarded for moving to the goal and staying there. We see this behaviour arise in the fixed, heat capacity 4 environment with no co-agent. When paired with an oracle co-agent, the agent performed worse than without this information. This is somewhat surprising---the information provided to the agent simplifies the problem, but performance declines. Here the information provided was not useful for maximizing return, and the agent needed to learn to ignore it. We expect that learning to ignore information is more complex than learning in the absence of the same information, and learning should take longer.  %

\subsection{How Important is the Choice of Control Learning Algorithm for Performance?}

\begin{figure*}[!th]
\centering
{\bf First 800 Episodes} \hfil {\bf All 5000 Episodes}\\
$\alpha_{gvf}=0.01$ \hfil $\alpha_{gvf}=0.1$  \hfil $\alpha_{gvf}=0.01$ \hfil $\alpha_{gvf}=0.1$\\
\ \\
\includegraphics[width=1.6in]{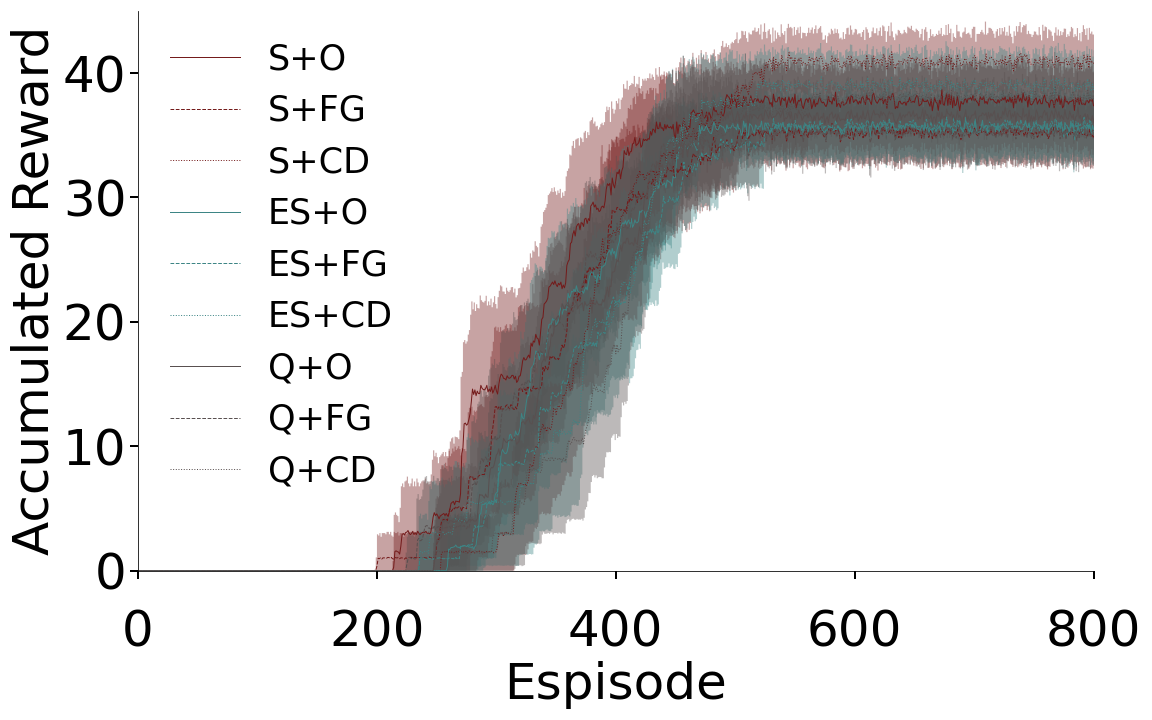}
\includegraphics[width=1.6in]{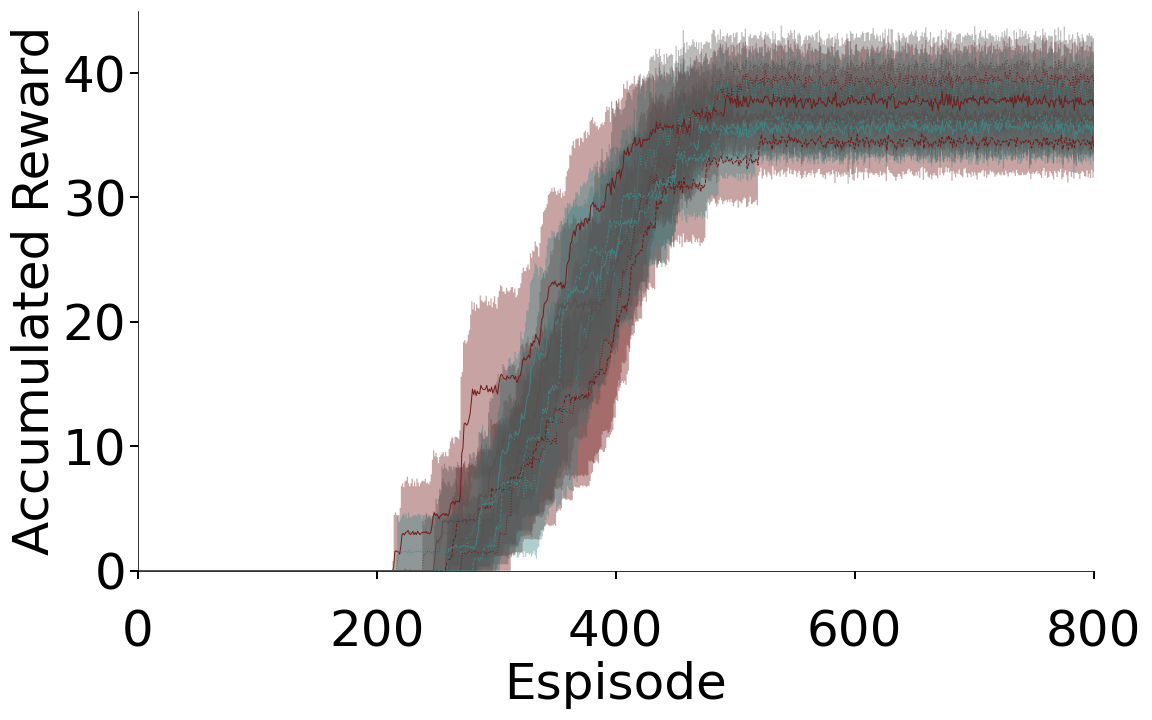}
\includegraphics[width=1.6in]{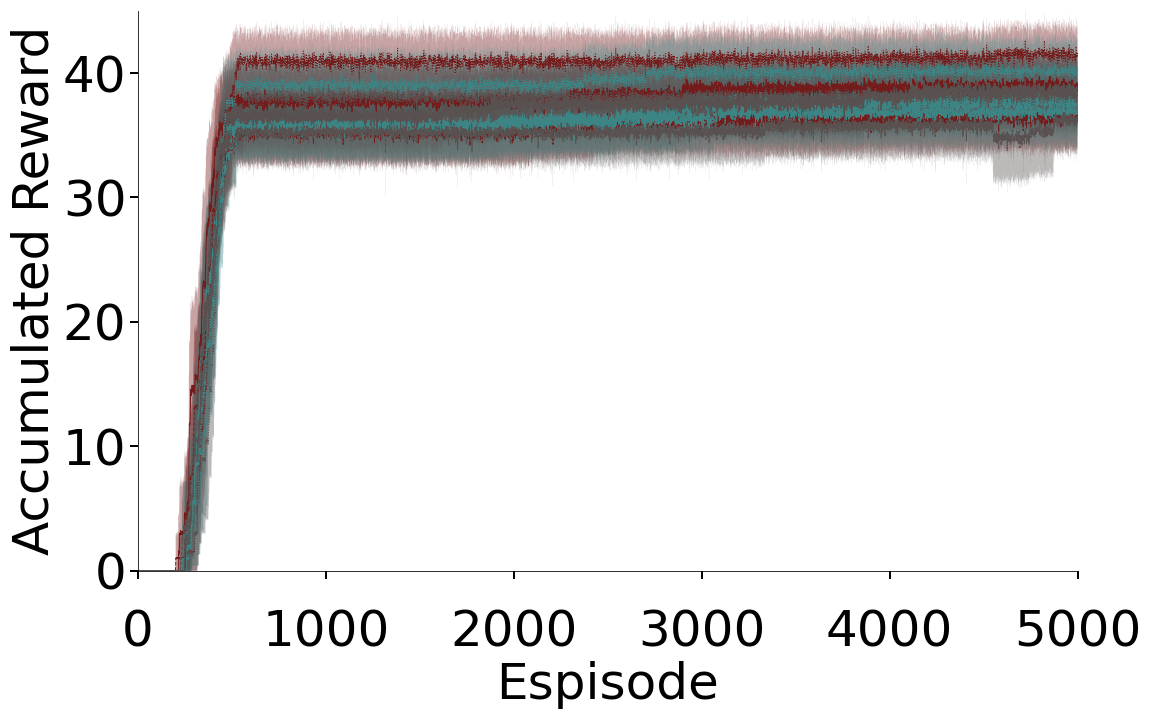}
\includegraphics[width=1.6in]{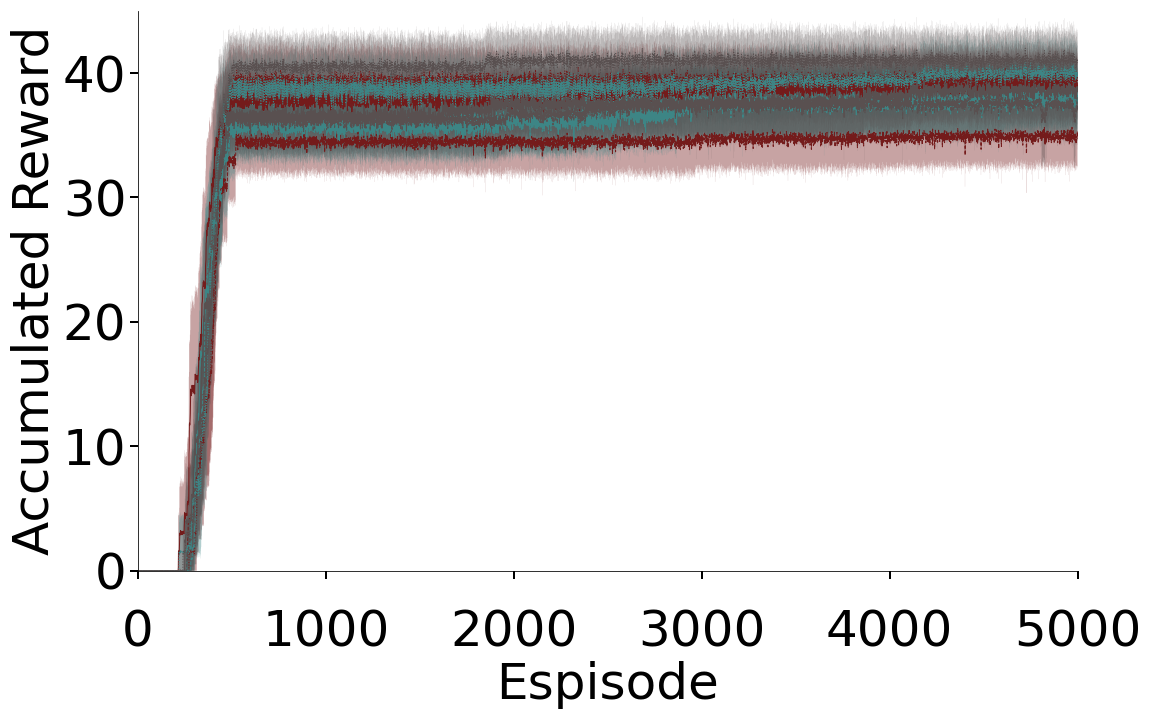}

(a) {\bf FIXED}\\
\ \\
\includegraphics[width=1.6in]{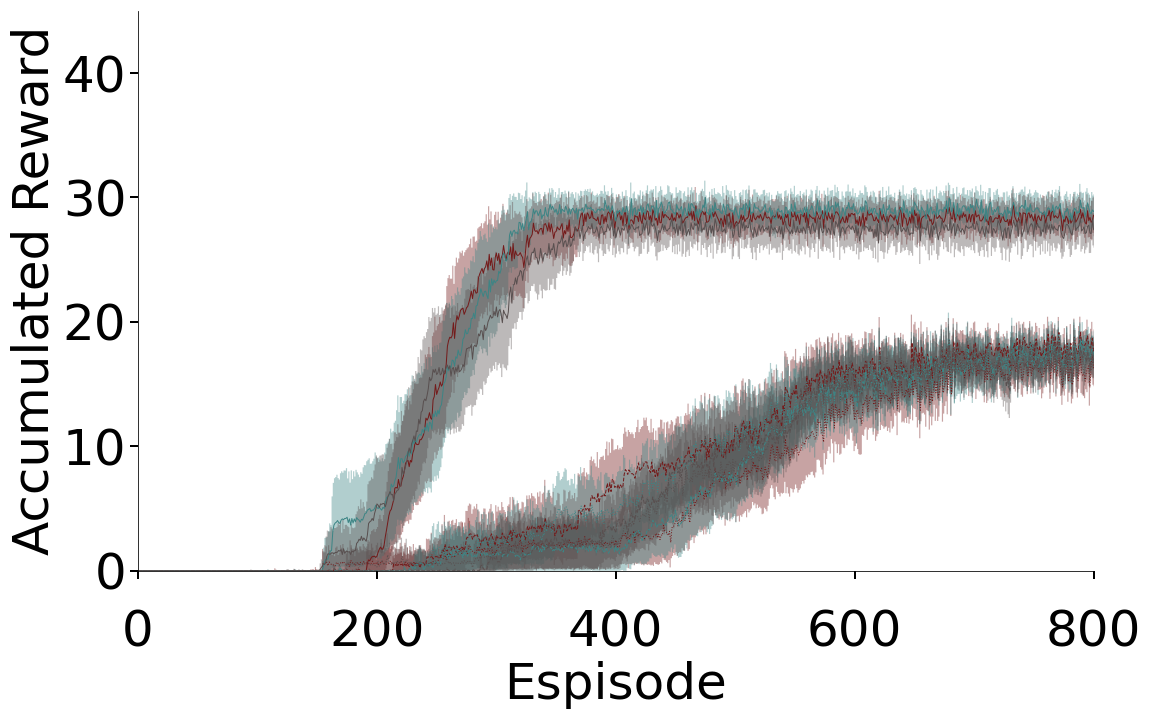}
\includegraphics[width=1.6in]{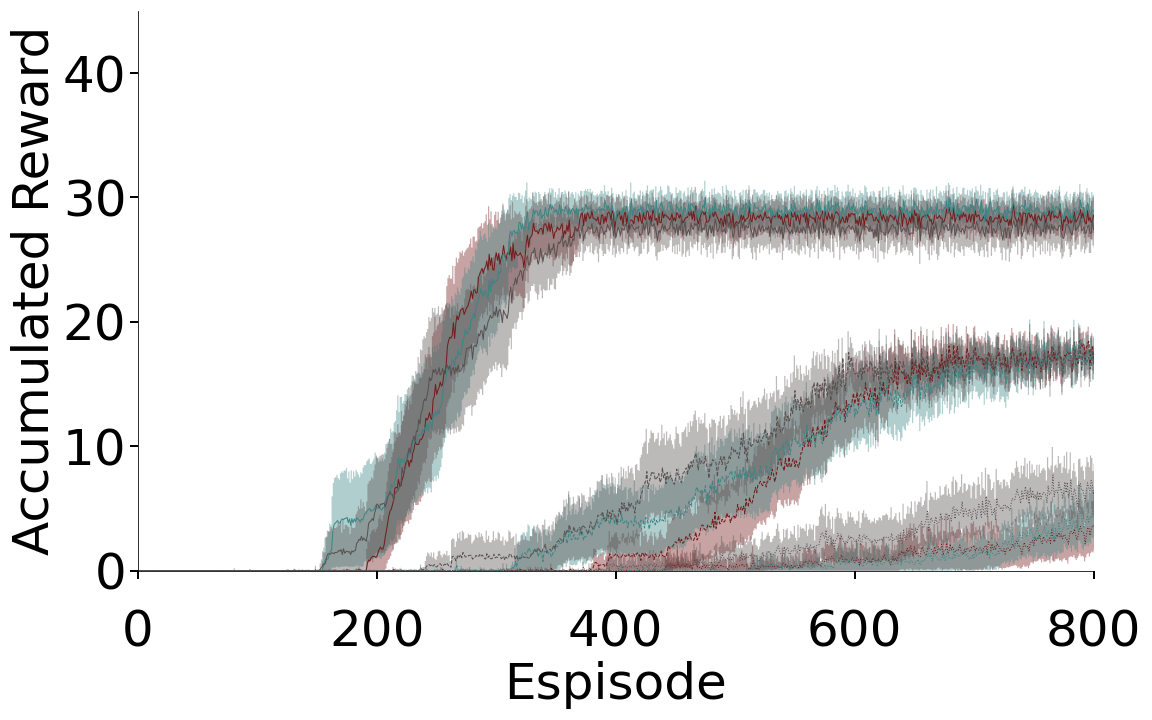}
\includegraphics[width=1.6in]{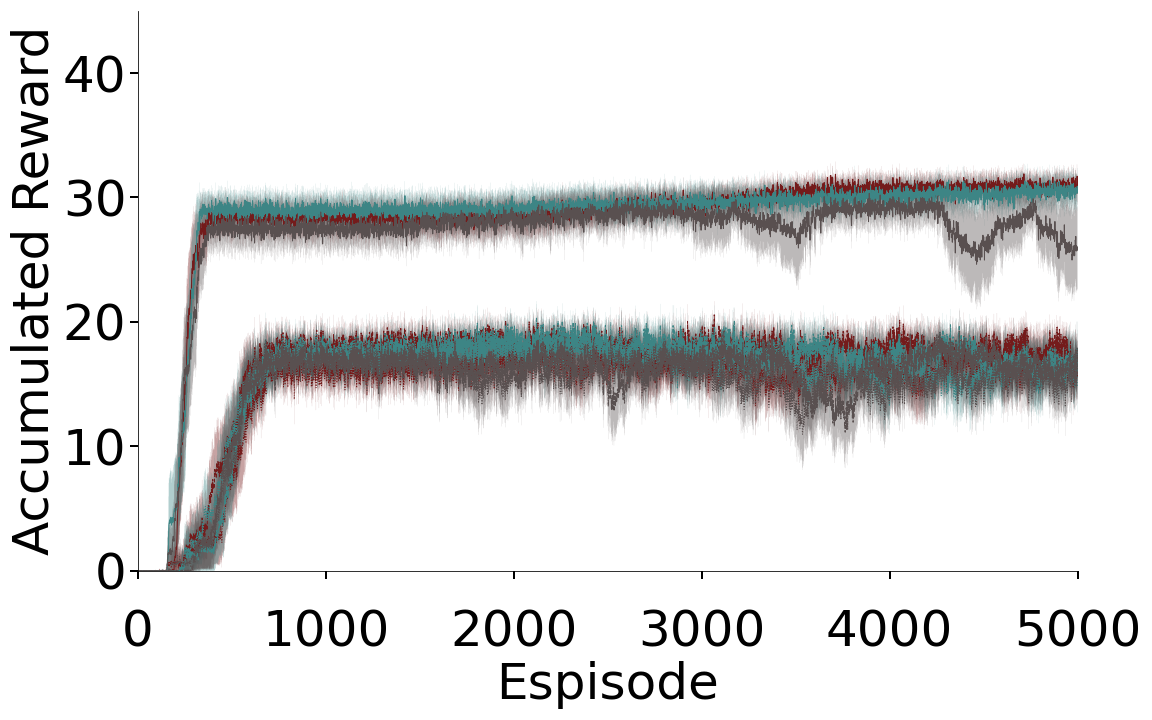}
\includegraphics[width=1.6in]{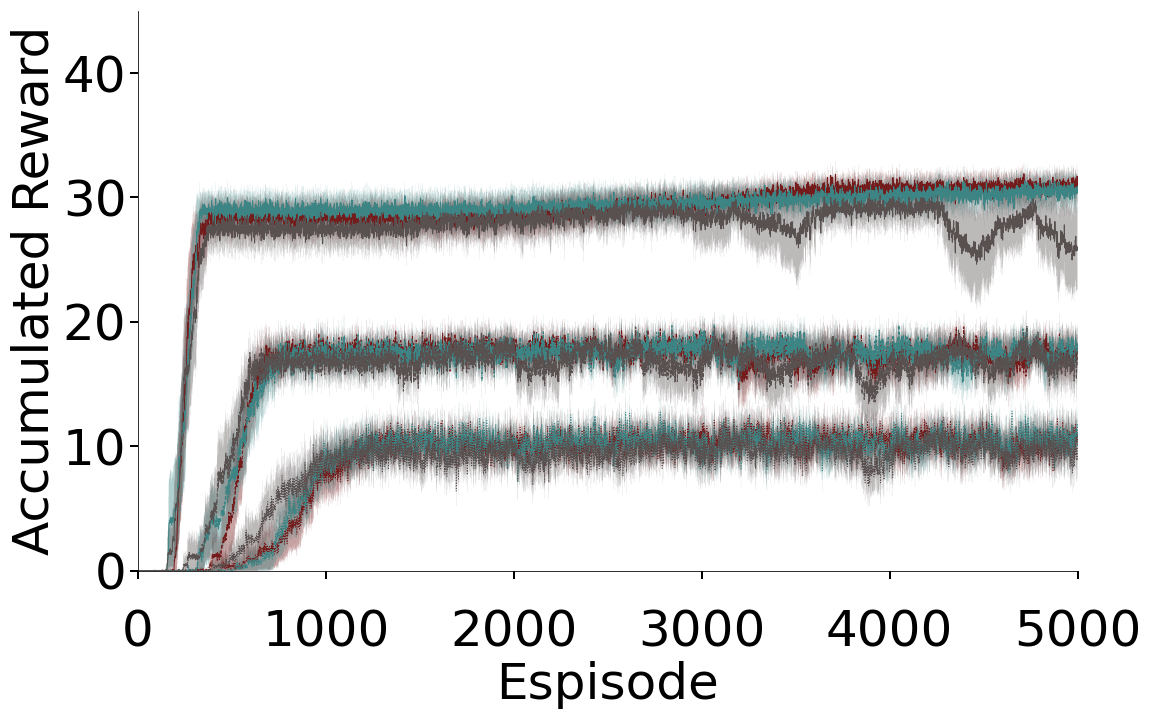}
(b) {\bf RANDOM}\\
\ \\
\includegraphics[width=1.6in]{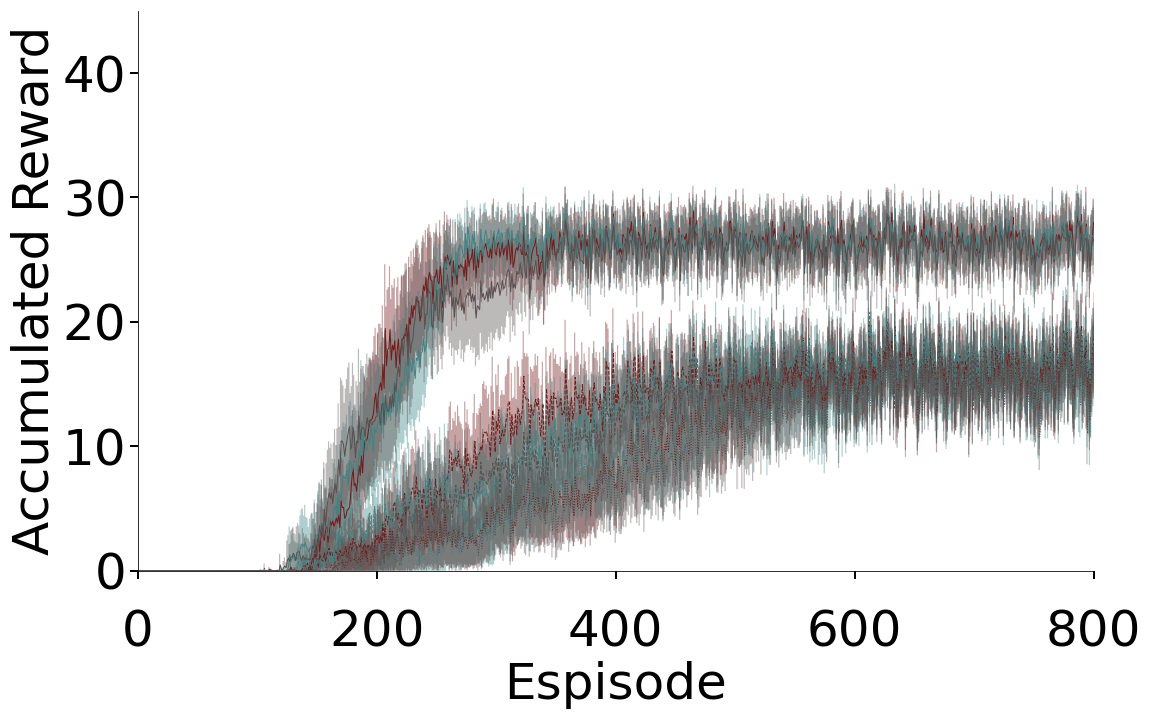}
\includegraphics[width=1.6in]{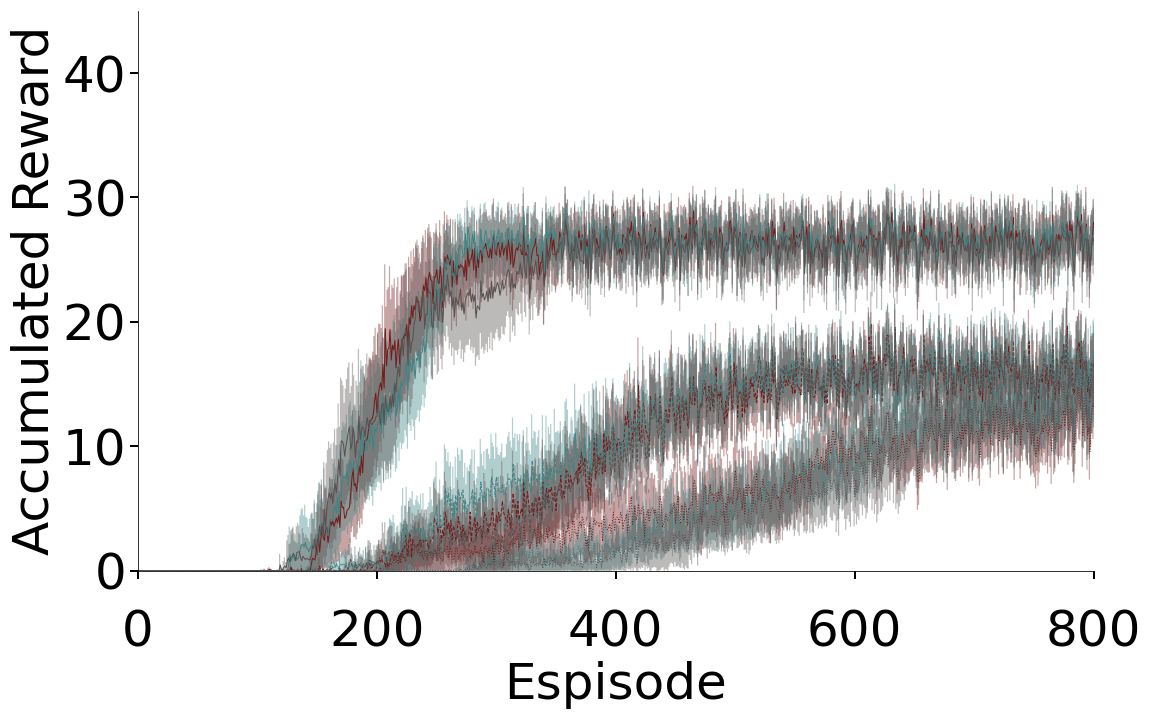}
\includegraphics[width=1.6in]{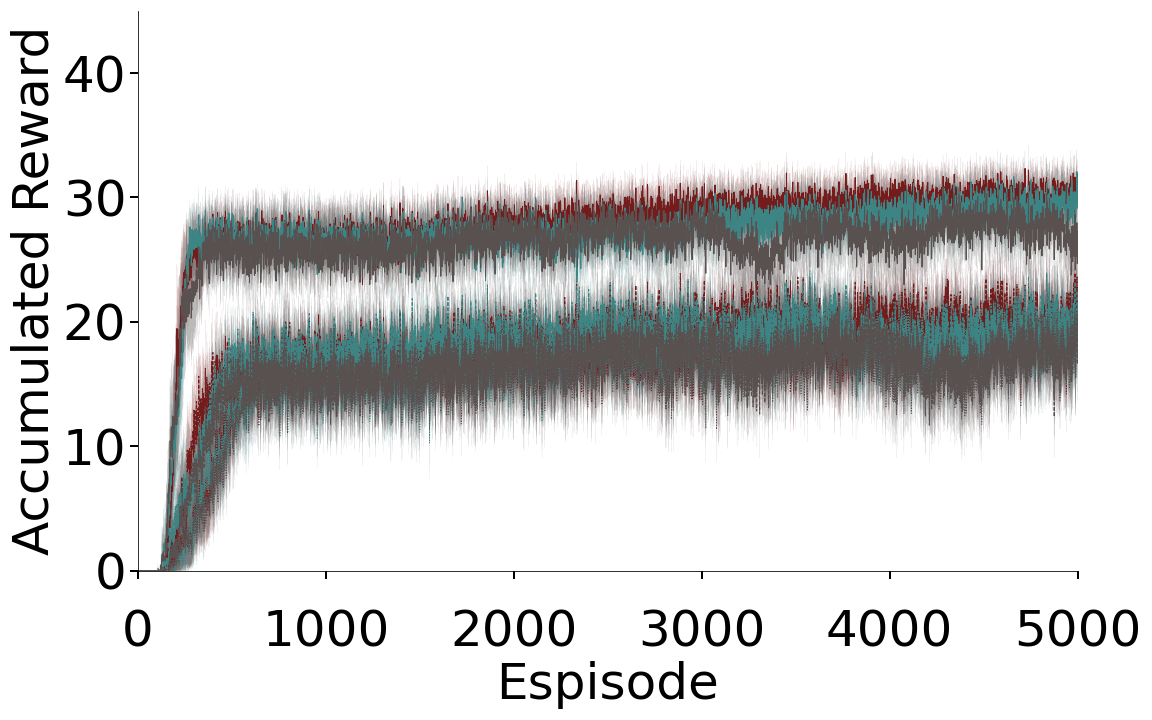}
\includegraphics[width=1.6in]{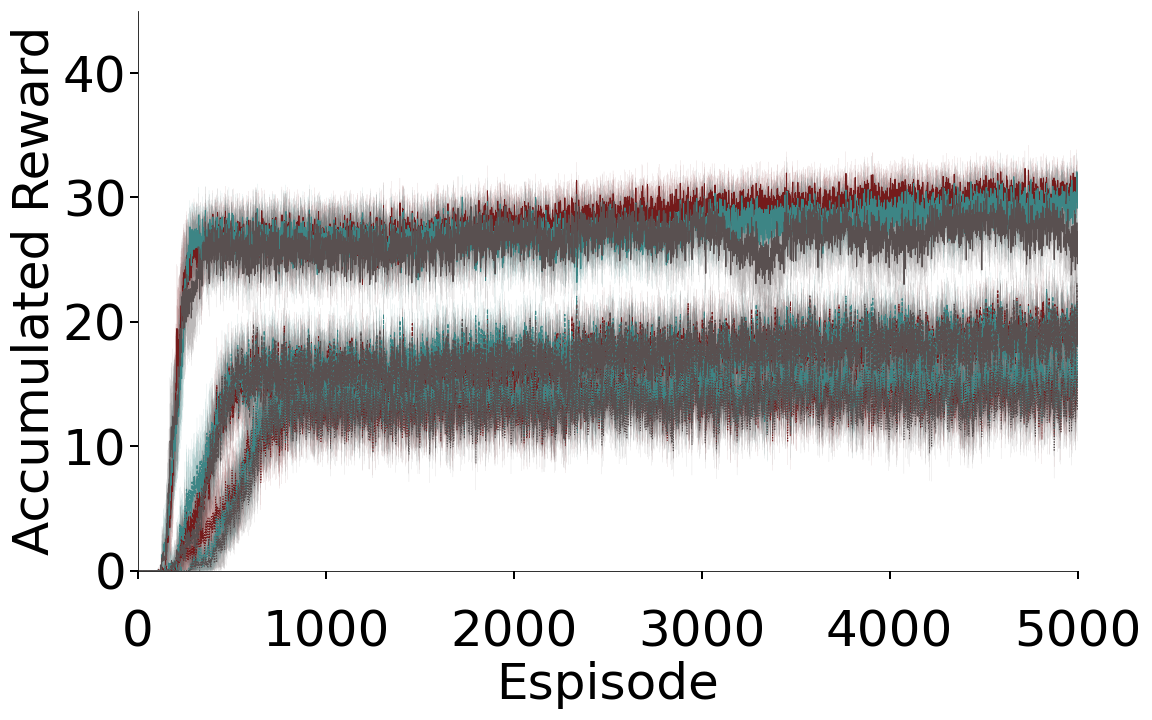}
(c) {\bf DRIFT}\\
\caption{{\bf Control learning algorithm had negligible impact on task performance}. Comparison of different agent learning algorithms over the first 800 episodes (8000 steps) and for the entire 5000 episodes (5M steps), averaged over 30 runs with a maximum heat capacity of 6, in (a) fixed, (b) random, and (c) drift conditions. Shown are Sarsa (red), Expected Sarsa (blue), and Q-Learning (grey), for the oracle-based co-agent (solid) and both the fixed-timescale and countdown GVF-based Pavlovian control co-agent (dashed and dotted, respectively) using a bit cascade representation. Shaded envelopes around traces indicate the standard error of the mean.}
\label{fig:alg-comparison}
\end{figure*}

Expected Sarsa was employed as the control learning algorithm throughout this study. We might expect that interactions between the agent and the co-agent could change due to the nature of the control learning update. We sought to understand the nature of these interactions by also comparing standard on-, and off-policy control algorithms, specifically the common algorithms Sarsa and Q-Learning (\cite{sutton2018}). Mainly, we aimed to see if there is compelling evidence to use one of these algorithms instead of Expected Sarsa.

The first of our additional of control learning algorithms, Sarsa, had a TD-error update that was computed as follows:

\begin{align}
\delta_t \gets R_{t+1} + \gamma w_t ^\intercal x(S_{t+1},A_{t+1}) -  w_t ^\intercal x(S_{t},A_{t})
\end{align}

In our second comparison control algorithm, Q($\lambda$), the off-policy nature of the error calculation is shown in the use of the maximum over all actions, written as:

\begin{equation}
\begin{split}
\delta_t \gets R_{t+1} + \gamma \max_a [w_t ^\intercal & x(S_{t+1},a)]\\
& -  w_t ^\intercal x(S_{t},A_{t})
\end{split}
\end{equation}

We aimed to examine any differences in these algorithms across the fixed, random, and drift conditions for cases where the control agent was coupled with a Pavlovia co-agent (accumulation and countdown cases). To highlight differences in any algorithms with respect to a proxy for the best possible co-agent, for this comparison we also included results for the oracle co-agent with perfect knowledge of the oncoming hazards (as described above). We chose to perform a focused study with the Bit Cascade representation based on the results in Figs.~\ref{fig:rep-fixed-gamma-comparison} and \ref{fig:rep-countdown-comparison}. The presence of degenerate policies in the heat capacity 4 condition resulted in restricting our focus further to the heat-capacity 6 condition. We fixed the agent learning rate at 0.01, and compared conditions across two co-agent learning rates, 0.01, and 0.1. Exploration was implemented with an epsilon-greedy policy with two considered epsilon values, 0.1 and 0.01.

Figure \ref{fig:alg-comparison} illustrates the differences between these conditions. As a key result we find that the choice of learning algorithm is dominated by other factors.  Pairing an agent with an oracle resulted in a cluster of high performance agents, with little variation between the control learning algorithms. We note the instability of Q-learning in the environment when paired with the oracle. Increasing epsilon to 0.1 removed this effect, but resulted in a lower asymptotic return. This appears to be an exploration-related issue, and we note the relation to the fact that agent weights are optimistically initialized to encourage exploration. The agent-oracle pairing was distinct from all other learning agent/co-agent combinations across all conditions, and independent of the control learning algorithm used. The choice of the co-agent learning rate $\alpha_{gvf}$ had a large effect on the performance of the accumulation and countdown GVF questions. High $\alpha_{gvf}$ resulted in clear separation in agent performance based on the question being asked. Low $\alpha_{gvf}$ reduced the performance difference resulting from question choice. In summary, the choice of control learning algorithm appeared to have a small effect on agent performance across conditions. Individual differences between algorithms do exist, as demonstrated by Q-Learning instability when paired with the oracle. These differences were smaller in magnitude in the settings we have focused on than differences induced by the hyper-parameter choices discussed above.

\section{Example Case Study of Human-Agent Interaction in Virtual Reality}
\label{sec:vr-experiments}

One of the main benefits of a machine co-agent is that, in principle, it is able to make predictions about the dynamics of the world that a human partner either cannot or does not want to compute on their own (possibly due to the difficulty or time-consuming nature of the computation, such as the case of cognitive offloading \citep{risko2016}, or the human's inability to sense relevant information). In order to convey the benefit of these predictions, it is natural that machine agents must be able to signal or otherwise communicate information to a human partner \citep{lazaridou2020,crandall2018}; such learned communication can be built upon relationships, and relationships can be built up through interaction \citep{scottphillips2009,scottphillips2014,knoblich2011}. Take for example your interactions with a wristwatch: if up to now it conveyed accurate time-information to you, you would have every reason to continue trusting its information the next time you consulted it. If its degree of competency degraded for some reason, and the information communicated were incorrect, you would quickly lose trust in the device and look to other sources for the information you need. Now suppose that your wristwatch was not designed to convey regular time intervals, but instead predict the onset of stochastically reoccurring events. How would your interactions with your wristwatch be affected by the fact that the device must continually learn, update, and change its behaviour {\em while you are using it}?

In this section we describe a pilot human-agent interaction study, investigating how time-based prediction agents can augment human predictions, and how the relationship between the human and agent develops over time as the co-agent develops competency. We extend the Frost Hollow domain described in previous sections to virtual reality (VR), and pair a human actor in that environment with a machine co-agent. Specifically, we pair the human with one of two co-agents (or with no co-agent as control). Each of these co-agents encode a different representation of time (a bit-cascade co-agent and a tile-coded-trace co-agent). We assess how a participant's performance and behaviour differs across co-agent types, using both quantitative and qualitative analyses. While the generality of any findings are limited by the use of a single expert participant, our aim is to discover interesting trends and themes that might deserve careful investigation with more participants in a future study.

\subsection{Human Interaction with\\Learning Systems}
\label{sec:prior-work}
Human interaction research regarding autonomous systems spans from early software interfaces for email and calendar applications \citep{Maes1994} to more complex and personal domains such as the control of prosthetic limbs \citep{embodied-cooperation,adaptive-artificial-limbs}, and has included a wide variety of automation techniques. Automation has traditionally been hand engineered to provide reliable performance, and therefore reliable human interaction. More recent machine learning systems are typically pre-trained before deployment, after which their parameters remain fixed. Research specifically involving interaction with continually learning algorithms has hitherto mainly focused on investigating agent learning dynamics using human interaction as part of the learning signal \citep{human-centered-rl-survey}. Autonomous systems that learn from human signals are important technologies, but system learning dynamics are inherently intertwined with interaction dynamics. \cite{power-to-the-people} convincingly argue the case for separating human interaction from agent learning in order to study ``how people actually interact---and want to interact---with learning systems''. They describe case studies involving people interacting with machine learning systems, and by specifically focusing on the human component of the interaction, they are able to discover novel modes of interaction, unforeseen obstacles, and unspoken assumptions about machine learners. A meta-review of factors that affect trust in human-robot interaction by \cite{hri-factors} suggests that system-specific factors such as behaviour, predictability, and failure rates greatly affect human trust in autonomous systems, justifying a system-specific investigation of human interaction with RL-based systems as distinct from other machine learning systems. One primary feature of the RL-based approach that we study in the present work that distinguishes it from other autonomous systems and warrants direct investigation is continual learning during the course of a task, and the effect that will have on human interaction.

 \subsection{Experimental Details and Parameters}

\begin{figure*}[ht]
\centering
\includegraphics[width=4.5in]{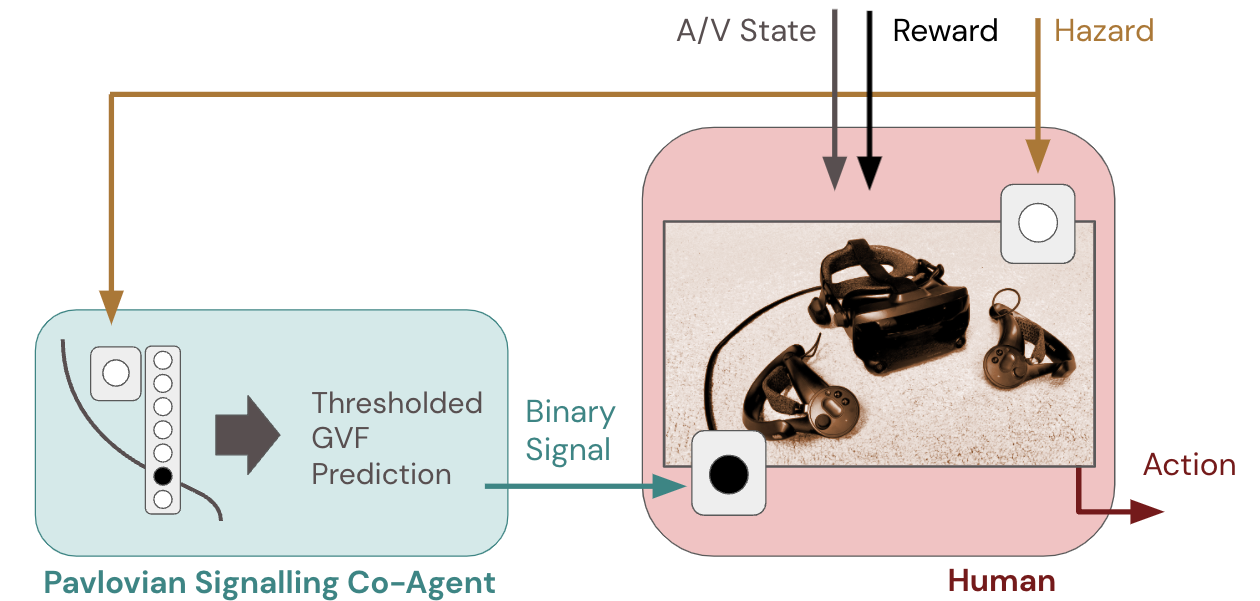}\\
\caption{{\bf Depiction of VR hardware and conceptual framework for human-agent interaction}. As in Sec. \ref{sec:control-experiments}  above, a Pavlovian control co-agent was paired with a control learner; in this case, the control learner was now a human who perceived state and signalling by way of a virtual reality headset and controllers, and moved through the environment (e.g., moved to gain heat or avoid hazards) using normal human locomotion. Co-agent signals and hazard presence were displayed to the human through on/off vibration of their hand-held controllers, while position in the world, reward, and other cues were presented in audio and visual form through the head-mounted display.}
\label{fig:vr-schematic}
\end{figure*}

\begin{figure*}[ht]
\centering
\includegraphics[height=1.9in]{Figures/FrostFP.PNG} \includegraphics[height=1.9in]{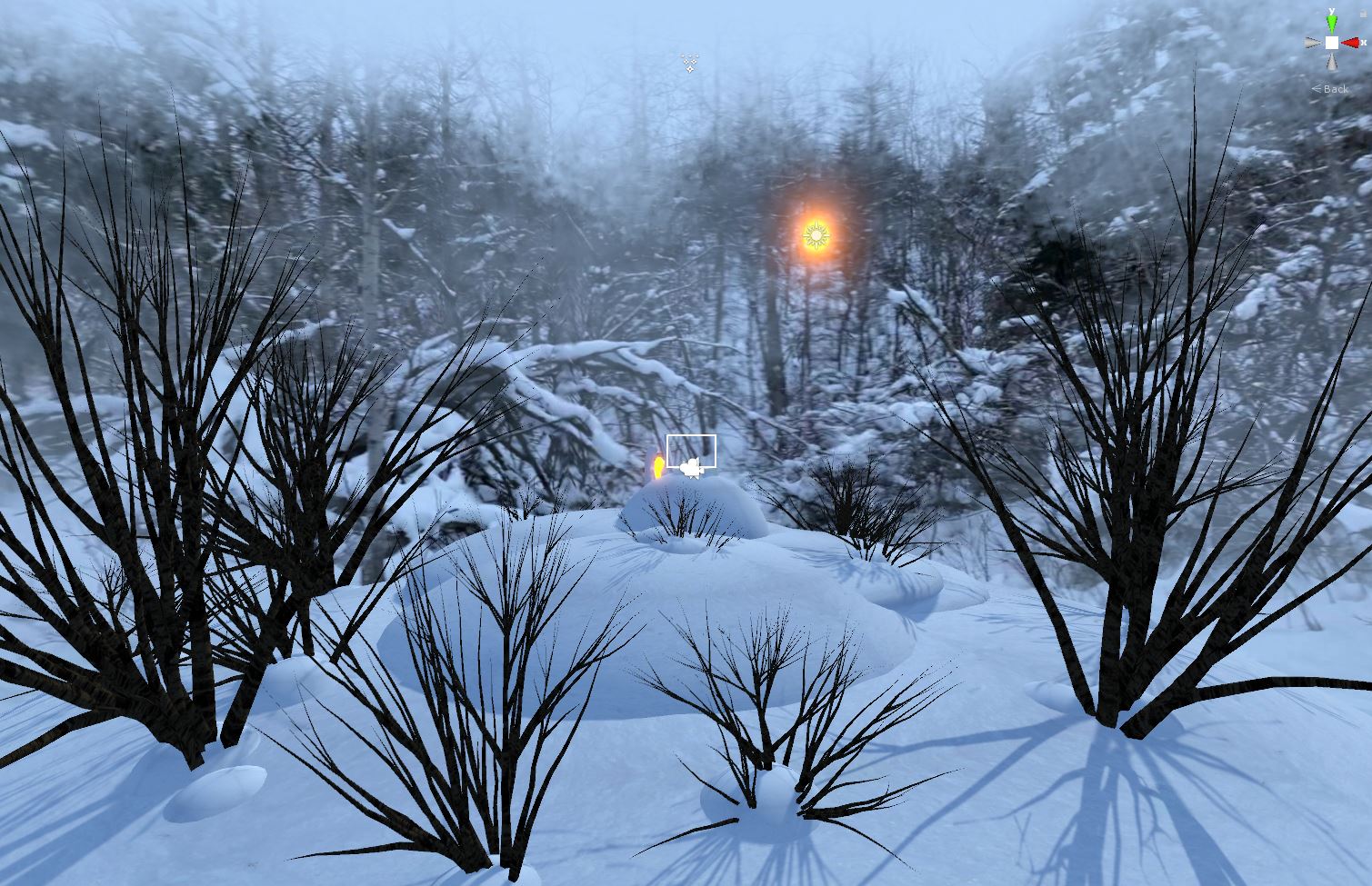}\\ 
{ (a) \hfil (b)}\\
\ \\
\includegraphics[height=1.9in]{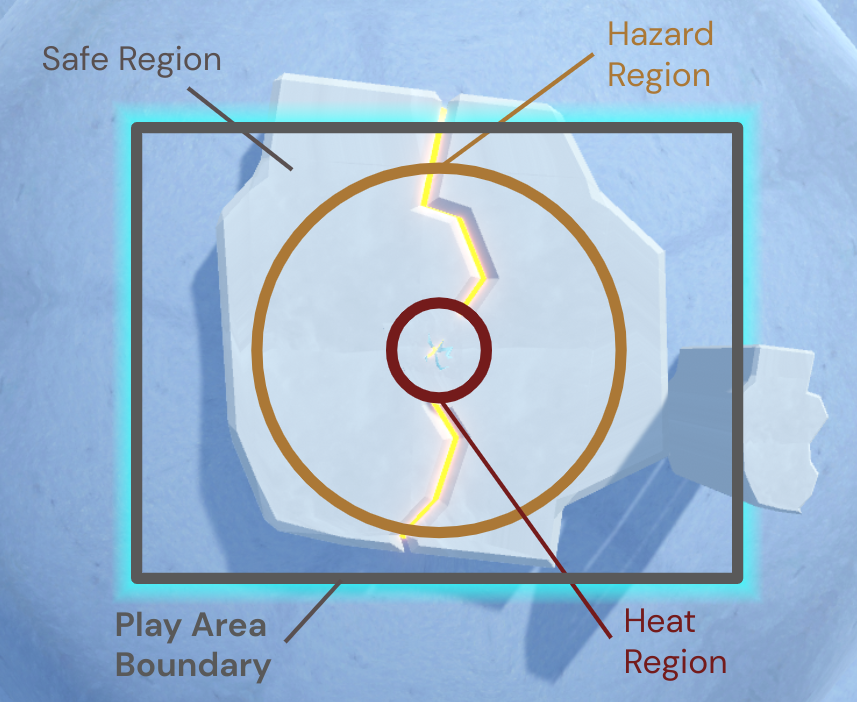}
\includegraphics[height=1.9in]{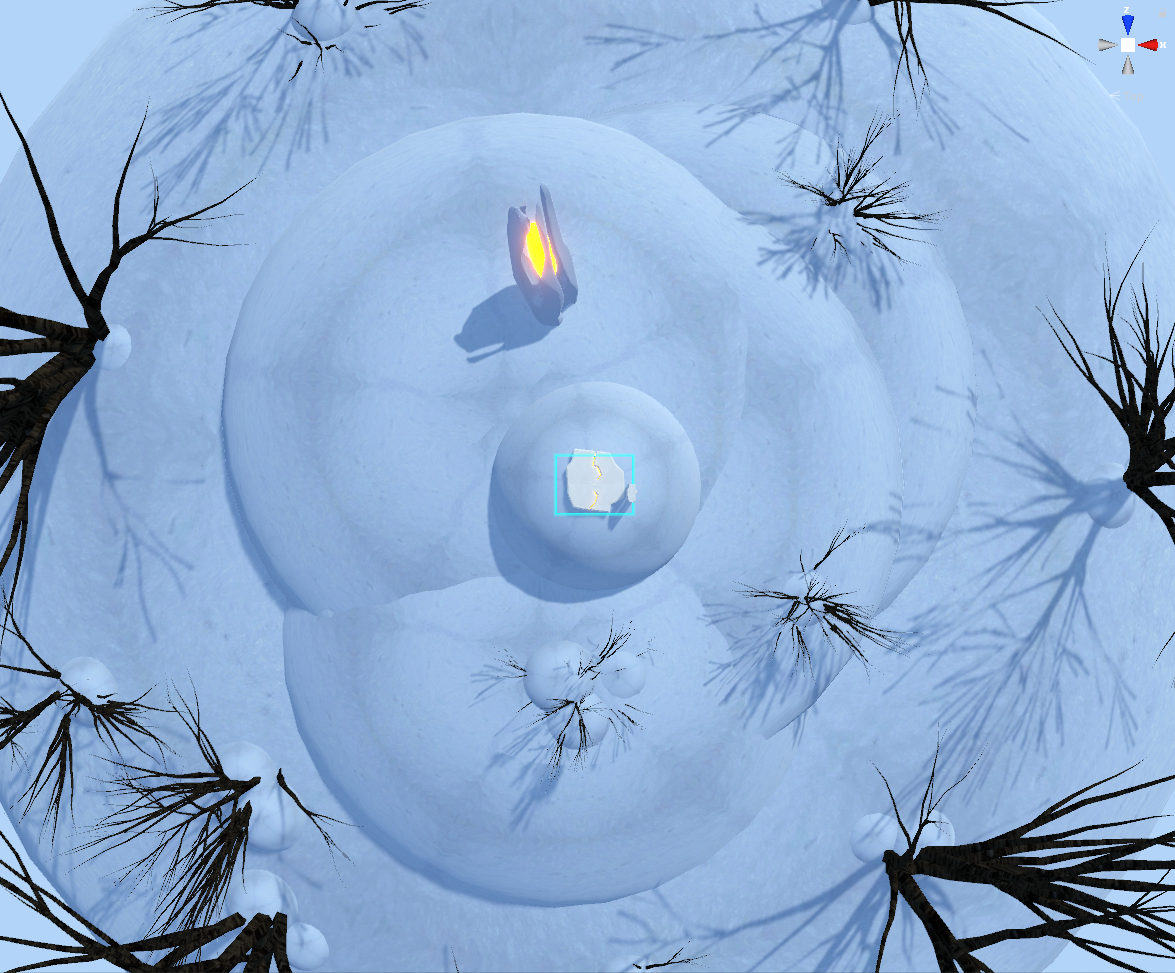}\\
{ (c) \hfil (d)}\\
\caption{{\bf Frost Hollow environment as implemented in virtual reality} for the human participant interaction case study: (a) first-person view of Frost Hollow from the centre of the playable domain, (b) far-field view of the field of play, (c) annotated top-down view of the field of play including heat and hazard zones, (d) far-field top-down view showing visual feedback elements such as the points-displaying crystal.}
\label{fig:vr-frosthollow}
\end{figure*}

We designed the VR variant of the Frost Hollow domain to mirror the abstract domain as closely as possible, while adapting it to time scales appropriate for human task learning. Further, for the Pavlovian signalling agents used in the experiments, we aimed to mirror their behaviour as closely as possible while adapting their temporal representations to the time scales appropriate for the VR implementation of the domain. One human participant interacted with this domain by way of a Valve Index headset and two handheld controllers (Valve Corporation, USA; headset max render rate of 144Hz), with this hardware shown in Fig.~\ref{fig:vr-schematic} (inset panel). Domain parameters, implementation details, and protocol details for the VR Frost Hollow domain are described in detail below.

Depicted in Fig.~\ref{fig:vr-frosthollow}, the VR Frost Hollow domain was implemented as a stand-alone scene in Unity 2019.2.17f1 (Unity Technologies, USA) with Steam VR (Valve Corporation, USA) as a three-dimensional winter world. An approximately 3m wide by 2m deep region was designed for participant activity, with the extremes of this region presented to the participant via Steam VR standard laser walls that render only upon the participant's proximity to the region's perimeter (Fig. \ref{fig:vr-frosthollow}c). This world had a base Unity time step length of approximately 8ms. While different in aspects of task dynamics and stimulus presentation, our VR protocol here roughly follows from the VR approach suggested in \cite{pilarski2019}.

{\bf Heat gain and loss}: The heat generation region at the centre of the participant's VR space---analogous to central state 3 in the abstract environment---was a circle 0.165m in radius, while the hazard region was a circle 1m in radius analogous to states 1-5 in the abstract domain. These regions are shown in Figure \ref{fig:vr-frosthollow}c. Participant location and orientation in space was exclusively defined as single-point position and rotation reported by their VR headset. Like the abstract domain, the participant aimed to fill a heat gauge (starting at 0.0 and accumulating to a maximum of 5.0); when this heat gauge was full, they could raise one of the VR controllers above the height of their head (measured as the height of the headset centroid) to empty their heat gauge and convert it into a single unit of stored heat (a ``point'' or unit of game reward). While within the heat region, the participant would collect approximately 0.1875 heat per second. As a result, they would need to stay in the heat region for at least 26.67s to fill their heat gauge. When the participant was in the hazard region during an active hazard pulse, they lost 0.2 heat per timestep---a substantial 25 heat per second---meaning any length of stay over 200ms in an active hazard would remove all accumulated heat.

{\bf Hazard dynamics and conditions}: Similar to the abstract domain, we also studied the three different inter-stimulus-interval (ISI) conditions: the fixed condition, the drift condition, and the random condition. To make the task more suitable for human participation, and to make the timing problem more challenging for the human player, the base ISI for the hazard pulses was set to 20s of real time (recall this to be the interval between pulse onset to pulse onset, set to 10 timesteps for the abstract domain)\footnote{In selecting the ISI for our human study, we note the commentary of \cite{paton2018} and others on the challenge of recognizing and generating temporal patterns in the range of tens of seconds.}; the hazard pulse length was a consistent 4s in all conditions. For the random condition, the inactive portion of the ISI was varied uniformly by [-4s, 6s] between 12s and 22s in length. For the drift condition, the inactive portion of the ISI was shifted by a uniform random amount between [-2s, 2s], with all values that would fall outside a minimum/maximum inactive ISI or 12s/22s cropped to the extremes of the range. As such, the total ISI for all trials fell between a maximum range of 16s and 26s, with each hazard pulse onset occurring between 12s and 22s after the last.

{\bf Machine learning co-agents}:  For this protocol, we presented the participant with three co-agent conditions following the temporal representation descriptions in Sec. \ref{sec:nexting-experiments}: {\em no co-agent}, a {\em bit cascade co-agent} (BC) with a state representation that advanced every 0.5s, and a {\em tile-coded trace co-agent} (TCT) with its trace tiled into a maximum of 40 bins and a per-step trace decay rate of 0.998. Both of these representations covered roughly the same number of bins when presented with an ISI of 20s (i.e., the number of active features was constant for both representations for this starting ISI). Re-implemented in Unity, the prediction-learning algorithms for these co-agents followed those described in Sec. \ref{sec:nexting-experiments} above, with parameters of the temporal-difference updates process being set for both co-agents at empirically determined values of $\alpha = 0.1$, $\lambda = 0.99$, and $\gamma = 0.99$ (for the accumulation prediction case, and as the pulse-based state conditional gamma for the countdown prediction case); values were selected and verified to ensure acceptable online learning speed over a 5min trial time. Co-agent tokens were generated (and thus signals were provided to the participant) when the co-agent's prediction of the hazard exceeded a threshold of 10 (compared to the threshold of 2.05 for the same prediction type in the abstract domain). While both prediction types, along with the oscillator and bias unit representations, were in fact run in the background for all trials of the experiment (even during the no-co-agent condition), acknowledging the significant length of the protocol we only selected the accumulation TCT and BC prediction to be conveyed as signals to the participant during the with-co-agent trials and thus as comparison conditions for this protocol. Co-agent-learned weights were not stored between trials and learning restarted fresh at the beginning of each trial from the same starting point (zero-fill weight initialization).

{\bf Visual task presentation}: The participant was presented with a first-person view of the Frost Hollow domain by way of their headset (Fig. \ref{fig:vr-frosthollow}a). As shown in Fig. \ref{fig:vr-frosthollow}, this was a small platform on top of rolling snow hills in a wrap-around forest background, with selected foreground trees and shrubs, a crystal artifact in their front-left field of view (Fig. \ref{fig:vr-frosthollow}a, left) that represented stored heat, and a visual presentation of a table where they could stand to pick up and place VR controllers and headset at the beginning and end of an experiment. Dynamic visual elements of the scene included snow that fell in the environment at different rates, randomly located pulses of yellow heat that rose from the platform while the participant was in the heat region, and a blue/white Unity bloom filter (colour shift increase in display intensity and light-source glow) that was applied to the participant's view when they were in the hazard region during an active hazard. The participant could also see when their heat gauge was between $~$4.5 and 5.0 units of heat by way of yellow light slowly filling the crack that ran through the centre of play area (Fig. \ref{fig:vr-frosthollow}c, centre line); to increase cognitive load, the participant by design could not see the status of their heat bar if it was not close to full (if they deemed it relevant, they would be forced to mentally track their own heat gain and loss). Upon converting a full heat gauge to stored heat (raising their controller over their head with a full heat gauge), there would be a visual flash and the crystal in front left of the participant would shift in intensity one step from black through orange to bright yellow to provide a permanent visual indicator of the points they had gained over the course of the trial.

{\bf Audio task presentation}: Task-relevant audio elements of the scene included a momentary sound at the time of hazard pulse onset  (abrupt and used only for this purpose), a chime to start and end the experimental trial, and an audio cue that played when a full heat gauge was stored by the participant. To further confound participant timing and increase cognitive load, a number of distractor audio cues with different onsets and length/periodicity were also introduced to the domain. These included a background wind audio sound that rose and fell and looped with $~$7s periodicity, and a collection of seven different audio samples of different lengths and intensities containing one or more bird calls and additional background wind sounds. Two distraction cues would be initiated after the beginning of each new ISI inactive period, with one starting at an uniform random time between inactive period start and 0.33 times the duration of that full pulse, and another starting randomly within 0.7 of the full pulse length; these audio cues had randomly varying volumes, were also randomly positioned in a 10m x 10m 3D space around the participant between 1--3m above the ground.

{\bf Vibrotactile task presentation:} Both VR controllers of the Valve Index were capable of providing vibrational feedback to the participant. During the full duration of the active period of the hazard pulse, the participant's left hand-held controller would vibrate at maximum intensity (regardless of whether they were inside or outside the hazard region.) For conditions where the participant was paired with a Pavlovian signalling co-agent, the threshold-based token (signal) from the co-agent to the participant was represented as a maximum intensity vibration to the right hand-held controller---if and only if the co-agent was signalling, the right controller would vibrate in the participant's hand.

\subsection{Participant Protocol and Instructions} 

For this case study we worked with a single participant (male, age 40, no history of sensorimotor impairments). Due to COVID-19 limitations in place for the duration of this work, we were unable to recruit external participants for this study, as intended and as per our approved human research ethics protocol for this work. {\em Our participant for this pilot case study was thus a member of the study team}. While this recruitment choice is a strong limitation in terms of introducing experimenter bias to the interpretation and execution of the provided trial protocol since our participant was an expert familiar with the domain of deployment, learning machines, and task dynamics, this choice has both disadvantages and advantages for the different comparisons in this study. We acknowledge these disadvantages and advantages over both the description of the protocol and analysis methods below, as well as our mediation strategies to blind the participant to key elements of the protocol.

The participant engaged with the Frost Hollow domain over the course of ten sessions, each consisting of nine trials that were five minutes in length (for a total of 90 trials, or roughly 10 hours of participant experimental time). Each individual session was conducted over the span of roughly 1h, with small breaks between each of the nine trials for the participant to remove the headset and if needed drink water or write down responses to qualitative prompts provided as part of the participant instructions (interpretation of qualitative feedback is detailed in Sec. \ref{sec:qual} below). Sessions began approximately one month after VR environment finalization, and were spread over a one month collection period, with one or two sessions per day on data collection days. This protocol was found to be slightly fatiguing physically and moderately fatiguing cognitively, depending on the trial.

As shown in the session and trial breakdown in Tab. \ref{table:protocol}, trials were identified as being either in the fixed (F), random (R), or drift (D) conditions; for each of these, the participant was paired with either no co-agent (N), the bit cascade co-agent (B), or the tile-coded trace co-agent (T). Due to the ease of identifying the no co-agent conditions, and the overall length of the experiment, FN, RN, and DN conditions were block randomized together as noted in Tab. \ref{table:protocol}, either starting or ending a session in alternating fashion. Finally, to also make identification of individual trials in the F and D conditions more challenging and to better cover the space of ISI lengths, starting ISI for the F and D conditions was set on a trial by trial basis according to the values in Tab. \ref{table:protocol-times}, while in the random condition the starting ISI was 16 seconds. 

The participant was instructed to begin each trial standing upright at the centre of the play area after donning the headset and controllers. They were free to look around the whole play area (unrestricted head movement and body pose) and use any preferred counting strategy or mental process to try to keep track of in-trial timing (i.e., pulse arrivals) but no external timing mechanisms. Further, when entering or exiting the hazard zone, the participant was instructed to as best as possible take three measured steps along their left/right axis to and from the heat zone (i.e., no running or jumping, stepping left to exit the heat region and right to re-enter it). As an expert in the task mechanics for both the abstract domain and the VR domain, the participant knew at experiment onset the visual, audio, and tactile presentations of the task, the heat gauge and reward mechanic and its presentation, and the objective of collecting stored heat ("point" as displayed on the in-game crystal). The participant also knew and had experienced but was not practiced in perceiving the three possible domain conditions (random, drift, fixed), and also knew and had experienced but was not practiced in perceiving the vibrational feedback from the hazard pulse and co-agent. Further, the participant knew to an expert level the mechanics of the BC and TCT prediction learning approaches, but had not practiced with or had experience perceiving stimulus from the final formulation of the two co-agents used in the trials.

\begin{table*}[th!]
\centering
\begin{tabular}{|c|ccccccccc|}
\hline
Session$\downarrow$/Trial$\rightarrow$ & 1 & 2 & 3 & 4 & 5 & 6 & 7 & 8 & 9\\
\hline
1 & {\bf FN} & {\bf RN} & {\bf DN} & FT & RT & DT & RB & DB & FB\\
2 & DB & FT & DT & RT & FB & RB & {\bf DN} & {\bf RN} & {\bf FN}\\
3 & {\bf RN} & {\bf FN} & {\bf DN} & DB & DT & FB & FT & RT & RB\\
4 & DT & FT & DB & RT & FB & RB & {\bf FN} & {\bf DN} & {\bf RN}\\
5 & {\bf DN} & {\bf FN} & {\bf RN} & DT & DB & RT & RB & FT & FB\\
6 & RT & DT & RB & FT & FB & DB & {\bf RN} & {\bf DN} & {\bf FN}\\
7 & {\bf FN} & {\bf RN} & {\bf DN} & DB & RT & RB & FT & FB & DT\\
8 & RB & DT & FB & DB & RT & FT & {\bf DN} & {\bf RN} & {\bf FN}\\
9 & {\bf RN} & {\bf DN} & {\bf FN} & RT & FT & DT & FB & DB & RB\\
10 & DT & RT & FB & RB & FT & DB & {\bf RN} & {\bf FN} & {\bf DN}\\
\hline
\end{tabular}
\caption{{\bf Protocol condition randomization scheme}. Control conditions with no co-agent shown in bold. Condition codes denoted as follows: F (fixed), R (random), D (drift), N (no co-agent), B (bit cascade co-agent), and T (tile-coded-trace co-agent).}
\label{table:protocol}
\end{table*}

\begin{table*}[th!]
\centering
\begin{tabular}{|c|ccccccccc|}
\hline
Session$\downarrow$/Trial$\rightarrow$ & 1 & 2 & 3 & 4 & 5 & 6 & 7 & 8 & 9\\
\hline
1 & 15.07 & {\bf 16.0} & 19.49 & 19.8 & {\bf 16.0} & 13.64 & {\bf 16.0} & 17.25 & 18.56\\
2 & 14.14 & 17.88 & 15.6 & {\bf 16.0} & 18.45 & {\bf 16.0} & 12.57 & {\bf 16.0} & 17.23\\
3 & {\bf 16.0} & 14.15 & 15.66 & 19.07 & 18.18 & 17.7 & 15.7 & {\bf 16.0} & {\bf 16.0}\\
4 & 12.97 & 15.67 & 19.49 & {\bf 16.0} & 19.04 & {\bf 16.0} & 12.38 & 12.96 & {\bf 16.0}\\
5 & 13.45 & 12.78 & {\bf 16.0} & 12.28 & 19.48 & {\bf 16.0} & {\bf 16.0} & 13.91 & 15.23\\
6 & {\bf 16.0} & 12.42 & {\bf 16.0} & 12.36 & 19.52 & 14.15 & {\bf 16.0} & 18.11 & 15.37\\
7 & 12.62 & {\bf 16.0} & 16.36 & 17.92 & {\bf 16.0} & {\bf 16.0} & 12.65 & 16.76 & 15.84\\
8 & {\bf 16.0} & 15.12 & 12.39 & 19.95 & {\bf 16.0} & 19.05 & 17.47 & {\bf 16.0} & 17.84\\
9 & {\bf 16.0} & 15.97 & 18.54 & {\bf 16.0} & 15.98 & 12.91 & 19.13 & 16.08 & {\bf 16.0}\\
10 & 14.13 & {\bf 16.0} & 17.64 & {\bf 16.0} & 15.63 & 18.61 & {\bf 16.0} & 14.09 & 12.98\\
\hline
\end{tabular}
\caption{{\bf Protocol initial ISI randomization scheme}. Bold indicates trials of the random condition, wherein the starting ISI was always 16.0 (and the ISI chosen randomly thereafter).}
\label{table:protocol-times}
\end{table*}

{\bf Data analysis and blinding strategy}: Our expectation was that the use of an expert participant with knowledge of the domain and co-agents would have the effect of largely removing or minimizing the effect of a participant learning curve over the multiple sessions and trials, increasing our ability to identify small performance and qualitative differences in the different conditions due to environmental changes and not due to participant learning errors, but increased the chance that the participant could uniquely identify either the co-agent being used or the exact experimental condition of a given trial (thus potentially injecting experimenter bias into the way they engaged with the task or perceived and used the agent feedback). 
Thus, to help preserve the benefits of a pre-trained familiar participant\footnote{n.b., The time course of participant learning of the task was not a key area of interest for the present study, though naturally of interest for follow-up studies.} while minimizing their ability to guess the nature of the task or co-agent, we put in place the following protocol elements to blind the participant as best as possible to the task. 

First, the participant never viewed the protocol orders in Tab. \ref{table:protocol}, nor the starting time values in Tab. \ref{table:protocol-times}. Further, the participant was not able to view the log files from their runs and had no contact with the resulting data or their analysis until their presentation in this manuscript; at the start of each trial, the only information visible to the participant was the session number and trial number entered into the experimental software to launch the trial. Analysis of the quantitative results was done by an independent study team member who designed the trial randomization scheme but did not participate in running the trial protocol. Finally, all qualitative observations written by the participant were given to a second independent study team member for coding, analysis, and critical discourse follow-up with the participant. Qualitative and quantitative analysis were conducted separately and without contact by other study team members or the participant, until a third phase of analysis wherein qualitative and quantitative results were analyzed together for common trends and disparities.

\subsection{Results: Quantitative Analysis of Single Human Case Study}
\label{sec:quant}
Conclusions drawn from quantitative analysis are limited in this study because of the single participant and limited number of trials. Statistical analyses were conducted to determine whether for {\em this participant} there were any differences in performance or behaviour across co-agent types. Data violated assumptions of normality in nearly every comparison, so non-parametric methods were used. For performance metrics data was grouped pair-wise by session, so Friedman's tests were conducted followed by Wilcoxon Signed-Rank tests with a Holm-\v{S}id\'{a}k correction for multiple comparisons. For goal region exit and re-entry timing, and signal-to-exit timing, data was analyzed on a per-pulse basis, and assumed independent; Kruskal-Wallis tests were conducted in these cases, as well as Mann-Whitney U-tests with a Holm-\v{S}id\'{a}k correction. Significance is reported in all cases at the family-wise $\alpha=0.05$ level. Results of the statistical analyses are reported in Tab.~ \ref{tab:significance}.

\begin{figure}
\centering
\includegraphics[width=\columnwidth]{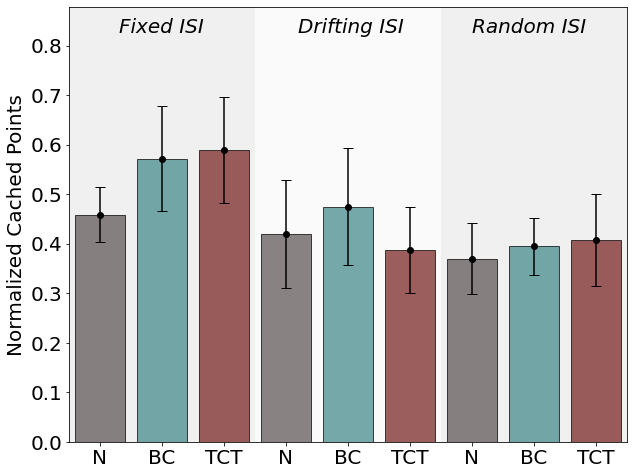}
\caption{{\bf Overall task performance}. Bars represent the mean over trials of total points cached, normalized by the maximum number of points possible. Error bars represent the 95\% confidence interval. N = no co-agent; BC = bit-cascade co-agent; TCT = tile-coded trace co-agent.}
\label{figure:h-a-mean-points-cached}
\end{figure}

\begin{figure}
\centering
\includegraphics[width=\columnwidth]{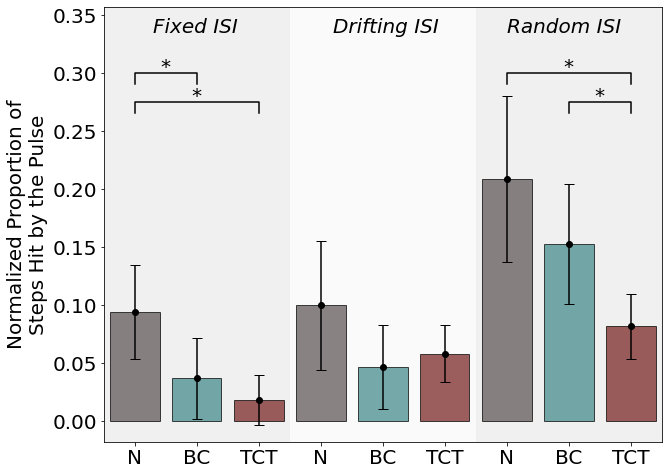}
\caption{{\bf Time-Steps hit by the hazard}. Bars represent the mean over trials of number of steps hit by the hazard pulse, normalized by the maximum possible. Error bars represent the 95\% confidence interval. N = no co-agent; BC = bit-cascade co-agent; TCT = tile-coded trace co-agent.}
\label{figure:h-a-mean-hitsteps}
\end{figure}

\begin{figure}
\centering
\includegraphics[width=\columnwidth]{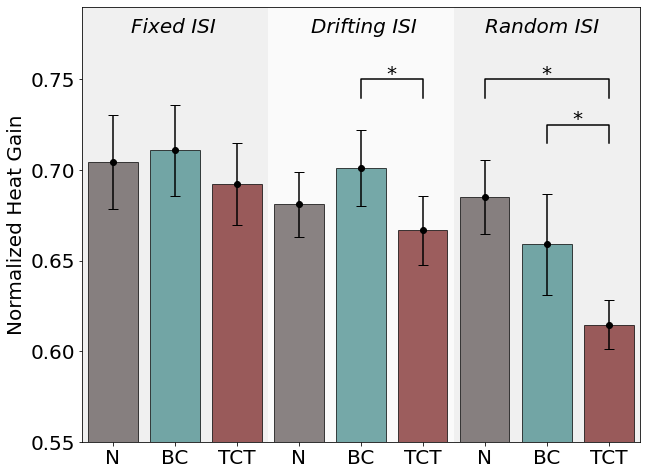}
\caption{{\bf Total heat gain}. Bars represent the mean over trials of total heat gain, normalized by the maximum possible. This corresponds to the amount of time spent in the goal region. Error bars represent the 95\% confidence interval. N = no co-agent; BC = bit-cascade co-agent; TCT = tile-coded trace co-agent.}
\label{figure:h-a-mean-heat-gain}
\end{figure}

{\bf Performance Metrics}:
Looking first at overall task performance (Fig.~\ref{figure:h-a-mean-points-cached}), we see a small (and not statistically significant) increase in performance in the fixed pulse-interval condition when the participant was paired with either co-agent. For the more difficult conditions where the ISI changes over the course of the trial, there is no clear difference in overall task performance depending on co-agent pairing. In general, these results suggest that overall task performance is not a clear indicator of any differences between human-co-agent pairings in this setting. Figure \ref{figure:h-a-mean-hitsteps} shows differences in the proportion of time-steps where the participant was hit by the hazard. In the fixed ISI condition, the participant spends less time being hit by the pulse when paired with either co-agent as compared to none. In the random ISI condition, the participant is hit by the pulse less when paired with the tile-coded trace co-agent than when paired with the bit-cascade co-agent, or no co-agent. Figure \ref{figure:h-a-mean-heat-gain} displays the participant's heat gain in each condition, which corresponds to the proportion of time spent in the goal region. Differences here appear in the more challenging conditions, where the participant spends less time in the goal region when paired with the tile-coded trace co-agent than when paired with the bit-cascade co-agent. Considering Figs.~ \ref{figure:h-a-mean-points-cached}, \ref{figure:h-a-mean-hitsteps}, and \ref{figure:h-a-mean-heat-gain} together, it appears that the participant engages in more cautious behaviour when paired with the tile-coded trace co-agent as compared to the bit-cascade co-agent (they gain less heat, and are hit by the hazard less often), while attaining comparable task performance. This result suggests possible differences in participant behaviour across co-agent pairings, which we consider in later sections on Movement Characteristics and Human-Agent Interaction.

\begin{figure}
\centering
\includegraphics[width=\columnwidth]{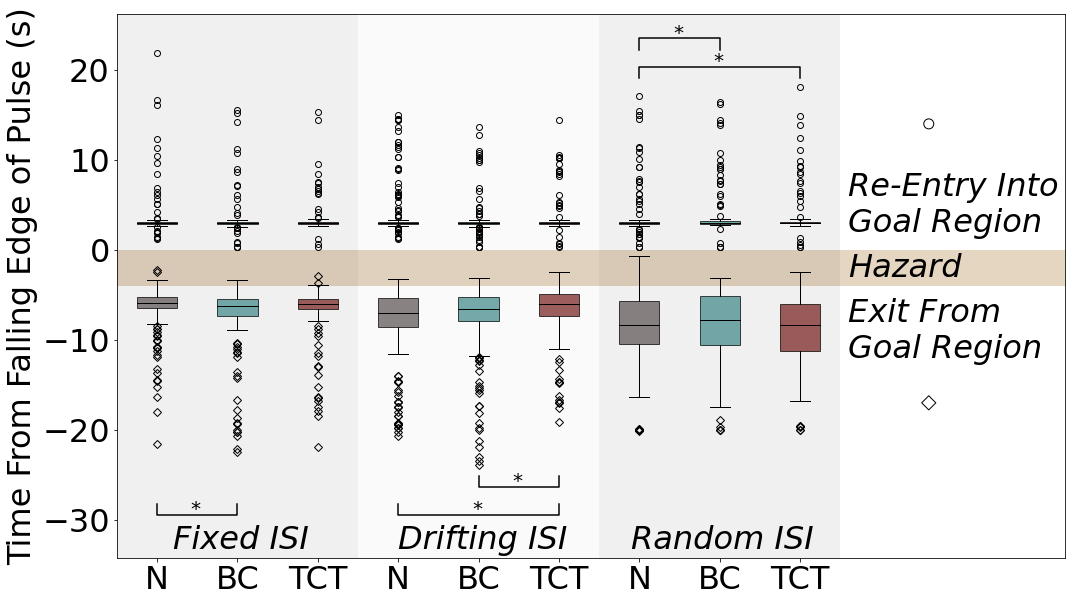}
\caption{{\bf Goal region exit and re-entry timing for each condition}. Individual data points are individual pulses within trials, and all timing information is synchronised around the falling edge of the hazard pulse. Time flows from bottom to top on this chart; the participant exits the goal region in advance of the hazard (below the yellow region), and re-enters the goal region after the hazard resolves (above the yellow region). Due to the spread of the data compared to the differences in group medians, statistical significance is not expected to correspond to practical significance.}
\label{figure:h-a-exit-and-re-entry-timing}
\end{figure}

{\bf Movement Characteristics}:
The participant's movement characteristics did not appreciably change across conditions or across trials. Instructed to move at a ``measured pace'', the participant maintained a reasonably consistent 0.89 m/s exit velocity (standard deviation $\pm$ 0.22 m/s) regardless of trial or condition. Similarly, position trajectories were not found to vary noticeably. The timing of exit and re-entry from and back to the goal region for each hazard pulse (Fig.~ \ref{figure:h-a-exit-and-re-entry-timing}) seems to vary much more substantially across ISI conditions than across co-agent types. Re-entry timing is very consistent relative to exit timing, which especially for the more difficult conditions can be quite variable. There were a few statistically significant differences found in median exit and re-entry times across co-agent pairings ($N > 140$ for each group), but the differences in the medians compared to the spread of the data suggests that these differences are not likely to be practically significant. 

\begin{figure}
\centering
\includegraphics[height=1.9in]{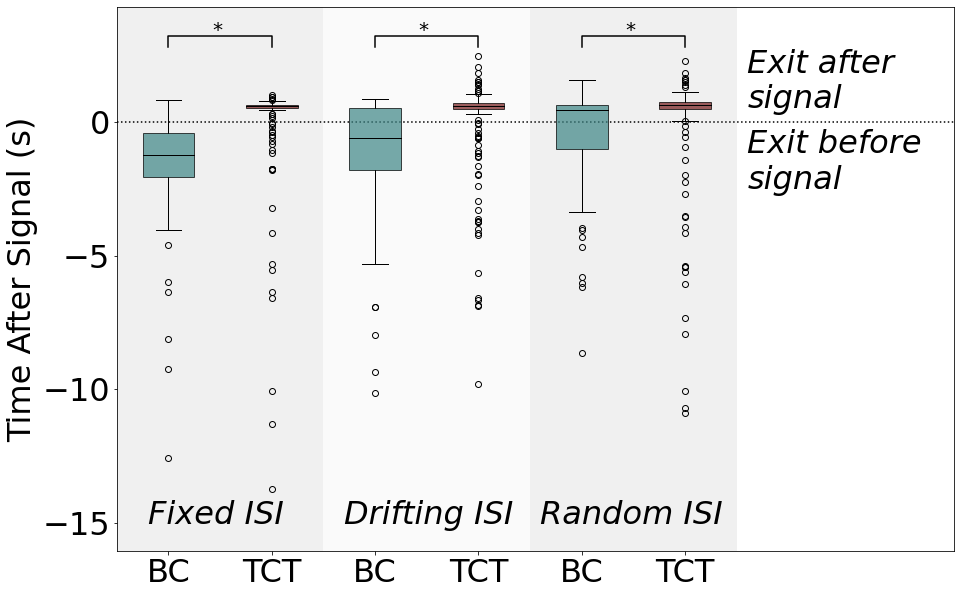}
\caption{{\bf Time interval between signal from co-agent and goal region exit}. Negative intervals indicate the participant leaving the goal region in advance of the co-agent's signal.}
\label{figure:h-a-signal-exit-interval-boxplot}
\end{figure}

\begin{figure*}
\centering
\includegraphics[width=\linewidth]{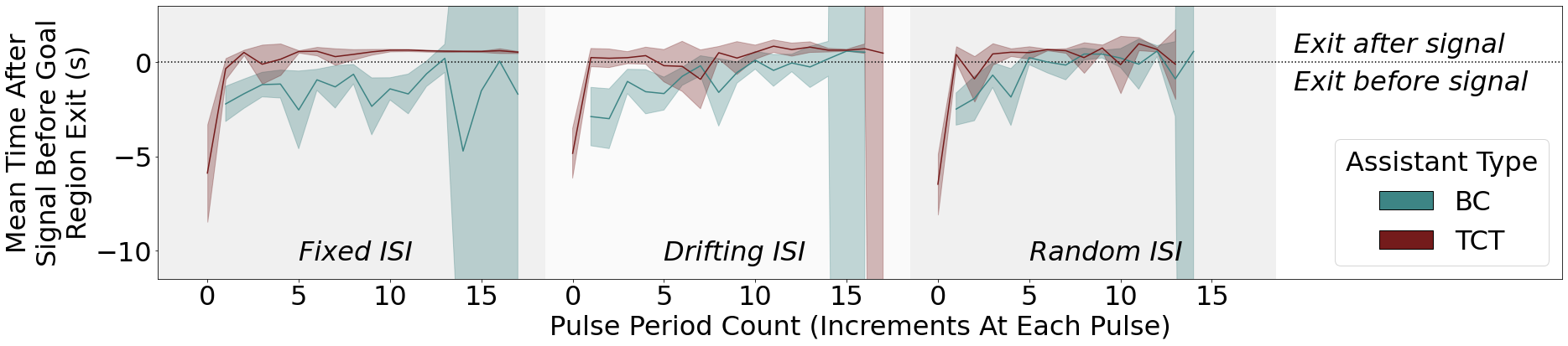}
\caption{{\bf Time interval between signal from co-agent and goal region exit}, shown as a trajectory over the length of trials. Data are shown as the mean (solid line) and 95\% confidence interval (shaded region) of the signal-to-exit interval for each pulse. Negative data indicate the participant leaving the goal region in advance of the co-agent's signal. Due to the randomization of the starting ISI (refer to Table \ref{table:protocol-times}) and fixed trial duration, some trials with shorter ISIs presented more pulses than others. This led to the occurrence of one or two trials with high pulse count (>14), resulting in the large confidence intervals at the ends of these plots.}
\label{figure:h-a-signal-exit-interval-over-trial-length}
\end{figure*}

{\bf Human-Agent Interaction}:
The length of time between the co-agent's signal and the participant's exit from the goal region is plotted in Fig.~\ref{figure:h-a-signal-exit-interval-boxplot}. A negative value on this chart indicates that the participant left the goal region before being cued by the co-agent. Here, we see that the participant exhibits clear behavioural differences when interacting with each co-agent ($p<<0.05$ in all comparisons). When paired with a co-agent using a tile-coded trace representation, the participant nearly always waits for the co-agent signal before leaving the goal region (data above the dotted line). When paired with a co-agent using a bit-cascade representation, the participant is much more likely to exit the goal region before the co-agent gives a signal. Expanding these box-plots out over trial time in Fig.~\ref{figure:h-a-signal-exit-interval-over-trial-length}, we can see the participant's response time relative to the co-agent signal change over the course of a trial for each of the conditions. In the fixed ISI condition, when paired with the tile-coded trace co-agent, the participant seems to move after the co-agent's cue as early as the second or third pulse of a trial. Under the same conditions, when paired with a bit-cascade co-agent, the participant relies entirely on their own internal timing. For the more difficult conditions, the participant eventually moves after the cue of either co-agent, but aligns their movements with the tile-coded trace co-agent's cue more readily than with the bit-cascade co-agent's cue. While it is tempting to interpret this feature of the data as the participant relying on the tile-coded trace co-agent's cue more than the bit-cascade co-agent's cue, there is insufficient evidence from these charts alone to conclude how the participant is using either signal, as we see in Sec. \ref{sec:qual}.

\begin{figure*}[!th]
\centering
\includegraphics[width=\linewidth]{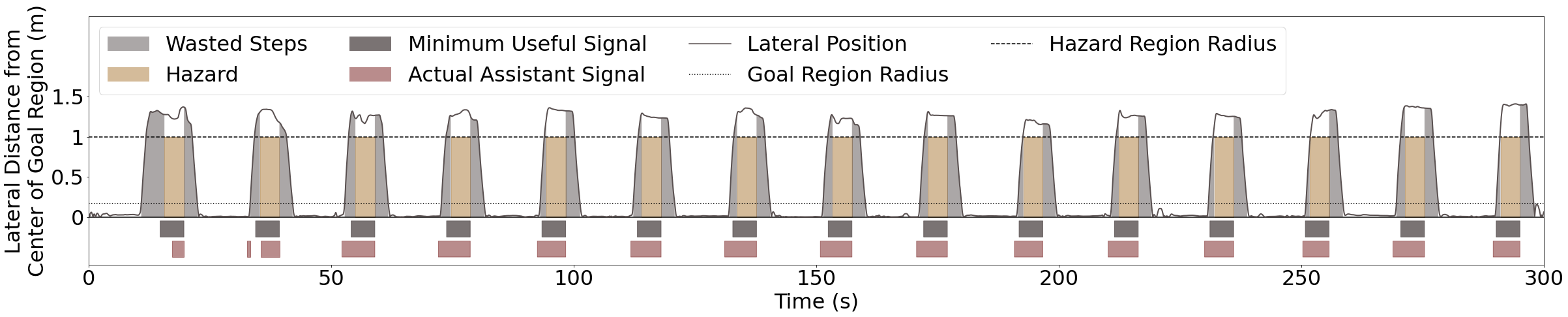}
\caption{{\bf A typical simple trial} (in this case fixed pulse interval condition, with tile-coded trace co-agent, from session 3).``Wasted steps'' are counted any time the participant is outside the goal region while the hazard pulse is inactive. ``Minimum Useful Signal'' corresponds to the signal that gives the participant exactly 0.89s lead-time before the hazard begins, calculated according to the participant's exit velocity.}
\label{figure:h-a-example-trajectory-simple}
\end{figure*}

\begin{figure*}[!th]
\centering
\includegraphics[width=\linewidth]{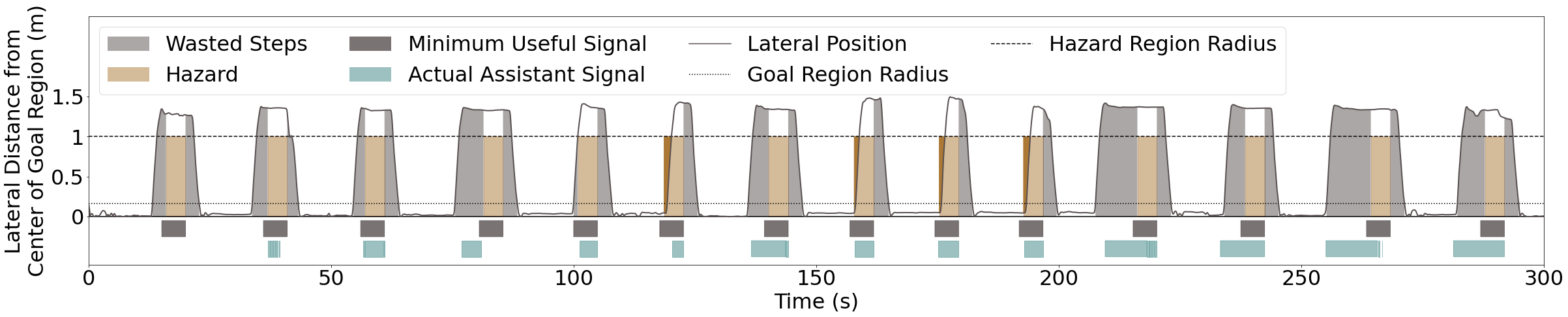}
\caption{{\bf A typical challenging trial} (in this case random pulse interval condition, with bit-cascade co-agent from session 3).``Wasted steps'' are counted any time the participant is outside the goal region while the hazard pulse is inactive. ``Minimum Useful Signal'' corresponds to the signal that gives the participant exactly 0.89s lead-time before the hazard begins, calculated according to the participant's exit velocity.}
\label{figure:h-a-example-trajectory-challenging}
\end{figure*}

To gain an intuitive sense of how the participant engages with the co-agent, we present two specific trials in detail: a typical ``simple'' trial (Fig. \ref{figure:h-a-example-trajectory-simple}, which corresponds to the fixed ISI condition, paired with the tile-coded trace co-agent of session 3), and a typical ``challenging'' trial (Fig. \ref{figure:h-a-example-trajectory-challenging}, which corresponds to the random ISI condition, paired with the bit-cascade co-agent of session 3). Beginning with the simple trial, we see the co-agent providing a useful (though inconsistent) signal beginning on the second hazard pulse, and reliably thereafter. The amount of wasted steps leading up to the hazard pulse diminishes to a narrow margin as the trial progresses. In the challenging trial we see the co-agent give its first useful signal only by the time of the fourth pulse. Until this time, the participant has been using their internal timing to determine when to leave the goal region in advance of the signal. The co-agent is then unable to give reliable signals for the next few pulses (pulses 5, 6, 8, 9, and 10), and the participant is hit by the hazard several times. By the 11th pulse, the participant resumes reliance on their internal timing to leave the goal region in advance of the co-agent signal, wasting many steps in order to avoid the hazard.

\begin{table*}[!th]
\centering
\begin{tabular}{|lr|>{\centering\arraybackslash}m{1.8cm}>{\centering\arraybackslash}m{1.8cm}>{\centering\arraybackslash}m{1.8cm}>{\centering\arraybackslash}m{1.8cm}>{\centering\arraybackslash}m{1.8cm}>{\centering\arraybackslash}m{1.8cm}|}
\hline
 & & \makecell{Points \\ Cached} & \makecell{Steps Hit \\ by Hazard} & \makecell{Heat Gain} & \makecell{Re-Entry \\ Timing} & \makecell{Exit \\ Timing} & \makecell{Signal-to-\\Exit Time} \\ \hline
\hline
\multicolumn{8}{|l|}{\bf{A Priori Tests}}\\
\hline
\hline
Fixed & $\chi^2(2)$ & 5.2 & \bf{14.0} & 1.4 & 0.9 & 5.5 &  - \\
 & $p$ & 0.0755 & \bf{0.0009} & 0.4966 & 0.6364 & 0.0632 &  - \\ \hline
Drifting & $\chi^2(2)$ & 1.4 & 1.4 & \bf{7.2} & 0.5 & \bf{11.5} &  - \\
 & $p$ & 0.4895 & 0.4966 & \bf{0.0273} & 0.7956 & \bf{0.0032} &  - \\ \hline
Random & $\chi^2(2)$ & 0.7 & \bf{11.4} & \bf{13.4} & \bf{11.9} & 2.1 &  - \\
 & $p$ & 0.7165 & \bf{0.0033} & \bf{0.0012} & \bf{0.0026} & 0.3540 &  - \\ \hline
\hline
\multicolumn{8}{|l|}{\bf{Post Hoc Tests (p-values)}}\\
\hline
\hline
Fixed & NvsB & 0.1415 & \bf{0.0432} & 0.9594 & 0.4518 & \bf{0.0398} &  - \\
 & NvsT & 0.1724 & \bf{0.0151} & 0.5550 & 0.4518 & 0.1632 &  - \\
 & BvsT & 0.7344 & 0.1763 & 0.4930 & 0.4633 & 0.1108 & \bf{0.0000} \\
\hline
Drifting & NvsB & 0.6831 & 0.4257 & 0.3642 & 0.6026 & 0.1247 &  - \\
 & NvsT & 0.6831 & 0.4881 & 0.3642 & 0.6026 & \bf{0.0018} &  - \\
 & BvsT & 0.5507 & 0.6465 & \bf{0.0206} & 0.6026 & \bf{0.0227} & \bf{0.0000} \\
\hline
Random & NvsB & 0.7990 & 0.2026 & 0.0593 & \bf{0.0037} & 0.3397 &  - \\
 & NvsT & 0.7990 & \bf{0.0278} & \bf{0.0151} & \bf{0.0037} & 0.3397 &  - \\
 & BvsT & 0.7990 & \bf{0.0278} & \bf{0.0329} & 0.4455 & 0.2159 & \bf{0.0000} \\
\hline
\end{tabular}
\caption{{\bf Results from statistical analysis of comparisons}. Significant results ($\alpha=0.05$) are indicated in bold text. Friedman's Tests ($\chi^2_{critical}(2)=6.20$) followed by Wilcoxon Signed-Rank tests were conducted for steps hit by hazard, points cached, and heat gain. Kruskal-Wallis tests ($\chi^2_{critical}(2)=5.99$) followed by Mann-Whitney U tests were conducted for exit and re-entry timing. Mann-Whitney U tests were conducted for signal-to-exit time.}
\label{tab:significance}
\end{table*}

\begin{figure*}
\centering
\includegraphics[width=\linewidth]{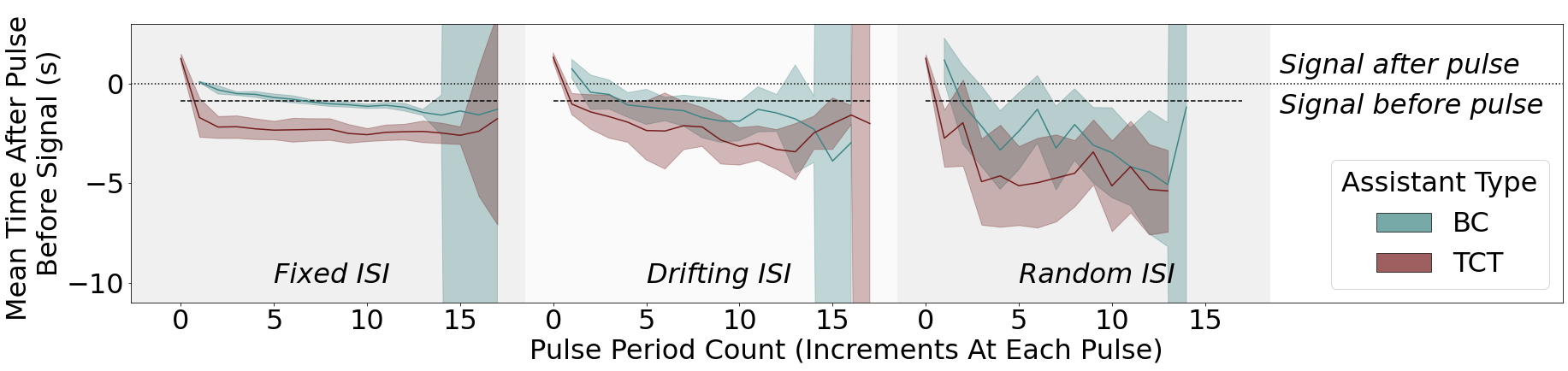}
\caption{{\bf Mean interval between signal from co-agent and pulse over trial length}. Shaded regions represent the 95\% confidence interval. The minimum useful lead time (shown as dashed line) before the hazard pulse (dotted line) is 0.89 seconds, corresponding to the time needed to exit the hazard region given the average speed of the participant during exit events. Due to the randomization of the starting ISI (refer to Table \ref{table:protocol-times}) and fixed trial duration, some trials with shorter ISIs presented more pulses than others. This led to the occurrence of one or two trials with high pulse count (>14), resulting in the large confidence intervals at the ends of these plots.}
\label{figure:h-a-learned-signal-timing}
\end{figure*}

\begin{figure*}
\centering
\includegraphics[width=\linewidth]{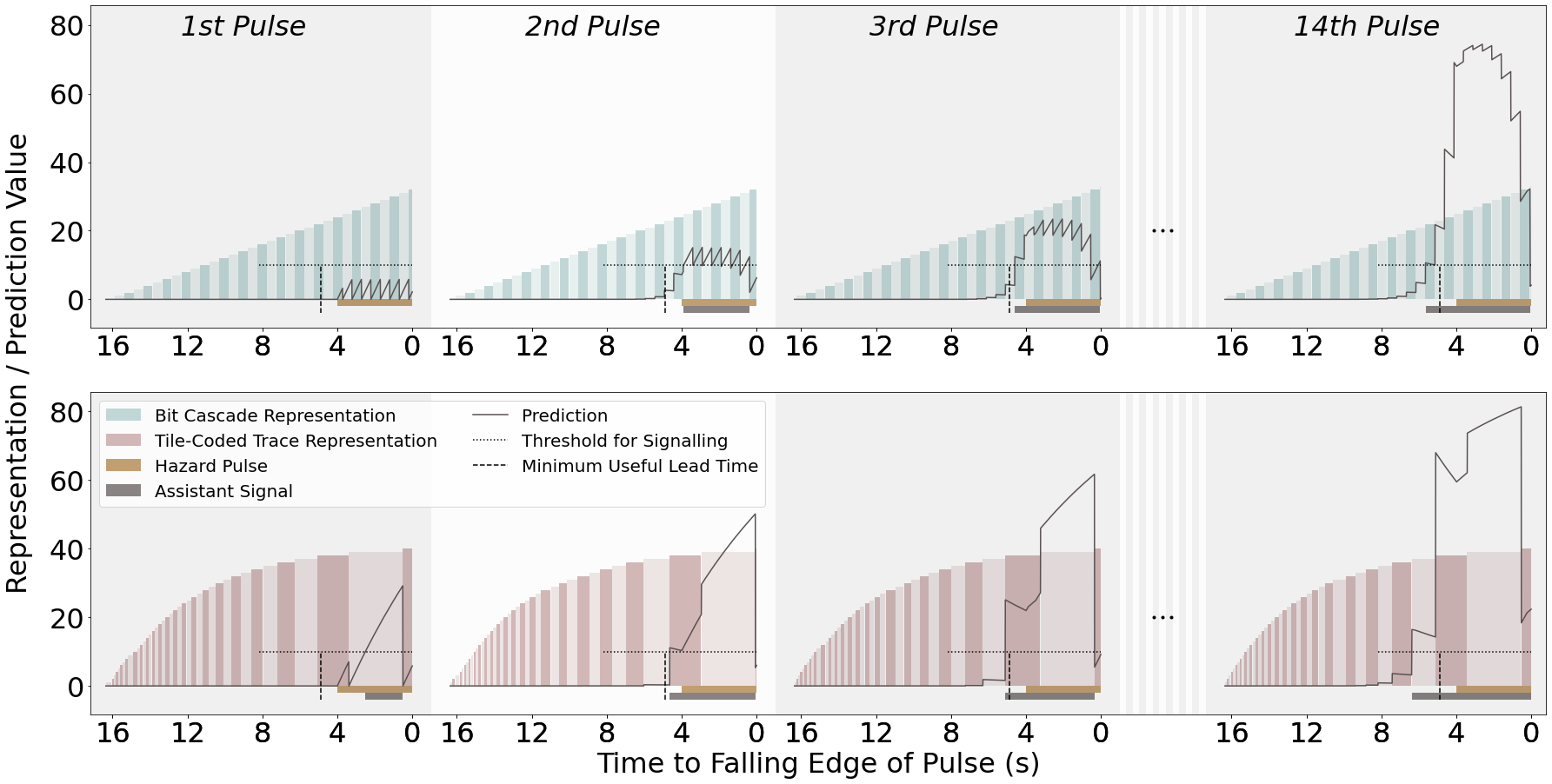}
\caption{{\bf Effect of representation and threshold choice on prediction and communication (fixed condition)}. Examining a single trial (here the fixed pulse interval condition of session 3, trial 2), we can observe the change in a co-agent's prediction over time (solid line). Shaded vertical bars depict each co-agent's representation of time; each representation bin is active for a period of time corresponding to that bin's width on this chart. Bars below the chart indicate the time for which the hazard pulse is active and the time for which the co-agent is signalling.}
\label{figure:h-a-effect-of-representation-simple}
\end{figure*}

\begin{figure*}
\centering
\includegraphics[width=\linewidth]{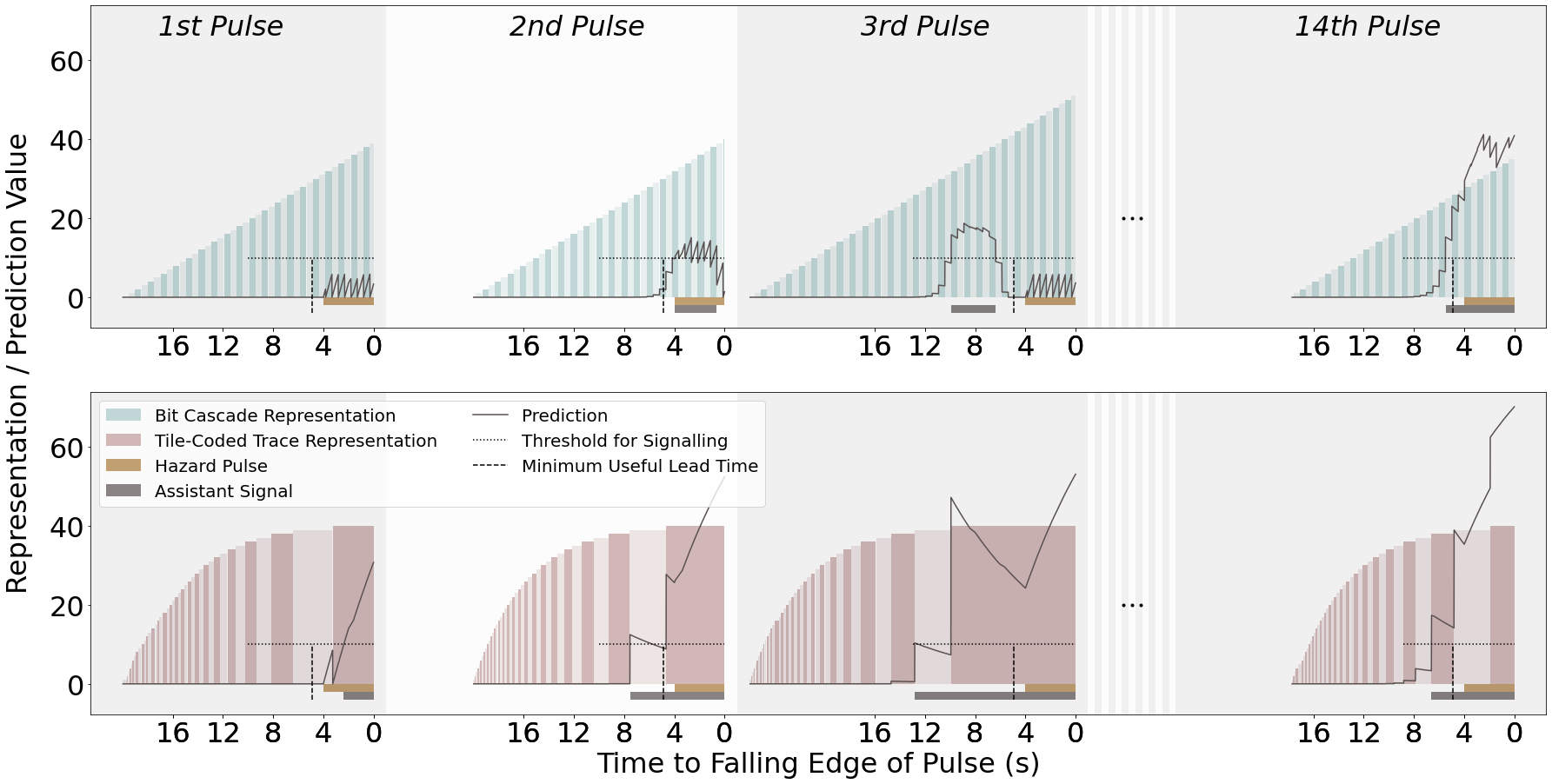}
\caption{{\bf Effect of representation and threshold choice on prediction and communication (random condition)}. Examining a single trial (here the random pulse interval condition of session 5, trial 3), we can observe the change in a co-agent's prediction over time (solid line). Shaded vertical bars depict each co-agent's representation of time; each representation bin is active for a period of time corresponding to that bin's width on this chart. Bars below the chart indicate the time for which the hazard pulse is active and the time for which the co-agent is signalling.}
\label{figure:h-a-effect-of-representation-challenging}
\end{figure*}

{\bf Co-agent Characteristics}:
We are able to examine co-agent learning directly because the co-agent's learning of the task does not depend in any way on the participant's actions. Figure \ref{figure:h-a-learned-signal-timing} shows the mean interval between the co-agent signal and the onset of the hazard pulse, for each pulse over the length of the trials. Recall that this can be interpreted as ``how long before the onset of the pulse did the co-agent's prediction of the pulse rise above the threshold for signalling?''. When the co-agent's cue is less than 0.89s before the hazard (above the dashed line), the signal doesn't give the participant enough time to react given how long it takes to leave the hazard region. In the simplest prediction task with fixed ISI, the bit cascade co-agent is unable to reliably give a useful signal (below the dashed line) until after about the sixth or seventh pulse while the tile-coded-trace co-agent is able to give a reliably useful signal after only the second pulse. More challenging conditions introduce more variance in these intervals, but the trend remains that the tile-coded trace co-agent provides useful signals earlier, and more reliably. The reason for this can be understood by examination of the representations and corresponding predictions made by each co-agent. Again for clarity we present data from two individual trials: an easily predictable case (Fig. \ref{figure:h-a-effect-of-representation-simple}, a typical fixed-ISI trial corresponding to session 3 trial 2) and a more challenging case (Fig. \ref{figure:h-a-effect-of-representation-challenging}, a typical random-ISI trial corresponding to session 5 trial 3). The regular, small binning of the bit-cascade representation means several features correspond to the hazard pulse. The value of each of these features must be updated in turn as they are activated, meaning that none of them are active for very long, resulting in the slowly and evenly rising saw-tooth crest in the prediction. By contrast, the tile-coded trace representation has large representational bins in the time span near the hazard. The few features here are active for much of the duration of the pulse, meaning that the predictions corresponding to these features can be updated rapidly. The more rapidly updated tile-coded trace predictions cross the signalling threshold sooner than those of the bit-cascade. In the more challenging condition (Fig. \ref{figure:h-a-effect-of-representation-challenging}), the effects of a changing ISI are clearly visible in the presentation of the third pulse. In this trial, the third pulse came along much later than the first two, and during a period of the bit-cascade representation that had not yet been activated. The tile-coded trace representation is better able to cope with this discrepancy, as the wide binning allows a previously activated feature to cover this time period.

The threshold and representation choices are therefore critical to co-agent signalling behaviour, and in turn the interaction between the co-agent and the participant. The threshold and representation bin-widths in this experiment were chosen considering late-trial performance, so that once the predictions stabilized both representations would give roughly equal notice before a pulse. A lower threshold or wider feature bins would likely have allowed the bit-cascade co-agent to provide reliably useful signals earlier in the trials. The bit-cascade co-agent is able to produce a more accurate prediction of the hazard than the tile-coded trace co-agent due to the finer feature binning in the pulse region (e.g., Fig. \ref{fig:nexting-long-isi-comp}). The trade-off for this accuracy comes in the form of slower early learning, which was found detrimental to the interaction between the co-agent and the human participant (a tension with the usual approach of seeking excellent asymptotic accuracy, perhaps at the expense of early learning speed).

\subsection{Qualitative Questions and Interpretation of Participant Feedback}
\label{sec:qual}

\begin{table*}[th!]
\centering
\begin{tabular}{|lp{6in}|}
\hline
& {\bf Experimenter-developed questions}\\
\hline
$\circ$ & Are you trying to figure out how the co-agent (and environment) work?\\
$\circ$ & For the whole trial?\\
$\circ$  & If not, did you figure it out or just start to trust it?\\
$\circ$  & After time, or successes? \\
$\circ$  & How much do you notice or think about the other agent at the beginning? The middle? The end?\\
\hline
& {\bf Participant-developed questions}\\
\hline
$\circ$  & Changes in when and how I counted: did I count from the start of the trial? Did I shift to just counting from the agent cue and not counting from the beginning? When did I shift between these and under what conditions or observations on timing?\\
$\circ$  & What co-agent behaviours did I like and not like?\\
$\circ$  & Adaptation rates: what were my expectations on response or learning times for co-agents?\\
$\circ$  & Did I think of co-agents as adaptive systems / predictors or not?\\
$\circ$  & What conditions did I lose confidence in the co-agent; when did I gain confidence?\\
$\circ$  & When did trust in the co-agent occur quickly?\\
\hline
\end{tabular}
\caption{{\bf Qualitative questions considered by the participant} for compiling notes during each session. Experimenter-developed questions were posed by the member of the study team conducting the qualitative analysis at the outset of the trials. Participant-developed questions were generated independently by the member of the study team acting as the participant, and evolved as the study progressed.}
\label{tab:qual-questions}
\end{table*}

In cases where a machine agent is interacting with a human, there are insights that can be gained from the user's experience through the use of qualitative methods. The participant was provided with questions to consider as they underwent the trials, and also initiated their own questions (Tab. \ref{tab:qual-questions}). Participant-generated questions allowed for the participant to contribute to the data being gathered based on their experience with the system, rather than being restricted to a priori questions. The participant made notes after each session. These notes were sent without review by the participant to a different author who analyzed them. The notes were analyzed in isolation of the quantitative data initially in order to minimize biases that may have arisen from seeing the quantitative results. The notes were made after each session (not trial), so they are broader senses of the interaction and unfortunately lack specific connections to the type of co-agent. We acknowledge that the participant is an author on this paper and an expert in the system used, and that the analysis of the notes taken was done by a person very familiar with the participant. These factors, along with the circumstances that prompted the use of only a single participant, ultimately mean that the findings of this portion cannot be generalized, but provide insight as to what direction to take in expanded studies; we expect they will illuminate what outcomes we might find possible or seek to directly test for in a wider study \citep{mook1983}.

Discourse analysis was the primary methodology used to analyze the notes made by the participant. The notes were examined for recurring sentiment and topics, and three themes emerged. These were trust, cognitive load, and use of the message. Following the initial analysis the participant discussed their notes and thoughts with the experimenter doing the analysis, raising additional points of interest. 

{\bf Trust}:
Trust was built most rapidly when the system learned quickly. In response to the prompt ``When did trust in the co-agent occur quickly'', the participant noted:  ``when it learned fast''. The participant found that when the system was demonstrably correct earlier in trial, indicated by the participant receiving the signal from the agent to move followed closely by the hazard indication, they tended to trust it more. Trust is foundational to the other themes that emerged from the participant's notes. Trust changed how the participant experienced the mental load:
``With trust in my co-agent, I can let [my] mind wander.'', and
trust also allowed unexpected uses of the signal:
``... the co-agent helps me feel like I have a lower bound of safety once it is trained, and then can choose my risk based on its feedback.''
It is reasonable to suggest that if the participant did not trust the agent they would not feel there was safety, that they could rely on it's input to make riskier decisions about the task.

User trust in a machine co-agent may be crucial to any other outcomes we seek from such a partnership. Future studies may benefit from a direct line of questioning that specifically focuses on trust. A possible challenge in investigating the development of trust is that trust is initially in the participant's thoughts, but once it is developed the participant no longer thinks about trust directly. However, the existence of that trust enables the participant to shift their cognitive load to the agent as well as being a key factor in the participant's use of the message in unexpected or unintended ways.

{\bf Cognitive load}:
The task was designed to make mental timing difficult, to encourage the use of the agent's signal. This difficulty is reflected in the participant's notes: 
``Counting in my head is fatiguing, and means I am unable to allow my mind to wander even for a moment. With a co-agent, I can let my mind wander and, even if I am still counting myself, can have a `heads up' if I need to pay attention again when it buzzes.''
On the trials with a co-agent the participant could let their focus lapse to some degree. This reduced the cognitive load on the participant by offloading some of the responsibility to the co-agent agent. The participant also reported that their count could be inaccurate. They reported that when caching the points (which is exciting and tense) their internal count may have accelerated. This feeling was based on when the participant received the co-agent cue earlier than expected. ``For co-agent conditions, I think I am too eager to off load the timing to the co-agent. I think I stop counting and follow the agent feedback cues earlier than I strategically should.''
This comment, coupled with a comment made by the participant at the top of the second session notes (``Another 8 sessions of this? Sigh.''), suggests that the participant wanted to offload the dull but cognitively intensive task on the agent, even in situations where trust hadn't yet been established in the system. Further explorations are required in order to determine whether established trust leads to cognitive off-loading, or whether the desire to off-load cognitive processes leads to premature trust. Most likely, the establishment of trust and the gradual reliance on the system are related processes that are developed simultaneously, as reflected in the participant's notes: ``...in some runs I found myself gradually stopping counting and starting to rely on the agent cues even without explicitly choosing to do so... I’d be part way into a pulse and realize (in a bit of a panic) that I had forgotten to start counting... but then would get the cue and move and it was fine.''
The transfer of cognitive load to the co-agent in this case was inadvertent, rather than a conscious choice. The element of panic suggests that the participant does not yet fully trust the system, but the system's competence encouraged further trust. This note also emphasizes that the participant is not relying on the system to do the task, but is working with the co-agent to accomplish the task. 

{\bf Use of the message}:
The machine agent was designed to signal the participant with its prediction of the onset of the upcoming wind gust. Despite knowing this, the participant occasionally used the prompt from the co-agent in ways that were not anticipated in the original design. 

One way the prompt of the co-agent was used was to help the participant determine what type of trial they were in (fixed, drift, or random). ``...evidence accumulation on my part for what kind of trial I was in was definitely helped by having a co-agent as off-board memory and a sanity check for what I perceived so far.'' Other statements made by the participant suggest that over time, they found it less important to determine what sort of trial they were in. Rather, they began to associate the prompt from the co-agent with the oncoming hazard in the way that the participant felt they could best solve the task: ``Even if I moved before the co-agent’s cue and did not rely on the agent in that regard, I might at times be seeing how the pulse and the agent buzz line up or have changed in their distance from each other, and use that info to modify how I respond to my own counting strategy.'' By making use of their greater knowledge of the system, the world, and the context of the task, the participant was able to leverage the agent signal as more than a simple cue to move. The participant began to associate the prompt with the hazard in a spatial sense; the participant got a sense for if the hazard was ``approaching'' or ``receding'' based on perceived time between prompt and hazard signals. 
In some instances, the participant was confident enough about their understanding of the prompt-to-hazard timing to engage in risky behaviour: ``I was at times racing the pulse; the co-agent would cue me but I would see the heat bar almost full and then gamble that it would fill fast enough before the [hazard] came, given what I knew about relationship between cue and future [hazard].''
A final example of non-standard use of the prompt came in cases with a particularly short hazard ISI, where the participant used the co-agent signal as a verification of their mental time-keeping: ``...even though I had agent help, it was not fast enough to be useful in advance of [hazard], so I mainly used it as a checksum.''

{\bf Discussion with Participant}:
Frequently during discussion, the participant mentioned using the signal from the co-agent as a reference to the impending event rather than a direct signal of it, or as a direction to move. Use of the message as a reference point rather than as a cue to move suggests that the participant made use of the co-agent's signal to inform their own tracking and planning of the task, rather than naively following its advice. Noting again that our participant is deeply familiar with learning systems and their limitations, it will be interesting to see in future studies whether more varied participants have similar interactions.

The participant noted that in early sessions, part of their strategy was to infer the type of trial they were interacting with (fixed, random, or drift), in order to inform their timing. However, after repeated interactions with the system, the participant found it became unimportant to try to identify the trial condition when choosing how they will act. Instead, they reported that they built a sense of the relation between the prompt and the hazard and acted according to that. Despite the participant's deep initial understanding of the task, the participant's approach changed over time, seemingly as a result of interactions with the co-agent. 

\subsection{Synthesis of Quantitative and\\Qualitative Results}

In both the quantitative and qualitative analyses, we see human trust of the co-agent emerging as an important theme. The participant's notes suggest that using the sign of the signal-to-exit interval (Fig. \ref{figure:h-a-signal-exit-interval-boxplot}) as an indicator of human trust might miss parts of the picture, since the participant makes use of the co-agent signal in other ways than as simply a cue to move. Other quantitative measures of trust should be sought, to corroborate this interpretation. One particular notion of intense trust called out in the participant notes (when the participant is ``racing'' the pulse, caching points after the co-agent signal but before the hazard) is also visible in the quantitative data. Of the 14 instances where a point caching event is recorded after a co-agent signal and before a hazard, 13 of these instances occurred when the participant was paired with the tile-coded trace co-agent. This could be interpreted as a strong measure of trust between the human and tile-coded trace co-agent, but it should also be noted that (as shown in Fig. \ref{figure:h-a-learned-signal-timing}) the tile-coded trace co-agent reliably gives more lead-time than necessary before the pulse, leaving time for pulse ``racing'' that the bit-cascade co-agent does not.

The qualitative analysis also explains some apparently contradictory aspects of the quantitative data. For example, the fact that the participant does not appear to use the bit-cascade co-agent in fixed-ISI trials (Fig. \ref{figure:h-a-signal-exit-interval-boxplot}) yet the participant is hit by the pulse significantly fewer times as compared to no co-agent (Fig. \ref{figure:h-a-mean-hitsteps}) can be explained by the participant's use of the agent signal as a ``sanity check for what [the participant] perceived so far'', or as a ``checksum'', rather than as the intended cue to leave the goal region.

\subsection{Follow-Up Studies}
Specific quantitative measures to assess human trust in the co-agent would be particularly informative for future studies, especially if such measures could assess changes in levels of trust over the course of a trial or across sessions. One such task modification might involve the introduction of a secondary, voluntary and cognitively demanding task that could be performed simultaneously while gathering heat. While engaged with the secondary task, the participant would need to place trust in the co-agent to keep track of the timing in the primary task. Another point of potential interest would be the introduction of an adaptive agent that blends features of the two representations tested in this study; an agent that adapts its feature-binning to allow for rapid early learning as well as late-trial accuracy may be an interesting candidate for study. 

For future time-based prediction experiments or applications involving human actors with machine co-agents, we make no particular recommendations about representation or threshold choices, as we understand these to be task-specific. We do however stress the importance of these choices, and recommend that they be made with both early and late learning stages in mind, and considering the interaction between the human and machine's actions.

\section{Discussion}
\label{sec:discussion}

As each independent empirical section (Sec. \ref{sec:nexting-experiments}--\ref{sec:vr-experiments}) contained relevant discussion to the results presented in the respective sections, in this final section of the manuscript we provide discussion on cross-cutting themes. In particular, we highlight general findings in Pavlovian signalling, comparisons to the relevant literature, and also cross-cutting observations that connect the different experiments presented above.

The main contribution of this manuscript is the in-depth exploration of what we here defined as Pavlovian signalling. At a high level, the results in this work support a clear value case for using Pavlovian signalling as a lens to study certain agent-agent relationships. As identified by \cite{pilarski2017}, we can frame different dyadic partnerships between agents in terms of the agency and capacity of the parties engaged in the interaction; while capacity of a partnership might be limited by having reduced agency by one of the parties, the simplicity of one partner provides the opportunity for fast learning of its behaviour by the other party \citep{pilarski2017}. This is the case with Pavlovian signalling: we see that, with no policy learning on the part of the co-agent, predictions may be learned rapidly (less than 500 steps), as seen in Figs. \ref{fig:token-fixed-gamma-comp} and \ref{fig:token-countdown-comp} and when distilled down to a low-bandwidth signal can be readily used for policy learning by a second control learning agent (also c.f. \cite{pilarski2012} for a discussion of the way bandwidth, latency, and explicitness impact agent/co-agent relationships).  

In addition to the learnability of the co-agent by the main agent, we see evidence that there are further benefits to Pavlovian signalling in that it assumes nothing about the internal structure or mutability of the agent; the agent can approach the signals from an ungrounded perspective, as in the full control learning machine agent in Sec. \ref{sec:control-experiments} or the human agent in Sec. \ref{sec:vr-experiments}. The agent can also approach the signals from a grounded perspective, as per the responsive, hard-wired agent introduced in Sec. \ref{sec:control-experiments}. Further, there is no reason to limit the co-agent to only observations of the environment as we have done in these studies; it is natural to think that information pertaining to the main agent and its adaptability can be used as well as inputs to the assitant, further increasing its ability to make relevant predictions and to begin to build what we have previously termed communicative capital \citep{pilarski2017}. As such, Pavlovian signalling appears to be a viable stepping stone between between fixed agent-agent relationships and fully, bidirectionally learned multiagent relationships.

We also note that, across studies in this work, that the interpretation of representations and how they relate to predictions and finally control behaviour appears to depend on the agency of the party receiving the signals. In the case of a Pavlovian signalling co-agent and machine agent (Sec. \ref{sec:control-experiments}), the lack of temporal aliasing in the bit cascade representation was observed to provide a level of consistency in the generated tokens that allowed for agents partnered with bit cascade co-agents to do better (fixed condition) or equal to (random and drift conditions) that of co-agent-agent combinations using a presentation that produced more temporal aliasing (Figs. \ref{fig:rep-comparison-boxplot}). However, in the case of a Pavlovian signalling co-agent paired with an agent that was able to model the co-agent and produce more complex decision making, tile-coded-trace representations with greater temporal aliasing proved more effective and more well received as signal inputs to the human agent (e.g., Fig. \ref{figure:h-a-signal-exit-interval-over-trial-length}). This is an interesting finding, and suggests that the interaction between temporal aliasing/resolution and agent capacity is more nuanced than we expected, and is an important element to study when formulating agent-agent partnerships. 

Returning to our basis in the neuroscience literature, we note that we largely designed our representations to act as population clocks with different proprieties. These clocks were then provided to prediction learners that either served as pattern timing prospective elements (accumulation predictions forecasting the incidence of future events in time) or interval timing elements with ramping model properties (countdown predictions forecasting the remaining interval until a future event). These predictions were used to inform reflexive action or more complex control, with their utility relating strongly to the amount and location of temporal aliasing observed in the prediction at different places within an ISI (c.f., Fig. \ref{fig:nexting-long-isi-comp} fixed condition). From the comparisons of accumulation and countdown predictions, we also see that our approximation of interval timing GVFs are more likely to get problematic over longer spans or with more variability in the environment than would approximations of pattern timing GVFs. Further, even what might be considered simple ramping cells (bias unit) prove to be important in some simple cases if the threshold for token generation is tuned well---i.e., tracking and not prediction also forms a viable ramping model for interval timing in our instance of Pavlovian signalling.

In total, we found evidence for our hypothesis that nexting on the part of a machine agent can be used in token formation to help bridge the gap in time between past stimulus and later decision points for a human or machine decision-making agent, improving decision making abilities of the agent in small or large ways. However, there was no clearly ``best'' temporal representation across the different studies and contexts, even if some representations allowed prediction formation that most closely matched the true cumulant return calculations of Eq. \ref{eq:gvf}. Or, said differently and to paraphrase \cite{kearney2021}, from our results we see that the accuracy of a co-agent's predictions do not necessarily dictate their usefulness in Pavlovian signalling for agent-agent interaction.

\subsection{Preliminary Evidence on more Complex Tokenization Approaches}
\label{sec:discussion-multitoken}

\begin{figure}[!t]
\centering
\includegraphics[width=\linewidth]{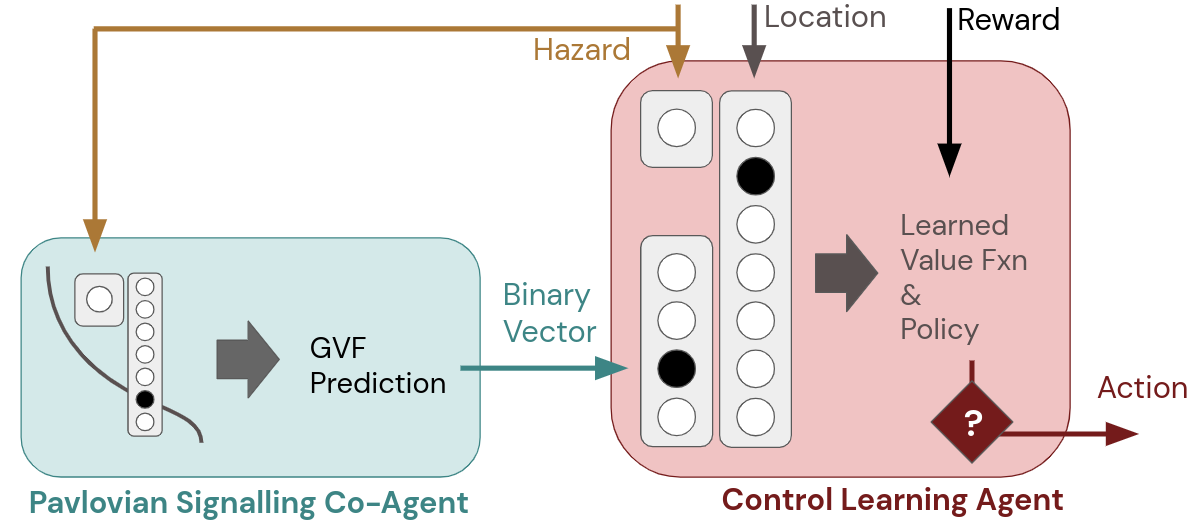}\
\caption{Schematic example of an agent/co-agent pairing for {\bf Pavlovian signalling with multiple tokens}. GVF predictions are here mapped to a vector of binary tokens according to a pre-determined coding scheme.}
\label{fig:multi-token-schematic}
\end{figure}

As noted in Sec. \ref{sec:nexting-experiments}, the Pavlovian signalling we propose in this work as a process for turning predictions into tokens is in no way limited to the generation of a single binary token. We may also consider cases where predictions are mapped in a hard-coded way to the activity of bits in a one-hot vector of arbitrary size, are hashed into a vector with multiple active features, or are mapped to a vector of scalar values (with one example shown schematically in Fig. \ref{fig:multi-token-schematic}). We also note that the mapping process (e.g., the threshold or other mapping function) may best be suited to real-time adaptation during the process of learning either on the part of the co-agent or by some pre-determined external mechanism. Indeed, the observations from the random and drift conditions in Figs. \ref{fig:token-fixed-gamma-comp} and \ref{fig:token-countdown-comp}, the trends noted for control learning in different representations for the abstract Frost Hollow environment, and the qualitative and quantitative human-machine findings all suggest there might be merit in online adaptation of a Pavlovian signalling approach. Online adaptation and multiple tokens are outside the scope of the present work (which the authors note is already quite lengthy), but we believe may serve as helpful doorways toward understanding more complex emergent communication learning processes between an agent and a co-agent.

To provide initial insight into a multi-token process, we also performed a preliminary study on one extension where tokens took the form of a one-hot vector tokenization scheme (implementing the approach for vectors of length 1, 3, and 5). Our findings showed that control learning agent performance on the Frost Hollow domain when using these tokens as part of the agent's state in place of the single binary token used in our studies above led to a stratification of agent performance levels, where more complex tokenization schemes led to potentially slower learning by the control agent, but higher overall asymptotic performance. We extrapolate from these anecdotal results that, not surprisingly, increasing the complexity of tokenization will lead to more complex agent/co-agent interaction and possible performance gains and losses depending on the implementation of such approaches. Evidence from \cite{edwards2016} also suggests that scalar presentation of predictive tokens with appropriate threshold for minimum and maximum token values may lead to actionable signals for the receiving agent. Evidence from \cite{pilarski2019} further suggests that token presentation based on context (e.g., location of stimulus) has possible merits for the decision-making agent.

We also note that it is unclear how tokenization complexity will impact co-agents learning and adapting their predictions and tokens over extended periods of time, as partnerships move beyond the limited episodic length studied in this work and embrace the full continual learning setting for both co-agent/machine-agent and co-agent/human contexts.

As is well described by \cite{scottphillips2014}, there is a clear route to progress from grounded signals to what he termed as full, ostensive-inferential communication; \cite{pezzulo2013} also describe the power of signalling in more elaborate acts of coordination and agent-agent alignment. We are therefore interested to explore how Pavlovian signalling can provide a functional bridge towards this more natural, expressive communication interface. \cite{pilarski2019} provided one view into this pathway, through a bidirectional learning relationship  wherein the agent (a human) made certain things visible for co-agent learning, and the co-agent subsequently learned when and how to make its Pavlovian signalling tokens visible to the agent. Tokens created in Pavlovian signalling with GVFs have the nice property that they are constructivist in nature and we could imagine them being created by a co-agent autonomously as opposed to specified by an external designer. Tokens are for the sender grounded in co-agent-centric (subjectively specified) GVF question parameters cumulant $c$, time scale $\gamma$, and policy $\pi$ and also in the mapping approach, in this case is parameterized by the threshold value $\tau$ used in token generation. For our specific case, this 4-element tuple

\begin{equation}
\lbrace c, \gamma, \pi, \tau \rbrace    
\end{equation}

is the grounding of the token, and it can be fully subjective to the sending agent and not require some reference to an objective frame of reference. This suggests a solid, understandable basis for more eloquent communication between co-agents and agents.

\begin{tcolorbox}[colback=yellow!7!red!5!white,colframe=yellow!35!red!25!white]
\textbf{Pavlovian signalling as implemented in this work} is a process wherein learned, temporally extended predictions in the form of generalized value functions are mapped via a fixed threshold to Boolean tokens intended for receipt by a decision-making agent, and where these signals are grounded for the sender in the cumulant $c$, time scale $\gamma$, policy $\pi$ and threshold value $\tau$ of their computational elements.
\end{tcolorbox}

\subsection{Non-linear Co-agents and\\Single-agent Alternatives}

A final, natural question relates to the form of the co-agent's prediction learning algorithm and architecture. For clarity, learning speed and stability in an online learning environment, we chose to specifically study co-agents with linear learning mechanisms. Non-linear learning, if efficient in an online context, promises even better representation of time and multifaceted temporal phenomena. For example, we might consider the clockwork RNN of \cite{koutnik2014} or an LSTM formulation as per \cite{rafiee2021}. These are viable areas for future work. 

Recently \cite{rafiee2021} investigated recurrent learning algorithms such as LSTMs on a suite of problems inspired by experiments in animal learning. The key question addressed in their work was how effectively a recurrent learning agent can represent time in trace conditioning. Trace conditioning is a form of classical conditioning that proceeds in repeated trials. In each trial the agent is presented with a binary CS that occurs for several timesteps (with a random unpredictable onset). After several time-steps (randomized but within some bounded interval), a binary US is presented to the agent. The time between CS onset and US onset is called the inter-stimulus interval or ISI. The agent's job is to predict the US based on the CS, just as a rabbit predicts the air-puff based on a tone. The challenge is that there are no sensory signals presented to the agent between the CS and the US; the agent must construct a representation that somehow encodes the passage of time. The major finding of Rafiee et al.'s study was that recurrent learning systems could accurately make trace conditioning predictions for short ISI. However, as ISI was increased, state-of-the-art algorithms could not learn accurate predictions of the US even as the amount of Back propagation through time was increased. 

Further, we might consider placing temporal representation elements, such as an LSTM, inside the main agent and not considering the co-agent as an explicit system. While there are cases where this could be (and verifiably is) an excellent choice for learning complex phenomena, we note that the ability to consider a co-agent separately gives us the capacity to study interactions with systems not under our control, e.g., a production level system, an external piece of hardware or software, and importantly as shown in Sec. \ref{sec:vr-experiments}, advanced biological learning agents of great relevance to daily life: humans beings.

\section{Conclusions}
\label{sec:conclusions}

In this work, we contributed a concrete definition and exploration of Pavlovian signalling as implemented via processes of GVF learning. Our findings, while preliminary in that they are derived from a single human-agent case series and related fundamental agent-agent experiments, suggest that {\em Pavlovian signalling by a co-agent can be learned very rapidly in single-shot real time deployment}, improves temporal decision-making of receiving human and machine control-learning agents, and opens a number of future avenues for using GVF learning in this way to augment, adapt, and potentially enhance human and machine perception, action, and cognition---especially in settings where the connection between the agent and co-agent is tight but perhaps low in bandwidth. When presented with information from a machine co-agent, we observed changes in both the timing of decisions of both human and machine agents (behavioural change, differences in sensorimotor trajectories, and reaction times) and also the quantitative outcomes of the timing task (score, number of mistakes). Future study is needed with larger environments or continual learning settings with distractors, and in tasks that more fully blend both time and space. In summary, we believe there is great opportunity for using Pavlovian signalling to understand agent-agent signalling and communication in complex tasks that unfold in both time and space, and as a useful way-point between hand-designed communication interfaces and full machine communication learning.

\subsection*{Acknowledgements}
The authors thank Kevin McKee, Kory Mathewson, Michael Bowling, Ola Kalinowska, Nathan Wispinski, Andrew Bolt, Alexander Zacherl, Drew Purves, Edward Hughes, Christopher Summerfield, Richard Sutton, Koray Kavukcuoglu, and Shanon Phelan, for many helpful conversations, discussions, and technical insights regarding this work, in addition to a number of other colleagues within DeepMind and at the University of Alberta. University of Alberta collaboration on this work by PMP, ASRP, and AW was supported, in part, by the Canada CIFAR AI Chairs program, the Canada Research Chairs program, the Natural Sciences and Engineering Research Council (NSERC), and the Alberta Machine Intelligence Institute (Amii). We also thank the authors and maintainers of Launchpad, which we used for our experiments~\citep{launchpad2021}. We further note that peer-reviewed versions of components of this paper have appeared or will appear in \cite{butcher2022} and \cite{brenneis2021}.

\newpage

\subsection*{Author Contribution Statement}

All authors contributed to the concept or design of the work and to the acquisition, analysis or interpretation of data; drafted the article or revised it critically for important intellectual content; approved the version to be published; and are in agreement to be accountable for all aspects of the work in ensuring that questions related to the accuracy or integrity of any part of the work are appropriately investigated and resolved.


\end{document}